\documentclass[11pt]{article}

\usepackage[letterpaper,textwidth=6.75in,textheight=9in,centering]{geometry}
\usepackage{times}
\usepackage{authblk}   
\usepackage{microtype}
\usepackage{graphicx}
\usepackage{booktabs}
\usepackage{multirow}
\usepackage{array}
\usepackage{enumitem}
\usepackage{float}
\usepackage{placeins}   
\usepackage{algorithm}
\usepackage{algorithmic}
\usepackage{xcolor}
\usepackage[round]{natbib}
\usepackage[font=small,labelfont=bf,skip=4pt]{caption}
\usepackage[colorlinks=true,linkcolor={blue!45!black},citecolor={blue!45!black},urlcolor={blue!45!black}]{hyperref}

\usepackage{amsmath}
\usepackage{amssymb}
\usepackage{mathtools}
\usepackage{amsthm}
\usepackage[capitalize,noabbrev]{cleveref}

\expandafter\def\csname v@C1.hb.qwen\endcsname{$+$0.75 [0.62, 0.86]}
\expandafter\def\csname v@C1p.hb.qwen\endcsname{0.0001}
\expandafter\def\csname v@C2.hb.qwen\endcsname{$+$0.63 [0.52, 0.72]}
\expandafter\def\csname v@C2p.hb.qwen\endcsname{0.0001}
\expandafter\def\csname v@C3.hb.qwen\endcsname{$+$0.11 [0.05, 0.18]}
\expandafter\def\csname v@C3p.hb.qwen\endcsname{0.0001}
\expandafter\def\csname v@C4.hb.qwen\endcsname{0.98 [$-$1.05, 2.21]}
\expandafter\def\csname v@C4p.hb.qwen\endcsname{0.10}
\expandafter\def\csname v@C1.pz.qwen\endcsname{$+$0.77 [0.72, 0.81]}
\expandafter\def\csname v@C1p.pz.qwen\endcsname{0.0001}
\expandafter\def\csname v@C2.pz.qwen\endcsname{$+$0.74 [0.69, 0.78]}
\expandafter\def\csname v@C2p.pz.qwen\endcsname{0.0001}
\expandafter\def\csname v@C3.pz.qwen\endcsname{$+$0.71 [0.65, 0.77]}
\expandafter\def\csname v@C3p.pz.qwen\endcsname{0.0001}
\expandafter\def\csname v@C1.mq.qwen\endcsname{$+$0.41 [0.36, 0.45]}
\expandafter\def\csname v@C1p.mq.qwen\endcsname{0.0001}
\expandafter\def\csname v@C2.mq.qwen\endcsname{$+$0.36 [0.31, 0.40]}
\expandafter\def\csname v@C2p.mq.qwen\endcsname{0.0001}
\expandafter\def\csname v@C3.mq.qwen\endcsname{$+$0.08 [0.05, 0.11]}
\expandafter\def\csname v@C3p.mq.qwen\endcsname{0.0001}
\expandafter\def\csname v@C1.hb.olmo\endcsname{$+$0.60 [0.50, 0.71]}
\expandafter\def\csname v@C1p.hb.olmo\endcsname{0.0001}
\expandafter\def\csname v@C2.hb.olmo\endcsname{$+$0.44 [0.33, 0.54]}
\expandafter\def\csname v@C2p.hb.olmo\endcsname{0.0001}
\expandafter\def\csname v@C3.hb.olmo\endcsname{$+$0.34 [0.22, 0.47]}
\expandafter\def\csname v@C3p.hb.olmo\endcsname{0.0001}
\expandafter\def\csname v@C4.hb.olmo\endcsname{0.61 [0.13, 0.96]}
\expandafter\def\csname v@C4p.hb.olmo\endcsname{0.01}
\expandafter\def\csname v@C1.pz.olmo\endcsname{$+$0.70 [0.66, 0.74]}
\expandafter\def\csname v@C1p.pz.olmo\endcsname{0.0001}
\expandafter\def\csname v@C2.pz.olmo\endcsname{$+$0.65 [0.61, 0.70]}
\expandafter\def\csname v@C2p.pz.olmo\endcsname{0.0001}
\expandafter\def\csname v@C3.pz.olmo\endcsname{$+$0.71 [0.66, 0.76]}
\expandafter\def\csname v@C3p.pz.olmo\endcsname{0.0001}
\expandafter\def\csname v@C1.mq.olmo\endcsname{$+$0.39 [0.34, 0.44]}
\expandafter\def\csname v@C1p.mq.olmo\endcsname{0.0001}
\expandafter\def\csname v@C2.mq.olmo\endcsname{$+$0.39 [0.35, 0.44]}
\expandafter\def\csname v@C2p.mq.olmo\endcsname{0.0001}
\expandafter\def\csname v@C3.mq.olmo\endcsname{$+$0.11 [0.07, 0.15]}
\expandafter\def\csname v@C3p.mq.olmo\endcsname{0.0001}
\expandafter\def\csname v@C1.min\endcsname{0.39}
\expandafter\def\csname v@C1.max\endcsname{0.77}
\expandafter\def\csname v@C1.pp.min\endcsname{39}
\expandafter\def\csname v@C1.pp.max\endcsname{77}
\expandafter\def\csname v@C2.min\endcsname{0.36}
\expandafter\def\csname v@C2.max\endcsname{0.74}
\expandafter\def\csname v@conf.pass\endcsname{19}
\expandafter\def\csname v@prereg.pass\endcsname{47}
\expandafter\def\csname v@prereg.n\endcsname{48}
\expandafter\def\csname v@eval.benchmarks\endcsname{3}
\expandafter\def\csname v@eval.families\endcsname{4}
\expandafter\def\csname v@eval.size.range\endcsname{4--32B}
\expandafter\def\csname v@eval.tok.savings.range\endcsname{10--54\%}
\expandafter\def\csname v@eval.tok.savings.mean\endcsname{36\%}
\expandafter\def\csname v@eval.accuracy.gain.pp\endcsname{64}
\expandafter\def\csname v@size.debate.range\endcsname{0.06--0.16}
\expandafter\def\csname v@size.evpc.range\endcsname{0.76--0.89}
\expandafter\def\csname v@size.gain.range\endcsname{$+$0.49 to $+$0.83}
\expandafter\def\csname v@size.gain.lo.min\endcsname{0.37}
\expandafter\def\csname v@size.tok.ratio.range\endcsname{0.61--0.83}
\expandafter\def\csname v@fil.cap.gap.abs.range\endcsname{0.68--0.86}
\expandafter\def\csname v@fil.ev.gap.range\endcsname{0.06--0.15}
\expandafter\def\csname v@fil.nodedup.gap.range\endcsname{0.31--0.74}
\expandafter\def\csname v@fil.hist.range\endcsname{0.05--0.22}
\expandafter\def\csname v@fil.digest.range\endcsname{0.02--0.09}
\expandafter\def\csname v@fil.cap.range\endcsname{0.01--0.12}
\expandafter\def\csname v@fil.hist.tok.ratio.range\endcsname{2.1--2.6$\times$}
\expandafter\def\csname v@fil.hist.debate.range\endcsname{$-$0.01 to $+$0.12}
\expandafter\def\csname v@fil.ev.hb.range\endcsname{0.70--0.75}
\expandafter\def\csname v@fil.evpc.hb.range\endcsname{0.78--0.86}
\expandafter\def\csname v@fil.capdedup.hb.range\endcsname{0.56--0.81}
\expandafter\def\csname v@fil.nodedup.hb.range\endcsname{0.42--0.53}
\expandafter\def\csname v@fil.ev.pz.range\endcsname{0.68--0.84}
\expandafter\def\csname v@fil.evpc.pz.range\endcsname{0.80--0.91}
\expandafter\def\csname v@fil.capdedup.pz.range\endcsname{0.25--0.48}
\expandafter\def\csname v@fil.nodedup.pz.range\endcsname{0.07--0.24}
\expandafter\def\csname v@fil.nodedup.tok.ratio.range\endcsname{2.4--4.7$\times$}
\expandafter\def\csname v@fil.ev.tok.more.range\endcsname{13--16}
\expandafter\def\csname v@gpuh.Z.a30\endcsname{254}
\expandafter\def\csname v@z5.own.range\endcsname{$-$0.51 to $-$0.35}
\expandafter\def\csname v@z5.claims.range\endcsname{$-$0.58 to $-$0.23}
\expandafter\def\csname v@z5.para.range\endcsname{$-$0.02 to 0.00}
\expandafter\def\csname v@z5.para.own.range\endcsname{$+$0.32 to $+$0.50}
\expandafter\def\csname v@z5.claims.tok.ratio.range\endcsname{1.4--1.6$\times$}
\expandafter\def\csname v@z5.view.wrong.range\endcsname{0.59--0.73}
\expandafter\def\csname v@z5.view.held.range\endcsname{0.38--0.68}
\expandafter\def\csname v@bound.msg.tokens\endcsname{200}
\expandafter\def\csname v@bound.inbox.16\endcsname{3,000}
\expandafter\def\csname v@bound.inbox.64\endcsname{12,600}
\expandafter\def\csname v@bound.context\endcsname{4,096}
\expandafter\def\csname v@bound.every.context.range\endcsname{8,192--16,384}
\expandafter\def\csname v@g5.hb.N4.debate\endcsname{0.39 [0.26, 0.53]}
\expandafter\def\csname v@g5.hb.N4.pool_evpc\endcsname{0.94 [0.87, 0.99]}
\expandafter\def\csname v@g5.hb.N4.full_iter\endcsname{0.94 [0.86, 0.99]}
\expandafter\def\csname v@g5.hb.N4.diff\endcsname{$+$0.55 [0.41, 0.68]}
\expandafter\def\csname v@g5.hb.N8.debate\endcsname{0.31 [0.20, 0.44]}
\expandafter\def\csname v@g5.hb.N8.pool_evpc\endcsname{0.93 [0.85, 0.98]}
\expandafter\def\csname v@g5.hb.N8.full_iter\endcsname{0.96 [0.91, 1.00]}
\expandafter\def\csname v@g5.hb.N8.diff\endcsname{$+$0.61 [0.47, 0.74]}
\expandafter\def\csname v@g5.hb.N16.debate\endcsname{0.27 [0.17, 0.38]}
\expandafter\def\csname v@g5.hb.N16.pool_evpc\endcsname{0.92 [0.83, 0.98]}
\expandafter\def\csname v@g5.hb.N16.full_iter\endcsname{0.96 [0.89, 1.00]}
\expandafter\def\csname v@g5.hb.N16.diff\endcsname{$+$0.65 [0.52, 0.76]}
\expandafter\def\csname v@g5.hb.N32.debate\endcsname{0.18 [0.09, 0.27]}
\expandafter\def\csname v@g5.hb.N32.pool_evpc\endcsname{0.91 [0.82, 0.98]}
\expandafter\def\csname v@g5.hb.N32.full_iter\endcsname{0.95 [0.86, 1.00]}
\expandafter\def\csname v@g5.hb.N32.diff\endcsname{$+$0.74 [0.63, 0.83]}
\expandafter\def\csname v@g5.diff.pooled\endcsname{$+$0.63 [0.53, 0.74]}
\expandafter\def\csname v@g5.diff.pp\endcsname{63}
\expandafter\def\csname v@g5.diff.est\endcsname{0.63}
\expandafter\def\csname v@g5.diff.p\endcsname{0.001}
\expandafter\def\csname v@g5.items\endcsname{37}
\expandafter\def\csname v@g5.excluded\endcsname{2}
\expandafter\def\csname v@g5.incomplete\endcsname{2}
\expandafter\def\csname v@g5.allruns\endcsname{$+$0.61 [0.50, 0.71] (41 tasks)}
\expandafter\def\csname v@g5.debate.span\endcsname{0.39 to 0.18}
\expandafter\def\csname v@g5.pool_evpc.range\endcsname{0.91--0.94}
\expandafter\def\csname v@g5.full_iter.range\endcsname{0.94--0.96}
\expandafter\def\csname v@g5.tok.savings.range\endcsname{23--40\%}
\expandafter\def\csname v@g5.usd.total\endcsname{23}
\expandafter\def\csname v@g5.calls.total\endcsname{214{,}611}
\expandafter\def\csname v@g5.pilot.diff\endcsname{$+$0.59 [0.45, 0.72]}
\expandafter\def\csname v@g5.pred.G5\endcsname{8 of 8}
\expandafter\def\csname v@g5.pred.G5T\endcsname{17 of 17}
\expandafter\def\csname v@g5h.hb.N4.debate\endcsname{0.48 [0.36, 0.61]}
\expandafter\def\csname v@g5h.hb.N4.pool_evpc\endcsname{0.96 [0.89, 1.00]}
\expandafter\def\csname v@g5h.hb.N4.full_iter\endcsname{0.91 [0.81, 0.98]}
\expandafter\def\csname v@g5h.hb.N4.diff\endcsname{$+$0.48 [0.35, 0.60]}
\expandafter\def\csname v@g5h.hb.N8.debate\endcsname{0.36 [0.24, 0.49]}
\expandafter\def\csname v@g5h.hb.N8.pool_evpc\endcsname{0.97 [0.90, 1.00]}
\expandafter\def\csname v@g5h.hb.N8.full_iter\endcsname{0.92 [0.83, 0.99]}
\expandafter\def\csname v@g5h.hb.N8.diff\endcsname{$+$0.60 [0.47, 0.73]}
\expandafter\def\csname v@g5h.hb.N16.debate\endcsname{0.34 [0.21, 0.47]}
\expandafter\def\csname v@g5h.hb.N16.pool_evpc\endcsname{0.95 [0.88, 1.00]}
\expandafter\def\csname v@g5h.hb.N16.full_iter\endcsname{0.92 [0.83, 0.99]}
\expandafter\def\csname v@g5h.hb.N16.diff\endcsname{$+$0.61 [0.48, 0.75]}
\expandafter\def\csname v@g5h.hb.N32.debate\endcsname{0.30 [0.17, 0.44]}
\expandafter\def\csname v@g5h.hb.N32.pool_evpc\endcsname{0.95 [0.88, 1.00]}
\expandafter\def\csname v@g5h.hb.N32.full_iter\endcsname{0.92 [0.83, 0.99]}
\expandafter\def\csname v@g5h.hb.N32.diff\endcsname{$+$0.65 [0.51, 0.78]}
\expandafter\def\csname v@g5h.diff.pooled\endcsname{$+$0.59 [0.47, 0.70]}
\expandafter\def\csname v@g5h.diff.pp\endcsname{59}
\expandafter\def\csname v@g5h.diff.est\endcsname{0.59}
\expandafter\def\csname v@g5h.diff.p\endcsname{0.001}
\expandafter\def\csname v@g5h.items\endcsname{36}
\expandafter\def\csname v@g5h.excluded\endcsname{3}
\expandafter\def\csname v@g5h.incomplete\endcsname{2}
\expandafter\def\csname v@g5h.allruns\endcsname{$+$0.53 [0.42, 0.65] (41 tasks)}
\expandafter\def\csname v@g5h.debate.span\endcsname{0.48 to 0.30}
\expandafter\def\csname v@g5h.pool_evpc.range\endcsname{0.95--0.97}
\expandafter\def\csname v@g5h.full_iter.range\endcsname{0.91--0.92}
\expandafter\def\csname v@g5h.tok.savings.range\endcsname{32--49\%}
\expandafter\def\csname v@g5h.usd.total\endcsname{51}
\expandafter\def\csname v@g5h.calls.total\endcsname{211{,}636}
\expandafter\def\csname v@g5h.pilot.diff\endcsname{$+$0.46 [0.31, 0.62]}
\expandafter\def\csname v@g5h.pred.G5H\endcsname{8 of 8}
\expandafter\def\csname v@g5h.pred.G5HT\endcsname{17 of 17}
\expandafter\def\csname v@D.hb.gemma.balanced.16\endcsname{0.43}
\expandafter\def\csname v@D.hb.gemma.debate.16\endcsname{0.34}
\expandafter\def\csname v@D.hb.gemma.debate_evpc.16\endcsname{0.94}
\expandafter\def\csname v@D.hb.gemma.full.16\endcsname{0.94}
\expandafter\def\csname v@D.hb.gemma.full_iter.16\endcsname{0.97}
\expandafter\def\csname v@D.hb.gemma.pool_evpc.16\endcsname{0.96}
\expandafter\def\csname v@D.hb.gemma.solo.16\endcsname{0.01}
\expandafter\def\csname v@D.hb.gemma.balanced.32\endcsname{0.41}
\expandafter\def\csname v@D.hb.gemma.debate.32\endcsname{0.28}
\expandafter\def\csname v@D.hb.gemma.debate_evpc.32\endcsname{0.88}
\expandafter\def\csname v@D.hb.gemma.full.32\endcsname{0.91}
\expandafter\def\csname v@D.hb.gemma.full_iter.32\endcsname{0.95}
\expandafter\def\csname v@D.hb.gemma.pool_evpc.32\endcsname{0.96}
\expandafter\def\csname v@D.hb.gemma.solo.32\endcsname{0.02}
\expandafter\def\csname v@D.hb.gemma.balanced.4\endcsname{0.56}
\expandafter\def\csname v@D.hb.gemma.debate.4\endcsname{0.55}
\expandafter\def\csname v@D.hb.gemma.debate_evpc.4\endcsname{0.96}
\expandafter\def\csname v@D.hb.gemma.full.4\endcsname{0.91}
\expandafter\def\csname v@D.hb.gemma.full_iter.4\endcsname{0.91}
\expandafter\def\csname v@D.hb.gemma.pool_evpc.4\endcsname{0.95}
\expandafter\def\csname v@D.hb.gemma.solo.4\endcsname{0.09}
\expandafter\def\csname v@D.hb.gemma.balanced.8\endcsname{0.50}
\expandafter\def\csname v@D.hb.gemma.debate.8\endcsname{0.39}
\expandafter\def\csname v@D.hb.gemma.debate_evpc.8\endcsname{0.93}
\expandafter\def\csname v@D.hb.gemma.full.8\endcsname{0.92}
\expandafter\def\csname v@D.hb.gemma.full_iter.8\endcsname{0.94}
\expandafter\def\csname v@D.hb.gemma.pool_evpc.8\endcsname{0.93}
\expandafter\def\csname v@D.hb.gemma.solo.8\endcsname{0.01}
\expandafter\def\csname v@D.pz.gemma.balanced.16\endcsname{0.32}
\expandafter\def\csname v@D.pz.gemma.debate.16\endcsname{0.25}
\expandafter\def\csname v@D.pz.gemma.debate_evpc.16\endcsname{0.82}
\expandafter\def\csname v@D.pz.gemma.full.16\endcsname{1.00}
\expandafter\def\csname v@D.pz.gemma.full_iter.16\endcsname{1.00}
\expandafter\def\csname v@D.pz.gemma.pool_evpc.16\endcsname{0.99}
\expandafter\def\csname v@D.pz.gemma.solo.16\endcsname{0.04}
\expandafter\def\csname v@D.pz.gemma.balanced.32\endcsname{0.29}
\expandafter\def\csname v@D.pz.gemma.debate.32\endcsname{0.21}
\expandafter\def\csname v@D.pz.gemma.debate_evpc.32\endcsname{0.25}
\expandafter\def\csname v@D.pz.gemma.full.32\endcsname{1.00}
\expandafter\def\csname v@D.pz.gemma.full_iter.32\endcsname{1.00}
\expandafter\def\csname v@D.pz.gemma.pool_evpc.32\endcsname{0.99}
\expandafter\def\csname v@D.pz.gemma.solo.32\endcsname{0.04}
\expandafter\def\csname v@D.pz.gemma.balanced.4\endcsname{0.41}
\expandafter\def\csname v@D.pz.gemma.debate.4\endcsname{0.37}
\expandafter\def\csname v@D.pz.gemma.debate_evpc.4\endcsname{1.00}
\expandafter\def\csname v@D.pz.gemma.full.4\endcsname{1.00}
\expandafter\def\csname v@D.pz.gemma.full_iter.4\endcsname{1.00}
\expandafter\def\csname v@D.pz.gemma.pool_evpc.4\endcsname{0.99}
\expandafter\def\csname v@D.pz.gemma.solo.4\endcsname{0.02}
\expandafter\def\csname v@D.pz.gemma.balanced.8\endcsname{0.34}
\expandafter\def\csname v@D.pz.gemma.debate.8\endcsname{0.31}
\expandafter\def\csname v@D.pz.gemma.debate_evpc.8\endcsname{1.00}
\expandafter\def\csname v@D.pz.gemma.full.8\endcsname{1.00}
\expandafter\def\csname v@D.pz.gemma.full_iter.8\endcsname{1.00}
\expandafter\def\csname v@D.pz.gemma.pool_evpc.8\endcsname{0.99}
\expandafter\def\csname v@D.pz.gemma.solo.8\endcsname{0.03}
\expandafter\def\csname v@D.hb.mini.balanced.16\endcsname{0.43}
\expandafter\def\csname v@D.hb.mini.debate.16\endcsname{0.36}
\expandafter\def\csname v@D.hb.mini.debate_evpc.16\endcsname{0.79}
\expandafter\def\csname v@D.hb.mini.full.16\endcsname{0.74}
\expandafter\def\csname v@D.hb.mini.full_iter.16\endcsname{0.70}
\expandafter\def\csname v@D.hb.mini.pool_evpc.16\endcsname{0.86}
\expandafter\def\csname v@D.hb.mini.solo.16\endcsname{0.00}
\expandafter\def\csname v@D.hb.mini.balanced.32\endcsname{0.33}
\expandafter\def\csname v@D.hb.mini.debate.32\endcsname{0.37}
\expandafter\def\csname v@D.hb.mini.debate_evpc.32\endcsname{0.55}
\expandafter\def\csname v@D.hb.mini.full.32\endcsname{0.74}
\expandafter\def\csname v@D.hb.mini.full_iter.32\endcsname{0.80}
\expandafter\def\csname v@D.hb.mini.pool_evpc.32\endcsname{0.87}
\expandafter\def\csname v@D.hb.mini.solo.32\endcsname{0.01}
\expandafter\def\csname v@D.hb.mini.balanced.4\endcsname{0.45}
\expandafter\def\csname v@D.hb.mini.debate.4\endcsname{0.43}
\expandafter\def\csname v@D.hb.mini.debate_evpc.4\endcsname{0.84}
\expandafter\def\csname v@D.hb.mini.full.4\endcsname{0.69}
\expandafter\def\csname v@D.hb.mini.full_iter.4\endcsname{0.66}
\expandafter\def\csname v@D.hb.mini.pool_evpc.4\endcsname{0.84}
\expandafter\def\csname v@D.hb.mini.solo.4\endcsname{0.09}
\expandafter\def\csname v@D.hb.mini.balanced.8\endcsname{0.49}
\expandafter\def\csname v@D.hb.mini.debate.8\endcsname{0.40}
\expandafter\def\csname v@D.hb.mini.debate_evpc.8\endcsname{0.84}
\expandafter\def\csname v@D.hb.mini.full.8\endcsname{0.74}
\expandafter\def\csname v@D.hb.mini.full_iter.8\endcsname{0.67}
\expandafter\def\csname v@D.hb.mini.pool_evpc.8\endcsname{0.82}
\expandafter\def\csname v@D.hb.mini.solo.8\endcsname{0.02}
\expandafter\def\csname v@D.pz.mini.balanced.16\endcsname{0.09}
\expandafter\def\csname v@D.pz.mini.debate.16\endcsname{0.09}
\expandafter\def\csname v@D.pz.mini.debate_evpc.16\endcsname{0.67}
\expandafter\def\csname v@D.pz.mini.full.16\endcsname{0.99}
\expandafter\def\csname v@D.pz.mini.full_iter.16\endcsname{0.98}
\expandafter\def\csname v@D.pz.mini.pool_evpc.16\endcsname{0.96}
\expandafter\def\csname v@D.pz.mini.solo.16\endcsname{0.00}
\expandafter\def\csname v@D.pz.mini.balanced.32\endcsname{0.11}
\expandafter\def\csname v@D.pz.mini.debate.32\endcsname{0.07}
\expandafter\def\csname v@D.pz.mini.debate_evpc.32\endcsname{0.04}
\expandafter\def\csname v@D.pz.mini.full.32\endcsname{1.00}
\expandafter\def\csname v@D.pz.mini.full_iter.32\endcsname{0.99}
\expandafter\def\csname v@D.pz.mini.pool_evpc.32\endcsname{0.98}
\expandafter\def\csname v@D.pz.mini.solo.32\endcsname{0.00}
\expandafter\def\csname v@D.pz.mini.balanced.4\endcsname{0.16}
\expandafter\def\csname v@D.pz.mini.debate.4\endcsname{0.11}
\expandafter\def\csname v@D.pz.mini.debate_evpc.4\endcsname{0.93}
\expandafter\def\csname v@D.pz.mini.full.4\endcsname{0.95}
\expandafter\def\csname v@D.pz.mini.full_iter.4\endcsname{0.94}
\expandafter\def\csname v@D.pz.mini.pool_evpc.4\endcsname{0.95}
\expandafter\def\csname v@D.pz.mini.solo.4\endcsname{0.01}
\expandafter\def\csname v@D.pz.mini.balanced.8\endcsname{0.12}
\expandafter\def\csname v@D.pz.mini.debate.8\endcsname{0.13}
\expandafter\def\csname v@D.pz.mini.debate_evpc.8\endcsname{0.93}
\expandafter\def\csname v@D.pz.mini.full.8\endcsname{0.99}
\expandafter\def\csname v@D.pz.mini.full_iter.8\endcsname{0.96}
\expandafter\def\csname v@D.pz.mini.pool_evpc.8\endcsname{0.95}
\expandafter\def\csname v@D.pz.mini.solo.8\endcsname{0.00}
\expandafter\def\csname v@D.hb.olmo.balanced.16\endcsname{0.20}
\expandafter\def\csname v@D.hb.olmo.countblind.16\endcsname{0.27}
\expandafter\def\csname v@D.hb.olmo.debate.16\endcsname{0.16}
\expandafter\def\csname v@D.hb.olmo.debate_anon.16\endcsname{0.13}
\expandafter\def\csname v@D.hb.olmo.debate_evpc.16\endcsname{0.71}
\expandafter\def\csname v@D.hb.olmo.exchange.16\endcsname{0.31}
\expandafter\def\csname v@D.hb.olmo.full.16\endcsname{0.48}
\expandafter\def\csname v@D.hb.olmo.full_iter.16\endcsname{0.67}
\expandafter\def\csname v@D.hb.olmo.pool_evpc.16\endcsname{0.79}
\expandafter\def\csname v@D.hb.olmo.shareall.16\endcsname{0.05}
\expandafter\def\csname v@D.hb.olmo.solo.16\endcsname{0.00}
\expandafter\def\csname v@D.hb.olmo.balanced.32\endcsname{0.22}
\expandafter\def\csname v@D.hb.olmo.countblind.32\endcsname{0.31}
\expandafter\def\csname v@D.hb.olmo.debate.32\endcsname{0.10}
\expandafter\def\csname v@D.hb.olmo.debate_anon.32\endcsname{0.12}
\expandafter\def\csname v@D.hb.olmo.debate_evpc.32\endcsname{0.43}
\expandafter\def\csname v@D.hb.olmo.exchange.32\endcsname{0.21}
\expandafter\def\csname v@D.hb.olmo.full.32\endcsname{0.57}
\expandafter\def\csname v@D.hb.olmo.full_iter.32\endcsname{0.70}
\expandafter\def\csname v@D.hb.olmo.pool_evpc.32\endcsname{0.78}
\expandafter\def\csname v@D.hb.olmo.shareall.32\endcsname{0.08}
\expandafter\def\csname v@D.hb.olmo.solo.32\endcsname{0.00}
\expandafter\def\csname v@D.hb.olmo.balanced.4\endcsname{0.28}
\expandafter\def\csname v@D.hb.olmo.countblind.4\endcsname{0.29}
\expandafter\def\csname v@D.hb.olmo.debate.4\endcsname{0.23}
\expandafter\def\csname v@D.hb.olmo.debate_anon.4\endcsname{0.24}
\expandafter\def\csname v@D.hb.olmo.debate_evpc.4\endcsname{0.75}
\expandafter\def\csname v@D.hb.olmo.exchange.4\endcsname{0.44}
\expandafter\def\csname v@D.hb.olmo.full.4\endcsname{0.52}
\expandafter\def\csname v@D.hb.olmo.full_iter.4\endcsname{0.63}
\expandafter\def\csname v@D.hb.olmo.pool_evpc.4\endcsname{0.74}
\expandafter\def\csname v@D.hb.olmo.shareall.4\endcsname{0.18}
\expandafter\def\csname v@D.hb.olmo.solo.4\endcsname{0.05}
\expandafter\def\csname v@D.hb.olmo.debate.64\endcsname{0.08}
\expandafter\def\csname v@D.hb.olmo.full.64\endcsname{0.60}
\expandafter\def\csname v@D.hb.olmo.full_iter.64\endcsname{0.70}
\expandafter\def\csname v@D.hb.olmo.pool_evpc.64\endcsname{0.80}
\expandafter\def\csname v@D.hb.olmo.solo.64\endcsname{0.00}
\expandafter\def\csname v@D.hb.olmo.balanced.8\endcsname{0.29}
\expandafter\def\csname v@D.hb.olmo.countblind.8\endcsname{0.23}
\expandafter\def\csname v@D.hb.olmo.debate.8\endcsname{0.16}
\expandafter\def\csname v@D.hb.olmo.debate_anon.8\endcsname{0.15}
\expandafter\def\csname v@D.hb.olmo.debate_evpc.8\endcsname{0.76}
\expandafter\def\csname v@D.hb.olmo.exchange.8\endcsname{0.39}
\expandafter\def\csname v@D.hb.olmo.full.8\endcsname{0.50}
\expandafter\def\csname v@D.hb.olmo.full_iter.8\endcsname{0.65}
\expandafter\def\csname v@D.hb.olmo.pool_evpc.8\endcsname{0.79}
\expandafter\def\csname v@D.hb.olmo.shareall.8\endcsname{0.10}
\expandafter\def\csname v@D.hb.olmo.solo.8\endcsname{0.00}
\expandafter\def\csname v@D.mq.olmo.balanced.16\endcsname{0.10}
\expandafter\def\csname v@D.mq.olmo.countblind.16\endcsname{0.09}
\expandafter\def\csname v@D.mq.olmo.debate.16\endcsname{0.10}
\expandafter\def\csname v@D.mq.olmo.debate_evpc.16\endcsname{0.45}
\expandafter\def\csname v@D.mq.olmo.full.16\endcsname{0.75}
\expandafter\def\csname v@D.mq.olmo.full_iter.16\endcsname{0.76}
\expandafter\def\csname v@D.mq.olmo.pool_evpc.16\endcsname{0.56}
\expandafter\def\csname v@D.mq.olmo.solo.16\endcsname{0.02}
\expandafter\def\csname v@D.mq.olmo.balanced.4\endcsname{0.19}
\expandafter\def\csname v@D.mq.olmo.countblind.4\endcsname{0.16}
\expandafter\def\csname v@D.mq.olmo.debate.4\endcsname{0.17}
\expandafter\def\csname v@D.mq.olmo.debate_evpc.4\endcsname{0.50}
\expandafter\def\csname v@D.mq.olmo.full.4\endcsname{0.60}
\expandafter\def\csname v@D.mq.olmo.full_iter.4\endcsname{0.66}
\expandafter\def\csname v@D.mq.olmo.pool_evpc.4\endcsname{0.49}
\expandafter\def\csname v@D.mq.olmo.solo.4\endcsname{0.07}
\expandafter\def\csname v@D.mq.olmo.balanced.8\endcsname{0.13}
\expandafter\def\csname v@D.mq.olmo.countblind.8\endcsname{0.14}
\expandafter\def\csname v@D.mq.olmo.debate.8\endcsname{0.14}
\expandafter\def\csname v@D.mq.olmo.debate_evpc.8\endcsname{0.50}
\expandafter\def\csname v@D.mq.olmo.full.8\endcsname{0.68}
\expandafter\def\csname v@D.mq.olmo.full_iter.8\endcsname{0.75}
\expandafter\def\csname v@D.mq.olmo.pool_evpc.8\endcsname{0.52}
\expandafter\def\csname v@D.mq.olmo.solo.8\endcsname{0.04}
\expandafter\def\csname v@D.pz.olmo.balanced.16\endcsname{0.16}
\expandafter\def\csname v@D.pz.olmo.countblind.16\endcsname{0.17}
\expandafter\def\csname v@D.pz.olmo.debate.16\endcsname{0.10}
\expandafter\def\csname v@D.pz.olmo.debate_evpc.16\endcsname{0.52}
\expandafter\def\csname v@D.pz.olmo.full.16\endcsname{0.94}
\expandafter\def\csname v@D.pz.olmo.full_iter.16\endcsname{0.93}
\expandafter\def\csname v@D.pz.olmo.pool_evpc.16\endcsname{0.81}
\expandafter\def\csname v@D.pz.olmo.solo.16\endcsname{0.03}
\expandafter\def\csname v@D.pz.olmo.balanced.32\endcsname{0.20}
\expandafter\def\csname v@D.pz.olmo.countblind.32\endcsname{0.15}
\expandafter\def\csname v@D.pz.olmo.debate.32\endcsname{0.10}
\expandafter\def\csname v@D.pz.olmo.debate_evpc.32\endcsname{0.11}
\expandafter\def\csname v@D.pz.olmo.full.32\endcsname{0.96}
\expandafter\def\csname v@D.pz.olmo.full_iter.32\endcsname{0.93}
\expandafter\def\csname v@D.pz.olmo.pool_evpc.32\endcsname{0.80}
\expandafter\def\csname v@D.pz.olmo.solo.32\endcsname{0.02}
\expandafter\def\csname v@D.pz.olmo.balanced.4\endcsname{0.16}
\expandafter\def\csname v@D.pz.olmo.countblind.4\endcsname{0.18}
\expandafter\def\csname v@D.pz.olmo.debate.4\endcsname{0.15}
\expandafter\def\csname v@D.pz.olmo.debate_evpc.4\endcsname{0.80}
\expandafter\def\csname v@D.pz.olmo.full.4\endcsname{0.91}
\expandafter\def\csname v@D.pz.olmo.full_iter.4\endcsname{0.85}
\expandafter\def\csname v@D.pz.olmo.pool_evpc.4\endcsname{0.79}
\expandafter\def\csname v@D.pz.olmo.solo.4\endcsname{0.05}
\expandafter\def\csname v@D.pz.olmo.balanced.8\endcsname{0.17}
\expandafter\def\csname v@D.pz.olmo.countblind.8\endcsname{0.18}
\expandafter\def\csname v@D.pz.olmo.debate.8\endcsname{0.14}
\expandafter\def\csname v@D.pz.olmo.debate_evpc.8\endcsname{0.80}
\expandafter\def\csname v@D.pz.olmo.full.8\endcsname{0.90}
\expandafter\def\csname v@D.pz.olmo.full_iter.8\endcsname{0.89}
\expandafter\def\csname v@D.pz.olmo.pool_evpc.8\endcsname{0.82}
\expandafter\def\csname v@D.pz.olmo.solo.8\endcsname{0.01}
\expandafter\def\csname v@D.hb.q32.debate.16\endcsname{0.16}
\expandafter\def\csname v@D.hb.q32.full.16\endcsname{0.83}
\expandafter\def\csname v@D.hb.q32.full_iter.16\endcsname{0.92}
\expandafter\def\csname v@D.hb.q32.pool_evpc.16\endcsname{0.89}
\expandafter\def\csname v@D.hb.q32.solo.16\endcsname{0.03}
\expandafter\def\csname v@D.hb.q32.debate.32\endcsname{0.13}
\expandafter\def\csname v@D.hb.q32.full.32\endcsname{0.85}
\expandafter\def\csname v@D.hb.q32.full_iter.32\endcsname{0.92}
\expandafter\def\csname v@D.hb.q32.pool_evpc.32\endcsname{0.89}
\expandafter\def\csname v@D.hb.q32.solo.32\endcsname{0.03}
\expandafter\def\csname v@D.hb.q32.debate.4\endcsname{0.31}
\expandafter\def\csname v@D.hb.q32.full.4\endcsname{0.83}
\expandafter\def\csname v@D.hb.q32.full_iter.4\endcsname{0.89}
\expandafter\def\csname v@D.hb.q32.pool_evpc.4\endcsname{0.91}
\expandafter\def\csname v@D.hb.q32.solo.4\endcsname{0.06}
\expandafter\def\csname v@D.hb.q32.debate.8\endcsname{0.20}
\expandafter\def\csname v@D.hb.q32.full.8\endcsname{0.83}
\expandafter\def\csname v@D.hb.q32.full_iter.8\endcsname{0.91}
\expandafter\def\csname v@D.hb.q32.pool_evpc.8\endcsname{0.90}
\expandafter\def\csname v@D.hb.q32.solo.8\endcsname{0.02}
\expandafter\def\csname v@D.hb.qwen.balanced.16\endcsname{0.13}
\expandafter\def\csname v@D.hb.qwen.countblind.16\endcsname{0.19}
\expandafter\def\csname v@D.hb.qwen.debate.16\endcsname{0.06}
\expandafter\def\csname v@D.hb.qwen.debate_anon.16\endcsname{0.08}
\expandafter\def\csname v@D.hb.qwen.debate_evpc.16\endcsname{0.82}
\expandafter\def\csname v@D.hb.qwen.exchange.16\endcsname{0.21}
\expandafter\def\csname v@D.hb.qwen.full.16\endcsname{0.61}
\expandafter\def\csname v@D.hb.qwen.full_iter.16\endcsname{0.74}
\expandafter\def\csname v@D.hb.qwen.pool_evpc.16\endcsname{0.86}
\expandafter\def\csname v@D.hb.qwen.shareall.16\endcsname{0.06}
\expandafter\def\csname v@D.hb.qwen.solo.16\endcsname{0.03}
\expandafter\def\csname v@D.hb.qwen.balanced.32\endcsname{0.13}
\expandafter\def\csname v@D.hb.qwen.countblind.32\endcsname{0.18}
\expandafter\def\csname v@D.hb.qwen.debate.32\endcsname{0.07}
\expandafter\def\csname v@D.hb.qwen.debate_anon.32\endcsname{0.04}
\expandafter\def\csname v@D.hb.qwen.debate_evpc.32\endcsname{0.75}
\expandafter\def\csname v@D.hb.qwen.exchange.32\endcsname{0.12}
\expandafter\def\csname v@D.hb.qwen.full.32\endcsname{0.60}
\expandafter\def\csname v@D.hb.qwen.full_iter.32\endcsname{0.73}
\expandafter\def\csname v@D.hb.qwen.pool_evpc.32\endcsname{0.84}
\expandafter\def\csname v@D.hb.qwen.shareall.32\endcsname{0.03}
\expandafter\def\csname v@D.hb.qwen.solo.32\endcsname{0.03}
\expandafter\def\csname v@D.hb.qwen.balanced.4\endcsname{0.18}
\expandafter\def\csname v@D.hb.qwen.countblind.4\endcsname{0.23}
\expandafter\def\csname v@D.hb.qwen.debate.4\endcsname{0.11}
\expandafter\def\csname v@D.hb.qwen.debate_anon.4\endcsname{0.17}
\expandafter\def\csname v@D.hb.qwen.debate_evpc.4\endcsname{0.81}
\expandafter\def\csname v@D.hb.qwen.exchange.4\endcsname{0.34}
\expandafter\def\csname v@D.hb.qwen.full.4\endcsname{0.63}
\expandafter\def\csname v@D.hb.qwen.full_iter.4\endcsname{0.70}
\expandafter\def\csname v@D.hb.qwen.pool_evpc.4\endcsname{0.80}
\expandafter\def\csname v@D.hb.qwen.shareall.4\endcsname{0.20}
\expandafter\def\csname v@D.hb.qwen.solo.4\endcsname{0.05}
\expandafter\def\csname v@D.hb.qwen.debate.64\endcsname{0.08}
\expandafter\def\csname v@D.hb.qwen.full.64\endcsname{0.60}
\expandafter\def\csname v@D.hb.qwen.full_iter.64\endcsname{0.73}
\expandafter\def\csname v@D.hb.qwen.pool_evpc.64\endcsname{0.84}
\expandafter\def\csname v@D.hb.qwen.solo.64\endcsname{0.03}
\expandafter\def\csname v@D.hb.qwen.balanced.8\endcsname{0.15}
\expandafter\def\csname v@D.hb.qwen.countblind.8\endcsname{0.19}
\expandafter\def\csname v@D.hb.qwen.debate.8\endcsname{0.06}
\expandafter\def\csname v@D.hb.qwen.debate_anon.8\endcsname{0.07}
\expandafter\def\csname v@D.hb.qwen.debate_evpc.8\endcsname{0.85}
\expandafter\def\csname v@D.hb.qwen.exchange.8\endcsname{0.24}
\expandafter\def\csname v@D.hb.qwen.full.8\endcsname{0.59}
\expandafter\def\csname v@D.hb.qwen.full_iter.8\endcsname{0.71}
\expandafter\def\csname v@D.hb.qwen.pool_evpc.8\endcsname{0.82}
\expandafter\def\csname v@D.hb.qwen.shareall.8\endcsname{0.07}
\expandafter\def\csname v@D.hb.qwen.solo.8\endcsname{0.02}
\expandafter\def\csname v@D.mq.qwen.balanced.16\endcsname{0.12}
\expandafter\def\csname v@D.mq.qwen.countblind.16\endcsname{0.14}
\expandafter\def\csname v@D.mq.qwen.debate.16\endcsname{0.11}
\expandafter\def\csname v@D.mq.qwen.debate_evpc.16\endcsname{0.46}
\expandafter\def\csname v@D.mq.qwen.full.16\endcsname{0.88}
\expandafter\def\csname v@D.mq.qwen.full_iter.16\endcsname{0.89}
\expandafter\def\csname v@D.mq.qwen.pool_evpc.16\endcsname{0.54}
\expandafter\def\csname v@D.mq.qwen.solo.16\endcsname{0.03}
\expandafter\def\csname v@D.mq.qwen.balanced.4\endcsname{0.20}
\expandafter\def\csname v@D.mq.qwen.countblind.4\endcsname{0.26}
\expandafter\def\csname v@D.mq.qwen.debate.4\endcsname{0.21}
\expandafter\def\csname v@D.mq.qwen.debate_evpc.4\endcsname{0.56}
\expandafter\def\csname v@D.mq.qwen.full.4\endcsname{0.84}
\expandafter\def\csname v@D.mq.qwen.full_iter.4\endcsname{0.85}
\expandafter\def\csname v@D.mq.qwen.pool_evpc.4\endcsname{0.57}
\expandafter\def\csname v@D.mq.qwen.solo.4\endcsname{0.08}
\expandafter\def\csname v@D.mq.qwen.balanced.8\endcsname{0.19}
\expandafter\def\csname v@D.mq.qwen.countblind.8\endcsname{0.20}
\expandafter\def\csname v@D.mq.qwen.debate.8\endcsname{0.13}
\expandafter\def\csname v@D.mq.qwen.debate_evpc.8\endcsname{0.56}
\expandafter\def\csname v@D.mq.qwen.full.8\endcsname{0.86}
\expandafter\def\csname v@D.mq.qwen.full_iter.8\endcsname{0.88}
\expandafter\def\csname v@D.mq.qwen.pool_evpc.8\endcsname{0.56}
\expandafter\def\csname v@D.mq.qwen.solo.8\endcsname{0.04}
\expandafter\def\csname v@D.pz.qwen.balanced.16\endcsname{0.16}
\expandafter\def\csname v@D.pz.qwen.countblind.16\endcsname{0.16}
\expandafter\def\csname v@D.pz.qwen.debate.16\endcsname{0.11}
\expandafter\def\csname v@D.pz.qwen.debate_evpc.16\endcsname{0.69}
\expandafter\def\csname v@D.pz.qwen.full.16\endcsname{0.94}
\expandafter\def\csname v@D.pz.qwen.full_iter.16\endcsname{0.95}
\expandafter\def\csname v@D.pz.qwen.pool_evpc.16\endcsname{0.90}
\expandafter\def\csname v@D.pz.qwen.solo.16\endcsname{0.12}
\expandafter\def\csname v@D.pz.qwen.balanced.32\endcsname{0.17}
\expandafter\def\csname v@D.pz.qwen.countblind.32\endcsname{0.16}
\expandafter\def\csname v@D.pz.qwen.debate.32\endcsname{0.10}
\expandafter\def\csname v@D.pz.qwen.debate_evpc.32\endcsname{0.20}
\expandafter\def\csname v@D.pz.qwen.full.32\endcsname{0.96}
\expandafter\def\csname v@D.pz.qwen.full_iter.32\endcsname{0.96}
\expandafter\def\csname v@D.pz.qwen.pool_evpc.32\endcsname{0.91}
\expandafter\def\csname v@D.pz.qwen.solo.32\endcsname{0.12}
\expandafter\def\csname v@D.pz.qwen.balanced.4\endcsname{0.15}
\expandafter\def\csname v@D.pz.qwen.countblind.4\endcsname{0.21}
\expandafter\def\csname v@D.pz.qwen.debate.4\endcsname{0.17}
\expandafter\def\csname v@D.pz.qwen.debate_evpc.4\endcsname{0.85}
\expandafter\def\csname v@D.pz.qwen.full.4\endcsname{0.89}
\expandafter\def\csname v@D.pz.qwen.full_iter.4\endcsname{0.93}
\expandafter\def\csname v@D.pz.qwen.pool_evpc.4\endcsname{0.87}
\expandafter\def\csname v@D.pz.qwen.solo.4\endcsname{0.11}
\expandafter\def\csname v@D.pz.qwen.balanced.8\endcsname{0.14}
\expandafter\def\csname v@D.pz.qwen.countblind.8\endcsname{0.18}
\expandafter\def\csname v@D.pz.qwen.debate.8\endcsname{0.12}
\expandafter\def\csname v@D.pz.qwen.debate_evpc.8\endcsname{0.88}
\expandafter\def\csname v@D.pz.qwen.full.8\endcsname{0.92}
\expandafter\def\csname v@D.pz.qwen.full_iter.8\endcsname{0.94}
\expandafter\def\csname v@D.pz.qwen.pool_evpc.8\endcsname{0.88}
\expandafter\def\csname v@D.pz.qwen.solo.8\endcsname{0.12}
\expandafter\def\csname v@D.max.debate\endcsname{0.23}
\expandafter\def\csname v@debsolo.max\endcsname{0.08}
\expandafter\def\csname v@Dr.hb.qwen.pool_evpc\endcsname{0.80--0.86}
\expandafter\def\csname v@Dr.hb.qwen.debate\endcsname{0.06--0.11}
\expandafter\def\csname v@Dr.hb.qwen.debate_evpc\endcsname{0.75--0.85}
\expandafter\def\csname v@Dr.hb.qwen.full_iter\endcsname{0.70--0.74}
\expandafter\def\csname v@Dr.hb.qwen.balanced\endcsname{0.13--0.18}
\expandafter\def\csname v@items.hb.qwen\endcsname{38}
\expandafter\def\csname v@Dr.pz.qwen.pool_evpc\endcsname{0.87--0.91}
\expandafter\def\csname v@Dr.pz.qwen.debate\endcsname{0.10--0.17}
\expandafter\def\csname v@Dr.pz.qwen.debate_evpc\endcsname{0.20--0.88}
\expandafter\def\csname v@Dr.pz.qwen.full_iter\endcsname{0.93--0.96}
\expandafter\def\csname v@Dr.pz.qwen.balanced\endcsname{0.14--0.17}
\expandafter\def\csname v@items.pz.qwen\endcsname{200}
\expandafter\def\csname v@Dr.mq.qwen.pool_evpc\endcsname{0.54--0.57}
\expandafter\def\csname v@Dr.mq.qwen.debate\endcsname{0.11--0.21}
\expandafter\def\csname v@Dr.mq.qwen.debate_evpc\endcsname{0.46--0.56}
\expandafter\def\csname v@Dr.mq.qwen.full_iter\endcsname{0.85--0.89}
\expandafter\def\csname v@Dr.mq.qwen.balanced\endcsname{0.12--0.20}
\expandafter\def\csname v@items.mq.qwen\endcsname{244}
\expandafter\def\csname v@Dr.hb.olmo.pool_evpc\endcsname{0.74--0.80}
\expandafter\def\csname v@Dr.hb.olmo.debate\endcsname{0.08--0.23}
\expandafter\def\csname v@Dr.hb.olmo.debate_evpc\endcsname{0.43--0.76}
\expandafter\def\csname v@Dr.hb.olmo.full_iter\endcsname{0.63--0.70}
\expandafter\def\csname v@Dr.hb.olmo.balanced\endcsname{0.20--0.29}
\expandafter\def\csname v@items.hb.olmo\endcsname{38}
\expandafter\def\csname v@Dr.pz.olmo.pool_evpc\endcsname{0.79--0.82}
\expandafter\def\csname v@Dr.pz.olmo.debate\endcsname{0.10--0.15}
\expandafter\def\csname v@Dr.pz.olmo.debate_evpc\endcsname{0.11--0.80}
\expandafter\def\csname v@Dr.pz.olmo.full_iter\endcsname{0.85--0.93}
\expandafter\def\csname v@Dr.pz.olmo.balanced\endcsname{0.16--0.20}
\expandafter\def\csname v@items.pz.olmo\endcsname{193}
\expandafter\def\csname v@Dr.mq.olmo.pool_evpc\endcsname{0.49--0.56}
\expandafter\def\csname v@Dr.mq.olmo.debate\endcsname{0.10--0.17}
\expandafter\def\csname v@Dr.mq.olmo.debate_evpc\endcsname{0.45--0.50}
\expandafter\def\csname v@Dr.mq.olmo.full_iter\endcsname{0.66--0.76}
\expandafter\def\csname v@Dr.mq.olmo.balanced\endcsname{0.10--0.19}
\expandafter\def\csname v@items.mq.olmo\endcsname{180}
\expandafter\def\csname v@Dr.hb.mini.pool_evpc\endcsname{0.82--0.87}
\expandafter\def\csname v@Dr.hb.mini.debate\endcsname{0.36--0.43}
\expandafter\def\csname v@Dr.hb.mini.debate_evpc\endcsname{0.55--0.84}
\expandafter\def\csname v@Dr.hb.mini.full_iter\endcsname{0.66--0.80}
\expandafter\def\csname v@Dr.hb.mini.balanced\endcsname{0.33--0.49}
\expandafter\def\csname v@items.hb.mini\endcsname{35}
\expandafter\def\csname v@Dr.pz.mini.pool_evpc\endcsname{0.95--0.98}
\expandafter\def\csname v@Dr.pz.mini.debate\endcsname{0.07--0.13}
\expandafter\def\csname v@Dr.pz.mini.debate_evpc\endcsname{0.04--0.93}
\expandafter\def\csname v@Dr.pz.mini.full_iter\endcsname{0.94--0.99}
\expandafter\def\csname v@Dr.pz.mini.balanced\endcsname{0.09--0.16}
\expandafter\def\csname v@items.pz.mini\endcsname{152}
\expandafter\def\csname v@Dr.hb.gemma.pool_evpc\endcsname{0.93--0.96}
\expandafter\def\csname v@Dr.hb.gemma.debate\endcsname{0.28--0.55}
\expandafter\def\csname v@Dr.hb.gemma.debate_evpc\endcsname{0.88--0.96}
\expandafter\def\csname v@Dr.hb.gemma.full_iter\endcsname{0.91--0.97}
\expandafter\def\csname v@Dr.hb.gemma.balanced\endcsname{0.41--0.56}
\expandafter\def\csname v@items.hb.gemma\endcsname{38}
\expandafter\def\csname v@Dr.pz.gemma.pool_evpc\endcsname{0.99--0.99}
\expandafter\def\csname v@Dr.pz.gemma.debate\endcsname{0.21--0.37}
\expandafter\def\csname v@Dr.pz.gemma.debate_evpc\endcsname{0.25--1.00}
\expandafter\def\csname v@Dr.pz.gemma.full_iter\endcsname{1.00--1.00}
\expandafter\def\csname v@Dr.pz.gemma.balanced\endcsname{0.29--0.41}
\expandafter\def\csname v@items.pz.gemma\endcsname{181}
\expandafter\def\csname v@Dr.hb.q32.pool_evpc\endcsname{0.89--0.91}
\expandafter\def\csname v@Dr.hb.q32.debate\endcsname{0.13--0.31}
\expandafter\def\csname v@Dr.hb.q32.full_iter\endcsname{0.89--0.92}
\expandafter\def\csname v@items.hb.q32\endcsname{38}
\expandafter\def\csname v@items.hb.min\endcsname{37}
\expandafter\def\csname v@items.hb.max\endcsname{39}
\expandafter\def\csname v@K.hb.qwen.fi\endcsname{$+$0.12 [0.01, 0.23]}
\expandafter\def\csname v@K.hb.qwen.fi.pt\endcsname{$+$0.12}
\expandafter\def\csname v@K.pz.qwen.fi\endcsname{$-$0.06 [$-$0.08, $-$0.04]}
\expandafter\def\csname v@K.pz.qwen.fi.pt\endcsname{$-$0.06}
\expandafter\def\csname v@K.mq.qwen.fi\endcsname{$-$0.32 [$-$0.36, $-$0.27]}
\expandafter\def\csname v@K.mq.qwen.fi.pt\endcsname{$-$0.32}
\expandafter\def\csname v@K.hb.olmo.fi\endcsname{$+$0.10 [$-$0.01, 0.21]}
\expandafter\def\csname v@K.hb.olmo.fi.pt\endcsname{$+$0.10}
\expandafter\def\csname v@K.pz.olmo.fi\endcsname{$-$0.10 [$-$0.13, $-$0.07]}
\expandafter\def\csname v@K.pz.olmo.fi.pt\endcsname{$-$0.10}
\expandafter\def\csname v@K.mq.olmo.fi\endcsname{$-$0.20 [$-$0.24, $-$0.16]}
\expandafter\def\csname v@K.mq.olmo.fi.pt\endcsname{$-$0.20}
\expandafter\def\csname v@K.below.range\endcsname{$-$0.06 to $-$0.32}
\expandafter\def\csname v@tok.hb.ratio\endcsname{0.61--0.65}
\expandafter\def\csname v@tok.pz.ratio\endcsname{0.55--0.58}
\expandafter\def\csname v@tok.mq.ratio\endcsname{0.68--0.74}
\expandafter\def\csname v@tok.range\endcsname{0.55--0.74}
\expandafter\def\csname v@tok.savings.range\endcsname{10--46\%}
\expandafter\def\csname v@late.qwen.pool.pct\endcsname{93\%}
\expandafter\def\csname v@late.qwen.debate.K3\endcsname{0.00}
\expandafter\def\csname v@late.qwen.debate.K0\endcsname{0.07}
\expandafter\def\csname v@late.qwen.pool.K0\endcsname{0.75}
\expandafter\def\csname v@late.qwen.pool.K3\endcsname{0.69}
\expandafter\def\csname v@late.olmo.pool.pct\endcsname{89\%}
\expandafter\def\csname v@late.olmo.debate.K3\endcsname{0.06}
\expandafter\def\csname v@late.olmo.debate.K0\endcsname{0.18}
\expandafter\def\csname v@late.olmo.pool.K0\endcsname{0.68}
\expandafter\def\csname v@late.olmo.pool.K3\endcsname{0.61}
\expandafter\def\csname v@q32gap.min\endcsname{0.49}
\expandafter\def\csname v@D16.range.debate\endcsname{0.06--0.16}
\expandafter\def\csname v@D16.range.balanced\endcsname{0.10--0.20}
\expandafter\def\csname v@D16.range.pool_evpc\endcsname{0.54--0.90}
\expandafter\def\csname v@D16.range.debate_evpc\endcsname{0.45--0.82}
\expandafter\def\csname v@D16.range.full_iter\endcsname{0.67--0.95}
\expandafter\def\csname v@D16.range.solo\endcsname{0.00--0.12}
\expandafter\def\csname v@q32gap.max\endcsname{0.71}
\expandafter\def\csname v@pred.T1.in\endcsname{140}
\expandafter\def\csname v@pred.T1.n\endcsname{208}
\expandafter\def\csname v@pred.T1.pct\endcsname{67\%}
\expandafter\def\csname v@pred.T2.in\endcsname{92}
\expandafter\def\csname v@pred.T2.n\endcsname{128}
\expandafter\def\csname v@pred.T2.pct\endcsname{72\%}
\expandafter\def\csname v@pred.T3.in\endcsname{85}
\expandafter\def\csname v@pred.T3.n\endcsname{96}
\expandafter\def\csname v@pred.T3.pct\endcsname{89\%}
\expandafter\def\csname v@pred.Texp.in\endcsname{64}
\expandafter\def\csname v@pred.Texp.n\endcsname{76}
\expandafter\def\csname v@pred.Texp.pct\endcsname{84\%}
\expandafter\def\csname v@pred.Texp.reach.in\endcsname{38}
\expandafter\def\csname v@pred.Texp.reach.n\endcsname{38}
\expandafter\def\csname v@pred.Texp.reach.pct\endcsname{100\%}
\expandafter\def\csname v@pred.Texp.A.in\endcsname{26}
\expandafter\def\csname v@pred.Texp.A.n\endcsname{38}
\expandafter\def\csname v@pred.Texp.A.pct\endcsname{68\%}
\expandafter\def\csname v@pred.Texp.rho.n\endcsname{19}
\expandafter\def\csname v@pred.Texp.rho.maxdiff\endcsname{0.023}
\expandafter\def\csname v@pred.T.in\endcsname{317}
\expandafter\def\csname v@pred.T.n\endcsname{432}
\expandafter\def\csname v@pred.G1.in\endcsname{211}
\expandafter\def\csname v@pred.G1.n\endcsname{236}
\expandafter\def\csname v@pred.G2.in\endcsname{38}
\expandafter\def\csname v@pred.G2.n\endcsname{48}
\expandafter\def\csname v@pred.G3.in\endcsname{24}
\expandafter\def\csname v@pred.G3.n\endcsname{24}
\expandafter\def\csname v@pred.G4.in\endcsname{224}
\expandafter\def\csname v@pred.G4.n\endcsname{244}
\expandafter\def\csname v@pred.H1.in\endcsname{125}
\expandafter\def\csname v@pred.H1.n\endcsname{136}
\expandafter\def\csname v@pred.H1d.in\endcsname{8}
\expandafter\def\csname v@pred.H1d.n\endcsname{8}
\expandafter\def\csname v@pred.H2.in\endcsname{76}
\expandafter\def\csname v@pred.H2.n\endcsname{80}
\expandafter\def\csname v@pred.H3.in\endcsname{24}
\expandafter\def\csname v@pred.H3.n\endcsname{36}
\expandafter\def\csname v@pred.H4.in\endcsname{4}
\expandafter\def\csname v@pred.H4.n\endcsname{4}
\expandafter\def\csname v@pred.H5.in\endcsname{110}
\expandafter\def\csname v@pred.H5.n\endcsname{110}
\expandafter\def\csname v@pred.I1.in\endcsname{20}
\expandafter\def\csname v@pred.I1.n\endcsname{24}
\expandafter\def\csname v@pred.I2.in\endcsname{12}
\expandafter\def\csname v@pred.I2.n\endcsname{12}
\expandafter\def\csname v@pred.I3.in\endcsname{16}
\expandafter\def\csname v@pred.I3.n\endcsname{16}
\expandafter\def\csname v@pred.I4.in\endcsname{16}
\expandafter\def\csname v@pred.I4.n\endcsname{16}
\expandafter\def\csname v@sel.ranked\endcsname{0.795}
\expandafter\def\csname v@sel.pool\endcsname{0.787}
\expandafter\def\csname v@gain.debate.range\endcsname{0.83--0.97}
\expandafter\def\csname v@gain.matched.range\endcsname{0.28--0.75}
\expandafter\def\csname v@gain.pool.range\endcsname{0.50--0.73}
\expandafter\def\csname v@gain.debate.inc1\endcsname{4}
\expandafter\def\csname v@gain.matched.below1\endcsname{6}
\expandafter\def\csname v@gain.matched.lower\endcsname{6}
\expandafter\def\csname v@gain.pool.below1\endcsname{6}
\expandafter\def\csname v@debate.drift.max\endcsname{0.15}
\expandafter\def\csname v@gain.full.range\endcsname{0.57--0.86}
\expandafter\def\csname v@sw.qwen.debate.4\endcsname{0.11}
\expandafter\def\csname v@sw.qwen.debate.8\endcsname{0.06}
\expandafter\def\csname v@sw.qwen.debate.16\endcsname{0.05}
\expandafter\def\csname v@sw.qwen.debate.32\endcsname{0.04}
\expandafter\def\csname v@swaway.qwen.debate\endcsname{0.00--0.02}
\expandafter\def\csname v@sw.qwen.matched.4\endcsname{0.19}
\expandafter\def\csname v@sw.qwen.matched.8\endcsname{0.19}
\expandafter\def\csname v@sw.qwen.matched.16\endcsname{0.20}
\expandafter\def\csname v@sw.qwen.matched.32\endcsname{0.19}
\expandafter\def\csname v@swaway.qwen.matched\endcsname{0.02--0.05}
\expandafter\def\csname v@sw.olmo.debate.4\endcsname{0.35}
\expandafter\def\csname v@sw.olmo.debate.8\endcsname{0.22}
\expandafter\def\csname v@sw.olmo.debate.16\endcsname{0.15}
\expandafter\def\csname v@sw.olmo.debate.32\endcsname{0.11}
\expandafter\def\csname v@swaway.olmo.debate\endcsname{0.00--0.09}
\expandafter\def\csname v@sw.olmo.matched.4\endcsname{0.58}
\expandafter\def\csname v@sw.olmo.matched.8\endcsname{0.63}
\expandafter\def\csname v@sw.olmo.matched.16\endcsname{0.74}
\expandafter\def\csname v@sw.olmo.matched.32\endcsname{0.71}
\expandafter\def\csname v@swaway.olmo.matched\endcsname{0.18--0.28}
\expandafter\def\csname v@swto.hb.qwen.debate.16\endcsname{0.04}
\expandafter\def\csname v@swaway.hb.qwen.debate.16\endcsname{0.01}
\expandafter\def\csname v@swto.hb.qwen.balanced.16\endcsname{0.15}
\expandafter\def\csname v@swaway.hb.qwen.balanced.16\endcsname{0.05}
\expandafter\def\csname v@swto.hb.qwen.pool_evpc.16\endcsname{0.82}
\expandafter\def\csname v@swaway.hb.qwen.pool_evpc.16\endcsname{0.06}
\expandafter\def\csname v@swto.pz.qwen.debate.16\endcsname{0.04}
\expandafter\def\csname v@swaway.pz.qwen.debate.16\endcsname{0.04}
\expandafter\def\csname v@swto.pz.qwen.balanced.16\endcsname{0.11}
\expandafter\def\csname v@swaway.pz.qwen.balanced.16\endcsname{0.06}
\expandafter\def\csname v@swto.pz.qwen.pool_evpc.16\endcsname{0.83}
\expandafter\def\csname v@swaway.pz.qwen.pool_evpc.16\endcsname{0.04}
\expandafter\def\csname v@swto.mq.qwen.debate.16\endcsname{0.11}
\expandafter\def\csname v@swaway.mq.qwen.debate.16\endcsname{0.01}
\expandafter\def\csname v@swto.mq.qwen.balanced.16\endcsname{0.14}
\expandafter\def\csname v@swaway.mq.qwen.balanced.16\endcsname{0.04}
\expandafter\def\csname v@swto.mq.qwen.pool_evpc.16\endcsname{0.54}
\expandafter\def\csname v@swaway.mq.qwen.pool_evpc.16\endcsname{0.03}
\expandafter\def\csname v@swto.hb.olmo.debate.16\endcsname{0.15}
\expandafter\def\csname v@swaway.hb.olmo.debate.16\endcsname{0.00}
\expandafter\def\csname v@swto.hb.olmo.balanced.16\endcsname{0.46}
\expandafter\def\csname v@swaway.hb.olmo.balanced.16\endcsname{0.28}
\expandafter\def\csname v@swto.hb.olmo.pool_evpc.16\endcsname{0.80}
\expandafter\def\csname v@swaway.hb.olmo.pool_evpc.16\endcsname{0.07}
\expandafter\def\csname v@swto.pz.olmo.debate.16\endcsname{0.08}
\expandafter\def\csname v@swaway.pz.olmo.debate.16\endcsname{0.03}
\expandafter\def\csname v@swto.pz.olmo.balanced.16\endcsname{0.33}
\expandafter\def\csname v@swaway.pz.olmo.balanced.16\endcsname{0.23}
\expandafter\def\csname v@swto.pz.olmo.pool_evpc.16\endcsname{0.78}
\expandafter\def\csname v@swaway.pz.olmo.pool_evpc.16\endcsname{0.04}
\expandafter\def\csname v@swto.mq.olmo.debate.16\endcsname{0.12}
\expandafter\def\csname v@swaway.mq.olmo.debate.16\endcsname{0.04}
\expandafter\def\csname v@swto.mq.olmo.balanced.16\endcsname{0.18}
\expandafter\def\csname v@swaway.mq.olmo.balanced.16\endcsname{0.11}
\expandafter\def\csname v@swto.mq.olmo.pool_evpc.16\endcsname{0.61}
\expandafter\def\csname v@swaway.mq.olmo.pool_evpc.16\endcsname{0.10}
\expandafter\def\csname v@swto.range.debate\endcsname{0.04--0.15}
\expandafter\def\csname v@swaway.range.debate\endcsname{0.00--0.04}
\expandafter\def\csname v@swto.range.balanced\endcsname{0.11--0.46}
\expandafter\def\csname v@swaway.range.balanced\endcsname{0.04--0.28}
\expandafter\def\csname v@swto.range.pool_evpc\endcsname{0.54--0.83}
\expandafter\def\csname v@swaway.range.pool_evpc\endcsname{0.03--0.10}
\expandafter\def\csname v@mod.strong.range\endcsname{0.74--0.94}
\expandafter\def\csname v@mod.weak.range\endcsname{0.54--0.68}
\expandafter\def\csname v@rd.hb.qwen.r1\endcsname{0.70}
\expandafter\def\csname v@rd.hb.qwen.r6\endcsname{0.80}
\expandafter\def\csname v@rd.hb.qwen.D\endcsname{0.86}
\expandafter\def\csname v@rd.hb.olmo.r1\endcsname{0.56}
\expandafter\def\csname v@rd.hb.olmo.r6\endcsname{0.73}
\expandafter\def\csname v@rd.hb.olmo.D\endcsname{0.79}
\expandafter\def\csname v@rd.pz.qwen.r1\endcsname{0.36}
\expandafter\def\csname v@rd.pz.qwen.r6\endcsname{0.77}
\expandafter\def\csname v@rd.pz.qwen.D\endcsname{0.90}
\expandafter\def\csname v@rd.pz.olmo.r1\endcsname{0.35}
\expandafter\def\csname v@rd.pz.olmo.r6\endcsname{0.70}
\expandafter\def\csname v@rd.pz.olmo.D\endcsname{0.81}
\expandafter\def\csname v@rd.mq.qwen.r1\endcsname{0.19}
\expandafter\def\csname v@rd.mq.qwen.r6\endcsname{0.44}
\expandafter\def\csname v@rd.mq.qwen.D\endcsname{0.54}
\expandafter\def\csname v@rd.mq.olmo.r1\endcsname{0.15}
\expandafter\def\csname v@rd.mq.olmo.r6\endcsname{0.40}
\expandafter\def\csname v@rd.mq.olmo.D\endcsname{0.56}
\expandafter\def\csname v@rd.gap.range\endcsname{0.17--0.53}
\expandafter\def\csname v@rd.rounds.range\endcsname{0.10--0.41}
\expandafter\def\csname v@rd.plur.range\endcsname{0.06--0.15}
\expandafter\def\csname v@cite.local.agree\endcsname{88--99\%}
\expandafter\def\csname v@cite.local.disc\endcsname{0.04--0.07}
\expandafter\def\csname v@cite.local.hb\endcsname{3}
\expandafter\def\csname v@cite.local.pz\endcsname{0}
\expandafter\def\csname v@cite.local.mq\endcsname{7}
\expandafter\def\csname v@g1.argued.hb.qwen\endcsname{$-$0.44 [$-$0.59, $-$0.30]}
\expandafter\def\csname v@g1.argued.pz.qwen\endcsname{$-$0.64 [$-$0.71, $-$0.55]}
\expandafter\def\csname v@g1.nodedup.qwen.4\endcsname{0.81}
\expandafter\def\csname v@g1.nodedup.qwen.32\endcsname{0.60}
\expandafter\def\csname v@g1.nodedup.gap.qwen\endcsname{$-$0.31 [$-$0.45, $-$0.18]}
\expandafter\def\csname v@g1.ev4.qwen\endcsname{$-$0.01 [$-$0.06, 0.03]}
\expandafter\def\csname v@g1.rep.qwen\endcsname{0.00 [0.00, 0.00]}
\expandafter\def\csname v@g1.pull.qwen.label\endcsname{$-$0.59}
\expandafter\def\csname v@g1.pull.qwen.argued\endcsname{3.95}
\expandafter\def\csname v@g1.pull.qwen.raw\endcsname{3.59}
\expandafter\def\csname v@g1.pull.qwen.pool_raw\endcsname{5.07}
\expandafter\def\csname v@g1.pull.qwen.hidden\endcsname{$-$0.63}
\expandafter\def\csname v@g1.pull.qwen.labeldiff\endcsname{$+$0.04}
\expandafter\def\csname v@g1f.qwen.16\endcsname{1.08 [0.90, 1.29]}
\expandafter\def\csname v@g1f.qwen.16.dev\endcsname{177}
\expandafter\def\csname v@g1f.qwen.32\endcsname{1.57 [1.36, 1.82]}
\expandafter\def\csname v@g1f.qwen.32.dev\endcsname{449}
\expandafter\def\csname v@g1.argued.hb.olmo\endcsname{$-$0.33 [$-$0.49, $-$0.17]}
\expandafter\def\csname v@g1.argued.pz.olmo\endcsname{$-$0.62 [$-$0.70, $-$0.55]}
\expandafter\def\csname v@g1.nodedup.olmo.4\endcsname{0.73}
\expandafter\def\csname v@g1.nodedup.olmo.32\endcsname{0.55}
\expandafter\def\csname v@g1.nodedup.gap.olmo\endcsname{$-$0.36 [$-$0.55, $-$0.17]}
\expandafter\def\csname v@g1.ev4.olmo\endcsname{$-$0.06 [$-$0.15, 0.02]}
\expandafter\def\csname v@g1.rep.olmo\endcsname{$-$0.05 [$-$0.10, 0.00]}
\expandafter\def\csname v@g1.pull.olmo.label\endcsname{0.88}
\expandafter\def\csname v@g1.pull.olmo.argued\endcsname{3.14}
\expandafter\def\csname v@g1.pull.olmo.raw\endcsname{3.40}
\expandafter\def\csname v@g1.pull.olmo.pool_raw\endcsname{3.53}
\expandafter\def\csname v@g1.pull.olmo.hidden\endcsname{0.33}
\expandafter\def\csname v@g1.pull.olmo.labeldiff\endcsname{$+$0.55}
\expandafter\def\csname v@g1f.olmo.16\endcsname{0.93 [0.87, 1.00]}
\expandafter\def\csname v@g1f.olmo.16.dev\endcsname{1,006}
\expandafter\def\csname v@g1f.olmo.32\endcsname{0.89 [0.85, 0.94]}
\expandafter\def\csname v@g1f.olmo.32.dev\endcsname{5,053}
\expandafter\def\csname v@g1.raw.range\endcsname{$-$0.06 to $+$0.07}
\expandafter\def\csname v@g2.qwen.pool\endcsname{0.88}
\expandafter\def\csname v@g2.qwen.hb_release\endcsname{0.15}
\expandafter\def\csname v@g2.qwen.exchange_hb\endcsname{0.54}
\expandafter\def\csname v@g2.qwen.revealall\endcsname{0.54}
\expandafter\def\csname v@g2.qwen.dar\endcsname{0.08}
\expandafter\def\csname v@g2.qwen.dar2\endcsname{0.14}
\expandafter\def\csname v@g2.qwen.w300\endcsname{$+$0.25 [0.14, 0.38]}
\expandafter\def\csname v@g2.olmo.pool\endcsname{0.88}
\expandafter\def\csname v@g2.olmo.hb_release\endcsname{0.22}
\expandafter\def\csname v@g2.olmo.exchange_hb\endcsname{0.37}
\expandafter\def\csname v@g2.olmo.revealall\endcsname{0.26}
\expandafter\def\csname v@g2.olmo.dar\endcsname{0.11}
\expandafter\def\csname v@g2.olmo.dar2\endcsname{0.09}
\expandafter\def\csname v@g2.olmo.w300\endcsname{0.00 [$-$0.10, 0.10]}
\expandafter\def\csname v@g2.gap.range\endcsname{0.34--0.80}
\expandafter\def\csname v@g3.hb.qwen\endcsname{$+$0.02 [0.00, 0.04]}
\expandafter\def\csname v@g3.gap.qwen\endcsname{17\%}
\expandafter\def\csname v@g3.hb.olmo\endcsname{$+$0.07 [0.04, 0.11]}
\expandafter\def\csname v@g3.gap.olmo\endcsname{43\%}
\expandafter\def\csname v@g3.pz.qwen\endcsname{$-$0.01 [$-$0.02, 0.00]}
\expandafter\def\csname v@g3.pz.olmo\endcsname{$+$0.03 [0.01, 0.04]}
\expandafter\def\csname v@g3.mq.qwen\endcsname{0.00 [0.00, 0.01]}
\expandafter\def\csname v@g3.mq.olmo\endcsname{$+$0.05 [0.03, 0.06]}
\expandafter\def\csname v@g3.other.range\endcsname{$-$0.01 to $+$0.05}
\expandafter\def\csname v@g4.qwen.pool_evpc.K0\endcsname{$+$0.78 [0.64, 0.90]}
\expandafter\def\csname v@g4.qwen.pool_evpc.K1\endcsname{$+$0.76 [0.61, 0.90]}
\expandafter\def\csname v@g4.qwen.pool_evpc.K2\endcsname{$+$0.76 [0.62, 0.89]}
\expandafter\def\csname v@g4.qwen.pool_evpc.K3\endcsname{$+$0.77 [0.63, 0.89]}
\expandafter\def\csname v@g4.qwen.pool_evpc.K4\endcsname{$+$0.73 [0.59, 0.86]}
\expandafter\def\csname v@g4.qwen.pool_evpc.K5\endcsname{$+$0.72 [0.57, 0.85]}
\expandafter\def\csname v@g4.qwen.pool_evpc.r1\endcsname{0.98 [0.89, 1.06]}
\expandafter\def\csname v@g4.qwen.pool_evpc.r2\endcsname{0.98 [0.89, 1.07]}
\expandafter\def\csname v@g4.qwen.pool_evpc.r3\endcsname{0.99 [0.89, 1.09]}
\expandafter\def\csname v@g4.qwen.pool_evpc.r4\endcsname{0.94 [0.84, 1.05]}
\expandafter\def\csname v@g4.qwen.pool_evpc.r5\endcsname{0.92 [0.83, 1.01]}
\expandafter\def\csname v@g4.qwen.balanced.K0\endcsname{$+$0.20 [0.11, 0.30]}
\expandafter\def\csname v@g4.qwen.balanced.K1\endcsname{$+$0.05 [$-$0.05, 0.16]}
\expandafter\def\csname v@g4.qwen.balanced.K2\endcsname{$-$0.04 [$-$0.15, 0.05]}
\expandafter\def\csname v@g4.qwen.balanced.K3\endcsname{$-$0.04 [$-$0.14, 0.05]}
\expandafter\def\csname v@g4.qwen.balanced.K4\endcsname{$-$0.04 [$-$0.13, 0.03]}
\expandafter\def\csname v@g4.qwen.balanced.K5\endcsname{$-$0.02 [$-$0.12, 0.07]}
\expandafter\def\csname v@g4.qwen.balanced.r1\endcsname{0.25 [$-$0.28, 0.88]}
\expandafter\def\csname v@g4.qwen.balanced.r2\endcsname{$-$0.22 [$-$0.93, 0.28]}
\expandafter\def\csname v@g4.qwen.balanced.r3\endcsname{$-$0.19 [$-$0.84, 0.29]}
\expandafter\def\csname v@g4.qwen.balanced.r4\endcsname{$-$0.22 [$-$0.86, 0.15]}
\expandafter\def\csname v@g4.qwen.balanced.r5\endcsname{$-$0.12 [$-$0.72, 0.34]}
\expandafter\def\csname v@g4.qwen.exchange.K0\endcsname{$+$0.24 [0.14, 0.36]}
\expandafter\def\csname v@g4.qwen.exchange.K1\endcsname{$+$0.29 [0.17, 0.42]}
\expandafter\def\csname v@g4.qwen.exchange.K2\endcsname{$+$0.24 [0.14, 0.36]}
\expandafter\def\csname v@g4.qwen.exchange.K3\endcsname{$+$0.21 [0.11, 0.32]}
\expandafter\def\csname v@g4.qwen.exchange.K4\endcsname{$+$0.20 [0.09, 0.31]}
\expandafter\def\csname v@g4.qwen.exchange.K5\endcsname{$+$0.20 [0.08, 0.33]}
\expandafter\def\csname v@g4.qwen.exchange.r1\endcsname{1.19 [0.83, 1.78]}
\expandafter\def\csname v@g4.qwen.exchange.r2\endcsname{1.00 [0.61, 1.64]}
\expandafter\def\csname v@g4.qwen.exchange.r3\endcsname{0.87 [0.56, 1.33]}
\expandafter\def\csname v@g4.qwen.exchange.r4\endcsname{0.83 [0.43, 1.39]}
\expandafter\def\csname v@g4.qwen.exchange.r5\endcsname{0.81 [0.34, 1.47]}
\expandafter\def\csname v@g4.qwen.shareall.K0\endcsname{$+$0.18 [0.09, 0.28]}
\expandafter\def\csname v@g4.qwen.shareall.K1\endcsname{$-$0.01 [$-$0.10, 0.08]}
\expandafter\def\csname v@g4.qwen.shareall.K2\endcsname{$-$0.01 [$-$0.09, 0.07]}
\expandafter\def\csname v@g4.qwen.shareall.K3\endcsname{$+$0.01 [$-$0.07, 0.09]}
\expandafter\def\csname v@g4.qwen.shareall.K4\endcsname{0.00 [$-$0.07, 0.07]}
\expandafter\def\csname v@g4.qwen.shareall.K5\endcsname{$-$0.03 [$-$0.08, 0.02]}
\expandafter\def\csname v@g4.qwen.shareall.r1\endcsname{$-$0.04 [$-$0.53, 0.65]}
\expandafter\def\csname v@g4.qwen.shareall.r2\endcsname{$-$0.07 [$-$0.62, 0.47]}
\expandafter\def\csname v@g4.qwen.shareall.r3\endcsname{0.07 [$-$0.39, 0.61]}
\expandafter\def\csname v@g4.qwen.shareall.r4\endcsname{0.01 [$-$0.44, 0.51]}
\expandafter\def\csname v@g4.qwen.shareall.r5\endcsname{$-$0.16 [$-$0.61, 0.14]}
\expandafter\def\csname v@g4.qwen.debate.K0\endcsname{$+$0.10 [0.03, 0.17]}
\expandafter\def\csname v@g4.qwen.debate.K1\endcsname{$+$0.01 [$-$0.07, 0.09]}
\expandafter\def\csname v@g4.qwen.debate.K2\endcsname{$-$0.01 [$-$0.12, 0.11]}
\expandafter\def\csname v@g4.qwen.debate.K3\endcsname{$-$0.04 [$-$0.10, 0.01]}
\expandafter\def\csname v@g4.qwen.debate.K4\endcsname{$-$0.05 [$-$0.11, 0.01]}
\expandafter\def\csname v@g4.qwen.debate.K5\endcsname{$-$0.06 [$-$0.13, 0.00]}
\expandafter\def\csname v@g4.olmo.pool_evpc.K0\endcsname{$+$0.70 [0.54, 0.83]}
\expandafter\def\csname v@g4.olmo.pool_evpc.K1\endcsname{$+$0.63 [0.44, 0.79]}
\expandafter\def\csname v@g4.olmo.pool_evpc.K2\endcsname{$+$0.65 [0.50, 0.78]}
\expandafter\def\csname v@g4.olmo.pool_evpc.K3\endcsname{$+$0.61 [0.46, 0.75]}
\expandafter\def\csname v@g4.olmo.pool_evpc.K4\endcsname{$+$0.60 [0.43, 0.75]}
\expandafter\def\csname v@g4.olmo.pool_evpc.K5\endcsname{$+$0.49 [0.32, 0.65]}
\expandafter\def\csname v@g4.olmo.pool_evpc.r1\endcsname{0.90 [0.76, 1.03]}
\expandafter\def\csname v@g4.olmo.pool_evpc.r2\endcsname{0.93 [0.82, 1.06]}
\expandafter\def\csname v@g4.olmo.pool_evpc.r3\endcsname{0.88 [0.77, 0.99]}
\expandafter\def\csname v@g4.olmo.pool_evpc.r4\endcsname{0.86 [0.69, 1.01]}
\expandafter\def\csname v@g4.olmo.pool_evpc.r5\endcsname{0.70 [0.53, 0.87]}
\expandafter\def\csname v@g4.olmo.balanced.K0\endcsname{$+$0.29 [0.17, 0.42]}
\expandafter\def\csname v@g4.olmo.balanced.K1\endcsname{$+$0.16 [0.02, 0.31]}
\expandafter\def\csname v@g4.olmo.balanced.K2\endcsname{$+$0.11 [$-$0.03, 0.25]}
\expandafter\def\csname v@g4.olmo.balanced.K3\endcsname{$+$0.08 [$-$0.03, 0.20]}
\expandafter\def\csname v@g4.olmo.balanced.K4\endcsname{$+$0.02 [$-$0.09, 0.13]}
\expandafter\def\csname v@g4.olmo.balanced.K5\endcsname{$-$0.04 [$-$0.14, 0.07]}
\expandafter\def\csname v@g4.olmo.balanced.r1\endcsname{0.55 [0.07, 1.09]}
\expandafter\def\csname v@g4.olmo.balanced.r2\endcsname{0.38 [$-$0.09, 1.00]}
\expandafter\def\csname v@g4.olmo.balanced.r3\endcsname{0.29 [$-$0.09, 0.77]}
\expandafter\def\csname v@g4.olmo.balanced.r4\endcsname{0.07 [$-$0.39, 0.42]}
\expandafter\def\csname v@g4.olmo.balanced.r5\endcsname{$-$0.12 [$-$0.57, 0.25]}
\expandafter\def\csname v@g4.olmo.exchange.K0\endcsname{$+$0.30 [0.10, 0.48]}
\expandafter\def\csname v@g4.olmo.exchange.K1\endcsname{$+$0.28 [0.06, 0.48]}
\expandafter\def\csname v@g4.olmo.exchange.K2\endcsname{$+$0.30 [0.16, 0.44]}
\expandafter\def\csname v@g4.olmo.exchange.K3\endcsname{$+$0.34 [0.15, 0.53]}
\expandafter\def\csname v@g4.olmo.exchange.K4\endcsname{$+$0.30 [0.14, 0.46]}
\expandafter\def\csname v@g4.olmo.exchange.K5\endcsname{$+$0.20 [0.02, 0.38]}
\expandafter\def\csname v@g4.olmo.exchange.r1\endcsname{0.92 [0.42, 1.41]}
\expandafter\def\csname v@g4.olmo.exchange.r2\endcsname{1.01 [0.71, 1.87]}
\expandafter\def\csname v@g4.olmo.exchange.r3\endcsname{1.15 [0.77, 1.95]}
\expandafter\def\csname v@g4.olmo.exchange.r4\endcsname{1.01 [0.66, 1.76]}
\expandafter\def\csname v@g4.olmo.exchange.r5\endcsname{0.67 [0.13, 1.05]}
\expandafter\def\csname v@g4.olmo.shareall.K0\endcsname{$+$0.18 [0.08, 0.28]}
\expandafter\def\csname v@g4.olmo.shareall.K1\endcsname{$+$0.05 [$-$0.09, 0.19]}
\expandafter\def\csname v@g4.olmo.shareall.K2\endcsname{$+$0.02 [$-$0.11, 0.13]}
\expandafter\def\csname v@g4.olmo.shareall.K3\endcsname{$-$0.07 [$-$0.20, 0.05]}
\expandafter\def\csname v@g4.olmo.shareall.K4\endcsname{$-$0.08 [$-$0.20, 0.02]}
\expandafter\def\csname v@g4.olmo.shareall.K5\endcsname{$-$0.11 [$-$0.21, $-$0.02]}
\expandafter\def\csname v@g4.olmo.shareall.r1\endcsname{0.29 [$-$0.67, 1.09]}
\expandafter\def\csname v@g4.olmo.shareall.r2\endcsname{0.09 [$-$0.95, 0.65]}
\expandafter\def\csname v@g4.olmo.shareall.r3\endcsname{$-$0.40 [$-$1.64, 0.22]}
\expandafter\def\csname v@g4.olmo.shareall.r4\endcsname{$-$0.46 [$-$1.65, 0.10]}
\expandafter\def\csname v@g4.olmo.shareall.r5\endcsname{$-$0.60 [$-$1.90, $-$0.08]}
\expandafter\def\csname v@g4.olmo.debate.K0\endcsname{$+$0.19 [0.09, 0.30]}
\expandafter\def\csname v@g4.olmo.debate.K1\endcsname{$+$0.14 [0.05, 0.22]}
\expandafter\def\csname v@g4.olmo.debate.K2\endcsname{$+$0.06 [$-$0.04, 0.15]}
\expandafter\def\csname v@g4.olmo.debate.K3\endcsname{$+$0.05 [$-$0.06, 0.16]}
\expandafter\def\csname v@g4.olmo.debate.K4\endcsname{$+$0.05 [$-$0.03, 0.14]}
\expandafter\def\csname v@g4.olmo.debate.K5\endcsname{0.00 [$-$0.09, 0.09]}
\expandafter\def\csname v@g4.olmo.debate.r1\endcsname{0.72 [0.24, 1.81]}
\expandafter\def\csname v@g4.olmo.debate.r2\endcsname{0.31 [$-$0.23, 1.08]}
\expandafter\def\csname v@g4.olmo.debate.r3\endcsname{0.28 [$-$0.31, 1.18]}
\expandafter\def\csname v@g4.olmo.debate.r4\endcsname{0.28 [$-$0.17, 1.03]}
\expandafter\def\csname v@g4.olmo.debate.r5\endcsname{0.00 [$-$0.60, 0.59]}
\expandafter\def\csname v@g4.mini.pool_evpc.K0\endcsname{$+$0.82 [0.72, 0.91]}
\expandafter\def\csname v@g4.mini.pool_evpc.K1\endcsname{$+$0.64 [0.49, 0.78]}
\expandafter\def\csname v@g4.mini.pool_evpc.K2\endcsname{$+$0.70 [0.57, 0.82]}
\expandafter\def\csname v@g4.mini.pool_evpc.K3\endcsname{$+$0.66 [0.54, 0.77]}
\expandafter\def\csname v@g4.mini.pool_evpc.K4\endcsname{$+$0.61 [0.50, 0.71]}
\expandafter\def\csname v@g4.mini.pool_evpc.K5\endcsname{$+$0.67 [0.52, 0.80]}
\expandafter\def\csname v@g4.mini.pool_evpc.r1\endcsname{0.78 [0.65, 0.89]}
\expandafter\def\csname v@g4.mini.pool_evpc.r2\endcsname{0.85 [0.75, 0.95]}
\expandafter\def\csname v@g4.mini.pool_evpc.r3\endcsname{0.81 [0.70, 0.91]}
\expandafter\def\csname v@g4.mini.pool_evpc.r4\endcsname{0.74 [0.65, 0.83]}
\expandafter\def\csname v@g4.mini.pool_evpc.r5\endcsname{0.81 [0.69, 0.92]}
\expandafter\def\csname v@g4.mini.balanced.K0\endcsname{$+$0.40 [0.30, 0.49]}
\expandafter\def\csname v@g4.mini.balanced.K1\endcsname{$+$0.30 [0.18, 0.41]}
\expandafter\def\csname v@g4.mini.balanced.K2\endcsname{$+$0.23 [0.11, 0.35]}
\expandafter\def\csname v@g4.mini.balanced.K3\endcsname{$+$0.17 [0.03, 0.31]}
\expandafter\def\csname v@g4.mini.balanced.K4\endcsname{$+$0.14 [0.01, 0.26]}
\expandafter\def\csname v@g4.mini.balanced.K5\endcsname{$+$0.05 [$-$0.09, 0.20]}
\expandafter\def\csname v@g4.mini.balanced.r1\endcsname{0.75 [0.45, 1.11]}
\expandafter\def\csname v@g4.mini.balanced.r2\endcsname{0.59 [0.31, 0.92]}
\expandafter\def\csname v@g4.mini.balanced.r3\endcsname{0.42 [0.09, 0.75]}
\expandafter\def\csname v@g4.mini.balanced.r4\endcsname{0.35 [0.03, 0.71]}
\expandafter\def\csname v@g4.mini.balanced.r5\endcsname{0.13 [$-$0.24, 0.54]}
\expandafter\def\csname v@g4.mini.exchange.K0\endcsname{$+$0.48 [0.31, 0.64]}
\expandafter\def\csname v@g4.mini.exchange.K1\endcsname{$+$0.51 [0.37, 0.64]}
\expandafter\def\csname v@g4.mini.exchange.K2\endcsname{$+$0.41 [0.27, 0.55]}
\expandafter\def\csname v@g4.mini.exchange.K3\endcsname{$+$0.30 [0.14, 0.46]}
\expandafter\def\csname v@g4.mini.exchange.K4\endcsname{$+$0.30 [0.13, 0.46]}
\expandafter\def\csname v@g4.mini.exchange.K5\endcsname{$+$0.17 [0.02, 0.33]}
\expandafter\def\csname v@g4.mini.exchange.r1\endcsname{1.06 [0.86, 1.40]}
\expandafter\def\csname v@g4.mini.exchange.r2\endcsname{0.86 [0.69, 1.10]}
\expandafter\def\csname v@g4.mini.exchange.r3\endcsname{0.63 [0.39, 0.87]}
\expandafter\def\csname v@g4.mini.exchange.r4\endcsname{0.62 [0.32, 0.91]}
\expandafter\def\csname v@g4.mini.exchange.r5\endcsname{0.36 [0.04, 0.62]}
\expandafter\def\csname v@g4.mini.shareall.K0\endcsname{$+$0.39 [0.27, 0.52]}
\expandafter\def\csname v@g4.mini.shareall.K1\endcsname{$+$0.20 [0.09, 0.31]}
\expandafter\def\csname v@g4.mini.shareall.K2\endcsname{$+$0.04 [$-$0.09, 0.16]}
\expandafter\def\csname v@g4.mini.shareall.K3\endcsname{$+$0.02 [$-$0.09, 0.12]}
\expandafter\def\csname v@g4.mini.shareall.K4\endcsname{$-$0.05 [$-$0.16, 0.05]}
\expandafter\def\csname v@g4.mini.shareall.K5\endcsname{$-$0.04 [$-$0.15, 0.08]}
\expandafter\def\csname v@g4.mini.shareall.r1\endcsname{0.53 [0.28, 0.80]}
\expandafter\def\csname v@g4.mini.shareall.r2\endcsname{0.09 [$-$0.28, 0.39]}
\expandafter\def\csname v@g4.mini.shareall.r3\endcsname{0.04 [$-$0.26, 0.29]}
\expandafter\def\csname v@g4.mini.shareall.r4\endcsname{$-$0.14 [$-$0.50, 0.12]}
\expandafter\def\csname v@g4.mini.shareall.r5\endcsname{$-$0.09 [$-$0.45, 0.19]}
\expandafter\def\csname v@g4.mini.debate.K0\endcsname{$+$0.31 [0.18, 0.43]}
\expandafter\def\csname v@g4.mini.debate.K1\endcsname{$+$0.11 [$-$0.01, 0.22]}
\expandafter\def\csname v@g4.mini.debate.K2\endcsname{$+$0.11 [$-$0.03, 0.25]}
\expandafter\def\csname v@g4.mini.debate.K3\endcsname{$+$0.08 [$-$0.03, 0.18]}
\expandafter\def\csname v@g4.mini.debate.K4\endcsname{$+$0.05 [$-$0.07, 0.18]}
\expandafter\def\csname v@g4.mini.debate.K5\endcsname{$+$0.02 [$-$0.10, 0.15]}
\expandafter\def\csname v@g4.mini.debate.r1\endcsname{0.36 [$-$0.03, 0.71]}
\expandafter\def\csname v@g4.mini.debate.r2\endcsname{0.37 [$-$0.09, 0.77]}
\expandafter\def\csname v@g4.mini.debate.r3\endcsname{0.25 [$-$0.09, 0.66]}
\expandafter\def\csname v@g4.mini.debate.r4\endcsname{0.17 [$-$0.27, 0.60]}
\expandafter\def\csname v@g4.mini.debate.r5\endcsname{0.05 [$-$0.37, 0.50]}
\expandafter\def\csname v@g4.gemma.pool_evpc.K0\endcsname{$+$0.91 [0.82, 0.97]}
\expandafter\def\csname v@g4.gemma.pool_evpc.K1\endcsname{$+$0.91 [0.84, 0.97]}
\expandafter\def\csname v@g4.gemma.pool_evpc.K2\endcsname{$+$0.90 [0.82, 0.96]}
\expandafter\def\csname v@g4.gemma.pool_evpc.K3\endcsname{$+$0.91 [0.83, 0.97]}
\expandafter\def\csname v@g4.gemma.pool_evpc.K4\endcsname{$+$0.89 [0.81, 0.95]}
\expandafter\def\csname v@g4.gemma.pool_evpc.K5\endcsname{$+$0.83 [0.73, 0.92]}
\expandafter\def\csname v@g4.gemma.pool_evpc.r1\endcsname{1.00 [0.92, 1.08]}
\expandafter\def\csname v@g4.gemma.pool_evpc.r2\endcsname{0.98 [0.91, 1.06]}
\expandafter\def\csname v@g4.gemma.pool_evpc.r3\endcsname{0.99 [0.92, 1.07]}
\expandafter\def\csname v@g4.gemma.pool_evpc.r4\endcsname{0.97 [0.90, 1.05]}
\expandafter\def\csname v@g4.gemma.pool_evpc.r5\endcsname{0.91 [0.81, 1.01]}
\expandafter\def\csname v@g4.gemma.balanced.K0\endcsname{$+$0.42 [0.30, 0.55]}
\expandafter\def\csname v@g4.gemma.balanced.K1\endcsname{$+$0.44 [0.29, 0.59]}
\expandafter\def\csname v@g4.gemma.balanced.K2\endcsname{$+$0.43 [0.29, 0.57]}
\expandafter\def\csname v@g4.gemma.balanced.K3\endcsname{$+$0.30 [0.16, 0.45]}
\expandafter\def\csname v@g4.gemma.balanced.K4\endcsname{$+$0.41 [0.27, 0.55]}
\expandafter\def\csname v@g4.gemma.balanced.K5\endcsname{$+$0.21 [0.09, 0.34]}
\expandafter\def\csname v@g4.gemma.balanced.r1\endcsname{1.05 [0.76, 1.42]}
\expandafter\def\csname v@g4.gemma.balanced.r2\endcsname{1.02 [0.71, 1.47]}
\expandafter\def\csname v@g4.gemma.balanced.r3\endcsname{0.72 [0.42, 1.00]}
\expandafter\def\csname v@g4.gemma.balanced.r4\endcsname{0.98 [0.71, 1.31]}
\expandafter\def\csname v@g4.gemma.balanced.r5\endcsname{0.51 [0.22, 0.87]}
\expandafter\def\csname v@g4.gemma.exchange.K0\endcsname{$+$0.72 [0.57, 0.85]}
\expandafter\def\csname v@g4.gemma.exchange.K1\endcsname{$+$0.76 [0.65, 0.85]}
\expandafter\def\csname v@g4.gemma.exchange.K2\endcsname{$+$0.66 [0.53, 0.78]}
\expandafter\def\csname v@g4.gemma.exchange.K3\endcsname{$+$0.60 [0.47, 0.73]}
\expandafter\def\csname v@g4.gemma.exchange.K4\endcsname{$+$0.54 [0.42, 0.66]}
\expandafter\def\csname v@g4.gemma.exchange.K5\endcsname{$+$0.45 [0.32, 0.57]}
\expandafter\def\csname v@g4.gemma.exchange.r1\endcsname{1.04 [0.90, 1.25]}
\expandafter\def\csname v@g4.gemma.exchange.r2\endcsname{0.91 [0.71, 1.15]}
\expandafter\def\csname v@g4.gemma.exchange.r3\endcsname{0.83 [0.64, 1.07]}
\expandafter\def\csname v@g4.gemma.exchange.r4\endcsname{0.75 [0.60, 0.94]}
\expandafter\def\csname v@g4.gemma.exchange.r5\endcsname{0.62 [0.46, 0.82]}
\expandafter\def\csname v@g4.gemma.shareall.K0\endcsname{$+$0.51 [0.39, 0.64]}
\expandafter\def\csname v@g4.gemma.shareall.K1\endcsname{$+$0.45 [0.29, 0.62]}
\expandafter\def\csname v@g4.gemma.shareall.K2\endcsname{$+$0.40 [0.23, 0.56]}
\expandafter\def\csname v@g4.gemma.shareall.K3\endcsname{$+$0.32 [0.17, 0.47]}
\expandafter\def\csname v@g4.gemma.shareall.K4\endcsname{$+$0.32 [0.17, 0.47]}
\expandafter\def\csname v@g4.gemma.shareall.K5\endcsname{$+$0.17 [0.05, 0.29]}
\expandafter\def\csname v@g4.gemma.shareall.r1\endcsname{0.89 [0.60, 1.17]}
\expandafter\def\csname v@g4.gemma.shareall.r2\endcsname{0.78 [0.52, 1.02]}
\expandafter\def\csname v@g4.gemma.shareall.r3\endcsname{0.63 [0.38, 0.89]}
\expandafter\def\csname v@g4.gemma.shareall.r4\endcsname{0.62 [0.36, 0.87]}
\expandafter\def\csname v@g4.gemma.shareall.r5\endcsname{0.33 [0.10, 0.55]}
\expandafter\def\csname v@g4.gemma.debate.K0\endcsname{$+$0.46 [0.33, 0.60]}
\expandafter\def\csname v@g4.gemma.debate.K1\endcsname{$+$0.44 [0.30, 0.58]}
\expandafter\def\csname v@g4.gemma.debate.K2\endcsname{$+$0.33 [0.19, 0.47]}
\expandafter\def\csname v@g4.gemma.debate.K3\endcsname{$+$0.31 [0.16, 0.45]}
\expandafter\def\csname v@g4.gemma.debate.K4\endcsname{$+$0.34 [0.20, 0.47]}
\expandafter\def\csname v@g4.gemma.debate.K5\endcsname{$+$0.17 [0.05, 0.31]}
\expandafter\def\csname v@g4.gemma.debate.r1\endcsname{0.95 [0.68, 1.33]}
\expandafter\def\csname v@g4.gemma.debate.r2\endcsname{0.73 [0.46, 1.02]}
\expandafter\def\csname v@g4.gemma.debate.r3\endcsname{0.67 [0.38, 0.98]}
\expandafter\def\csname v@g4.gemma.debate.r4\endcsname{0.74 [0.47, 1.04]}
\expandafter\def\csname v@g4.gemma.debate.r5\endcsname{0.38 [0.11, 0.63]}
\expandafter\def\csname v@lk.qwen.debate.K0\endcsname{$+$0.07 [0.02, 0.13]}
\expandafter\def\csname v@lk.qwen.debate.K3\endcsname{0.00 [$-$0.10, 0.08]}
\expandafter\def\csname v@lk.qwen.pool.K0\endcsname{$+$0.75 [0.62, 0.85]}
\expandafter\def\csname v@lk.qwen.pool.K3\endcsname{$+$0.69 [0.57, 0.80]}
\expandafter\def\csname v@lk.olmo.debate.K0\endcsname{$+$0.18 [0.10, 0.25]}
\expandafter\def\csname v@lk.olmo.debate.K3\endcsname{$+$0.06 [0.00, 0.13]}
\expandafter\def\csname v@lk.olmo.pool.K0\endcsname{$+$0.68 [0.56, 0.80]}
\expandafter\def\csname v@lk.olmo.pool.K3\endcsname{$+$0.61 [0.50, 0.71]}
\expandafter\def\csname v@g4.pool.kept\endcsname{0.74--1.00}
\expandafter\def\csname v@g.claims.held\endcsname{22}
\expandafter\def\csname v@g.claims.failed\endcsname{9}
\expandafter\def\csname v@h1.hb16.qwen.pool_evpc\endcsname{0.88}
\expandafter\def\csname v@h1.hb16.qwen.debate_evpc\endcsname{0.88}
\expandafter\def\csname v@h1.hb16.qwen.balanced_evpc_rand\endcsname{0.88}
\expandafter\def\csname v@h1.hb16.qwen.balanced_evpc\endcsname{0.88}
\expandafter\def\csname v@h1.hb16.qwen.pool_evpc_raw\endcsname{0.44}
\expandafter\def\csname v@h1.hb16.qwen.debate_evpc_raw\endcsname{0.43}
\expandafter\def\csname v@h1.hb16.qwen.balanced_nu05_evpc_raw\endcsname{0.62}
\expandafter\def\csname v@h1.hb16.qwen.balanced_evpc_rand_raw\endcsname{0.66}
\expandafter\def\csname v@h1.hb16.qwen.balanced_evpc_raw\endcsname{0.71}
\expandafter\def\csname v@h1.hb16.qwen.debate\endcsname{0.08}
\expandafter\def\csname v@h1.hb16.qwen.balanced\endcsname{0.20}
\expandafter\def\csname v@h1.hb16.qwen.solo\endcsname{0.02}
\expandafter\def\csname v@h1.hb16.qwen.bal_vs_pool\endcsname{$+$0.28 [0.17, 0.40]}
\expandafter\def\csname v@h1.hb16.qwen.bal_vs_pool.pt\endcsname{$+$0.28}
\expandafter\def\csname v@h1.hb16.qwen.rand_vs_deb\endcsname{$+$0.23 [0.11, 0.35]}
\expandafter\def\csname v@h1.hb16.qwen.rand_vs_deb.pt\endcsname{$+$0.23}
\expandafter\def\csname v@h1.hb16.qwen.rank\endcsname{$+$0.05 [$-$0.03, 0.13]}
\expandafter\def\csname v@h1.hb16.qwen.rank.pt\endcsname{$+$0.05}
\expandafter\def\csname v@h1.hb16.qwen.cost\endcsname{$+$0.16}
\expandafter\def\csname v@h1.hb16.qwen.costpool\endcsname{$+$0.44}
\expandafter\def\csname v@h1.hb16.qwen.gain.pool_evpc_raw\endcsname{1.15 [1.00, 1.31]}
\expandafter\def\csname v@h1.hb16.qwen.gain.pool_evpc_raw.pt\endcsname{1.15}
\expandafter\def\csname v@h1.hb16.qwen.swto.pool_evpc_raw\endcsname{0.42}
\expandafter\def\csname v@h1.hb16.qwen.swaway.pool_evpc_raw\endcsname{0.00}
\expandafter\def\csname v@h1.hb16.qwen.gain.balanced_evpc_raw\endcsname{0.67 [0.50, 0.82]}
\expandafter\def\csname v@h1.hb16.qwen.gain.balanced_evpc_raw.pt\endcsname{0.67}
\expandafter\def\csname v@h1.hb16.qwen.swto.balanced_evpc_raw\endcsname{0.71}
\expandafter\def\csname v@h1.hb16.qwen.swaway.balanced_evpc_raw\endcsname{0.02}
\expandafter\def\csname v@h1.hb16.qwen.gain.debate_evpc_raw\endcsname{0.97 [0.84, 1.08]}
\expandafter\def\csname v@h1.hb16.qwen.gain.debate_evpc_raw.pt\endcsname{0.97}
\expandafter\def\csname v@h1.hb16.qwen.swto.debate_evpc_raw\endcsname{0.44}
\expandafter\def\csname v@h1.hb16.qwen.swaway.debate_evpc_raw\endcsname{0.03}
\expandafter\def\csname v@h1.hb16.qwen.gain.balanced_evpc_rand_raw\endcsname{0.61 [0.45, 0.77]}
\expandafter\def\csname v@h1.hb16.qwen.gain.balanced_evpc_rand_raw.pt\endcsname{0.61}
\expandafter\def\csname v@h1.hb16.qwen.swto.balanced_evpc_rand_raw\endcsname{0.67}
\expandafter\def\csname v@h1.hb16.qwen.swaway.balanced_evpc_rand_raw\endcsname{0.03}
\expandafter\def\csname v@h1.hb32.qwen.pool_evpc\endcsname{0.91}
\expandafter\def\csname v@h1.hb32.qwen.debate_evpc\endcsname{0.79}
\expandafter\def\csname v@h1.hb32.qwen.balanced_evpc_rand\endcsname{0.85}
\expandafter\def\csname v@h1.hb32.qwen.balanced_evpc\endcsname{0.91}
\expandafter\def\csname v@h1.hb32.qwen.pool_evpc_raw\endcsname{0.31}
\expandafter\def\csname v@h1.hb32.qwen.debate_evpc_raw\endcsname{0.34}
\expandafter\def\csname v@h1.hb32.qwen.balanced_evpc_rand_raw\endcsname{0.57}
\expandafter\def\csname v@h1.hb32.qwen.balanced_evpc_raw\endcsname{0.78}
\expandafter\def\csname v@h1.hb32.qwen.debate\endcsname{0.05}
\expandafter\def\csname v@h1.hb32.qwen.balanced\endcsname{0.22}
\expandafter\def\csname v@h1.hb32.qwen.solo\endcsname{0.01}
\expandafter\def\csname v@h1.hb32.qwen.bal_vs_pool\endcsname{$+$0.47 [0.31, 0.62]}
\expandafter\def\csname v@h1.hb32.qwen.bal_vs_pool.pt\endcsname{$+$0.47}
\expandafter\def\csname v@h1.hb32.qwen.rand_vs_deb\endcsname{$+$0.22 [0.13, 0.32]}
\expandafter\def\csname v@h1.hb32.qwen.rand_vs_deb.pt\endcsname{$+$0.22}
\expandafter\def\csname v@h1.hb32.qwen.rank\endcsname{$+$0.21 [0.11, 0.32]}
\expandafter\def\csname v@h1.hb32.qwen.rank.pt\endcsname{$+$0.21}
\expandafter\def\csname v@h1.hb32.qwen.cost\endcsname{$+$0.13}
\expandafter\def\csname v@h1.hb32.qwen.costpool\endcsname{$+$0.59}
\expandafter\def\csname v@h1.hb32.qwen.gain.pool_evpc_raw\endcsname{1.20 [0.98, 1.41]}
\expandafter\def\csname v@h1.hb32.qwen.gain.pool_evpc_raw.pt\endcsname{1.20}
\expandafter\def\csname v@h1.hb32.qwen.swto.pool_evpc_raw\endcsname{0.31}
\expandafter\def\csname v@h1.hb32.qwen.swaway.pool_evpc_raw\endcsname{0.01}
\expandafter\def\csname v@h1.hb32.qwen.gain.balanced_evpc_raw\endcsname{0.60 [0.48, 0.72]}
\expandafter\def\csname v@h1.hb32.qwen.gain.balanced_evpc_raw.pt\endcsname{0.60}
\expandafter\def\csname v@h1.hb32.qwen.swto.balanced_evpc_raw\endcsname{0.79}
\expandafter\def\csname v@h1.hb32.qwen.swaway.balanced_evpc_raw\endcsname{0.02}
\expandafter\def\csname v@h1.hb32.qwen.gain.debate_evpc_raw\endcsname{0.97 [0.83, 1.12]}
\expandafter\def\csname v@h1.hb32.qwen.gain.debate_evpc_raw.pt\endcsname{0.97}
\expandafter\def\csname v@h1.hb32.qwen.swto.debate_evpc_raw\endcsname{0.34}
\expandafter\def\csname v@h1.hb32.qwen.swaway.debate_evpc_raw\endcsname{0.01}
\expandafter\def\csname v@h1.hb32.qwen.gain.balanced_evpc_rand_raw\endcsname{0.72 [0.62, 0.82]}
\expandafter\def\csname v@h1.hb32.qwen.gain.balanced_evpc_rand_raw.pt\endcsname{0.72}
\expandafter\def\csname v@h1.hb32.qwen.swto.balanced_evpc_rand_raw\endcsname{0.64}
\expandafter\def\csname v@h1.hb32.qwen.swaway.balanced_evpc_rand_raw\endcsname{0.09}
\expandafter\def\csname v@h1.pz16.qwen.pool_evpc\endcsname{0.85}
\expandafter\def\csname v@h1.pz16.qwen.debate_evpc\endcsname{0.60}
\expandafter\def\csname v@h1.pz16.qwen.balanced_evpc_rand\endcsname{0.56}
\expandafter\def\csname v@h1.pz16.qwen.balanced_evpc\endcsname{0.83}
\expandafter\def\csname v@h1.pz16.qwen.pool_evpc_raw\endcsname{0.22}
\expandafter\def\csname v@h1.pz16.qwen.debate_evpc_raw\endcsname{0.28}
\expandafter\def\csname v@h1.pz16.qwen.balanced_nu05_evpc_raw\endcsname{0.27}
\expandafter\def\csname v@h1.pz16.qwen.balanced_evpc_rand_raw\endcsname{0.26}
\expandafter\def\csname v@h1.pz16.qwen.balanced_evpc_raw\endcsname{0.34}
\expandafter\def\csname v@h1.pz16.qwen.debate\endcsname{0.10}
\expandafter\def\csname v@h1.pz16.qwen.balanced\endcsname{0.15}
\expandafter\def\csname v@h1.pz16.qwen.solo\endcsname{0.08}
\expandafter\def\csname v@h1.pz16.qwen.bal_vs_pool\endcsname{$+$0.12 [0.06, 0.17]}
\expandafter\def\csname v@h1.pz16.qwen.bal_vs_pool.pt\endcsname{$+$0.12}
\expandafter\def\csname v@h1.pz16.qwen.rand_vs_deb\endcsname{$-$0.01 [$-$0.05, 0.02]}
\expandafter\def\csname v@h1.pz16.qwen.rand_vs_deb.pt\endcsname{$-$0.01}
\expandafter\def\csname v@h1.pz16.qwen.rank\endcsname{$+$0.07 [0.03, 0.12]}
\expandafter\def\csname v@h1.pz16.qwen.rank.pt\endcsname{$+$0.07}
\expandafter\def\csname v@h1.pz16.qwen.cost\endcsname{$+$0.49}
\expandafter\def\csname v@h1.pz16.qwen.costpool\endcsname{$+$0.64}
\expandafter\def\csname v@h1.pz16.qwen.gain.pool_evpc_raw\endcsname{0.85 [0.64, 1.04]}
\expandafter\def\csname v@h1.pz16.qwen.gain.pool_evpc_raw.pt\endcsname{0.85}
\expandafter\def\csname v@h1.pz16.qwen.swto.pool_evpc_raw\endcsname{0.16}
\expandafter\def\csname v@h1.pz16.qwen.swaway.pool_evpc_raw\endcsname{0.01}
\expandafter\def\csname v@h1.pz16.qwen.gain.balanced_evpc_raw\endcsname{0.44 [0.30, 0.56]}
\expandafter\def\csname v@h1.pz16.qwen.gain.balanced_evpc_raw.pt\endcsname{0.44}
\expandafter\def\csname v@h1.pz16.qwen.swto.balanced_evpc_raw\endcsname{0.35}
\expandafter\def\csname v@h1.pz16.qwen.swaway.balanced_evpc_raw\endcsname{0.08}
\expandafter\def\csname v@h1.pz16.qwen.gain.debate_evpc_raw\endcsname{0.92 [0.81, 1.03]}
\expandafter\def\csname v@h1.pz16.qwen.gain.debate_evpc_raw.pt\endcsname{0.92}
\expandafter\def\csname v@h1.pz16.qwen.swto.debate_evpc_raw\endcsname{0.22}
\expandafter\def\csname v@h1.pz16.qwen.swaway.debate_evpc_raw\endcsname{0.01}
\expandafter\def\csname v@h1.pz16.qwen.gain.balanced_evpc_rand_raw\endcsname{0.63 [0.50, 0.75]}
\expandafter\def\csname v@h1.pz16.qwen.gain.balanced_evpc_rand_raw.pt\endcsname{0.63}
\expandafter\def\csname v@h1.pz16.qwen.swto.balanced_evpc_rand_raw\endcsname{0.28}
\expandafter\def\csname v@h1.pz16.qwen.swaway.balanced_evpc_rand_raw\endcsname{0.08}
\expandafter\def\csname v@h1.late16.qwen.pool_evpc\endcsname{0.90}
\expandafter\def\csname v@h1.late16.qwen.debate_evpc\endcsname{0.80}
\expandafter\def\csname v@h1.late16.qwen.balanced_evpc\endcsname{0.90}
\expandafter\def\csname v@h1.late16.qwen.pool_evpc_raw\endcsname{0.09}
\expandafter\def\csname v@h1.late16.qwen.debate_evpc_raw\endcsname{0.11}
\expandafter\def\csname v@h1.late16.qwen.balanced_evpc_rand_raw\endcsname{0.27}
\expandafter\def\csname v@h1.late16.qwen.balanced_evpc_raw\endcsname{0.28}
\expandafter\def\csname v@h1.late16.qwen.debate\endcsname{0.02}
\expandafter\def\csname v@h1.late16.qwen.solo\endcsname{0.02}
\expandafter\def\csname v@h1.late16.qwen.bal_vs_pool\endcsname{$+$0.19 [0.08, 0.30]}
\expandafter\def\csname v@h1.late16.qwen.bal_vs_pool.pt\endcsname{$+$0.19}
\expandafter\def\csname v@h1.late16.qwen.rand_vs_deb\endcsname{$+$0.15 [0.07, 0.24]}
\expandafter\def\csname v@h1.late16.qwen.rand_vs_deb.pt\endcsname{$+$0.15}
\expandafter\def\csname v@h1.late16.qwen.rank\endcsname{$+$0.02 [$-$0.07, 0.10]}
\expandafter\def\csname v@h1.late16.qwen.rank.pt\endcsname{$+$0.02}
\expandafter\def\csname v@h1.late16.qwen.cost\endcsname{$+$0.61}
\expandafter\def\csname v@h1.late16.qwen.costpool\endcsname{$+$0.81}
\expandafter\def\csname v@h1.late16.qwen.gain.pool_evpc_raw\endcsname{1.48 [1.31, 1.78]}
\expandafter\def\csname v@h1.late16.qwen.gain.pool_evpc_raw.pt\endcsname{1.48}
\expandafter\def\csname v@h1.late16.qwen.swto.pool_evpc_raw\endcsname{0.09}
\expandafter\def\csname v@h1.late16.qwen.swaway.pool_evpc_raw\endcsname{0.00}
\expandafter\def\csname v@h1.late16.qwen.gain.balanced_evpc_raw\endcsname{0.64 [0.29, 0.90]}
\expandafter\def\csname v@h1.late16.qwen.gain.balanced_evpc_raw.pt\endcsname{0.64}
\expandafter\def\csname v@h1.late16.qwen.swto.balanced_evpc_raw\endcsname{0.28}
\expandafter\def\csname v@h1.late16.qwen.swaway.balanced_evpc_raw\endcsname{0.00}
\expandafter\def\csname v@h1.late16.qwen.gain.debate_evpc_raw\endcsname{1.22 [0.57, 1.55]}
\expandafter\def\csname v@h1.late16.qwen.gain.debate_evpc_raw.pt\endcsname{1.22}
\expandafter\def\csname v@h1.late16.qwen.swto.debate_evpc_raw\endcsname{0.12}
\expandafter\def\csname v@h1.late16.qwen.swaway.debate_evpc_raw\endcsname{0.00}
\expandafter\def\csname v@h1.late16.qwen.gain.balanced_evpc_rand_raw\endcsname{0.87 [0.25, 1.14]}
\expandafter\def\csname v@h1.late16.qwen.gain.balanced_evpc_rand_raw.pt\endcsname{0.87}
\expandafter\def\csname v@h1.late16.qwen.swto.balanced_evpc_rand_raw\endcsname{0.28}
\expandafter\def\csname v@h1.late16.qwen.swaway.balanced_evpc_rand_raw\endcsname{0.02}
\expandafter\def\csname v@h1.kernel.qwen.pool_evpc.K0\endcsname{$+$0.86 [0.75, 0.96]}
\expandafter\def\csname v@h1.kernel.qwen.pool_evpc.K0.pt\endcsname{$+$0.86}
\expandafter\def\csname v@h1.kernel.qwen.pool_evpc.K\endcsname{$+$0.89 [0.79, 0.96]}
\expandafter\def\csname v@h1.kernel.qwen.pool_evpc.K.pt\endcsname{$+$0.89}
\expandafter\def\csname v@h1.kernel.qwen.pool_evpc.ratio\endcsname{1.02 [0.96, 1.11]}
\expandafter\def\csname v@h1.kernel.qwen.pool_evpc.ratio.pt\endcsname{1.02}
\expandafter\def\csname v@h1.kernel.qwen.pool_evpc_raw.K0\endcsname{$+$0.42 [0.29, 0.56]}
\expandafter\def\csname v@h1.kernel.qwen.pool_evpc_raw.K0.pt\endcsname{$+$0.42}
\expandafter\def\csname v@h1.kernel.qwen.pool_evpc_raw.K\endcsname{$+$0.08 [0.02, 0.16]}
\expandafter\def\csname v@h1.kernel.qwen.pool_evpc_raw.K.pt\endcsname{$+$0.08}
\expandafter\def\csname v@h1.kernel.qwen.pool_evpc_raw.ratio\endcsname{0.19 [0.04, 0.39]}
\expandafter\def\csname v@h1.kernel.qwen.pool_evpc_raw.ratio.pt\endcsname{0.19}
\expandafter\def\csname v@h1.kernel.qwen.balanced_evpc_raw.K0\endcsname{$+$0.70 [0.56, 0.83]}
\expandafter\def\csname v@h1.kernel.qwen.balanced_evpc_raw.K0.pt\endcsname{$+$0.70}
\expandafter\def\csname v@h1.kernel.qwen.balanced_evpc_raw.K\endcsname{$+$0.27 [0.13, 0.41]}
\expandafter\def\csname v@h1.kernel.qwen.balanced_evpc_raw.K.pt\endcsname{$+$0.27}
\expandafter\def\csname v@h1.kernel.qwen.balanced_evpc_raw.ratio\endcsname{0.38 [0.20, 0.56]}
\expandafter\def\csname v@h1.kernel.qwen.balanced_evpc_raw.ratio.pt\endcsname{0.38}
\expandafter\def\csname v@h1.kernel.qwen.debate_evpc_raw.K0\endcsname{$+$0.42 [0.27, 0.57]}
\expandafter\def\csname v@h1.kernel.qwen.debate_evpc_raw.K0.pt\endcsname{$+$0.42}
\expandafter\def\csname v@h1.kernel.qwen.debate_evpc_raw.K\endcsname{$+$0.10 [0.04, 0.17]}
\expandafter\def\csname v@h1.kernel.qwen.debate_evpc_raw.K.pt\endcsname{$+$0.10}
\expandafter\def\csname v@h1.kernel.qwen.debate_evpc_raw.ratio\endcsname{0.24 [0.11, 0.39]}
\expandafter\def\csname v@h1.kernel.qwen.debate_evpc_raw.ratio.pt\endcsname{0.24}
\expandafter\def\csname v@h1.kernel.qwen.balanced_evpc_rand_raw.K0\endcsname{$+$0.65 [0.50, 0.79]}
\expandafter\def\csname v@h1.kernel.qwen.balanced_evpc_rand_raw.K0.pt\endcsname{$+$0.65}
\expandafter\def\csname v@h1.kernel.qwen.balanced_evpc_rand_raw.K\endcsname{$+$0.25 [0.14, 0.38]}
\expandafter\def\csname v@h1.kernel.qwen.balanced_evpc_rand_raw.K.pt\endcsname{$+$0.25}
\expandafter\def\csname v@h1.kernel.qwen.balanced_evpc_rand_raw.ratio\endcsname{0.39 [0.24, 0.55]}
\expandafter\def\csname v@h1.kernel.qwen.balanced_evpc_rand_raw.ratio.pt\endcsname{0.39}
\expandafter\def\csname v@h1.hb16.olmo.pool_evpc\endcsname{0.88}
\expandafter\def\csname v@h1.hb16.olmo.debate_evpc\endcsname{0.81}
\expandafter\def\csname v@h1.hb16.olmo.balanced_evpc_rand\endcsname{0.80}
\expandafter\def\csname v@h1.hb16.olmo.balanced_evpc\endcsname{0.90}
\expandafter\def\csname v@h1.hb16.olmo.pool_evpc_raw\endcsname{0.54}
\expandafter\def\csname v@h1.hb16.olmo.debate_evpc_raw\endcsname{0.46}
\expandafter\def\csname v@h1.hb16.olmo.balanced_nu05_evpc_raw\endcsname{0.52}
\expandafter\def\csname v@h1.hb16.olmo.balanced_evpc_rand_raw\endcsname{0.54}
\expandafter\def\csname v@h1.hb16.olmo.balanced_evpc_raw\endcsname{0.63}
\expandafter\def\csname v@h1.hb16.olmo.debate\endcsname{0.21}
\expandafter\def\csname v@h1.hb16.olmo.balanced\endcsname{0.31}
\expandafter\def\csname v@h1.hb16.olmo.solo\endcsname{0.11}
\expandafter\def\csname v@h1.hb16.olmo.bal_vs_pool\endcsname{$+$0.09 [$-$0.02, 0.19]}
\expandafter\def\csname v@h1.hb16.olmo.bal_vs_pool.pt\endcsname{$+$0.09}
\expandafter\def\csname v@h1.hb16.olmo.rand_vs_deb\endcsname{$+$0.08 [$-$0.07, 0.23]}
\expandafter\def\csname v@h1.hb16.olmo.rand_vs_deb.pt\endcsname{$+$0.08}
\expandafter\def\csname v@h1.hb16.olmo.rank\endcsname{$+$0.09 [$-$0.04, 0.21]}
\expandafter\def\csname v@h1.hb16.olmo.rank.pt\endcsname{$+$0.09}
\expandafter\def\csname v@h1.hb16.olmo.cost\endcsname{$+$0.27}
\expandafter\def\csname v@h1.hb16.olmo.costpool\endcsname{$+$0.33}
\expandafter\def\csname v@h1.hb16.olmo.gain.pool_evpc_raw\endcsname{0.83 [0.76, 0.91]}
\expandafter\def\csname v@h1.hb16.olmo.gain.pool_evpc_raw.pt\endcsname{0.83}
\expandafter\def\csname v@h1.hb16.olmo.swto.pool_evpc_raw\endcsname{0.48}
\expandafter\def\csname v@h1.hb16.olmo.swaway.pool_evpc_raw\endcsname{0.06}
\expandafter\def\csname v@h1.hb16.olmo.gain.balanced_evpc_raw\endcsname{0.19 [0.07, 0.30]}
\expandafter\def\csname v@h1.hb16.olmo.gain.balanced_evpc_raw.pt\endcsname{0.19}
\expandafter\def\csname v@h1.hb16.olmo.swto.balanced_evpc_raw\endcsname{1.04}
\expandafter\def\csname v@h1.hb16.olmo.swaway.balanced_evpc_raw\endcsname{0.55}
\expandafter\def\csname v@h1.hb16.olmo.gain.debate_evpc_raw\endcsname{0.87 [0.75, 0.95]}
\expandafter\def\csname v@h1.hb16.olmo.gain.debate_evpc_raw.pt\endcsname{0.87}
\expandafter\def\csname v@h1.hb16.olmo.swto.debate_evpc_raw\endcsname{0.43}
\expandafter\def\csname v@h1.hb16.olmo.swaway.debate_evpc_raw\endcsname{0.09}
\expandafter\def\csname v@h1.hb16.olmo.gain.balanced_evpc_rand_raw\endcsname{0.37 [0.26, 0.47]}
\expandafter\def\csname v@h1.hb16.olmo.gain.balanced_evpc_rand_raw.pt\endcsname{0.37}
\expandafter\def\csname v@h1.hb16.olmo.swto.balanced_evpc_rand_raw\endcsname{0.83}
\expandafter\def\csname v@h1.hb16.olmo.swaway.balanced_evpc_rand_raw\endcsname{0.43}
\expandafter\def\csname v@h1.hb32.olmo.pool_evpc\endcsname{0.92}
\expandafter\def\csname v@h1.hb32.olmo.debate_evpc\endcsname{0.57}
\expandafter\def\csname v@h1.hb32.olmo.balanced_evpc_rand\endcsname{0.62}
\expandafter\def\csname v@h1.hb32.olmo.balanced_evpc\endcsname{0.90}
\expandafter\def\csname v@h1.hb32.olmo.pool_evpc_raw\endcsname{0.48}
\expandafter\def\csname v@h1.hb32.olmo.debate_evpc_raw\endcsname{0.41}
\expandafter\def\csname v@h1.hb32.olmo.balanced_evpc_rand_raw\endcsname{0.46}
\expandafter\def\csname v@h1.hb32.olmo.balanced_evpc_raw\endcsname{0.72}
\expandafter\def\csname v@h1.hb32.olmo.debate\endcsname{0.21}
\expandafter\def\csname v@h1.hb32.olmo.balanced\endcsname{0.34}
\expandafter\def\csname v@h1.hb32.olmo.solo\endcsname{0.09}
\expandafter\def\csname v@h1.hb32.olmo.bal_vs_pool\endcsname{$+$0.23 [0.11, 0.36]}
\expandafter\def\csname v@h1.hb32.olmo.bal_vs_pool.pt\endcsname{$+$0.23}
\expandafter\def\csname v@h1.hb32.olmo.rand_vs_deb\endcsname{$+$0.05 [$-$0.06, 0.17]}
\expandafter\def\csname v@h1.hb32.olmo.rand_vs_deb.pt\endcsname{$+$0.05}
\expandafter\def\csname v@h1.hb32.olmo.rank\endcsname{$+$0.26 [0.13, 0.39]}
\expandafter\def\csname v@h1.hb32.olmo.rank.pt\endcsname{$+$0.26}
\expandafter\def\csname v@h1.hb32.olmo.cost\endcsname{$+$0.18}
\expandafter\def\csname v@h1.hb32.olmo.costpool\endcsname{$+$0.43}
\expandafter\def\csname v@h1.hb32.olmo.gain.pool_evpc_raw\endcsname{0.66 [0.56, 0.77]}
\expandafter\def\csname v@h1.hb32.olmo.gain.pool_evpc_raw.pt\endcsname{0.66}
\expandafter\def\csname v@h1.hb32.olmo.swto.pool_evpc_raw\endcsname{0.49}
\expandafter\def\csname v@h1.hb32.olmo.swaway.pool_evpc_raw\endcsname{0.10}
\expandafter\def\csname v@h1.hb32.olmo.gain.balanced_evpc_raw\endcsname{0.20 [0.09, 0.29]}
\expandafter\def\csname v@h1.hb32.olmo.gain.balanced_evpc_raw.pt\endcsname{0.20}
\expandafter\def\csname v@h1.hb32.olmo.swto.balanced_evpc_raw\endcsname{1.08}
\expandafter\def\csname v@h1.hb32.olmo.swaway.balanced_evpc_raw\endcsname{0.47}
\expandafter\def\csname v@h1.hb32.olmo.gain.debate_evpc_raw\endcsname{0.92 [0.82, 1.03]}
\expandafter\def\csname v@h1.hb32.olmo.gain.debate_evpc_raw.pt\endcsname{0.92}
\expandafter\def\csname v@h1.hb32.olmo.swto.debate_evpc_raw\endcsname{0.38}
\expandafter\def\csname v@h1.hb32.olmo.swaway.debate_evpc_raw\endcsname{0.06}
\expandafter\def\csname v@h1.hb32.olmo.gain.balanced_evpc_rand_raw\endcsname{0.41 [0.29, 0.51]}
\expandafter\def\csname v@h1.hb32.olmo.gain.balanced_evpc_rand_raw.pt\endcsname{0.41}
\expandafter\def\csname v@h1.hb32.olmo.swto.balanced_evpc_rand_raw\endcsname{0.97}
\expandafter\def\csname v@h1.hb32.olmo.swaway.balanced_evpc_rand_raw\endcsname{0.65}
\expandafter\def\csname v@h1.pz16.olmo.pool_evpc\endcsname{0.75}
\expandafter\def\csname v@h1.pz16.olmo.debate_evpc\endcsname{0.44}
\expandafter\def\csname v@h1.pz16.olmo.balanced_evpc_rand\endcsname{0.36}
\expandafter\def\csname v@h1.pz16.olmo.balanced_evpc\endcsname{0.70}
\expandafter\def\csname v@h1.pz16.olmo.pool_evpc_raw\endcsname{0.13}
\expandafter\def\csname v@h1.pz16.olmo.debate_evpc_raw\endcsname{0.15}
\expandafter\def\csname v@h1.pz16.olmo.balanced_nu05_evpc_raw\endcsname{0.17}
\expandafter\def\csname v@h1.pz16.olmo.balanced_evpc_rand_raw\endcsname{0.19}
\expandafter\def\csname v@h1.pz16.olmo.balanced_evpc_raw\endcsname{0.19}
\expandafter\def\csname v@h1.pz16.olmo.debate\endcsname{0.08}
\expandafter\def\csname v@h1.pz16.olmo.balanced\endcsname{0.14}
\expandafter\def\csname v@h1.pz16.olmo.solo\endcsname{0.04}
\expandafter\def\csname v@h1.pz16.olmo.bal_vs_pool\endcsname{$+$0.07 [0.03, 0.11]}
\expandafter\def\csname v@h1.pz16.olmo.bal_vs_pool.pt\endcsname{$+$0.07}
\expandafter\def\csname v@h1.pz16.olmo.rand_vs_deb\endcsname{$+$0.05 [$-$0.01, 0.10]}
\expandafter\def\csname v@h1.pz16.olmo.rand_vs_deb.pt\endcsname{$+$0.05}
\expandafter\def\csname v@h1.pz16.olmo.rank\endcsname{0.00 [$-$0.04, 0.04]}
\expandafter\def\csname v@h1.pz16.olmo.rank.pt\endcsname{0.00}
\expandafter\def\csname v@h1.pz16.olmo.cost\endcsname{$+$0.51}
\expandafter\def\csname v@h1.pz16.olmo.costpool\endcsname{$+$0.62}
\expandafter\def\csname v@h1.pz16.olmo.gain.pool_evpc_raw\endcsname{0.84 [0.64, 0.99]}
\expandafter\def\csname v@h1.pz16.olmo.gain.pool_evpc_raw.pt\endcsname{0.84}
\expandafter\def\csname v@h1.pz16.olmo.swto.pool_evpc_raw\endcsname{0.11}
\expandafter\def\csname v@h1.pz16.olmo.swaway.pool_evpc_raw\endcsname{0.01}
\expandafter\def\csname v@h1.pz16.olmo.gain.balanced_evpc_raw\endcsname{0.20 [0.08, 0.29]}
\expandafter\def\csname v@h1.pz16.olmo.gain.balanced_evpc_raw.pt\endcsname{0.20}
\expandafter\def\csname v@h1.pz16.olmo.swto.balanced_evpc_raw\endcsname{0.40}
\expandafter\def\csname v@h1.pz16.olmo.swaway.balanced_evpc_raw\endcsname{0.25}
\expandafter\def\csname v@h1.pz16.olmo.gain.debate_evpc_raw\endcsname{0.92 [0.78, 1.05]}
\expandafter\def\csname v@h1.pz16.olmo.gain.debate_evpc_raw.pt\endcsname{0.92}
\expandafter\def\csname v@h1.pz16.olmo.swto.debate_evpc_raw\endcsname{0.13}
\expandafter\def\csname v@h1.pz16.olmo.swaway.debate_evpc_raw\endcsname{0.02}
\expandafter\def\csname v@h1.pz16.olmo.gain.balanced_evpc_rand_raw\endcsname{0.35 [0.21, 0.47]}
\expandafter\def\csname v@h1.pz16.olmo.gain.balanced_evpc_rand_raw.pt\endcsname{0.35}
\expandafter\def\csname v@h1.pz16.olmo.swto.balanced_evpc_rand_raw\endcsname{0.35}
\expandafter\def\csname v@h1.pz16.olmo.swaway.balanced_evpc_rand_raw\endcsname{0.20}
\expandafter\def\csname v@h1.late16.olmo.pool_evpc\endcsname{0.85}
\expandafter\def\csname v@h1.late16.olmo.debate_evpc\endcsname{0.71}
\expandafter\def\csname v@h1.late16.olmo.balanced_evpc\endcsname{0.87}
\expandafter\def\csname v@h1.late16.olmo.pool_evpc_raw\endcsname{0.23}
\expandafter\def\csname v@h1.late16.olmo.debate_evpc_raw\endcsname{0.20}
\expandafter\def\csname v@h1.late16.olmo.balanced_evpc_rand_raw\endcsname{0.40}
\expandafter\def\csname v@h1.late16.olmo.balanced_evpc_raw\endcsname{0.44}
\expandafter\def\csname v@h1.late16.olmo.debate\endcsname{0.16}
\expandafter\def\csname v@h1.late16.olmo.solo\endcsname{0.10}
\expandafter\def\csname v@h1.late16.olmo.bal_vs_pool\endcsname{$+$0.21 [0.09, 0.32]}
\expandafter\def\csname v@h1.late16.olmo.bal_vs_pool.pt\endcsname{$+$0.21}
\expandafter\def\csname v@h1.late16.olmo.rand_vs_deb\endcsname{$+$0.20 [0.07, 0.34]}
\expandafter\def\csname v@h1.late16.olmo.rand_vs_deb.pt\endcsname{$+$0.20}
\expandafter\def\csname v@h1.late16.olmo.rank\endcsname{$+$0.04 [$-$0.06, 0.14]}
\expandafter\def\csname v@h1.late16.olmo.rank.pt\endcsname{$+$0.04}
\expandafter\def\csname v@h1.late16.olmo.cost\endcsname{$+$0.43}
\expandafter\def\csname v@h1.late16.olmo.costpool\endcsname{$+$0.61}
\expandafter\def\csname v@h1.late16.olmo.gain.pool_evpc_raw\endcsname{0.75 [0.50, 0.94]}
\expandafter\def\csname v@h1.late16.olmo.gain.pool_evpc_raw.pt\endcsname{0.75}
\expandafter\def\csname v@h1.late16.olmo.swto.pool_evpc_raw\endcsname{0.22}
\expandafter\def\csname v@h1.late16.olmo.swaway.pool_evpc_raw\endcsname{0.08}
\expandafter\def\csname v@h1.late16.olmo.gain.balanced_evpc_raw\endcsname{0.44 [0.29, 0.57]}
\expandafter\def\csname v@h1.late16.olmo.gain.balanced_evpc_raw.pt\endcsname{0.44}
\expandafter\def\csname v@h1.late16.olmo.swto.balanced_evpc_raw\endcsname{0.62}
\expandafter\def\csname v@h1.late16.olmo.swaway.balanced_evpc_raw\endcsname{0.30}
\expandafter\def\csname v@h1.late16.olmo.gain.debate_evpc_raw\endcsname{0.95 [0.63, 1.10]}
\expandafter\def\csname v@h1.late16.olmo.gain.debate_evpc_raw.pt\endcsname{0.95}
\expandafter\def\csname v@h1.late16.olmo.swto.debate_evpc_raw\endcsname{0.19}
\expandafter\def\csname v@h1.late16.olmo.swaway.debate_evpc_raw\endcsname{0.08}
\expandafter\def\csname v@h1.late16.olmo.gain.balanced_evpc_rand_raw\endcsname{0.37 [0.24, 0.49]}
\expandafter\def\csname v@h1.late16.olmo.gain.balanced_evpc_rand_raw.pt\endcsname{0.37}
\expandafter\def\csname v@h1.late16.olmo.swto.balanced_evpc_rand_raw\endcsname{0.66}
\expandafter\def\csname v@h1.late16.olmo.swaway.balanced_evpc_rand_raw\endcsname{0.36}
\expandafter\def\csname v@h1.kernel.olmo.pool_evpc.K0\endcsname{$+$0.77 [0.59, 0.92]}
\expandafter\def\csname v@h1.kernel.olmo.pool_evpc.K0.pt\endcsname{$+$0.77}
\expandafter\def\csname v@h1.kernel.olmo.pool_evpc.K\endcsname{$+$0.74 [0.56, 0.90]}
\expandafter\def\csname v@h1.kernel.olmo.pool_evpc.K.pt\endcsname{$+$0.74}
\expandafter\def\csname v@h1.kernel.olmo.pool_evpc.ratio\endcsname{0.97 [0.90, 1.04]}
\expandafter\def\csname v@h1.kernel.olmo.pool_evpc.ratio.pt\endcsname{0.97}
\expandafter\def\csname v@h1.kernel.olmo.pool_evpc_raw.K0\endcsname{$+$0.43 [0.26, 0.60]}
\expandafter\def\csname v@h1.kernel.olmo.pool_evpc_raw.K0.pt\endcsname{$+$0.43}
\expandafter\def\csname v@h1.kernel.olmo.pool_evpc_raw.K\endcsname{$+$0.13 [0.02, 0.24]}
\expandafter\def\csname v@h1.kernel.olmo.pool_evpc_raw.K.pt\endcsname{$+$0.13}
\expandafter\def\csname v@h1.kernel.olmo.pool_evpc_raw.ratio\endcsname{0.30 [0.07, 0.49]}
\expandafter\def\csname v@h1.kernel.olmo.pool_evpc_raw.ratio.pt\endcsname{0.30}
\expandafter\def\csname v@h1.kernel.olmo.balanced_evpc_raw.K0\endcsname{$+$0.52 [0.33, 0.69]}
\expandafter\def\csname v@h1.kernel.olmo.balanced_evpc_raw.K0.pt\endcsname{$+$0.52}
\expandafter\def\csname v@h1.kernel.olmo.balanced_evpc_raw.K\endcsname{$+$0.34 [0.19, 0.48]}
\expandafter\def\csname v@h1.kernel.olmo.balanced_evpc_raw.K.pt\endcsname{$+$0.34}
\expandafter\def\csname v@h1.kernel.olmo.balanced_evpc_raw.ratio\endcsname{0.65 [0.43, 0.90]}
\expandafter\def\csname v@h1.kernel.olmo.balanced_evpc_raw.ratio.pt\endcsname{0.65}
\expandafter\def\csname v@h1.kernel.olmo.debate_evpc_raw.K0\endcsname{$+$0.35 [0.21, 0.49]}
\expandafter\def\csname v@h1.kernel.olmo.debate_evpc_raw.K0.pt\endcsname{$+$0.35}
\expandafter\def\csname v@h1.kernel.olmo.debate_evpc_raw.K\endcsname{$+$0.09 [0.00, 0.20]}
\expandafter\def\csname v@h1.kernel.olmo.debate_evpc_raw.K.pt\endcsname{$+$0.09}
\expandafter\def\csname v@h1.kernel.olmo.debate_evpc_raw.ratio\endcsname{0.27 [0.00, 0.55]}
\expandafter\def\csname v@h1.kernel.olmo.debate_evpc_raw.ratio.pt\endcsname{0.27}
\expandafter\def\csname v@h1.kernel.olmo.balanced_evpc_rand_raw.K0\endcsname{$+$0.43 [0.24, 0.60]}
\expandafter\def\csname v@h1.kernel.olmo.balanced_evpc_rand_raw.K0.pt\endcsname{$+$0.43}
\expandafter\def\csname v@h1.kernel.olmo.balanced_evpc_rand_raw.K\endcsname{$+$0.30 [0.14, 0.45]}
\expandafter\def\csname v@h1.kernel.olmo.balanced_evpc_rand_raw.K.pt\endcsname{$+$0.30}
\expandafter\def\csname v@h1.kernel.olmo.balanced_evpc_rand_raw.ratio\endcsname{0.69 [0.42, 0.98]}
\expandafter\def\csname v@h1.kernel.olmo.balanced_evpc_rand_raw.ratio.pt\endcsname{0.69}
\expandafter\def\csname v@h1.hb16.gemma.rand_vs_deb.pt\endcsname{$+$0.11}
\expandafter\def\csname v@h1.hb16.gemma.rank.pt\endcsname{$+$0.08}
\expandafter\def\csname v@h1.hb16.gemma.bal_vs_pool\endcsname{$+$0.14 [0.01, 0.28]}
\expandafter\def\csname v@h1.hb16.gemma.bal_vs_pool.pt\endcsname{$+$0.14}
\expandafter\def\csname v@h1.hb16.mini.rand_vs_deb.pt\endcsname{$-$0.04}
\expandafter\def\csname v@h1.hb16.mini.rank.pt\endcsname{$-$0.05}
\expandafter\def\csname v@h1.hb16.mini.bal_vs_pool\endcsname{$-$0.12 [$-$0.23, $-$0.01]}
\expandafter\def\csname v@h1.hb16.mini.bal_vs_pool.pt\endcsname{$-$0.12}
\expandafter\def\csname v@h1.bal_vs_pool.range\endcsname{0.07 to 0.47}
\expandafter\def\csname v@h1.rand_vs_deb.range\endcsname{$-$0.01 to $+$0.23}
\expandafter\def\csname v@h1.costratio.range\endcsname{0.22 to 0.82}
\expandafter\def\csname v@h1.rank.range\endcsname{0.00 to 0.26}
\expandafter\def\csname v@h1.size.range\endcsname{$-$0.08 to $+$0.06}
\expandafter\def\csname v@h2.hb.qwen.tok.pool_evpc_cs2\endcsname{22k}
\expandafter\def\csname v@h2.hb.qwen.tok.pool_evpc_cs4\endcsname{42k}
\expandafter\def\csname v@h2.hb.qwen.tok.pool_evpc_ch4\endcsname{40k}
\expandafter\def\csname v@h2.hb.qwen.pool_evpc\endcsname{0.88}
\expandafter\def\csname v@h2.hb.qwen.pool_evpc_cs1\endcsname{0.85}
\expandafter\def\csname v@h2.hb.qwen.pool_evpc_cs1.d\endcsname{$-$0.03}
\expandafter\def\csname v@h2.hb.qwen.pool_evpc_cs2\endcsname{0.86}
\expandafter\def\csname v@h2.hb.qwen.pool_evpc_cs2.d\endcsname{$-$0.02}
\expandafter\def\csname v@h2.hb.qwen.pool_evpc_cs4\endcsname{0.82}
\expandafter\def\csname v@h2.hb.qwen.pool_evpc_cs4.d\endcsname{$-$0.06}
\expandafter\def\csname v@h2.hb.qwen.pool_evpc_ch4\endcsname{0.89}
\expandafter\def\csname v@h2.hb.qwen.pool_evpc_ch4.d\endcsname{$+$0.01}
\expandafter\def\csname v@h2.hb.qwen.pool_evpc_cs4ch4\endcsname{0.82}
\expandafter\def\csname v@h2.hb.qwen.pool_evpc_cs4ch4.d\endcsname{$-$0.06}
\expandafter\def\csname v@h2.hb.qwen.pool_ev\endcsname{0.85}
\expandafter\def\csname v@h2.hb.qwen.pool_ev.d\endcsname{$-$0.03}
\expandafter\def\csname v@h2.hb.qwen.pool_ev_nodedup\endcsname{0.70}
\expandafter\def\csname v@h2.hb.qwen.pool_ev_nodedup.d\endcsname{$-$0.18}
\expandafter\def\csname v@h2.pzs.qwen.pool_evpc\endcsname{0.87}
\expandafter\def\csname v@h2.pzs.qwen.pool_evpc_cs1\endcsname{0.69}
\expandafter\def\csname v@h2.pzs.qwen.pool_evpc_cs1.d\endcsname{$-$0.18}
\expandafter\def\csname v@h2.pzs.qwen.pool_evpc_cs2\endcsname{0.53}
\expandafter\def\csname v@h2.pzs.qwen.pool_evpc_cs2.d\endcsname{$-$0.34}
\expandafter\def\csname v@h2.pzs.qwen.pool_evpc_cs4\endcsname{0.16}
\expandafter\def\csname v@h2.pzs.qwen.pool_evpc_cs4.d\endcsname{$-$0.71}
\expandafter\def\csname v@h2.pzs.qwen.pool_evpc_ch4\endcsname{0.88}
\expandafter\def\csname v@h2.pzs.qwen.pool_evpc_ch4.d\endcsname{$+$0.01}
\expandafter\def\csname v@h2.pzs.qwen.pool_evpc_cs4ch4\endcsname{0.73}
\expandafter\def\csname v@h2.pzs.qwen.pool_evpc_cs4ch4.d\endcsname{$-$0.14}
\expandafter\def\csname v@h2.pzs.qwen.pool_ev\endcsname{0.77}
\expandafter\def\csname v@h2.pzs.qwen.pool_ev.d\endcsname{$-$0.10}
\expandafter\def\csname v@h2.pzs.qwen.pool_ev_nodedup\endcsname{0.01}
\expandafter\def\csname v@h2.pzs.qwen.pool_ev_nodedup.d\endcsname{$-$0.86}
\expandafter\def\csname v@h2.pzw.qwen.pool_evpc\endcsname{0.84}
\expandafter\def\csname v@h2.pzw.qwen.pool_evpc_cs1\endcsname{0.69}
\expandafter\def\csname v@h2.pzw.qwen.pool_evpc_cs1.d\endcsname{$-$0.15}
\expandafter\def\csname v@h2.pzw.qwen.pool_evpc_cs2\endcsname{0.62}
\expandafter\def\csname v@h2.pzw.qwen.pool_evpc_cs2.d\endcsname{$-$0.21}
\expandafter\def\csname v@h2.pzw.qwen.pool_evpc_cs4\endcsname{0.52}
\expandafter\def\csname v@h2.pzw.qwen.pool_evpc_cs4.d\endcsname{$-$0.32}
\expandafter\def\csname v@h2.pzw.qwen.pool_evpc_ch4\endcsname{0.90}
\expandafter\def\csname v@h2.pzw.qwen.pool_evpc_ch4.d\endcsname{$+$0.07}
\expandafter\def\csname v@h2.pzw.qwen.pool_evpc_cs4ch4\endcsname{0.71}
\expandafter\def\csname v@h2.pzw.qwen.pool_evpc_cs4ch4.d\endcsname{$-$0.12}
\expandafter\def\csname v@h2.pzw.qwen.pool_ev\endcsname{0.75}
\expandafter\def\csname v@h2.pzw.qwen.pool_ev.d\endcsname{$-$0.09}
\expandafter\def\csname v@h2.pzw.qwen.pool_ev_nodedup\endcsname{0.34}
\expandafter\def\csname v@h2.pzw.qwen.pool_ev_nodedup.d\endcsname{$-$0.50}
\expandafter\def\csname v@h2.pz.qwen.tok.pool_evpc_cs2\endcsname{17k}
\expandafter\def\csname v@h2.pz.qwen.tok.pool_evpc_cs4\endcsname{33k}
\expandafter\def\csname v@h2.pz.qwen.tok.pool_evpc_ch4\endcsname{22k}
\expandafter\def\csname v@h2.pz.qwen.pool_evpc\endcsname{0.85}
\expandafter\def\csname v@h2.pz.qwen.pool_evpc_cs1\endcsname{0.69}
\expandafter\def\csname v@h2.pz.qwen.pool_evpc_cs1.d\endcsname{$-$0.17}
\expandafter\def\csname v@h2.pz.qwen.pool_evpc_cs2\endcsname{0.58}
\expandafter\def\csname v@h2.pz.qwen.pool_evpc_cs2.d\endcsname{$-$0.28}
\expandafter\def\csname v@h2.pz.qwen.pool_evpc_cs4\endcsname{0.34}
\expandafter\def\csname v@h2.pz.qwen.pool_evpc_cs4.d\endcsname{$-$0.51}
\expandafter\def\csname v@h2.pz.qwen.pool_evpc_ch4\endcsname{0.89}
\expandafter\def\csname v@h2.pz.qwen.pool_evpc_ch4.d\endcsname{$+$0.04}
\expandafter\def\csname v@h2.pz.qwen.pool_evpc_cs4ch4\endcsname{0.72}
\expandafter\def\csname v@h2.pz.qwen.pool_evpc_cs4ch4.d\endcsname{$-$0.13}
\expandafter\def\csname v@h2.pz.qwen.pool_ev\endcsname{0.76}
\expandafter\def\csname v@h2.pz.qwen.pool_ev.d\endcsname{$-$0.09}
\expandafter\def\csname v@h2.pz.qwen.pool_ev_nodedup\endcsname{0.18}
\expandafter\def\csname v@h2.pz.qwen.pool_ev_nodedup.d\endcsname{$-$0.68}
\expandafter\def\csname v@h2.hb.qwen.cs4.ci\endcsname{$-$0.06 [$-$0.15, 0.01]}
\expandafter\def\csname v@h2.pz.qwen.cs4.ci\endcsname{$-$0.51 [$-$0.59, $-$0.43]}
\expandafter\def\csname v@h2.hb.qwen.ch4.ci\endcsname{$+$0.01 [$-$0.05, 0.07]}
\expandafter\def\csname v@h2.pz.qwen.ch4.ci\endcsname{$+$0.04 [0.00, 0.07]}
\expandafter\def\csname v@h2.hb.qwen.offset.ci\endcsname{0.00 [0.00, 0.00]}
\expandafter\def\csname v@h2.pz.qwen.offset.ci\endcsname{$+$0.38 [0.30, 0.45]}
\expandafter\def\csname v@h2.hb.olmo.tok.pool_evpc_cs2\endcsname{23k}
\expandafter\def\csname v@h2.hb.olmo.tok.pool_evpc_cs4\endcsname{43k}
\expandafter\def\csname v@h2.hb.olmo.tok.pool_evpc_ch4\endcsname{37k}
\expandafter\def\csname v@h2.hb.olmo.pool_evpc\endcsname{0.88}
\expandafter\def\csname v@h2.hb.olmo.pool_evpc_cs1\endcsname{0.89}
\expandafter\def\csname v@h2.hb.olmo.pool_evpc_cs1.d\endcsname{$+$0.02}
\expandafter\def\csname v@h2.hb.olmo.pool_evpc_cs2\endcsname{0.81}
\expandafter\def\csname v@h2.hb.olmo.pool_evpc_cs2.d\endcsname{$-$0.06}
\expandafter\def\csname v@h2.hb.olmo.pool_evpc_cs4\endcsname{0.73}
\expandafter\def\csname v@h2.hb.olmo.pool_evpc_cs4.d\endcsname{$-$0.14}
\expandafter\def\csname v@h2.hb.olmo.pool_evpc_ch4\endcsname{0.90}
\expandafter\def\csname v@h2.hb.olmo.pool_evpc_ch4.d\endcsname{$+$0.02}
\expandafter\def\csname v@h2.hb.olmo.pool_evpc_cs4ch4\endcsname{0.76}
\expandafter\def\csname v@h2.hb.olmo.pool_evpc_cs4ch4.d\endcsname{$-$0.12}
\expandafter\def\csname v@h2.hb.olmo.pool_ev\endcsname{0.86}
\expandafter\def\csname v@h2.hb.olmo.pool_ev.d\endcsname{$-$0.02}
\expandafter\def\csname v@h2.hb.olmo.pool_ev_nodedup\endcsname{0.54}
\expandafter\def\csname v@h2.hb.olmo.pool_ev_nodedup.d\endcsname{$-$0.34}
\expandafter\def\csname v@h2.pzs.olmo.pool_evpc\endcsname{0.76}
\expandafter\def\csname v@h2.pzs.olmo.pool_evpc_cs1\endcsname{0.44}
\expandafter\def\csname v@h2.pzs.olmo.pool_evpc_cs1.d\endcsname{$-$0.32}
\expandafter\def\csname v@h2.pzs.olmo.pool_evpc_cs2\endcsname{0.17}
\expandafter\def\csname v@h2.pzs.olmo.pool_evpc_cs2.d\endcsname{$-$0.59}
\expandafter\def\csname v@h2.pzs.olmo.pool_evpc_cs4\endcsname{0.03}
\expandafter\def\csname v@h2.pzs.olmo.pool_evpc_cs4.d\endcsname{$-$0.73}
\expandafter\def\csname v@h2.pzs.olmo.pool_evpc_ch4\endcsname{0.81}
\expandafter\def\csname v@h2.pzs.olmo.pool_evpc_ch4.d\endcsname{$+$0.05}
\expandafter\def\csname v@h2.pzs.olmo.pool_evpc_cs4ch4\endcsname{0.55}
\expandafter\def\csname v@h2.pzs.olmo.pool_evpc_cs4ch4.d\endcsname{$-$0.22}
\expandafter\def\csname v@h2.pzs.olmo.pool_ev\endcsname{0.62}
\expandafter\def\csname v@h2.pzs.olmo.pool_ev.d\endcsname{$-$0.14}
\expandafter\def\csname v@h2.pzs.olmo.pool_ev_nodedup\endcsname{0.00}
\expandafter\def\csname v@h2.pzs.olmo.pool_ev_nodedup.d\endcsname{$-$0.76}
\expandafter\def\csname v@h2.pzw.olmo.pool_evpc\endcsname{0.74}
\expandafter\def\csname v@h2.pzw.olmo.pool_evpc_cs1\endcsname{0.50}
\expandafter\def\csname v@h2.pzw.olmo.pool_evpc_cs1.d\endcsname{$-$0.24}
\expandafter\def\csname v@h2.pzw.olmo.pool_evpc_cs2\endcsname{0.37}
\expandafter\def\csname v@h2.pzw.olmo.pool_evpc_cs2.d\endcsname{$-$0.37}
\expandafter\def\csname v@h2.pzw.olmo.pool_evpc_cs4\endcsname{0.22}
\expandafter\def\csname v@h2.pzw.olmo.pool_evpc_cs4.d\endcsname{$-$0.53}
\expandafter\def\csname v@h2.pzw.olmo.pool_evpc_ch4\endcsname{0.73}
\expandafter\def\csname v@h2.pzw.olmo.pool_evpc_ch4.d\endcsname{$-$0.02}
\expandafter\def\csname v@h2.pzw.olmo.pool_evpc_cs4ch4\endcsname{0.57}
\expandafter\def\csname v@h2.pzw.olmo.pool_evpc_cs4ch4.d\endcsname{$-$0.18}
\expandafter\def\csname v@h2.pzw.olmo.pool_ev\endcsname{0.59}
\expandafter\def\csname v@h2.pzw.olmo.pool_ev.d\endcsname{$-$0.15}
\expandafter\def\csname v@h2.pzw.olmo.pool_ev_nodedup\endcsname{0.13}
\expandafter\def\csname v@h2.pzw.olmo.pool_ev_nodedup.d\endcsname{$-$0.61}
\expandafter\def\csname v@h2.pz.olmo.tok.pool_evpc_cs2\endcsname{15k}
\expandafter\def\csname v@h2.pz.olmo.tok.pool_evpc_cs4\endcsname{30k}
\expandafter\def\csname v@h2.pz.olmo.tok.pool_evpc_ch4\endcsname{18k}
\expandafter\def\csname v@h2.pz.olmo.pool_evpc\endcsname{0.75}
\expandafter\def\csname v@h2.pz.olmo.pool_evpc_cs1\endcsname{0.47}
\expandafter\def\csname v@h2.pz.olmo.pool_evpc_cs1.d\endcsname{$-$0.28}
\expandafter\def\csname v@h2.pz.olmo.pool_evpc_cs2\endcsname{0.27}
\expandafter\def\csname v@h2.pz.olmo.pool_evpc_cs2.d\endcsname{$-$0.48}
\expandafter\def\csname v@h2.pz.olmo.pool_evpc_cs4\endcsname{0.12}
\expandafter\def\csname v@h2.pz.olmo.pool_evpc_cs4.d\endcsname{$-$0.63}
\expandafter\def\csname v@h2.pz.olmo.pool_evpc_ch4\endcsname{0.77}
\expandafter\def\csname v@h2.pz.olmo.pool_evpc_ch4.d\endcsname{$+$0.02}
\expandafter\def\csname v@h2.pz.olmo.pool_evpc_cs4ch4\endcsname{0.56}
\expandafter\def\csname v@h2.pz.olmo.pool_evpc_cs4ch4.d\endcsname{$-$0.20}
\expandafter\def\csname v@h2.pz.olmo.pool_ev\endcsname{0.61}
\expandafter\def\csname v@h2.pz.olmo.pool_ev.d\endcsname{$-$0.15}
\expandafter\def\csname v@h2.pz.olmo.pool_ev_nodedup\endcsname{0.07}
\expandafter\def\csname v@h2.pz.olmo.pool_ev_nodedup.d\endcsname{$-$0.68}
\expandafter\def\csname v@h2.hb.olmo.cs4.ci\endcsname{$-$0.14 [$-$0.27, $-$0.02]}
\expandafter\def\csname v@h2.pz.olmo.cs4.ci\endcsname{$-$0.63 [$-$0.71, $-$0.55]}
\expandafter\def\csname v@h2.hb.olmo.ch4.ci\endcsname{$+$0.02 [$-$0.04, 0.09]}
\expandafter\def\csname v@h2.pz.olmo.ch4.ci\endcsname{$+$0.02 [$-$0.03, 0.06]}
\expandafter\def\csname v@h2.hb.olmo.offset.ci\endcsname{$+$0.02 [$-$0.05, 0.09]}
\expandafter\def\csname v@h2.pz.olmo.offset.ci\endcsname{$+$0.43 [0.36, 0.51]}
\expandafter\def\csname v@h2.ch4.range\endcsname{$+$0.01 to $+$0.04}
\expandafter\def\csname v@h2.offset.range\endcsname{$+$0.38 to $+$0.43}
\expandafter\def\csname v@h3.qwen.pool.range\endcsname{0.85--0.89}
\expandafter\def\csname v@h3.qwen.pool.4\endcsname{0.89}
\expandafter\def\csname v@h3.qwen.nd.4\endcsname{0.45}
\expandafter\def\csname v@h3.qwen.gap.4\endcsname{$-$0.44 [$-$0.52, $-$0.36]}
\expandafter\def\csname v@h3.qwen.pool.8\endcsname{0.87}
\expandafter\def\csname v@h3.qwen.nd.8\endcsname{0.27}
\expandafter\def\csname v@h3.qwen.gap.8\endcsname{$-$0.60 [$-$0.68, $-$0.52]}
\expandafter\def\csname v@h3.qwen.pool.16\endcsname{0.85}
\expandafter\def\csname v@h3.qwen.nd.16\endcsname{0.18}
\expandafter\def\csname v@h3.qwen.gap.16\endcsname{$-$0.68 [$-$0.76, $-$0.60]}
\expandafter\def\csname v@h3.qwen.pool.32\endcsname{0.88}
\expandafter\def\csname v@h3.qwen.nd.32\endcsname{0.16}
\expandafter\def\csname v@h3.qwen.gap.32\endcsname{$-$0.72 [$-$0.80, $-$0.65]}
\expandafter\def\csname v@h3.olmo.pool.range\endcsname{0.75--0.78}
\expandafter\def\csname v@h3.olmo.pool.4\endcsname{0.76}
\expandafter\def\csname v@h3.olmo.nd.4\endcsname{0.20}
\expandafter\def\csname v@h3.olmo.gap.4\endcsname{$-$0.56 [$-$0.63, $-$0.49]}
\expandafter\def\csname v@h3.olmo.pool.8\endcsname{0.78}
\expandafter\def\csname v@h3.olmo.nd.8\endcsname{0.11}
\expandafter\def\csname v@h3.olmo.gap.8\endcsname{$-$0.66 [$-$0.74, $-$0.58]}
\expandafter\def\csname v@h3.olmo.pool.16\endcsname{0.75}
\expandafter\def\csname v@h3.olmo.nd.16\endcsname{0.07}
\expandafter\def\csname v@h3.olmo.gap.16\endcsname{$-$0.68 [$-$0.76, $-$0.61]}
\expandafter\def\csname v@h3.olmo.pool.32\endcsname{0.78}
\expandafter\def\csname v@h3.olmo.nd.32\endcsname{0.05}
\expandafter\def\csname v@h3.olmo.gap.32\endcsname{$-$0.74 [$-$0.81, $-$0.66]}
\expandafter\def\csname v@an.test.qwen.solo\endcsname{0.80}
\expandafter\def\csname v@an.test.qwen.debate\endcsname{0.87}
\expandafter\def\csname v@an.test.qwen.balanced\endcsname{0.68}
\expandafter\def\csname v@an.test.qwen.pool_evpc\endcsname{0.09}
\expandafter\def\csname v@an.test.qwen.full_iter\endcsname{0.12}
\expandafter\def\csname v@an.pair.qwen.debate\endcsname{$+$0.78 [0.65, 0.89]}
\expandafter\def\csname v@an.pair.qwen.full\endcsname{$-$0.03 [$-$0.12, 0.05]}
\expandafter\def\csname v@an.pair.qwen.balanced\endcsname{$+$0.59 [0.46, 0.72]}
\expandafter\def\csname v@an.test.olmo.solo\endcsname{0.79}
\expandafter\def\csname v@an.test.olmo.debate\endcsname{0.67}
\expandafter\def\csname v@an.test.olmo.balanced\endcsname{0.32}
\expandafter\def\csname v@an.test.olmo.pool_evpc\endcsname{0.10}
\expandafter\def\csname v@an.test.olmo.full_iter\endcsname{0.10}
\expandafter\def\csname v@an.pair.olmo.debate\endcsname{$+$0.57 [0.45, 0.68]}
\expandafter\def\csname v@an.pair.olmo.full\endcsname{$+$0.01 [$-$0.07, 0.09]}
\expandafter\def\csname v@an.pair.olmo.balanced\endcsname{$+$0.22 [0.10, 0.35]}
\expandafter\def\csname v@an.test.gemma.solo\endcsname{0.82}
\expandafter\def\csname v@an.test.gemma.debate\endcsname{0.60}
\expandafter\def\csname v@an.test.gemma.balanced\endcsname{0.40}
\expandafter\def\csname v@an.test.gemma.pool_evpc\endcsname{0.03}
\expandafter\def\csname v@an.test.gemma.full_iter\endcsname{0.01}
\expandafter\def\csname v@an.test.mini.solo\endcsname{0.70}
\expandafter\def\csname v@an.test.mini.debate\endcsname{0.41}
\expandafter\def\csname v@an.test.mini.balanced\endcsname{0.13}
\expandafter\def\csname v@an.test.mini.pool_evpc\endcsname{0.04}
\expandafter\def\csname v@an.test.mini.full_iter\endcsname{0.01}
\expandafter\def\csname v@an.hb.qwen.solo\endcsname{0.74}
\expandafter\def\csname v@an.hb.qwen.debate\endcsname{0.86}
\expandafter\def\csname v@an.hb.qwen.pool_evpc\endcsname{0.07}
\expandafter\def\csname v@an.hb.qwen.pool_evpc_cs4\endcsname{0.10}
\expandafter\def\csname v@an.hb.qwen.pool_evpc_raw\endcsname{0.55}
\expandafter\def\csname v@an.hb.qwen.debate_evpc_raw\endcsname{0.53}
\expandafter\def\csname v@an.hb.qwen.balanced_evpc_rand_raw\endcsname{0.29}
\expandafter\def\csname v@an.hb.qwen.balanced_evpc_raw\endcsname{0.25}
\expandafter\def\csname v@an.hb.olmo.solo\endcsname{0.66}
\expandafter\def\csname v@an.hb.olmo.debate\endcsname{0.54}
\expandafter\def\csname v@an.hb.olmo.pool_evpc\endcsname{0.03}
\expandafter\def\csname v@an.hb.olmo.pool_evpc_cs4\endcsname{0.11}
\expandafter\def\csname v@an.hb.olmo.pool_evpc_raw\endcsname{0.34}
\expandafter\def\csname v@an.hb.olmo.debate_evpc_raw\endcsname{0.34}
\expandafter\def\csname v@an.hb.olmo.balanced_evpc_rand_raw\endcsname{0.18}
\expandafter\def\csname v@an.hb.olmo.balanced_evpc_raw\endcsname{0.14}
\expandafter\def\csname v@an.hb.early.range\endcsname{0.30--0.43}
\expandafter\def\csname v@red.hb.qwen.debate.0\endcsname{0.84}
\expandafter\def\csname v@red.hb.qwen.debate.5\endcsname{0.48}
\expandafter\def\csname v@red.hb.qwen.debate.drop\endcsname{0.36}
\expandafter\def\csname v@red.hb.qwen.pool_evpc.0\endcsname{0.84}
\expandafter\def\csname v@red.hb.qwen.pool_evpc.5\endcsname{0.63}
\expandafter\def\csname v@red.hb.qwen.pool_evpc.drop\endcsname{0.21}
\expandafter\def\csname v@red.hb.qwen.pool_evpc_raw.0\endcsname{0.84}
\expandafter\def\csname v@red.hb.qwen.pool_evpc_raw.5\endcsname{0.43}
\expandafter\def\csname v@red.hb.qwen.pool_evpc_raw.drop\endcsname{0.41}
\expandafter\def\csname v@red.hb.qwen.debate_evpc_raw.0\endcsname{0.84}
\expandafter\def\csname v@red.hb.qwen.debate_evpc_raw.5\endcsname{0.49}
\expandafter\def\csname v@red.hb.qwen.debate_evpc_raw.drop\endcsname{0.34}
\expandafter\def\csname v@red.hb.qwen.balanced_evpc_rand_raw.0\endcsname{0.84}
\expandafter\def\csname v@red.hb.qwen.balanced_evpc_rand_raw.5\endcsname{0.47}
\expandafter\def\csname v@red.hb.qwen.balanced_evpc_rand_raw.drop\endcsname{0.37}
\expandafter\def\csname v@red.hb.qwen.balanced_evpc_raw.0\endcsname{0.84}
\expandafter\def\csname v@red.hb.qwen.balanced_evpc_raw.5\endcsname{0.54}
\expandafter\def\csname v@red.hb.qwen.balanced_evpc_raw.drop\endcsname{0.30}
\expandafter\def\csname v@red.hb.olmo.debate.0\endcsname{0.81}
\expandafter\def\csname v@red.hb.olmo.debate.5\endcsname{0.37}
\expandafter\def\csname v@red.hb.olmo.debate.drop\endcsname{0.44}
\expandafter\def\csname v@red.hb.olmo.pool_evpc.0\endcsname{0.81}
\expandafter\def\csname v@red.hb.olmo.pool_evpc.5\endcsname{0.55}
\expandafter\def\csname v@red.hb.olmo.pool_evpc.drop\endcsname{0.26}
\expandafter\def\csname v@red.hb.olmo.pool_evpc_raw.0\endcsname{0.72}
\expandafter\def\csname v@red.hb.olmo.pool_evpc_raw.5\endcsname{0.32}
\expandafter\def\csname v@red.hb.olmo.pool_evpc_raw.drop\endcsname{0.39}
\expandafter\def\csname v@red.hb.olmo.debate_evpc_raw.0\endcsname{0.72}
\expandafter\def\csname v@red.hb.olmo.debate_evpc_raw.5\endcsname{0.30}
\expandafter\def\csname v@red.hb.olmo.debate_evpc_raw.drop\endcsname{0.42}
\expandafter\def\csname v@red.hb.olmo.balanced_evpc_rand_raw.0\endcsname{0.72}
\expandafter\def\csname v@red.hb.olmo.balanced_evpc_rand_raw.5\endcsname{0.33}
\expandafter\def\csname v@red.hb.olmo.balanced_evpc_rand_raw.drop\endcsname{0.39}
\expandafter\def\csname v@red.hb.olmo.balanced_evpc_raw.0\endcsname{0.72}
\expandafter\def\csname v@red.hb.olmo.balanced_evpc_raw.5\endcsname{0.38}
\expandafter\def\csname v@red.hb.olmo.balanced_evpc_raw.drop\endcsname{0.33}
\expandafter\def\csname v@red.hb.ratio.range\endcsname{3.1--3.3}
\expandafter\def\csname v@an.pz.qwen.solo\endcsname{0.64}
\expandafter\def\csname v@an.pz.qwen.debate\endcsname{0.72}
\expandafter\def\csname v@an.pz.qwen.pool_evpc\endcsname{0.05}
\expandafter\def\csname v@an.pz.qwen.pool_evpc_cs4\endcsname{0.31}
\expandafter\def\csname v@an.pz.qwen.pool_evpc_raw\endcsname{0.75}
\expandafter\def\csname v@an.pz.qwen.debate_evpc_raw\endcsname{0.69}
\expandafter\def\csname v@an.pz.qwen.balanced_evpc_rand_raw\endcsname{0.65}
\expandafter\def\csname v@an.pz.qwen.balanced_evpc_raw\endcsname{0.60}
\expandafter\def\csname v@an.pz.olmo.solo\endcsname{0.64}
\expandafter\def\csname v@an.pz.olmo.debate\endcsname{0.84}
\expandafter\def\csname v@an.pz.olmo.pool_evpc\endcsname{0.08}
\expandafter\def\csname v@an.pz.olmo.pool_evpc_cs4\endcsname{0.49}
\expandafter\def\csname v@an.pz.olmo.pool_evpc_raw\endcsname{0.82}
\expandafter\def\csname v@an.pz.olmo.debate_evpc_raw\endcsname{0.77}
\expandafter\def\csname v@an.pz.olmo.balanced_evpc_rand_raw\endcsname{0.44}
\expandafter\def\csname v@an.pz.olmo.balanced_evpc_raw\endcsname{0.43}
\expandafter\def\csname v@an.pz.early.range\endcsname{0.65--0.77}
\expandafter\def\csname v@red.pz.qwen.debate.0\endcsname{0.60}
\expandafter\def\csname v@red.pz.qwen.debate.5\endcsname{0.30}
\expandafter\def\csname v@red.pz.qwen.debate.drop\endcsname{0.30}
\expandafter\def\csname v@red.pz.qwen.pool_evpc.0\endcsname{0.60}
\expandafter\def\csname v@red.pz.qwen.pool_evpc.5\endcsname{0.59}
\expandafter\def\csname v@red.pz.qwen.pool_evpc.drop\endcsname{0.01}
\expandafter\def\csname v@red.pz.qwen.pool_evpc_raw.0\endcsname{0.59}
\expandafter\def\csname v@red.pz.qwen.pool_evpc_raw.5\endcsname{0.38}
\expandafter\def\csname v@red.pz.qwen.pool_evpc_raw.drop\endcsname{0.22}
\expandafter\def\csname v@red.pz.qwen.debate_evpc_raw.0\endcsname{0.59}
\expandafter\def\csname v@red.pz.qwen.debate_evpc_raw.5\endcsname{0.40}
\expandafter\def\csname v@red.pz.qwen.debate_evpc_raw.drop\endcsname{0.20}
\expandafter\def\csname v@red.pz.qwen.balanced_evpc_rand_raw.0\endcsname{0.59}
\expandafter\def\csname v@red.pz.qwen.balanced_evpc_rand_raw.5\endcsname{0.38}
\expandafter\def\csname v@red.pz.qwen.balanced_evpc_rand_raw.drop\endcsname{0.21}
\expandafter\def\csname v@red.pz.qwen.balanced_evpc_raw.0\endcsname{0.59}
\expandafter\def\csname v@red.pz.qwen.balanced_evpc_raw.5\endcsname{0.42}
\expandafter\def\csname v@red.pz.qwen.balanced_evpc_raw.drop\endcsname{0.17}
\expandafter\def\csname v@red.pz.olmo.debate.0\endcsname{0.64}
\expandafter\def\csname v@red.pz.olmo.debate.5\endcsname{0.11}
\expandafter\def\csname v@red.pz.olmo.debate.drop\endcsname{0.53}
\expandafter\def\csname v@red.pz.olmo.pool_evpc.0\endcsname{0.64}
\expandafter\def\csname v@red.pz.olmo.pool_evpc.5\endcsname{0.39}
\expandafter\def\csname v@red.pz.olmo.pool_evpc.drop\endcsname{0.24}
\expandafter\def\csname v@red.pz.olmo.pool_evpc_raw.0\endcsname{0.64}
\expandafter\def\csname v@red.pz.olmo.pool_evpc_raw.5\endcsname{0.15}
\expandafter\def\csname v@red.pz.olmo.pool_evpc_raw.drop\endcsname{0.50}
\expandafter\def\csname v@red.pz.olmo.debate_evpc_raw.0\endcsname{0.64}
\expandafter\def\csname v@red.pz.olmo.debate_evpc_raw.5\endcsname{0.17}
\expandafter\def\csname v@red.pz.olmo.debate_evpc_raw.drop\endcsname{0.48}
\expandafter\def\csname v@red.pz.olmo.balanced_evpc_rand_raw.0\endcsname{0.64}
\expandafter\def\csname v@red.pz.olmo.balanced_evpc_rand_raw.5\endcsname{0.17}
\expandafter\def\csname v@red.pz.olmo.balanced_evpc_rand_raw.drop\endcsname{0.47}
\expandafter\def\csname v@red.pz.olmo.balanced_evpc_raw.0\endcsname{0.64}
\expandafter\def\csname v@red.pz.olmo.balanced_evpc_raw.5\endcsname{0.17}
\expandafter\def\csname v@red.pz.olmo.balanced_evpc_raw.drop\endcsname{0.47}
\expandafter\def\csname v@red.pz.ratio.range\endcsname{4.3--4.8}
\expandafter\def\csname v@gainchk.debate.fe\endcsname{0.83--0.97}
\expandafter\def\csname v@gainchk.debate.po\endcsname{0.99--1.16}
\expandafter\def\csname v@gainchk.pool.fe\endcsname{0.50--0.73}
\expandafter\def\csname v@gainchk.pool.po\endcsname{0.77--0.97}
\expandafter\def\csname v@gainchk.items.min\endcsname{8}
\expandafter\def\csname v@h.claims.held\endcsname{22}
\expandafter\def\csname v@h.claims.failed\endcsname{9}
\expandafter\def\csname v@i.qwen.16.dev_mix_p2.solo\endcsname{0.09}
\expandafter\def\csname v@i.qwen.16.dev_mix_p2.solo.D0\endcsname{0.09}
\expandafter\def\csname v@i.olmo.16.dev_mix_p2.solo\endcsname{0.18}
\expandafter\def\csname v@i.olmo.16.dev_mix_p2.solo.D0\endcsname{0.18}
\expandafter\def\csname v@i.qwen.16.dev_mix_p2.debate\endcsname{0.16}
\expandafter\def\csname v@i.qwen.16.dev_mix_p2.debate.D0\endcsname{0.09}
\expandafter\def\csname v@i.olmo.16.dev_mix_p2.debate\endcsname{0.32}
\expandafter\def\csname v@i.olmo.16.dev_mix_p2.debate.D0\endcsname{0.18}
\expandafter\def\csname v@i.qwen.16.dev_mix_p2.debate_evpc\endcsname{0.92}
\expandafter\def\csname v@i.qwen.16.dev_mix_p2.debate_evpc.D0\endcsname{0.09}
\expandafter\def\csname v@i.olmo.16.dev_mix_p2.debate_evpc\endcsname{0.80}
\expandafter\def\csname v@i.olmo.16.dev_mix_p2.debate_evpc.D0\endcsname{0.18}
\expandafter\def\csname v@i.qwen.16.dev_mix_p2.pool_evpc\endcsname{0.93}
\expandafter\def\csname v@i.qwen.16.dev_mix_p2.pool_evpc.D0\endcsname{0.09}
\expandafter\def\csname v@i.olmo.16.dev_mix_p2.pool_evpc\endcsname{0.94}
\expandafter\def\csname v@i.olmo.16.dev_mix_p2.pool_evpc.D0\endcsname{0.18}
\expandafter\def\csname v@i.qwen.16.dev_mix_p2.full_iter\endcsname{0.90}
\expandafter\def\csname v@i.qwen.16.dev_mix_p2.full_iter.D0\endcsname{0.78}
\expandafter\def\csname v@i.olmo.16.dev_mix_p2.full_iter\endcsname{0.89}
\expandafter\def\csname v@i.olmo.16.dev_mix_p2.full_iter.D0\endcsname{0.70}
\expandafter\def\csname v@i.qwen.16.dev_swap_p2.pool_evpc\endcsname{0.94}
\expandafter\def\csname v@i.qwen.16.dev_swap_p2.pool_evpc.D0\endcsname{0.96}
\expandafter\def\csname v@i.olmo.16.dev_swap_p2.pool_evpc\endcsname{0.84}
\expandafter\def\csname v@i.olmo.16.dev_swap_p2.pool_evpc.D0\endcsname{0.97}
\expandafter\def\csname v@i.qwen.16.dev_swap_p2.pool_evpc_cs4\endcsname{0.96}
\expandafter\def\csname v@i.qwen.16.dev_swap_p2.pool_evpc_cs4.D0\endcsname{0.96}
\expandafter\def\csname v@i.olmo.16.dev_swap_p2.pool_evpc_cs4\endcsname{0.95}
\expandafter\def\csname v@i.olmo.16.dev_swap_p2.pool_evpc_cs4.D0\endcsname{0.97}
\expandafter\def\csname v@i.qwen.16.dev_swap_p2.pool_evpc_ch4\endcsname{0.94}
\expandafter\def\csname v@i.qwen.16.dev_swap_p2.pool_evpc_ch4.D0\endcsname{0.96}
\expandafter\def\csname v@i.olmo.16.dev_swap_p2.pool_evpc_ch4\endcsname{0.73}
\expandafter\def\csname v@i.olmo.16.dev_swap_p2.pool_evpc_ch4.D0\endcsname{0.97}
\expandafter\def\csname v@i.qwen.16.dev_replicate_p2.pool_evpc\endcsname{0.96}
\expandafter\def\csname v@i.qwen.16.dev_replicate_p2.pool_evpc.D0\endcsname{0.11}
\expandafter\def\csname v@i.olmo.16.dev_replicate_p2.pool_evpc\endcsname{0.95}
\expandafter\def\csname v@i.olmo.16.dev_replicate_p2.pool_evpc.D0\endcsname{0.16}
\expandafter\def\csname v@i.qwen.16.dev_replicate_p2.pool_ev\endcsname{0.94}
\expandafter\def\csname v@i.qwen.16.dev_replicate_p2.pool_ev.D0\endcsname{0.11}
\expandafter\def\csname v@i.olmo.16.dev_replicate_p2.pool_ev\endcsname{0.95}
\expandafter\def\csname v@i.olmo.16.dev_replicate_p2.pool_ev.D0\endcsname{0.16}
\expandafter\def\csname v@i.qwen.32.dev_replicate_p2.pool_evpc\endcsname{0.96}
\expandafter\def\csname v@i.qwen.32.dev_replicate_p2.pool_evpc.D0\endcsname{0.10}
\expandafter\def\csname v@i.olmo.32.dev_replicate_p2.pool_evpc\endcsname{0.98}
\expandafter\def\csname v@i.olmo.32.dev_replicate_p2.pool_evpc.D0\endcsname{0.16}
\expandafter\def\csname v@i.qwen.32.dev_replicate_p2.pool_ev\endcsname{0.95}
\expandafter\def\csname v@i.qwen.32.dev_replicate_p2.pool_ev.D0\endcsname{0.10}
\expandafter\def\csname v@i.olmo.32.dev_replicate_p2.pool_ev\endcsname{0.94}
\expandafter\def\csname v@i.olmo.32.dev_replicate_p2.pool_ev.D0\endcsname{0.16}
\expandafter\def\csname v@i.qwen.4.dev_p2_w300.debate\endcsname{0.53}
\expandafter\def\csname v@i.qwen.4.dev_p2_w300.debate.D0\endcsname{0.13}
\expandafter\def\csname v@i.olmo.4.dev_p2_w300.debate\endcsname{0.40}
\expandafter\def\csname v@i.olmo.4.dev_p2_w300.debate.D0\endcsname{0.16}
\expandafter\def\csname v@i.qwen.8.dev_p2_w300.debate\endcsname{0.26}
\expandafter\def\csname v@i.qwen.8.dev_p2_w300.debate.D0\endcsname{0.04}
\expandafter\def\csname v@i.olmo.8.dev_p2_w300.debate\endcsname{0.26}
\expandafter\def\csname v@i.olmo.8.dev_p2_w300.debate.D0\endcsname{0.07}
\expandafter\def\csname v@i.qwen.16.dev_p2_w300.debate\endcsname{0.19}
\expandafter\def\csname v@i.qwen.16.dev_p2_w300.debate.D0\endcsname{0.01}
\expandafter\def\csname v@i.olmo.16.dev_p2_w300.debate\endcsname{0.23}
\expandafter\def\csname v@i.olmo.16.dev_p2_w300.debate.D0\endcsname{0.04}
\expandafter\def\csname v@i.qwen.32.dev_p2_w300.debate\endcsname{0.19}
\expandafter\def\csname v@i.qwen.32.dev_p2_w300.debate.D0\endcsname{0.01}
\expandafter\def\csname v@i.olmo.32.dev_p2_w300.debate\endcsname{0.11}
\expandafter\def\csname v@i.olmo.32.dev_p2_w300.debate.D0\endcsname{0.04}
\expandafter\def\csname v@i.qwen.8.dev_p2_w300.pool_evpc\endcsname{0.96}
\expandafter\def\csname v@i.qwen.8.dev_p2_w300.pool_evpc.D0\endcsname{0.04}
\expandafter\def\csname v@i.olmo.8.dev_p2_w300.pool_evpc\endcsname{0.93}
\expandafter\def\csname v@i.olmo.8.dev_p2_w300.pool_evpc.D0\endcsname{0.07}
\expandafter\def\csname v@i.qwen.32.dev_p2_w300.pool_evpc\endcsname{0.96}
\expandafter\def\csname v@i.qwen.32.dev_p2_w300.pool_evpc.D0\endcsname{0.01}
\expandafter\def\csname v@i.olmo.32.dev_p2_w300.pool_evpc\endcsname{0.95}
\expandafter\def\csname v@i.olmo.32.dev_p2_w300.pool_evpc.D0\endcsname{0.04}
\expandafter\def\csname v@i.qwen.16.dev_mix_p2.pool_evpc-full_iter\endcsname{$+$0.03 [$-$0.05, 0.14]}
\expandafter\def\csname v@i.qwen.16.dev_mix_p2.pool_evpc-full_iter.pt\endcsname{$+$0.03}
\expandafter\def\csname v@i.qwen.16.dev_mix_p2.pool_evpc-debate\endcsname{$+$0.78 [0.64, 0.90]}
\expandafter\def\csname v@i.qwen.16.dev_mix_p2.pool_evpc-debate.pt\endcsname{$+$0.78}
\expandafter\def\csname v@i.qwen.16.dev_mix_p2.pool_evpc-debate_evpc\endcsname{$+$0.01 [0.00, 0.03]}
\expandafter\def\csname v@i.qwen.16.dev_mix_p2.pool_evpc-debate_evpc.pt\endcsname{$+$0.01}
\expandafter\def\csname v@i.qwen.16.dev_mix_p2.debate-solo\endcsname{$+$0.07 [0.02, 0.11]}
\expandafter\def\csname v@i.qwen.16.dev_mix_p2.debate-solo.pt\endcsname{$+$0.07}
\expandafter\def\csname v@i.olmo.16.dev_mix_p2.pool_evpc-full_iter\endcsname{$+$0.05 [$-$0.05, 0.15]}
\expandafter\def\csname v@i.olmo.16.dev_mix_p2.pool_evpc-full_iter.pt\endcsname{$+$0.05}
\expandafter\def\csname v@i.olmo.16.dev_mix_p2.pool_evpc-debate\endcsname{$+$0.62 [0.43, 0.79]}
\expandafter\def\csname v@i.olmo.16.dev_mix_p2.pool_evpc-debate.pt\endcsname{$+$0.62}
\expandafter\def\csname v@i.olmo.16.dev_mix_p2.pool_evpc-debate_evpc\endcsname{$+$0.14 [0.06, 0.22]}
\expandafter\def\csname v@i.olmo.16.dev_mix_p2.pool_evpc-debate_evpc.pt\endcsname{$+$0.14}
\expandafter\def\csname v@i.olmo.16.dev_mix_p2.debate-solo\endcsname{$+$0.14 [0.02, 0.27]}
\expandafter\def\csname v@i.olmo.16.dev_mix_p2.debate-solo.pt\endcsname{$+$0.14}
\expandafter\def\csname v@i.qwen.16.dev_swap_p2.pool_evpc_cs4-pool_evpc\endcsname{$+$0.02 [0.00, 0.05]}
\expandafter\def\csname v@i.qwen.16.dev_swap_p2.pool_evpc_cs4-pool_evpc.pt\endcsname{$+$0.02}
\expandafter\def\csname v@i.qwen.16.dev_swap_p2.pool_evpc_ch4-pool_evpc\endcsname{0.00 [$-$0.04, 0.05]}
\expandafter\def\csname v@i.qwen.16.dev_swap_p2.pool_evpc_ch4-pool_evpc.pt\endcsname{0.00}
\expandafter\def\csname v@i.qwen.16.dev_swap_p2.pool_evpc_ch4-pool_evpc_cs4\endcsname{$-$0.02 [$-$0.05, 0.01]}
\expandafter\def\csname v@i.qwen.16.dev_swap_p2.pool_evpc_ch4-pool_evpc_cs4.pt\endcsname{$-$0.02}
\expandafter\def\csname v@i.olmo.16.dev_swap_p2.pool_evpc_cs4-pool_evpc\endcsname{$+$0.11 [0.00, 0.22]}
\expandafter\def\csname v@i.olmo.16.dev_swap_p2.pool_evpc_cs4-pool_evpc.pt\endcsname{$+$0.11}
\expandafter\def\csname v@i.olmo.16.dev_swap_p2.pool_evpc_ch4-pool_evpc\endcsname{$-$0.11 [$-$0.22, $-$0.02]}
\expandafter\def\csname v@i.olmo.16.dev_swap_p2.pool_evpc_ch4-pool_evpc.pt\endcsname{$-$0.11}
\expandafter\def\csname v@i.olmo.16.dev_swap_p2.pool_evpc_ch4-pool_evpc_cs4\endcsname{$-$0.22 [$-$0.36, $-$0.10]}
\expandafter\def\csname v@i.olmo.16.dev_swap_p2.pool_evpc_ch4-pool_evpc_cs4.pt\endcsname{$-$0.22}
\expandafter\def\csname v@i.qwen.16.dev_p2.pool_evpc_cs4-pool_evpc\endcsname{$-$0.06 [$-$0.15, 0.01]}
\expandafter\def\csname v@i.qwen.16.dev_p2.pool_evpc_cs4-pool_evpc.pt\endcsname{$-$0.06}
\expandafter\def\csname v@i.qwen.16.dev_p2.pool_evpc_ch4-pool_evpc\endcsname{$+$0.01 [$-$0.05, 0.07]}
\expandafter\def\csname v@i.qwen.16.dev_p2.pool_evpc_ch4-pool_evpc.pt\endcsname{$+$0.01}
\expandafter\def\csname v@i.olmo.16.dev_p2.pool_evpc_cs4-pool_evpc\endcsname{$-$0.14 [$-$0.27, $-$0.02]}
\expandafter\def\csname v@i.olmo.16.dev_p2.pool_evpc_cs4-pool_evpc.pt\endcsname{$-$0.14}
\expandafter\def\csname v@i.olmo.16.dev_p2.pool_evpc_ch4-pool_evpc\endcsname{$+$0.02 [$-$0.04, 0.09]}
\expandafter\def\csname v@i.olmo.16.dev_p2.pool_evpc_ch4-pool_evpc.pt\endcsname{$+$0.02}
\expandafter\def\csname v@i.qwen.16.dev_replicate_p2.pool_ev-pool_evpc\endcsname{$-$0.02 [$-$0.06, 0.00]}
\expandafter\def\csname v@i.qwen.16.dev_replicate_p2.pool_ev-pool_evpc.pt\endcsname{$-$0.02}
\expandafter\def\csname v@i.qwen.16.dev_replicate_p2.pool_ev_nodedup-pool_evpc\endcsname{$-$0.19 [$-$0.32, $-$0.08]}
\expandafter\def\csname v@i.qwen.16.dev_replicate_p2.pool_ev_nodedup-pool_evpc.pt\endcsname{$-$0.19}
\expandafter\def\csname v@i.olmo.16.dev_replicate_p2.pool_ev-pool_evpc\endcsname{$+$0.01 [$-$0.03, 0.06]}
\expandafter\def\csname v@i.olmo.16.dev_replicate_p2.pool_ev-pool_evpc.pt\endcsname{$+$0.01}
\expandafter\def\csname v@i.olmo.16.dev_replicate_p2.pool_ev_nodedup-pool_evpc\endcsname{$-$0.16 [$-$0.31, $-$0.01]}
\expandafter\def\csname v@i.olmo.16.dev_replicate_p2.pool_ev_nodedup-pool_evpc.pt\endcsname{$-$0.16}
\expandafter\def\csname v@i.qwen.32.dev_replicate_p2.pool_ev-pool_evpc\endcsname{$-$0.01 [$-$0.03, 0.00]}
\expandafter\def\csname v@i.qwen.32.dev_replicate_p2.pool_ev-pool_evpc.pt\endcsname{$-$0.01}
\expandafter\def\csname v@i.qwen.32.dev_replicate_p2.pool_ev_nodedup-pool_evpc\endcsname{$-$0.27 [$-$0.45, $-$0.11]}
\expandafter\def\csname v@i.qwen.32.dev_replicate_p2.pool_ev_nodedup-pool_evpc.pt\endcsname{$-$0.27}
\expandafter\def\csname v@i.olmo.32.dev_replicate_p2.pool_ev-pool_evpc\endcsname{$-$0.04 [$-$0.11, 0.01]}
\expandafter\def\csname v@i.olmo.32.dev_replicate_p2.pool_ev-pool_evpc.pt\endcsname{$-$0.04}
\expandafter\def\csname v@i.olmo.32.dev_replicate_p2.pool_ev_nodedup-pool_evpc\endcsname{$-$0.24 [$-$0.41, $-$0.09]}
\expandafter\def\csname v@i.olmo.32.dev_replicate_p2.pool_ev_nodedup-pool_evpc.pt\endcsname{$-$0.24}
\expandafter\def\csname v@i.qwen.16.dev_p2.pool_ev-pool_evpc\endcsname{$-$0.03 [$-$0.11, 0.03]}
\expandafter\def\csname v@i.qwen.16.dev_p2.pool_ev-pool_evpc.pt\endcsname{$-$0.03}
\expandafter\def\csname v@i.qwen.16.dev_p2.pool_ev_nodedup-pool_evpc\endcsname{$-$0.18 [$-$0.30, $-$0.07]}
\expandafter\def\csname v@i.qwen.16.dev_p2.pool_ev_nodedup-pool_evpc.pt\endcsname{$-$0.18}
\expandafter\def\csname v@i.olmo.16.dev_p2.pool_ev-pool_evpc\endcsname{$-$0.02 [$-$0.07, 0.05]}
\expandafter\def\csname v@i.olmo.16.dev_p2.pool_ev-pool_evpc.pt\endcsname{$-$0.02}
\expandafter\def\csname v@i.olmo.16.dev_p2.pool_ev_nodedup-pool_evpc\endcsname{$-$0.34 [$-$0.49, $-$0.19]}
\expandafter\def\csname v@i.olmo.16.dev_p2.pool_ev_nodedup-pool_evpc.pt\endcsname{$-$0.34}
\expandafter\def\csname v@i.qwen.32.dev_p2.pool_ev-pool_evpc\endcsname{$-$0.05 [$-$0.12, 0.00]}
\expandafter\def\csname v@i.qwen.32.dev_p2.pool_ev-pool_evpc.pt\endcsname{$-$0.05}
\expandafter\def\csname v@i.qwen.32.dev_p2.pool_ev_nodedup-pool_evpc\endcsname{$-$0.31 [$-$0.45, $-$0.18]}
\expandafter\def\csname v@i.qwen.32.dev_p2.pool_ev_nodedup-pool_evpc.pt\endcsname{$-$0.31}
\expandafter\def\csname v@i.olmo.32.dev_p2.pool_ev-pool_evpc\endcsname{$-$0.04 [$-$0.10, 0.01]}
\expandafter\def\csname v@i.olmo.32.dev_p2.pool_ev-pool_evpc.pt\endcsname{$-$0.04}
\expandafter\def\csname v@i.olmo.32.dev_p2.pool_ev_nodedup-pool_evpc\endcsname{$-$0.36 [$-$0.55, $-$0.17]}
\expandafter\def\csname v@i.olmo.32.dev_p2.pool_ev_nodedup-pool_evpc.pt\endcsname{$-$0.36}
\expandafter\def\csname v@i.qwen.8.dev_p2_w300.pool_evpc-debate\endcsname{$+$0.70 [0.54, 0.83]}
\expandafter\def\csname v@i.qwen.8.dev_p2_w300.pool_evpc-debate.pt\endcsname{$+$0.70}
\expandafter\def\csname v@i.olmo.8.dev_p2_w300.pool_evpc-debate\endcsname{$+$0.68 [0.56, 0.79]}
\expandafter\def\csname v@i.olmo.8.dev_p2_w300.pool_evpc-debate.pt\endcsname{$+$0.68}
\expandafter\def\csname v@i.qwen.32.dev_p2_w300.pool_evpc-debate\endcsname{$+$0.77 [0.62, 0.90]}
\expandafter\def\csname v@i.qwen.32.dev_p2_w300.pool_evpc-debate.pt\endcsname{$+$0.77}
\expandafter\def\csname v@i.olmo.32.dev_p2_w300.pool_evpc-debate\endcsname{$+$0.83 [0.68, 0.96]}
\expandafter\def\csname v@i.olmo.32.dev_p2_w300.pool_evpc-debate.pt\endcsname{$+$0.83}
\expandafter\def\csname v@i.qwen.cross.I1.solo.16\endcsname{$+$0.07 [0.00, 0.18]}
\expandafter\def\csname v@i.qwen.cross.I1.pool_evpc.16\endcsname{$+$0.05 [$-$0.02, 0.14]}
\expandafter\def\csname v@i.qwen.cross.I4.debate.32\endcsname{$+$0.14 [0.01, 0.28]}
\expandafter\def\csname v@i.olmo.cross.I1.solo.16\endcsname{$+$0.07 [0.00, 0.18]}
\expandafter\def\csname v@i.olmo.cross.I1.pool_evpc.16\endcsname{$+$0.06 [0.01, 0.12]}
\expandafter\def\csname v@i.olmo.cross.I4.debate.32\endcsname{$-$0.09 [$-$0.24, 0.06]}
\expandafter\def\csname v@pred.I1.inf.n\endcsname{16}
\expandafter\def\csname v@pred.I1.inf.in\endcsname{12}
\expandafter\def\csname v@pred.I1.wide.n\endcsname{8}
\expandafter\def\csname v@pred.I2.inf.n\endcsname{12}
\expandafter\def\csname v@pred.I2.inf.in\endcsname{12}
\expandafter\def\csname v@pred.I2.wide.n\endcsname{0}
\expandafter\def\csname v@pred.I3.inf.n\endcsname{16}
\expandafter\def\csname v@pred.I3.inf.in\endcsname{16}
\expandafter\def\csname v@pred.I3.wide.n\endcsname{0}
\expandafter\def\csname v@pred.I4.inf.n\endcsname{16}
\expandafter\def\csname v@pred.I4.inf.in\endcsname{16}
\expandafter\def\csname v@pred.I4.wide.n\endcsname{0}
\expandafter\def\csname v@gsim.fe.bias.range\endcsname{$-$0.05 to $-$0.04}
\expandafter\def\csname v@gsim.po.low.range\endcsname{0.87--0.88}
\expandafter\def\csname v@gsim.0.2.fe\endcsname{0.15}
\expandafter\def\csname v@gsim.0.2.po\endcsname{0.87}
\expandafter\def\csname v@gsim.0.4.fe\endcsname{0.36}
\expandafter\def\csname v@gsim.0.4.po\endcsname{0.88}
\expandafter\def\csname v@gsim.0.6.fe\endcsname{0.56}
\expandafter\def\csname v@gsim.0.6.po\endcsname{0.87}
\expandafter\def\csname v@gsim.0.8.fe\endcsname{0.75}
\expandafter\def\csname v@gsim.0.8.po\endcsname{0.87}
\expandafter\def\csname v@gsim.1.0.fe\endcsname{0.95}
\expandafter\def\csname v@gsim.1.0.po\endcsname{1.00}
\expandafter\def\csname v@gsim.1.15.fe\endcsname{1.11}
\expandafter\def\csname v@gsim.1.15.po\endcsname{1.15}
\expandafter\def\csname v@gsim.all.low.range\endcsname{0.00--0.06}
\expandafter\def\csname v@gsim.floor.bias.range\endcsname{$-$0.05 to $-$0.01}
\expandafter\def\csname v@gsim.floor.width.range\endcsname{0.15--0.39}
\expandafter\def\csname v@gsim.floor.trans.range\endcsname{148--361}
\expandafter\def\csname v@gsim.small.bias.range\endcsname{$-$0.06 to 0.00}
\expandafter\def\csname v@gsim.small.width.range\endcsname{0.27--0.74}
\expandafter\def\csname v@gsim.small.trans.range\endcsname{48--122}
\expandafter\def\csname v@dwell.test.qwen.debate\endcsname{0.54}
\expandafter\def\csname v@dwell.test.qwen.debate.ci\endcsname{0.54 [0.49, 0.59]}
\expandafter\def\csname v@dwell.test.qwen.debate.rev\endcsname{0.00}
\expandafter\def\csname v@dwell.test.qwen.debate.rev.ci\endcsname{0.00 [0.00, 0.00]}
\expandafter\def\csname v@dwell.test.qwen.debate.msw\endcsname{0.21}
\expandafter\def\csname v@dwell.test.qwen.debate.msw.ci\endcsname{0.21 [0.15, 0.27]}
\expandafter\def\csname v@dwell.test.qwen.debate.mst\endcsname{0.88}
\expandafter\def\csname v@dwell.test.qwen.debate.mst.ci\endcsname{0.88 [0.84, 0.92]}
\expandafter\def\csname v@dwell.test.qwen.debate.sw\endcsname{0.27}
\expandafter\def\csname v@dwell.test.qwen.debate.sw.ci\endcsname{0.27 [0.16, 0.39]}
\expandafter\def\csname v@dwell.test.qwen.balanced\endcsname{0.62}
\expandafter\def\csname v@dwell.test.qwen.balanced.ci\endcsname{0.62 [0.56, 0.67]}
\expandafter\def\csname v@dwell.test.qwen.balanced.rev\endcsname{0.05}
\expandafter\def\csname v@dwell.test.qwen.balanced.rev.ci\endcsname{0.05 [0.00, 0.10]}
\expandafter\def\csname v@dwell.test.qwen.balanced.msw\endcsname{0.28}
\expandafter\def\csname v@dwell.test.qwen.balanced.msw.ci\endcsname{0.28 [0.20, 0.37]}
\expandafter\def\csname v@dwell.test.qwen.balanced.mst\endcsname{0.76}
\expandafter\def\csname v@dwell.test.qwen.balanced.mst.ci\endcsname{0.76 [0.69, 0.82]}
\expandafter\def\csname v@dwell.test.qwen.balanced.sw\endcsname{0.41}
\expandafter\def\csname v@dwell.test.qwen.balanced.sw.ci\endcsname{0.41 [0.26, 0.57]}
\expandafter\def\csname v@dwell.test.qwen.pool_evpc\endcsname{0.69}
\expandafter\def\csname v@dwell.test.qwen.pool_evpc.ci\endcsname{0.69 [0.59, 0.79]}
\expandafter\def\csname v@dwell.test.qwen.pool_evpc.rev\endcsname{0.17}
\expandafter\def\csname v@dwell.test.qwen.pool_evpc.rev.ci\endcsname{0.17 [0.00, 0.29]}
\expandafter\def\csname v@dwell.test.qwen.pool_evpc.msw\endcsname{0.23}
\expandafter\def\csname v@dwell.test.qwen.pool_evpc.msw.ci\endcsname{0.23 [0.12, 0.36]}
\expandafter\def\csname v@dwell.test.qwen.pool_evpc.mst\endcsname{0.72}
\expandafter\def\csname v@dwell.test.qwen.pool_evpc.mst.ci\endcsname{0.72 [0.64, 0.80]}
\expandafter\def\csname v@dwell.test.qwen.pool_evpc.sw\endcsname{0.26}
\expandafter\def\csname v@dwell.test.qwen.pool_evpc.sw.ci\endcsname{0.26 [0.12, 0.42]}
\expandafter\def\csname v@dwell.dev.qwen.debate\endcsname{0.52}
\expandafter\def\csname v@dwell.dev.qwen.debate.ci\endcsname{0.52 [0.42, 0.61]}
\expandafter\def\csname v@dwell.dev.qwen.debate.rev\endcsname{0.08}
\expandafter\def\csname v@dwell.dev.qwen.debate.rev.ci\endcsname{0.08 [0.00, 0.16]}
\expandafter\def\csname v@dwell.dev.qwen.debate.msw\endcsname{0.20}
\expandafter\def\csname v@dwell.dev.qwen.debate.msw.ci\endcsname{0.20 [0.13, 0.26]}
\expandafter\def\csname v@dwell.dev.qwen.debate.mst\endcsname{0.85}
\expandafter\def\csname v@dwell.dev.qwen.debate.mst.ci\endcsname{0.85 [0.78, 0.90]}
\expandafter\def\csname v@dwell.dev.qwen.debate.sw\endcsname{0.28}
\expandafter\def\csname v@dwell.dev.qwen.debate.sw.ci\endcsname{0.28 [0.15, 0.43]}
\expandafter\def\csname v@dwell.dev.qwen.balanced\endcsname{0.56}
\expandafter\def\csname v@dwell.dev.qwen.balanced.ci\endcsname{0.56 [0.52, 0.62]}
\expandafter\def\csname v@dwell.dev.qwen.balanced.rev\endcsname{0.05}
\expandafter\def\csname v@dwell.dev.qwen.balanced.rev.ci\endcsname{0.05 [0.00, 0.12]}
\expandafter\def\csname v@dwell.dev.qwen.balanced.msw\endcsname{0.21}
\expandafter\def\csname v@dwell.dev.qwen.balanced.msw.ci\endcsname{0.21 [0.15, 0.28]}
\expandafter\def\csname v@dwell.dev.qwen.balanced.mst\endcsname{0.68}
\expandafter\def\csname v@dwell.dev.qwen.balanced.mst.ci\endcsname{0.68 [0.59, 0.76]}
\expandafter\def\csname v@dwell.dev.qwen.balanced.sw\endcsname{0.69}
\expandafter\def\csname v@dwell.dev.qwen.balanced.sw.ci\endcsname{0.69 [0.50, 0.88]}
\expandafter\def\csname v@dwell.dev.qwen.pool_evpc\endcsname{0.41}
\expandafter\def\csname v@dwell.dev.qwen.pool_evpc.ci\endcsname{0.41 [0.00, 0.46]}
\expandafter\def\csname v@dwell.dev.qwen.pool_evpc.rev\endcsname{0.00}
\expandafter\def\csname v@dwell.dev.qwen.pool_evpc.rev.ci\endcsname{0.00 [0.00, 0.00]}
\expandafter\def\csname v@dwell.dev.qwen.pool_evpc.msw\endcsname{0.30}
\expandafter\def\csname v@dwell.dev.qwen.pool_evpc.msw.ci\endcsname{0.30 [0.00, 0.78]}
\expandafter\def\csname v@dwell.dev.qwen.pool_evpc.mst\endcsname{0.81}
\expandafter\def\csname v@dwell.dev.qwen.pool_evpc.mst.ci\endcsname{0.81 [0.74, 0.87]}
\expandafter\def\csname v@dwell.dev.qwen.pool_evpc.sw\endcsname{0.04}
\expandafter\def\csname v@dwell.dev.qwen.pool_evpc.sw.ci\endcsname{0.04 [0.00, 0.09]}
\expandafter\def\csname v@dwell.dev.qwen.pool_evpc_raw\endcsname{0.54}
\expandafter\def\csname v@dwell.dev.qwen.pool_evpc_raw.ci\endcsname{0.54 [0.49, 0.59]}
\expandafter\def\csname v@dwell.dev.qwen.pool_evpc_raw.rev\endcsname{0.00}
\expandafter\def\csname v@dwell.dev.qwen.pool_evpc_raw.rev.ci\endcsname{0.00 [0.00, 0.00]}
\expandafter\def\csname v@dwell.dev.qwen.pool_evpc_raw.msw\endcsname{0.21}
\expandafter\def\csname v@dwell.dev.qwen.pool_evpc_raw.msw.ci\endcsname{0.21 [0.15, 0.28]}
\expandafter\def\csname v@dwell.dev.qwen.pool_evpc_raw.mst\endcsname{0.84}
\expandafter\def\csname v@dwell.dev.qwen.pool_evpc_raw.mst.ci\endcsname{0.84 [0.78, 0.89]}
\expandafter\def\csname v@dwell.dev.qwen.pool_evpc_raw.sw\endcsname{0.39}
\expandafter\def\csname v@dwell.dev.qwen.pool_evpc_raw.sw.ci\endcsname{0.39 [0.22, 0.58]}
\expandafter\def\csname v@dwell.dev.qwen.debate_evpc_raw\endcsname{0.59}
\expandafter\def\csname v@dwell.dev.qwen.debate_evpc_raw.ci\endcsname{0.59 [0.54, 0.65]}
\expandafter\def\csname v@dwell.dev.qwen.debate_evpc_raw.rev\endcsname{0.06}
\expandafter\def\csname v@dwell.dev.qwen.debate_evpc_raw.rev.ci\endcsname{0.06 [0.00, 0.14]}
\expandafter\def\csname v@dwell.dev.qwen.debate_evpc_raw.msw\endcsname{0.20}
\expandafter\def\csname v@dwell.dev.qwen.debate_evpc_raw.msw.ci\endcsname{0.20 [0.16, 0.24]}
\expandafter\def\csname v@dwell.dev.qwen.debate_evpc_raw.mst\endcsname{0.76}
\expandafter\def\csname v@dwell.dev.qwen.debate_evpc_raw.mst.ci\endcsname{0.76 [0.71, 0.82]}
\expandafter\def\csname v@dwell.dev.qwen.debate_evpc_raw.sw\endcsname{0.75}
\expandafter\def\csname v@dwell.dev.qwen.debate_evpc_raw.sw.ci\endcsname{0.75 [0.55, 0.97]}
\expandafter\def\csname v@dwell.dev.qwen.balanced_evpc_rand_raw\endcsname{0.61}
\expandafter\def\csname v@dwell.dev.qwen.balanced_evpc_rand_raw.ci\endcsname{0.61 [0.57, 0.66]}
\expandafter\def\csname v@dwell.dev.qwen.balanced_evpc_rand_raw.rev\endcsname{0.01}
\expandafter\def\csname v@dwell.dev.qwen.balanced_evpc_rand_raw.rev.ci\endcsname{0.01 [0.00, 0.05]}
\expandafter\def\csname v@dwell.dev.qwen.balanced_evpc_rand_raw.msw\endcsname{0.28}
\expandafter\def\csname v@dwell.dev.qwen.balanced_evpc_rand_raw.msw.ci\endcsname{0.28 [0.20, 0.37]}
\expandafter\def\csname v@dwell.dev.qwen.balanced_evpc_rand_raw.mst\endcsname{0.75}
\expandafter\def\csname v@dwell.dev.qwen.balanced_evpc_rand_raw.mst.ci\endcsname{0.75 [0.69, 0.81]}
\expandafter\def\csname v@dwell.dev.qwen.balanced_evpc_rand_raw.sw\endcsname{0.75}
\expandafter\def\csname v@dwell.dev.qwen.balanced_evpc_rand_raw.sw.ci\endcsname{0.75 [0.51, 1.00]}
\expandafter\def\csname v@dwell.dev.qwen.balanced_evpc_raw\endcsname{0.60}
\expandafter\def\csname v@dwell.dev.qwen.balanced_evpc_raw.ci\endcsname{0.60 [0.55, 0.66]}
\expandafter\def\csname v@dwell.dev.qwen.balanced_evpc_raw.rev\endcsname{0.05}
\expandafter\def\csname v@dwell.dev.qwen.balanced_evpc_raw.rev.ci\endcsname{0.05 [0.00, 0.11]}
\expandafter\def\csname v@dwell.dev.qwen.balanced_evpc_raw.msw\endcsname{0.22}
\expandafter\def\csname v@dwell.dev.qwen.balanced_evpc_raw.msw.ci\endcsname{0.22 [0.18, 0.26]}
\expandafter\def\csname v@dwell.dev.qwen.balanced_evpc_raw.mst\endcsname{0.78}
\expandafter\def\csname v@dwell.dev.qwen.balanced_evpc_raw.mst.ci\endcsname{0.78 [0.72, 0.83]}
\expandafter\def\csname v@dwell.dev.qwen.balanced_evpc_raw.sw\endcsname{0.46}
\expandafter\def\csname v@dwell.dev.qwen.balanced_evpc_raw.sw.ci\endcsname{0.46 [0.29, 0.64]}
\expandafter\def\csname v@dwell.test.olmo.debate\endcsname{0.51}
\expandafter\def\csname v@dwell.test.olmo.debate.ci\endcsname{0.51 [0.47, 0.54]}
\expandafter\def\csname v@dwell.test.olmo.debate.rev\endcsname{0.00}
\expandafter\def\csname v@dwell.test.olmo.debate.rev.ci\endcsname{0.00 [0.00, 0.00]}
\expandafter\def\csname v@dwell.test.olmo.debate.msw\endcsname{0.24}
\expandafter\def\csname v@dwell.test.olmo.debate.msw.ci\endcsname{0.24 [0.18, 0.30]}
\expandafter\def\csname v@dwell.test.olmo.debate.mst\endcsname{0.73}
\expandafter\def\csname v@dwell.test.olmo.debate.mst.ci\endcsname{0.73 [0.68, 0.78]}
\expandafter\def\csname v@dwell.test.olmo.debate.sw\endcsname{0.52}
\expandafter\def\csname v@dwell.test.olmo.debate.sw.ci\endcsname{0.52 [0.38, 0.65]}
\expandafter\def\csname v@dwell.test.olmo.balanced\endcsname{0.91}
\expandafter\def\csname v@dwell.test.olmo.balanced.ci\endcsname{0.91 [0.86, 0.96]}
\expandafter\def\csname v@dwell.test.olmo.balanced.rev\endcsname{0.36}
\expandafter\def\csname v@dwell.test.olmo.balanced.rev.ci\endcsname{0.36 [0.29, 0.43]}
\expandafter\def\csname v@dwell.test.olmo.balanced.msw\endcsname{0.27}
\expandafter\def\csname v@dwell.test.olmo.balanced.msw.ci\endcsname{0.27 [0.23, 0.33]}
\expandafter\def\csname v@dwell.test.olmo.balanced.mst\endcsname{0.55}
\expandafter\def\csname v@dwell.test.olmo.balanced.mst.ci\endcsname{0.55 [0.48, 0.64]}
\expandafter\def\csname v@dwell.test.olmo.balanced.sw\endcsname{1.60}
\expandafter\def\csname v@dwell.test.olmo.balanced.sw.ci\endcsname{1.60 [1.31, 1.90]}
\expandafter\def\csname v@dwell.test.olmo.pool_evpc\endcsname{0.70}
\expandafter\def\csname v@dwell.test.olmo.pool_evpc.ci\endcsname{0.70 [0.60, 0.80]}
\expandafter\def\csname v@dwell.test.olmo.pool_evpc.rev\endcsname{0.14}
\expandafter\def\csname v@dwell.test.olmo.pool_evpc.rev.ci\endcsname{0.14 [0.06, 0.24]}
\expandafter\def\csname v@dwell.test.olmo.pool_evpc.msw\endcsname{0.19}
\expandafter\def\csname v@dwell.test.olmo.pool_evpc.msw.ci\endcsname{0.19 [0.14, 0.23]}
\expandafter\def\csname v@dwell.test.olmo.pool_evpc.mst\endcsname{0.63}
\expandafter\def\csname v@dwell.test.olmo.pool_evpc.mst.ci\endcsname{0.63 [0.56, 0.69]}
\expandafter\def\csname v@dwell.test.olmo.pool_evpc.sw\endcsname{0.55}
\expandafter\def\csname v@dwell.test.olmo.pool_evpc.sw.ci\endcsname{0.55 [0.34, 0.79]}
\expandafter\def\csname v@dwell.dev.olmo.debate\endcsname{0.58}
\expandafter\def\csname v@dwell.dev.olmo.debate.ci\endcsname{0.58 [0.52, 0.65]}
\expandafter\def\csname v@dwell.dev.olmo.debate.rev\endcsname{0.09}
\expandafter\def\csname v@dwell.dev.olmo.debate.rev.ci\endcsname{0.09 [0.02, 0.17]}
\expandafter\def\csname v@dwell.dev.olmo.debate.msw\endcsname{0.24}
\expandafter\def\csname v@dwell.dev.olmo.debate.msw.ci\endcsname{0.24 [0.19, 0.28]}
\expandafter\def\csname v@dwell.dev.olmo.debate.mst\endcsname{0.67}
\expandafter\def\csname v@dwell.dev.olmo.debate.mst.ci\endcsname{0.67 [0.60, 0.75]}
\expandafter\def\csname v@dwell.dev.olmo.debate.sw\endcsname{0.69}
\expandafter\def\csname v@dwell.dev.olmo.debate.sw.ci\endcsname{0.69 [0.46, 0.94]}
\expandafter\def\csname v@dwell.dev.olmo.balanced\endcsname{0.86}
\expandafter\def\csname v@dwell.dev.olmo.balanced.ci\endcsname{0.86 [0.80, 0.92]}
\expandafter\def\csname v@dwell.dev.olmo.balanced.rev\endcsname{0.24}
\expandafter\def\csname v@dwell.dev.olmo.balanced.rev.ci\endcsname{0.24 [0.18, 0.31]}
\expandafter\def\csname v@dwell.dev.olmo.balanced.msw\endcsname{0.23}
\expandafter\def\csname v@dwell.dev.olmo.balanced.msw.ci\endcsname{0.23 [0.20, 0.27]}
\expandafter\def\csname v@dwell.dev.olmo.balanced.mst\endcsname{0.49}
\expandafter\def\csname v@dwell.dev.olmo.balanced.mst.ci\endcsname{0.49 [0.41, 0.57]}
\expandafter\def\csname v@dwell.dev.olmo.balanced.sw\endcsname{1.76}
\expandafter\def\csname v@dwell.dev.olmo.balanced.sw.ci\endcsname{1.76 [1.41, 2.11]}
\expandafter\def\csname v@dwell.dev.olmo.pool_evpc\endcsname{0.61}
\expandafter\def\csname v@dwell.dev.olmo.pool_evpc.ci\endcsname{0.61 [0.48, 0.73]}
\expandafter\def\csname v@dwell.dev.olmo.pool_evpc.rev\endcsname{0.11}
\expandafter\def\csname v@dwell.dev.olmo.pool_evpc.rev.ci\endcsname{0.11 [0.00, 0.20]}
\expandafter\def\csname v@dwell.dev.olmo.pool_evpc.msw\endcsname{0.25}
\expandafter\def\csname v@dwell.dev.olmo.pool_evpc.msw.ci\endcsname{0.25 [0.13, 0.38]}
\expandafter\def\csname v@dwell.dev.olmo.pool_evpc.mst\endcsname{0.68}
\expandafter\def\csname v@dwell.dev.olmo.pool_evpc.mst.ci\endcsname{0.68 [0.61, 0.75]}
\expandafter\def\csname v@dwell.dev.olmo.pool_evpc.sw\endcsname{0.32}
\expandafter\def\csname v@dwell.dev.olmo.pool_evpc.sw.ci\endcsname{0.32 [0.18, 0.50]}
\expandafter\def\csname v@dwell.dev.olmo.pool_evpc_raw\endcsname{0.56}
\expandafter\def\csname v@dwell.dev.olmo.pool_evpc_raw.ci\endcsname{0.56 [0.50, 0.61]}
\expandafter\def\csname v@dwell.dev.olmo.pool_evpc_raw.rev\endcsname{0.07}
\expandafter\def\csname v@dwell.dev.olmo.pool_evpc_raw.rev.ci\endcsname{0.07 [0.02, 0.13]}
\expandafter\def\csname v@dwell.dev.olmo.pool_evpc_raw.msw\endcsname{0.28}
\expandafter\def\csname v@dwell.dev.olmo.pool_evpc_raw.msw.ci\endcsname{0.28 [0.23, 0.35]}
\expandafter\def\csname v@dwell.dev.olmo.pool_evpc_raw.mst\endcsname{0.68}
\expandafter\def\csname v@dwell.dev.olmo.pool_evpc_raw.mst.ci\endcsname{0.68 [0.63, 0.73]}
\expandafter\def\csname v@dwell.dev.olmo.pool_evpc_raw.sw\endcsname{0.79}
\expandafter\def\csname v@dwell.dev.olmo.pool_evpc_raw.sw.ci\endcsname{0.79 [0.57, 1.01]}
\expandafter\def\csname v@dwell.dev.olmo.debate_evpc_raw\endcsname{0.61}
\expandafter\def\csname v@dwell.dev.olmo.debate_evpc_raw.ci\endcsname{0.61 [0.54, 0.67]}
\expandafter\def\csname v@dwell.dev.olmo.debate_evpc_raw.rev\endcsname{0.10}
\expandafter\def\csname v@dwell.dev.olmo.debate_evpc_raw.rev.ci\endcsname{0.10 [0.04, 0.16]}
\expandafter\def\csname v@dwell.dev.olmo.debate_evpc_raw.msw\endcsname{0.19}
\expandafter\def\csname v@dwell.dev.olmo.debate_evpc_raw.msw.ci\endcsname{0.19 [0.16, 0.21]}
\expandafter\def\csname v@dwell.dev.olmo.debate_evpc_raw.mst\endcsname{0.65}
\expandafter\def\csname v@dwell.dev.olmo.debate_evpc_raw.mst.ci\endcsname{0.65 [0.59, 0.71]}
\expandafter\def\csname v@dwell.dev.olmo.debate_evpc_raw.sw\endcsname{0.98}
\expandafter\def\csname v@dwell.dev.olmo.debate_evpc_raw.sw.ci\endcsname{0.98 [0.77, 1.21]}
\expandafter\def\csname v@dwell.dev.olmo.balanced_evpc_rand_raw\endcsname{0.81}
\expandafter\def\csname v@dwell.dev.olmo.balanced_evpc_rand_raw.ci\endcsname{0.81 [0.75, 0.88]}
\expandafter\def\csname v@dwell.dev.olmo.balanced_evpc_rand_raw.rev\endcsname{0.27}
\expandafter\def\csname v@dwell.dev.olmo.balanced_evpc_rand_raw.rev.ci\endcsname{0.27 [0.17, 0.35]}
\expandafter\def\csname v@dwell.dev.olmo.balanced_evpc_rand_raw.msw\endcsname{0.24}
\expandafter\def\csname v@dwell.dev.olmo.balanced_evpc_rand_raw.msw.ci\endcsname{0.24 [0.21, 0.29]}
\expandafter\def\csname v@dwell.dev.olmo.balanced_evpc_rand_raw.mst\endcsname{0.46}
\expandafter\def\csname v@dwell.dev.olmo.balanced_evpc_rand_raw.mst.ci\endcsname{0.46 [0.39, 0.54]}
\expandafter\def\csname v@dwell.dev.olmo.balanced_evpc_rand_raw.sw\endcsname{1.69}
\expandafter\def\csname v@dwell.dev.olmo.balanced_evpc_rand_raw.sw.ci\endcsname{1.69 [1.38, 1.99]}
\expandafter\def\csname v@dwell.dev.olmo.balanced_evpc_raw\endcsname{0.90}
\expandafter\def\csname v@dwell.dev.olmo.balanced_evpc_raw.ci\endcsname{0.90 [0.84, 0.97]}
\expandafter\def\csname v@dwell.dev.olmo.balanced_evpc_raw.rev\endcsname{0.32}
\expandafter\def\csname v@dwell.dev.olmo.balanced_evpc_raw.rev.ci\endcsname{0.32 [0.25, 0.38]}
\expandafter\def\csname v@dwell.dev.olmo.balanced_evpc_raw.msw\endcsname{0.25}
\expandafter\def\csname v@dwell.dev.olmo.balanced_evpc_raw.msw.ci\endcsname{0.25 [0.21, 0.29]}
\expandafter\def\csname v@dwell.dev.olmo.balanced_evpc_raw.mst\endcsname{0.47}
\expandafter\def\csname v@dwell.dev.olmo.balanced_evpc_raw.mst.ci\endcsname{0.47 [0.41, 0.53]}
\expandafter\def\csname v@dwell.dev.olmo.balanced_evpc_raw.sw\endcsname{1.91}
\expandafter\def\csname v@dwell.dev.olmo.balanced_evpc_raw.sw.ci\endcsname{1.91 [1.64, 2.22]}
\expandafter\def\csname v@red.mix.qwen.debate.hidden\endcsname{0.55}
\expandafter\def\csname v@red.mix.qwen.debate.moved\endcsname{0.32}
\expandafter\def\csname v@red.mix.qwen.debate_evpc.hidden\endcsname{0.64}
\expandafter\def\csname v@red.mix.qwen.debate_evpc.moved\endcsname{0.37}
\expandafter\def\csname v@red.mix.qwen.pool_evpc.hidden\endcsname{0.61}
\expandafter\def\csname v@red.mix.qwen.pool_evpc.moved\endcsname{0.28}
\expandafter\def\csname v@red.mix.olmo.debate.hidden\endcsname{0.40}
\expandafter\def\csname v@red.mix.olmo.debate.moved\endcsname{0.23}
\expandafter\def\csname v@red.mix.olmo.debate_evpc.hidden\endcsname{0.51}
\expandafter\def\csname v@red.mix.olmo.debate_evpc.moved\endcsname{0.30}
\expandafter\def\csname v@red.mix.olmo.pool_evpc.hidden\endcsname{0.50}
\expandafter\def\csname v@red.mix.olmo.pool_evpc.moved\endcsname{0.24}
\expandafter\def\csname v@an.str.qwen.strong.pool_evpc.decoy\endcsname{0.15}
\expandafter\def\csname v@an.str.qwen.strong.pool_evpc_cs4.decoy\endcsname{0.66}
\expandafter\def\csname v@an.str.qwen.strong.pool_evpc_ch4.decoy\endcsname{0.13}
\expandafter\def\csname v@an.str.qwen.weak.pool_evpc.decoy\endcsname{0.13}
\expandafter\def\csname v@an.str.qwen.weak.pool_evpc_cs4.decoy\endcsname{0.27}
\expandafter\def\csname v@an.str.qwen.weak.pool_evpc_ch4.decoy\endcsname{0.11}
\expandafter\def\csname v@an.str.olmo.strong.pool_evpc.decoy\endcsname{0.13}
\expandafter\def\csname v@an.str.olmo.strong.pool_evpc_cs4.decoy\endcsname{0.77}
\expandafter\def\csname v@an.str.olmo.strong.pool_evpc_ch4.decoy\endcsname{0.10}
\expandafter\def\csname v@an.str.olmo.weak.pool_evpc.decoy\endcsname{0.16}
\expandafter\def\csname v@an.str.olmo.weak.pool_evpc_cs4.decoy\endcsname{0.53}
\expandafter\def\csname v@an.str.olmo.weak.pool_evpc_ch4.decoy\endcsname{0.14}
\expandafter\def\csname v@i.claims.held\endcsname{15}
\expandafter\def\csname v@i.claims.failed\endcsname{1}
\expandafter\def\csname v@ei.runs\endcsname{117,484}
\expandafter\def\csname v@ei.inboxes\endcsname{6,206,736}
\expandafter\def\csname v@ei.sentences\endcsname{46,184,115}
\expandafter\def\csname v@ei.allocations\endcsname{4,046,736}
\expandafter\def\csname v@ei.qwen.pool.S.E.4\endcsname{1.9}
\expandafter\def\csname v@ei.qwen.pool.S.I.4\endcsname{1.9}
\expandafter\def\csname v@ei.qwen.pool.S.n.4\endcsname{0.0}
\expandafter\def\csname v@ei.qwen.pool.H.E.4\endcsname{0.84}
\expandafter\def\csname v@ei.qwen.pool.H.I.4\endcsname{0.03}
\expandafter\def\csname v@ei.qwen.pool.H.n.4\endcsname{0.81}
\expandafter\def\csname v@ei.qwen.tok.E.4\endcsname{230}
\expandafter\def\csname v@ei.qwen.tok.I.4\endcsname{166}
\expandafter\def\csname v@ei.qwen.tok.n.4\endcsname{64}
\expandafter\def\csname v@ei.qwen.pool.S.E.8\endcsname{4.4}
\expandafter\def\csname v@ei.qwen.pool.S.I.8\endcsname{4.4}
\expandafter\def\csname v@ei.qwen.pool.S.n.8\endcsname{0.0}
\expandafter\def\csname v@ei.qwen.pool.H.E.8\endcsname{0.84}
\expandafter\def\csname v@ei.qwen.pool.H.I.8\endcsname{0.03}
\expandafter\def\csname v@ei.qwen.pool.H.n.8\endcsname{0.80}
\expandafter\def\csname v@ei.qwen.tok.E.8\endcsname{456}
\expandafter\def\csname v@ei.qwen.tok.I.8\endcsname{382}
\expandafter\def\csname v@ei.qwen.tok.n.8\endcsname{75}
\expandafter\def\csname v@ei.qwen.pool.S.E.16\endcsname{9.6}
\expandafter\def\csname v@ei.qwen.pool.S.I.16\endcsname{9.6}
\expandafter\def\csname v@ei.qwen.pool.S.n.16\endcsname{0.0}
\expandafter\def\csname v@ei.qwen.pool.H.E.16\endcsname{0.83}
\expandafter\def\csname v@ei.qwen.pool.H.I.16\endcsname{0.04}
\expandafter\def\csname v@ei.qwen.pool.H.n.16\endcsname{0.80}
\expandafter\def\csname v@ei.qwen.tok.E.16\endcsname{910}
\expandafter\def\csname v@ei.qwen.tok.I.16\endcsname{830}
\expandafter\def\csname v@ei.qwen.tok.n.16\endcsname{80}
\expandafter\def\csname v@ei.qwen.pool.S.E.32\endcsname{20.1}
\expandafter\def\csname v@ei.qwen.pool.S.I.32\endcsname{20.1}
\expandafter\def\csname v@ei.qwen.pool.S.n.32\endcsname{0.0}
\expandafter\def\csname v@ei.qwen.pool.H.E.32\endcsname{0.84}
\expandafter\def\csname v@ei.qwen.pool.H.I.32\endcsname{0.04}
\expandafter\def\csname v@ei.qwen.pool.H.n.32\endcsname{0.80}
\expandafter\def\csname v@ei.qwen.tok.E.32\endcsname{1,814}
\expandafter\def\csname v@ei.qwen.tok.I.32\endcsname{1,732}
\expandafter\def\csname v@ei.qwen.tok.n.32\endcsname{83}
\expandafter\def\csname v@ei.qwen.pool.S.E.64\endcsname{41.3}
\expandafter\def\csname v@ei.qwen.pool.S.I.64\endcsname{41.3}
\expandafter\def\csname v@ei.qwen.pool.S.n.64\endcsname{0.0}
\expandafter\def\csname v@ei.qwen.pool.H.E.64\endcsname{0.84}
\expandafter\def\csname v@ei.qwen.pool.H.I.64\endcsname{0.04}
\expandafter\def\csname v@ei.qwen.pool.H.n.64\endcsname{0.81}
\expandafter\def\csname v@ei.qwen.tok.E.64\endcsname{3,641}
\expandafter\def\csname v@ei.qwen.tok.I.64\endcsname{3,556}
\expandafter\def\csname v@ei.qwen.tok.n.64\endcsname{85}
\expandafter\def\csname v@ei.qwen.six.S.E.4\endcsname{1.9}
\expandafter\def\csname v@ei.qwen.six.S.I.4\endcsname{1.9}
\expandafter\def\csname v@ei.qwen.six.S.n.4\endcsname{0.0}
\expandafter\def\csname v@ei.qwen.six.H.E.4\endcsname{0.84}
\expandafter\def\csname v@ei.qwen.six.H.I.4\endcsname{0.03}
\expandafter\def\csname v@ei.qwen.six.H.n.4\endcsname{0.81}
\expandafter\def\csname v@ei.qwen.six.S.E.8\endcsname{3.8}
\expandafter\def\csname v@ei.qwen.six.S.I.8\endcsname{3.8}
\expandafter\def\csname v@ei.qwen.six.S.n.8\endcsname{0.0}
\expandafter\def\csname v@ei.qwen.six.H.E.8\endcsname{0.72}
\expandafter\def\csname v@ei.qwen.six.H.I.8\endcsname{0.03}
\expandafter\def\csname v@ei.qwen.six.H.n.8\endcsname{0.70}
\expandafter\def\csname v@ei.qwen.six.S.E.16\endcsname{4.0}
\expandafter\def\csname v@ei.qwen.six.S.I.16\endcsname{4.0}
\expandafter\def\csname v@ei.qwen.six.S.n.16\endcsname{0.0}
\expandafter\def\csname v@ei.qwen.six.H.E.16\endcsname{0.34}
\expandafter\def\csname v@ei.qwen.six.H.I.16\endcsname{0.01}
\expandafter\def\csname v@ei.qwen.six.H.n.16\endcsname{0.33}
\expandafter\def\csname v@ei.qwen.six.S.E.32\endcsname{4.0}
\expandafter\def\csname v@ei.qwen.six.S.I.32\endcsname{4.0}
\expandafter\def\csname v@ei.qwen.six.S.n.32\endcsname{0.0}
\expandafter\def\csname v@ei.qwen.six.H.E.32\endcsname{0.17}
\expandafter\def\csname v@ei.qwen.six.H.I.32\endcsname{0.00}
\expandafter\def\csname v@ei.qwen.six.H.n.32\endcsname{0.17}
\expandafter\def\csname v@ei.qwen.debate.S.E.4\endcsname{1.4}
\expandafter\def\csname v@ei.qwen.debate.S.I.4\endcsname{0.0}
\expandafter\def\csname v@ei.qwen.debate.S.n.4\endcsname{1.4}
\expandafter\def\csname v@ei.qwen.debate.H.E.4\endcsname{0.55}
\expandafter\def\csname v@ei.qwen.debate.H.I.4\endcsname{0.00}
\expandafter\def\csname v@ei.qwen.debate.H.n.4\endcsname{0.55}
\expandafter\def\csname v@ei.qwen.debate.S.E.8\endcsname{2.8}
\expandafter\def\csname v@ei.qwen.debate.S.I.8\endcsname{0.0}
\expandafter\def\csname v@ei.qwen.debate.S.n.8\endcsname{2.8}
\expandafter\def\csname v@ei.qwen.debate.H.E.8\endcsname{0.43}
\expandafter\def\csname v@ei.qwen.debate.H.I.8\endcsname{0.00}
\expandafter\def\csname v@ei.qwen.debate.H.n.8\endcsname{0.43}
\expandafter\def\csname v@ei.qwen.debate.S.E.16\endcsname{2.9}
\expandafter\def\csname v@ei.qwen.debate.S.I.16\endcsname{0.0}
\expandafter\def\csname v@ei.qwen.debate.S.n.16\endcsname{2.9}
\expandafter\def\csname v@ei.qwen.debate.H.E.16\endcsname{0.20}
\expandafter\def\csname v@ei.qwen.debate.H.I.16\endcsname{0.00}
\expandafter\def\csname v@ei.qwen.debate.H.n.16\endcsname{0.20}
\expandafter\def\csname v@ei.qwen.debate.S.E.32\endcsname{3.0}
\expandafter\def\csname v@ei.qwen.debate.S.I.32\endcsname{0.0}
\expandafter\def\csname v@ei.qwen.debate.S.n.32\endcsname{3.0}
\expandafter\def\csname v@ei.qwen.debate.H.E.32\endcsname{0.10}
\expandafter\def\csname v@ei.qwen.debate.H.I.32\endcsname{0.00}
\expandafter\def\csname v@ei.qwen.debate.H.n.32\endcsname{0.10}
\expandafter\def\csname v@ei.qwen.debate.S.E.64\endcsname{3.0}
\expandafter\def\csname v@ei.qwen.debate.S.I.64\endcsname{0.0}
\expandafter\def\csname v@ei.qwen.debate.S.n.64\endcsname{3.0}
\expandafter\def\csname v@ei.qwen.debate.H.E.64\endcsname{0.05}
\expandafter\def\csname v@ei.qwen.debate.H.I.64\endcsname{0.00}
\expandafter\def\csname v@ei.qwen.debate.H.n.64\endcsname{0.05}
\expandafter\def\csname v@ei.qwen.balanced.S.E.4\endcsname{2.8}
\expandafter\def\csname v@ei.qwen.balanced.S.I.4\endcsname{0.0}
\expandafter\def\csname v@ei.qwen.balanced.S.n.4\endcsname{2.8}
\expandafter\def\csname v@ei.qwen.balanced.H.E.4\endcsname{1.10}
\expandafter\def\csname v@ei.qwen.balanced.H.I.4\endcsname{0.00}
\expandafter\def\csname v@ei.qwen.balanced.H.n.4\endcsname{1.10}
\expandafter\def\csname v@ei.qwen.balanced.S.E.8\endcsname{2.8}
\expandafter\def\csname v@ei.qwen.balanced.S.I.8\endcsname{0.0}
\expandafter\def\csname v@ei.qwen.balanced.S.n.8\endcsname{2.8}
\expandafter\def\csname v@ei.qwen.balanced.H.E.8\endcsname{0.49}
\expandafter\def\csname v@ei.qwen.balanced.H.I.8\endcsname{0.00}
\expandafter\def\csname v@ei.qwen.balanced.H.n.8\endcsname{0.49}
\expandafter\def\csname v@ei.qwen.balanced.S.E.16\endcsname{2.8}
\expandafter\def\csname v@ei.qwen.balanced.S.I.16\endcsname{0.0}
\expandafter\def\csname v@ei.qwen.balanced.S.n.16\endcsname{2.8}
\expandafter\def\csname v@ei.qwen.balanced.H.E.16\endcsname{0.26}
\expandafter\def\csname v@ei.qwen.balanced.H.I.16\endcsname{0.00}
\expandafter\def\csname v@ei.qwen.balanced.H.n.16\endcsname{0.26}
\expandafter\def\csname v@ei.qwen.balanced.S.E.32\endcsname{2.9}
\expandafter\def\csname v@ei.qwen.balanced.S.I.32\endcsname{0.0}
\expandafter\def\csname v@ei.qwen.balanced.S.n.32\endcsname{2.9}
\expandafter\def\csname v@ei.qwen.balanced.H.E.32\endcsname{0.17}
\expandafter\def\csname v@ei.qwen.balanced.H.I.32\endcsname{0.00}
\expandafter\def\csname v@ei.qwen.balanced.H.n.32\endcsname{0.17}
\expandafter\def\csname v@ei.qwen.pool.S.E.seq\endcsname{1.9 / 4.4 / 9.6 / 20.1 / 41.3}
\expandafter\def\csname v@ei.qwen.pool.H.n.range\endcsname{0.80--0.81}
\expandafter\def\csname v@ei.qwen.debate.S.n.range\endcsname{1.4--3.0}
\expandafter\def\csname v@ei.qwen.debate.H.n.seq\endcsname{0.55 / 0.43 / 0.20 / 0.10 / 0.05}
\expandafter\def\csname v@ei.qwen.balanced.S.n.range\endcsname{2.8--2.9}
\expandafter\def\csname v@ei.qwen.balanced.H.n.seq\endcsname{1.10 / 0.49 / 0.26 / 0.17}
\expandafter\def\csname v@ei.qwen.pzdev.nd.S.n.4\endcsname{2.4}
\expandafter\def\csname v@ei.qwen.pzdev.nd.S.n.8\endcsname{5.8}
\expandafter\def\csname v@ei.qwen.pzdev.nd.S.n.16\endcsname{12.4}
\expandafter\def\csname v@ei.qwen.pzdev.nd.S.n.32\endcsname{25.7}
\expandafter\def\csname v@ei.qwen.nodedup.S.n.seq\endcsname{2.0 / 10.3 / 21.6}
\expandafter\def\csname v@ei.qwen.tok.cut.64\endcsname{98\%}
\expandafter\def\csname v@ei.qwen.pz.pool.H.n.range\endcsname{0.78--0.80}
\expandafter\def\csname v@ei.qwen.pz.debate.H.n.seq\endcsname{0.46 / 0.34 / 0.15 / 0.07}
\expandafter\def\csname v@ei.qwen.mq.pool.H.n.range\endcsname{0.27--0.29}
\expandafter\def\csname v@ei.qwen.mq.debate.H.n.seq\endcsname{0.19 / 0.14 / 0.06}
\expandafter\def\csname v@ei.qwen.hbdev.pool.S.E.16\endcsname{9.9}
\expandafter\def\csname v@ei.qwen.pzdev.pool.S.E.16\endcsname{12.1}
\expandafter\def\csname v@ei.olmo.pool.S.E.4\endcsname{1.5}
\expandafter\def\csname v@ei.olmo.pool.S.I.4\endcsname{1.5}
\expandafter\def\csname v@ei.olmo.pool.S.n.4\endcsname{0.0}
\expandafter\def\csname v@ei.olmo.pool.H.E.4\endcsname{0.71}
\expandafter\def\csname v@ei.olmo.pool.H.I.4\endcsname{0.02}
\expandafter\def\csname v@ei.olmo.pool.H.n.4\endcsname{0.70}
\expandafter\def\csname v@ei.olmo.tok.E.4\endcsname{187}
\expandafter\def\csname v@ei.olmo.tok.I.4\endcsname{133}
\expandafter\def\csname v@ei.olmo.tok.n.4\endcsname{54}
\expandafter\def\csname v@ei.olmo.pool.S.E.8\endcsname{3.6}
\expandafter\def\csname v@ei.olmo.pool.S.I.8\endcsname{3.6}
\expandafter\def\csname v@ei.olmo.pool.S.n.8\endcsname{0.0}
\expandafter\def\csname v@ei.olmo.pool.H.E.8\endcsname{0.74}
\expandafter\def\csname v@ei.olmo.pool.H.I.8\endcsname{0.03}
\expandafter\def\csname v@ei.olmo.pool.H.n.8\endcsname{0.71}
\expandafter\def\csname v@ei.olmo.tok.E.8\endcsname{377}
\expandafter\def\csname v@ei.olmo.tok.I.8\endcsname{312}
\expandafter\def\csname v@ei.olmo.tok.n.8\endcsname{65}
\expandafter\def\csname v@ei.olmo.pool.S.E.16\endcsname{7.8}
\expandafter\def\csname v@ei.olmo.pool.S.I.16\endcsname{7.8}
\expandafter\def\csname v@ei.olmo.pool.S.n.16\endcsname{0.0}
\expandafter\def\csname v@ei.olmo.pool.H.E.16\endcsname{0.71}
\expandafter\def\csname v@ei.olmo.pool.H.I.16\endcsname{0.03}
\expandafter\def\csname v@ei.olmo.pool.H.n.16\endcsname{0.68}
\expandafter\def\csname v@ei.olmo.tok.E.16\endcsname{743}
\expandafter\def\csname v@ei.olmo.tok.I.16\endcsname{676}
\expandafter\def\csname v@ei.olmo.tok.n.16\endcsname{67}
\expandafter\def\csname v@ei.olmo.pool.S.E.32\endcsname{16.2}
\expandafter\def\csname v@ei.olmo.pool.S.I.32\endcsname{16.2}
\expandafter\def\csname v@ei.olmo.pool.S.n.32\endcsname{0.0}
\expandafter\def\csname v@ei.olmo.pool.H.E.32\endcsname{0.73}
\expandafter\def\csname v@ei.olmo.pool.H.I.32\endcsname{0.03}
\expandafter\def\csname v@ei.olmo.pool.H.n.32\endcsname{0.70}
\expandafter\def\csname v@ei.olmo.tok.E.32\endcsname{1,478}
\expandafter\def\csname v@ei.olmo.tok.I.32\endcsname{1,406}
\expandafter\def\csname v@ei.olmo.tok.n.32\endcsname{72}
\expandafter\def\csname v@ei.olmo.pool.S.E.64\endcsname{32.9}
\expandafter\def\csname v@ei.olmo.pool.S.I.64\endcsname{32.9}
\expandafter\def\csname v@ei.olmo.pool.S.n.64\endcsname{0.0}
\expandafter\def\csname v@ei.olmo.pool.H.E.64\endcsname{0.72}
\expandafter\def\csname v@ei.olmo.pool.H.I.64\endcsname{0.03}
\expandafter\def\csname v@ei.olmo.pool.H.n.64\endcsname{0.69}
\expandafter\def\csname v@ei.olmo.tok.E.64\endcsname{2,923}
\expandafter\def\csname v@ei.olmo.tok.I.64\endcsname{2,851}
\expandafter\def\csname v@ei.olmo.tok.n.64\endcsname{71}
\expandafter\def\csname v@ei.olmo.six.S.E.4\endcsname{1.5}
\expandafter\def\csname v@ei.olmo.six.S.I.4\endcsname{1.5}
\expandafter\def\csname v@ei.olmo.six.S.n.4\endcsname{0.0}
\expandafter\def\csname v@ei.olmo.six.H.E.4\endcsname{0.71}
\expandafter\def\csname v@ei.olmo.six.H.I.4\endcsname{0.02}
\expandafter\def\csname v@ei.olmo.six.H.n.4\endcsname{0.69}
\expandafter\def\csname v@ei.olmo.six.S.E.8\endcsname{3.1}
\expandafter\def\csname v@ei.olmo.six.S.I.8\endcsname{3.1}
\expandafter\def\csname v@ei.olmo.six.S.n.8\endcsname{0.0}
\expandafter\def\csname v@ei.olmo.six.H.E.8\endcsname{0.64}
\expandafter\def\csname v@ei.olmo.six.H.I.8\endcsname{0.02}
\expandafter\def\csname v@ei.olmo.six.H.n.8\endcsname{0.62}
\expandafter\def\csname v@ei.olmo.six.S.E.16\endcsname{3.3}
\expandafter\def\csname v@ei.olmo.six.S.I.16\endcsname{3.3}
\expandafter\def\csname v@ei.olmo.six.S.n.16\endcsname{0.0}
\expandafter\def\csname v@ei.olmo.six.H.E.16\endcsname{0.30}
\expandafter\def\csname v@ei.olmo.six.H.I.16\endcsname{0.01}
\expandafter\def\csname v@ei.olmo.six.H.n.16\endcsname{0.29}
\expandafter\def\csname v@ei.olmo.six.S.E.32\endcsname{3.5}
\expandafter\def\csname v@ei.olmo.six.S.I.32\endcsname{3.5}
\expandafter\def\csname v@ei.olmo.six.S.n.32\endcsname{0.0}
\expandafter\def\csname v@ei.olmo.six.H.E.32\endcsname{0.15}
\expandafter\def\csname v@ei.olmo.six.H.I.32\endcsname{0.00}
\expandafter\def\csname v@ei.olmo.six.H.n.32\endcsname{0.15}
\expandafter\def\csname v@ei.olmo.debate.S.E.4\endcsname{0.8}
\expandafter\def\csname v@ei.olmo.debate.S.I.4\endcsname{0.0}
\expandafter\def\csname v@ei.olmo.debate.S.n.4\endcsname{0.8}
\expandafter\def\csname v@ei.olmo.debate.H.E.4\endcsname{0.34}
\expandafter\def\csname v@ei.olmo.debate.H.I.4\endcsname{0.00}
\expandafter\def\csname v@ei.olmo.debate.H.n.4\endcsname{0.34}
\expandafter\def\csname v@ei.olmo.debate.S.E.8\endcsname{1.6}
\expandafter\def\csname v@ei.olmo.debate.S.I.8\endcsname{0.0}
\expandafter\def\csname v@ei.olmo.debate.S.n.8\endcsname{1.6}
\expandafter\def\csname v@ei.olmo.debate.H.E.8\endcsname{0.26}
\expandafter\def\csname v@ei.olmo.debate.H.I.8\endcsname{0.00}
\expandafter\def\csname v@ei.olmo.debate.H.n.8\endcsname{0.26}
\expandafter\def\csname v@ei.olmo.debate.S.E.16\endcsname{1.7}
\expandafter\def\csname v@ei.olmo.debate.S.I.16\endcsname{0.0}
\expandafter\def\csname v@ei.olmo.debate.S.n.16\endcsname{1.7}
\expandafter\def\csname v@ei.olmo.debate.H.E.16\endcsname{0.12}
\expandafter\def\csname v@ei.olmo.debate.H.I.16\endcsname{0.00}
\expandafter\def\csname v@ei.olmo.debate.H.n.16\endcsname{0.12}
\expandafter\def\csname v@ei.olmo.debate.S.E.32\endcsname{1.7}
\expandafter\def\csname v@ei.olmo.debate.S.I.32\endcsname{0.0}
\expandafter\def\csname v@ei.olmo.debate.S.n.32\endcsname{1.7}
\expandafter\def\csname v@ei.olmo.debate.H.E.32\endcsname{0.06}
\expandafter\def\csname v@ei.olmo.debate.H.I.32\endcsname{0.00}
\expandafter\def\csname v@ei.olmo.debate.H.n.32\endcsname{0.06}
\expandafter\def\csname v@ei.olmo.debate.S.E.64\endcsname{1.7}
\expandafter\def\csname v@ei.olmo.debate.S.I.64\endcsname{0.0}
\expandafter\def\csname v@ei.olmo.debate.S.n.64\endcsname{1.7}
\expandafter\def\csname v@ei.olmo.debate.H.E.64\endcsname{0.03}
\expandafter\def\csname v@ei.olmo.debate.H.I.64\endcsname{0.00}
\expandafter\def\csname v@ei.olmo.debate.H.n.64\endcsname{0.03}
\expandafter\def\csname v@ei.olmo.balanced.S.E.4\endcsname{1.6}
\expandafter\def\csname v@ei.olmo.balanced.S.I.4\endcsname{0.0}
\expandafter\def\csname v@ei.olmo.balanced.S.n.4\endcsname{1.6}
\expandafter\def\csname v@ei.olmo.balanced.H.E.4\endcsname{0.70}
\expandafter\def\csname v@ei.olmo.balanced.H.I.4\endcsname{0.00}
\expandafter\def\csname v@ei.olmo.balanced.H.n.4\endcsname{0.70}
\expandafter\def\csname v@ei.olmo.balanced.S.E.8\endcsname{1.7}
\expandafter\def\csname v@ei.olmo.balanced.S.I.8\endcsname{0.0}
\expandafter\def\csname v@ei.olmo.balanced.S.n.8\endcsname{1.7}
\expandafter\def\csname v@ei.olmo.balanced.H.E.8\endcsname{0.31}
\expandafter\def\csname v@ei.olmo.balanced.H.I.8\endcsname{0.00}
\expandafter\def\csname v@ei.olmo.balanced.H.n.8\endcsname{0.31}
\expandafter\def\csname v@ei.olmo.balanced.S.E.16\endcsname{1.7}
\expandafter\def\csname v@ei.olmo.balanced.S.I.16\endcsname{0.0}
\expandafter\def\csname v@ei.olmo.balanced.S.n.16\endcsname{1.7}
\expandafter\def\csname v@ei.olmo.balanced.H.E.16\endcsname{0.16}
\expandafter\def\csname v@ei.olmo.balanced.H.I.16\endcsname{0.00}
\expandafter\def\csname v@ei.olmo.balanced.H.n.16\endcsname{0.16}
\expandafter\def\csname v@ei.olmo.balanced.S.E.32\endcsname{1.8}
\expandafter\def\csname v@ei.olmo.balanced.S.I.32\endcsname{0.0}
\expandafter\def\csname v@ei.olmo.balanced.S.n.32\endcsname{1.8}
\expandafter\def\csname v@ei.olmo.balanced.H.E.32\endcsname{0.10}
\expandafter\def\csname v@ei.olmo.balanced.H.I.32\endcsname{0.00}
\expandafter\def\csname v@ei.olmo.balanced.H.n.32\endcsname{0.10}
\expandafter\def\csname v@ei.olmo.pool.S.E.seq\endcsname{1.5 / 3.6 / 7.8 / 16.2 / 32.9}
\expandafter\def\csname v@ei.olmo.pool.H.n.range\endcsname{0.68--0.71}
\expandafter\def\csname v@ei.olmo.debate.S.n.range\endcsname{0.8--1.7}
\expandafter\def\csname v@ei.olmo.debate.H.n.seq\endcsname{0.34 / 0.26 / 0.12 / 0.06 / 0.03}
\expandafter\def\csname v@ei.olmo.balanced.S.n.range\endcsname{1.6--1.8}
\expandafter\def\csname v@ei.olmo.balanced.H.n.seq\endcsname{0.70 / 0.31 / 0.16 / 0.10}
\expandafter\def\csname v@ei.olmo.pzdev.nd.S.n.4\endcsname{2.2}
\expandafter\def\csname v@ei.olmo.pzdev.nd.S.n.8\endcsname{5.4}
\expandafter\def\csname v@ei.olmo.pzdev.nd.S.n.16\endcsname{11.6}
\expandafter\def\csname v@ei.olmo.pzdev.nd.S.n.32\endcsname{24.3}
\expandafter\def\csname v@ei.olmo.nodedup.S.n.seq\endcsname{1.7 / 9.6 / 20.5}
\expandafter\def\csname v@ei.olmo.tok.cut.64\endcsname{98\%}
\expandafter\def\csname v@ei.olmo.pz.pool.H.n.range\endcsname{0.69--0.71}
\expandafter\def\csname v@ei.olmo.pz.debate.H.n.seq\endcsname{0.24 / 0.18 / 0.07 / 0.04}
\expandafter\def\csname v@ei.olmo.mq.pool.H.n.range\endcsname{0.26--0.27}
\expandafter\def\csname v@ei.olmo.mq.debate.H.n.seq\endcsname{0.12 / 0.09 / 0.04}
\expandafter\def\csname v@ei.olmo.hbdev.pool.S.E.16\endcsname{8.3}
\expandafter\def\csname v@ei.olmo.pzdev.pool.S.E.16\endcsname{11.1}
\expandafter\def\csname v@ei.nu.debate.range\endcsname{0.97--1.04}
\expandafter\def\csname v@ei.nu.equal.range\endcsname{0.00}
\expandafter\def\csname v@ei.nu.half.range\endcsname{0.57--0.66}
\expandafter\def\csname v@ei.words.hb\endcsname{10 of 336}
\expandafter\def\csname v@ei.words.pz.range\endcsname{24--25\%}
\expandafter\def\csname v@ei.words.mq.range\endcsname{1\%}
\expandafter\def\csname v@ei.cut.max\endcsname{1.7\%}
\expandafter\def\csname v@ei.cut.variants.max\endcsname{2.6\%}
\expandafter\def\csname v@ei.cut.oracle.max\endcsname{11\%}
\expandafter\def\csname v@ei.anchor.hb\endcsname{79\% / 87\%}
\expandafter\def\csname v@ei.anchor.pz\endcsname{4\% / 6\%}
\expandafter\def\csname v@ei.anchor.mq\endcsname{79\% / 85\%}
\expandafter\def\csname v@kf.cap.range\endcsname{0.04--0.16}
\expandafter\def\csname v@kf.nodedup.hb.range\endcsname{0.54--0.70}
\expandafter\def\csname v@kf.nodedup.pz.range\endcsname{0.05--0.18}
\expandafter\def\csname v@kf.capdedup.hb.range\endcsname{0.76--0.86}
\expandafter\def\csname v@kf.ev.hb.range\endcsname{0.85--0.88}
\expandafter\def\csname v@kf.evpc.hb.range\endcsname{0.88--0.92}
\expandafter\def\csname v@kf.evpc.hb.evcells.range\endcsname{0.88--0.92}
\expandafter\def\csname v@kf.capdedup.pz.range\endcsname{0.20--0.44}
\expandafter\def\csname v@kf.ev.pz.range\endcsname{0.61--0.76}
\expandafter\def\csname v@kf.evpc.pz.range\endcsname{0.75--0.88}
\expandafter\def\csname v@kf.evpc.pz.evcells.range\endcsname{0.75--0.85}
\expandafter\def\csname v@kf.cap.gap.range\endcsname{$+$0.71 to $+$0.86}
\expandafter\def\csname v@kf.cap.gap.abs.range\endcsname{0.71--0.86}
\expandafter\def\csname v@kf.cap.pred.below\endcsname{4}
\expandafter\def\csname v@kf.cap.pred.n\endcsname{8}
\expandafter\def\csname v@kf.debate.range\endcsname{0.05--0.21}
\expandafter\def\csname v@kf.delay.r2.range\endcsname{0.05--0.61}
\expandafter\def\csname v@kf.evpc.r2.range\endcsname{0.57--0.88}
\expandafter\def\csname v@kf.delay.r3.range\endcsname{0.41--0.86}
\expandafter\def\csname v@kf.evpc.r3.range\endcsname{0.65--0.89}
\expandafter\def\csname v@kf.delay.final.range\endcsname{0.71--0.92}
\expandafter\def\csname v@kf.delay.evpc.range\endcsname{0.75--0.92}
\expandafter\def\csname v@kf.delay.n\endcsname{8}
\expandafter\def\csname v@kf.delay.endgap.range\endcsname{$-$0.01 to $+$0.07}
\expandafter\def\csname v@kf.delay.lag.range\endcsname{$-$0.04 to $+$0.03}
\expandafter\def\csname v@kf.k3.facts.range\endcsname{0.61--0.84}
\expandafter\def\csname v@kf.k3.factspert.range\endcsname{0.58--0.84}
\expandafter\def\csname v@kf.k3.args.range\endcsname{0.06--0.15}
\expandafter\def\csname v@kf.k3.argspert.range\endcsname{0.00--0.16}
\expandafter\def\csname v@kf.k3.nofb.range\endcsname{0.05--0.20}
\expandafter\def\csname v@kf.k3.factsdiff.range\endcsname{$-$0.06 to $+$0.06}
\expandafter\def\csname v@kf.k3.n\endcsname{4}
\expandafter\def\csname v@kf.k3.alone.range\endcsname{0.02--0.14}
\expandafter\def\csname v@J3a.hb.qwen\endcsname{$+$0.69 [0.56, 0.81]}
\expandafter\def\csname v@J3b.hb.qwen\endcsname{$+$0.52 [0.41, 0.64]}
\expandafter\def\csname v@J3a.pz.qwen\endcsname{$+$0.60 [0.56, 0.64]}
\expandafter\def\csname v@J3b.pz.qwen\endcsname{$+$0.46 [0.40, 0.52]}
\expandafter\def\csname v@J3a.hb.olmo\endcsname{$+$0.63 [0.52, 0.74]}
\expandafter\def\csname v@J3b.hb.olmo\endcsname{$+$0.59 [0.47, 0.71]}
\expandafter\def\csname v@J3a.pz.olmo\endcsname{$+$0.61 [0.57, 0.65]}
\expandafter\def\csname v@J3b.pz.olmo\endcsname{$+$0.59 [0.54, 0.64]}
\expandafter\def\csname v@J3a.min\endcsname{0.60}
\expandafter\def\csname v@J3a.max\endcsname{0.69}
\expandafter\def\csname v@J3b.min\endcsname{0.46}
\expandafter\def\csname v@J3b.max\endcsname{0.59}
\expandafter\def\csname v@J3.pass\endcsname{8}
\expandafter\def\csname v@J3a.pp.range\endcsname{60--69}
\expandafter\def\csname v@j3.evpc.range\endcsname{0.57--0.69}
\expandafter\def\csname v@j3.evpc.min\endcsname{0.57}
\expandafter\def\csname v@j3.raw.range\endcsname{0.01--0.07}
\expandafter\def\csname v@j3.raw.max\endcsname{0.07}
\expandafter\def\csname v@j3.nodedup.range\endcsname{0.06--0.43}
\expandafter\def\csname v@j3.alone.range\endcsname{0.00--0.06}
\expandafter\def\csname v@j3.tok.range\endcsname{89--94\%}
\expandafter\def\csname v@j3.gain.allruns.range\endcsname{0.58--0.65}
\expandafter\def\csname v@j3.gain.mean.pp\endcsname{61}
\expandafter\def\csname v@j3.tok.savings.mean\endcsname{8\%}
\expandafter\def\csname v@j3.peerA.range\endcsname{0.67--0.78}
\expandafter\def\csname v@j3.full.hb.range\endcsname{0.48--0.60}
\expandafter\def\csname v@j3.full.pz.range\endcsname{0.76--0.81}
\expandafter\def\csname v@j3.hb.below.range\endcsname{0.16--0.21}
\expandafter\def\csname v@pred.J3.width\endcsname{0.43}
\expandafter\def\csname v@j3.r1.hb.range\endcsname{0.33--0.62}
\expandafter\def\csname v@j3.evpc.hb.range\endcsname{0.57--0.68}
\expandafter\def\csname v@j3.r1.pz.range\endcsname{0.34--0.38}
\expandafter\def\csname v@j3.evpc.pz.range\endcsname{0.65--0.69}
\expandafter\def\csname v@j3.full.share.hb\endcsname{0.20--0.49}
\expandafter\def\csname v@j3.full.share.pz\endcsname{0.07--0.17}
\expandafter\def\csname v@j3.full.raw.range\endcsname{0.00--0.14}
\expandafter\def\csname v@j3.full.evpc.range\endcsname{0.71--0.87}
\expandafter\def\csname v@j3.full.nodedup.range\endcsname{0.01--0.51}
\expandafter\def\csname v@j3.full.raw.r1.range\endcsname{0.00--0.11}
\expandafter\def\csname v@j3.full.evpc.r1.range\endcsname{0.47--0.79}
\expandafter\def\csname v@pred.J3.in\endcsname{60}
\expandafter\def\csname v@pred.J3.n\endcsname{60}
\expandafter\def\csname v@j1.evpc\endcsname{0.83}
\expandafter\def\csname v@j1.raw\endcsname{0.05}
\expandafter\def\csname v@j1.weigh\endcsname{0.06}
\expandafter\def\csname v@j1.weigh.diff\endcsname{$+$0.01 [0.00, 0.03]}
\expandafter\def\csname v@j2.gain.range\endcsname{0.57--0.82}
\expandafter\def\csname v@pred.J2.in\endcsname{57}
\expandafter\def\csname v@pred.J2.n\endcsname{57}
\expandafter\def\csname v@gpuh.J.a30\endcsname{64}
\expandafter\def\csname v@gpuh.J.dev\endcsname{22}
\expandafter\def\csname v@gpuh.J.test\endcsname{43}
\expandafter\def\csname v@kv.miss.evpc.range\endcsname{0.71--0.83}
\expandafter\def\csname v@kv.miss.cap.range\endcsname{0.02--0.06}
\expandafter\def\csname v@kv.miss.capdedup.range\endcsname{0.29--0.37}
\expandafter\def\csname v@k3.did.range\endcsname{$-$0.14 to $+$0.05}
\expandafter\def\csname v@k3b.nofb.range\endcsname{0.55--0.85}
\expandafter\def\csname v@k3b.diff.range\endcsname{$-$0.09 to $+$0.07}
\expandafter\def\csname v@pred.K.in\endcsname{44}
\expandafter\def\csname v@pred.K.n\endcsname{52}
\expandafter\def\csname v@gpuh.K.a30\endcsname{34}
\expandafter\def\csname v@r.facts.c0.range\endcsname{0.47--0.88}
\expandafter\def\csname v@r.facts.c8.range\endcsname{0.08--0.72}
\expandafter\def\csname v@r.facts.c84.range\endcsname{0.13--0.74}
\expandafter\def\csname v@r.facts.drop.range\endcsname{0.16--0.48}
\expandafter\def\csname v@r.argued.18.range\endcsname{0.15--0.41}
\expandafter\def\csname v@r.argued.81.range\endcsname{0.98--1.00}
\expandafter\def\csname v@r.claim.1\endcsname{0}
\expandafter\def\csname v@r.claim.2\endcsname{3}
\expandafter\def\csname v@r.claim.3\endcsname{4}
\expandafter\def\csname v@r.claim.4\endcsname{4}
\expandafter\def\csname v@gpuh.R.a30\endcsname{1.9}
\expandafter\def\csname v@u.last.range\endcsname{$-$0.15 to $-$0.02}
\expandafter\def\csname v@u.own.range\endcsname{$-$0.49 to $-$0.29}
\expandafter\def\csname v@u.ownlast.range\endcsname{$-$0.49 to $-$0.25}
\expandafter\def\csname v@u.ownlast.deb.range\endcsname{$+$0.22 to $+$0.62}
\expandafter\def\csname v@u.ownnd.range\endcsname{$+$0.24 to $+$0.59}
\expandafter\def\csname v@u.first.range\endcsname{$-$0.13 to $-$0.01}
\expandafter\def\csname v@u.ownnd.deb.range\endcsname{$-$0.04 to 0.00}
\expandafter\def\csname v@u.evpc.deb.range\endcsname{$+$0.68 to $+$0.88}
\expandafter\def\csname v@u2.para.range\endcsname{$-$0.09 to $-$0.01}
\expandafter\def\csname v@u2.para.hb.qwen\endcsname{$-$0.01 [$-$0.03, 0.01]}
\expandafter\def\csname v@u2.para.pz.qwen\endcsname{$-$0.01 [$-$0.03, 0.01]}
\expandafter\def\csname v@u2.para.hb.olmo\endcsname{$-$0.09 [$-$0.16, $-$0.03]}
\expandafter\def\csname v@u2.para.pz.olmo\endcsname{$-$0.03 [$-$0.06, 0.01]}
\expandafter\def\csname v@u2.para.own.range\endcsname{$+$0.29 to $+$0.48}
\expandafter\def\csname v@u2.para.own.hb.qwen\endcsname{$+$0.29 [0.18, 0.40]}
\expandafter\def\csname v@u2.para.own.pz.qwen\endcsname{$+$0.48 [0.41, 0.55]}
\expandafter\def\csname v@u2.para.own.hb.olmo\endcsname{$+$0.38 [0.25, 0.51]}
\expandafter\def\csname v@u2.para.own.pz.olmo\endcsname{$+$0.44 [0.37, 0.51]}
\expandafter\def\csname v@u3.claims.own.range\endcsname{$-$0.11 to $+$0.20}
\expandafter\def\csname v@u3.gap.range\endcsname{$+$0.25 to $+$0.57}
\expandafter\def\csname v@u16.own.range\endcsname{$-$0.49 to $-$0.29}
\expandafter\def\csname v@u16.last.range\endcsname{$-$0.16 to $-$0.01}
\expandafter\def\csname v@u16.ownnd.range\endcsname{$+$0.22 to $+$0.58}
\expandafter\def\csname v@u16.evpc.deb.range\endcsname{$+$0.65 to $+$0.88}
\expandafter\def\csname v@u16.para.range\endcsname{$-$0.07 to 0.00}
\expandafter\def\csname v@u16.para.own.range\endcsname{$+$0.27 to $+$0.45}
\expandafter\def\csname v@u16.claims.own.range\endcsname{$-$0.11 to $+$0.20}
\expandafter\def\csname v@u.repro.maxdiff\endcsname{0.05}
\expandafter\def\csname v@u.ownnd.tok.ratio.debate.range\endcsname{1.4--3.0$\times$}
\expandafter\def\csname v@u.ownnd.tok.ratio.evpc.range\endcsname{2.4--5.2$\times$}
\expandafter\def\csname v@u3.tok.ratio.range\endcsname{1.4--1.6$\times$}
\expandafter\def\csname v@u.qwen.hb.16.debate\endcsname{0.03}
\expandafter\def\csname v@u.qwen.hb.16.pool_ev_own_nodedup\endcsname{0.04}
\expandafter\def\csname v@u.qwen.hb.16.pool_evpc_ownfirst\endcsname{0.45}
\expandafter\def\csname v@u.qwen.hb.16.pool_evpc_own\endcsname{0.62}
\expandafter\def\csname v@u.qwen.hb.16.pool_evpc_own_last\endcsname{0.68}
\expandafter\def\csname v@u.qwen.hb.16.pool_evpc_claims\endcsname{0.66}
\expandafter\def\csname v@u.qwen.hb.16.pool_evpc_para\endcsname{0.89}
\expandafter\def\csname v@u.qwen.hb.16.pool_evpc_last\endcsname{0.88}
\expandafter\def\csname v@u.qwen.hb.16.pool_evpc\endcsname{0.91}
\expandafter\def\csname v@u.qwen.hb.32.debate\endcsname{0.04}
\expandafter\def\csname v@u.qwen.hb.32.pool_ev_own_nodedup\endcsname{0.01}
\expandafter\def\csname v@u.qwen.hb.32.pool_evpc_ownfirst\endcsname{0.52}
\expandafter\def\csname v@u.qwen.hb.32.pool_evpc_own\endcsname{0.61}
\expandafter\def\csname v@u.qwen.hb.32.pool_evpc_own_last\endcsname{0.64}
\expandafter\def\csname v@u.qwen.hb.32.pool_evpc_para\endcsname{0.92}
\expandafter\def\csname v@u.qwen.hb.32.pool_evpc_last\endcsname{0.88}
\expandafter\def\csname v@u.qwen.hb.32.pool_evpc\endcsname{0.92}
\expandafter\def\csname v@u.qwen.pz.16.debate\endcsname{0.11}
\expandafter\def\csname v@u.qwen.pz.16.pool_ev_own_nodedup\endcsname{0.12}
\expandafter\def\csname v@u.qwen.pz.16.pool_evpc_ownfirst\endcsname{0.37}
\expandafter\def\csname v@u.qwen.pz.16.pool_evpc_own\endcsname{0.41}
\expandafter\def\csname v@u.qwen.pz.16.pool_evpc_own_last\endcsname{0.37}
\expandafter\def\csname v@u.qwen.pz.16.pool_evpc_claims\endcsname{0.49}
\expandafter\def\csname v@u.qwen.pz.16.pool_evpc_para\endcsname{0.86}
\expandafter\def\csname v@u.qwen.pz.16.pool_evpc_last\endcsname{0.85}
\expandafter\def\csname v@u.qwen.pz.16.pool_evpc\endcsname{0.86}
\expandafter\def\csname v@u.qwen.pz.32.debate\endcsname{0.09}
\expandafter\def\csname v@u.qwen.pz.32.pool_ev_own_nodedup\endcsname{0.09}
\expandafter\def\csname v@u.qwen.pz.32.pool_evpc_ownfirst\endcsname{0.39}
\expandafter\def\csname v@u.qwen.pz.32.pool_evpc_own\endcsname{0.36}
\expandafter\def\csname v@u.qwen.pz.32.pool_evpc_own_last\endcsname{0.40}
\expandafter\def\csname v@u.qwen.pz.32.pool_evpc_para\endcsname{0.88}
\expandafter\def\csname v@u.qwen.pz.32.pool_evpc_last\endcsname{0.85}
\expandafter\def\csname v@u.qwen.pz.32.pool_evpc\endcsname{0.89}
\expandafter\def\csname v@u.olmo.hb.16.debate\endcsname{0.24}
\expandafter\def\csname v@u.olmo.hb.16.pool_ev_own_nodedup\endcsname{0.17}
\expandafter\def\csname v@u.olmo.hb.16.pool_evpc_ownfirst\endcsname{0.33}
\expandafter\def\csname v@u.olmo.hb.16.pool_evpc_own\endcsname{0.40}
\expandafter\def\csname v@u.olmo.hb.16.pool_evpc_own_last\endcsname{0.49}
\expandafter\def\csname v@u.olmo.hb.16.pool_evpc_claims\endcsname{0.60}
\expandafter\def\csname v@u.olmo.hb.16.pool_evpc_para\endcsname{0.82}
\expandafter\def\csname v@u.olmo.hb.16.pool_evpc_last\endcsname{0.85}
\expandafter\def\csname v@u.olmo.hb.16.pool_evpc\endcsname{0.89}
\expandafter\def\csname v@u.olmo.hb.32.debate\endcsname{0.21}
\expandafter\def\csname v@u.olmo.hb.32.pool_ev_own_nodedup\endcsname{0.20}
\expandafter\def\csname v@u.olmo.hb.32.pool_evpc_ownfirst\endcsname{0.45}
\expandafter\def\csname v@u.olmo.hb.32.pool_evpc_own\endcsname{0.49}
\expandafter\def\csname v@u.olmo.hb.32.pool_evpc_own_last\endcsname{0.49}
\expandafter\def\csname v@u.olmo.hb.32.pool_evpc_para\endcsname{0.83}
\expandafter\def\csname v@u.olmo.hb.32.pool_evpc_last\endcsname{0.83}
\expandafter\def\csname v@u.olmo.hb.32.pool_evpc\endcsname{0.95}
\expandafter\def\csname v@u.olmo.pz.16.debate\endcsname{0.10}
\expandafter\def\csname v@u.olmo.pz.16.pool_ev_own_nodedup\endcsname{0.06}
\expandafter\def\csname v@u.olmo.pz.16.pool_evpc_ownfirst\endcsname{0.21}
\expandafter\def\csname v@u.olmo.pz.16.pool_evpc_own\endcsname{0.31}
\expandafter\def\csname v@u.olmo.pz.16.pool_evpc_own_last\endcsname{0.31}
\expandafter\def\csname v@u.olmo.pz.16.pool_evpc_claims\endcsname{0.19}
\expandafter\def\csname v@u.olmo.pz.16.pool_evpc_para\endcsname{0.72}
\expandafter\def\csname v@u.olmo.pz.16.pool_evpc_last\endcsname{0.61}
\expandafter\def\csname v@u.olmo.pz.16.pool_evpc\endcsname{0.76}
\expandafter\def\csname v@u.olmo.pz.32.debate\endcsname{0.09}
\expandafter\def\csname v@u.olmo.pz.32.pool_ev_own_nodedup\endcsname{0.06}
\expandafter\def\csname v@u.olmo.pz.32.pool_evpc_ownfirst\endcsname{0.21}
\expandafter\def\csname v@u.olmo.pz.32.pool_evpc_own\endcsname{0.30}
\expandafter\def\csname v@u.olmo.pz.32.pool_evpc_own_last\endcsname{0.30}
\expandafter\def\csname v@u.olmo.pz.32.pool_evpc_para\endcsname{0.76}
\expandafter\def\csname v@u.olmo.pz.32.pool_evpc_last\endcsname{0.64}
\expandafter\def\csname v@u.olmo.pz.32.pool_evpc\endcsname{0.78}
\expandafter\def\csname v@u.view.wrong.range\endcsname{0.50--0.73}
\expandafter\def\csname v@u.view.held.range\endcsname{0.32--0.67}
\expandafter\def\csname v@u.view.conclusion.range\endcsname{0.13--0.62}
\expandafter\def\csname v@gpuh.agmn\endcsname{55}
\expandafter\def\csname v@x.qwen.screen.d2.full\endcsname{0.98}
\expandafter\def\csname v@x.qwen.screen.d2.candidates\endcsname{224}
\expandafter\def\csname v@x.qwen.screen.d2.kept\endcsname{105}
\expandafter\def\csname v@x.qwen.screen.d4.full\endcsname{0.86}
\expandafter\def\csname v@x.qwen.screen.d4.candidates\endcsname{112}
\expandafter\def\csname v@x.qwen.screen.d4.kept\endcsname{91}
\expandafter\def\csname v@x.qwen.screen.d8.full\endcsname{0.66}
\expandafter\def\csname v@x.qwen.screen.d8.candidates\endcsname{320}
\expandafter\def\csname v@x.qwen.screen.d8.kept\endcsname{173}
\expandafter\def\csname v@x.qwen.screen.d16.full\endcsname{0.34}
\expandafter\def\csname v@x.qwen.screen.d16.candidates\endcsname{672}
\expandafter\def\csname v@x.qwen.screen.d16.kept\endcsname{112}
\expandafter\def\csname v@x.olmo.screen.d2.full\endcsname{0.88}
\expandafter\def\csname v@x.olmo.screen.d2.candidates\endcsname{224}
\expandafter\def\csname v@x.olmo.screen.d2.kept\endcsname{140}
\expandafter\def\csname v@x.olmo.screen.d4.full\endcsname{0.77}
\expandafter\def\csname v@x.olmo.screen.d4.candidates\endcsname{112}
\expandafter\def\csname v@x.olmo.screen.d4.kept\endcsname{86}
\expandafter\def\csname v@x.olmo.screen.d8.full\endcsname{0.57}
\expandafter\def\csname v@x.olmo.screen.d8.candidates\endcsname{320}
\expandafter\def\csname v@x.olmo.screen.d8.kept\endcsname{150}
\expandafter\def\csname v@x.olmo.screen.d16.full\endcsname{0.36}
\expandafter\def\csname v@x.olmo.screen.d16.candidates\endcsname{672}
\expandafter\def\csname v@x.olmo.screen.d16.kept\endcsname{142}
\expandafter\def\csname v@x.qwen.8.debate\endcsname{0.08}
\expandafter\def\csname v@x.qwen.8.pool_ev_nodedup\endcsname{0.61}
\expandafter\def\csname v@x.qwen.8.pool_evpc\endcsname{0.99}
\expandafter\def\csname v@x.qwen.8.full_iter\endcsname{1.00}
\expandafter\def\csname v@x.qwen.16.debate\endcsname{0.00}
\expandafter\def\csname v@x.qwen.16.pool_ev_nodedup\endcsname{0.31}
\expandafter\def\csname v@x.qwen.16.pool_evpc\endcsname{0.84}
\expandafter\def\csname v@x.qwen.16.full_iter\endcsname{0.98}
\expandafter\def\csname v@x.qwen.32.debate\endcsname{0.00}
\expandafter\def\csname v@x.qwen.32.pool_ev_nodedup\endcsname{0.00}
\expandafter\def\csname v@x.qwen.32.pool_evpc\endcsname{0.60}
\expandafter\def\csname v@x.qwen.32.full_iter\endcsname{1.00}
\expandafter\def\csname v@x.qwen.64.debate\endcsname{0.00}
\expandafter\def\csname v@x.qwen.64.pool_ev_nodedup\endcsname{0.00}
\expandafter\def\csname v@x.qwen.64.pool_evpc\endcsname{0.17}
\expandafter\def\csname v@x.qwen.64.full_iter\endcsname{0.90}
\expandafter\def\csname v@x.qwen.debate.range\endcsname{0.00--0.08}
\expandafter\def\csname v@x.qwen.pool_ev_nodedup.range\endcsname{0.00--0.61}
\expandafter\def\csname v@x.qwen.pool_evpc.range\endcsname{0.17--0.99}
\expandafter\def\csname v@x.qwen.full_iter.range\endcsname{0.90--1.00}
\expandafter\def\csname v@x.olmo.8.debate\endcsname{0.12}
\expandafter\def\csname v@x.olmo.8.pool_ev_nodedup\endcsname{0.58}
\expandafter\def\csname v@x.olmo.8.pool_evpc\endcsname{0.80}
\expandafter\def\csname v@x.olmo.8.full_iter\endcsname{1.00}
\expandafter\def\csname v@x.olmo.16.debate\endcsname{0.00}
\expandafter\def\csname v@x.olmo.16.pool_ev_nodedup\endcsname{0.15}
\expandafter\def\csname v@x.olmo.16.pool_evpc\endcsname{0.61}
\expandafter\def\csname v@x.olmo.16.full_iter\endcsname{1.00}
\expandafter\def\csname v@x.olmo.32.debate\endcsname{0.00}
\expandafter\def\csname v@x.olmo.32.pool_ev_nodedup\endcsname{0.00}
\expandafter\def\csname v@x.olmo.32.pool_evpc\endcsname{0.29}
\expandafter\def\csname v@x.olmo.32.full_iter\endcsname{0.99}
\expandafter\def\csname v@x.olmo.64.debate\endcsname{0.00}
\expandafter\def\csname v@x.olmo.64.pool_ev_nodedup\endcsname{0.00}
\expandafter\def\csname v@x.olmo.64.pool_evpc\endcsname{0.02}
\expandafter\def\csname v@x.olmo.64.full_iter\endcsname{1.00}
\expandafter\def\csname v@x.olmo.debate.range\endcsname{0.00--0.12}
\expandafter\def\csname v@x.olmo.pool_ev_nodedup.range\endcsname{0.00--0.58}
\expandafter\def\csname v@x.olmo.pool_evpc.range\endcsname{0.02--0.80}
\expandafter\def\csname v@x.olmo.full_iter.range\endcsname{0.99--1.00}
\expandafter\def\csname v@x.qwen.64.evpc.debate\endcsname{$+$0.17 [0.08, 0.28]}
\expandafter\def\csname v@x.qwen.64.evpc.nodedup\endcsname{$+$0.17 [0.08, 0.28]}
\expandafter\def\csname v@x.qwen.64.tok.debate\endcsname{378k}
\expandafter\def\csname v@x.qwen.64.tok.pool_ev_nodedup\endcsname{1,073k}
\expandafter\def\csname v@x.qwen.64.tok.pool_evpc\endcsname{292k}
\expandafter\def\csname v@x.qwen.64.tok.full_iter\endcsname{292k}
\expandafter\def\csname v@x.olmo.64.evpc.debate\endcsname{$+$0.02 [0.00, 0.05]}
\expandafter\def\csname v@x.olmo.64.evpc.nodedup\endcsname{$+$0.02 [0.00, 0.05]}
\expandafter\def\csname v@x.olmo.64.tok.debate\endcsname{314k}
\expandafter\def\csname v@x.olmo.64.tok.pool_ev_nodedup\endcsname{973k}
\expandafter\def\csname v@x.olmo.64.tok.pool_evpc\endcsname{283k}
\expandafter\def\csname v@x.olmo.64.tok.full_iter\endcsname{303k}
\expandafter\def\csname v@x.evpc.debate.range\endcsname{$+$0.02 to $+$0.91}
\expandafter\def\csname v@x.full_iter.range\endcsname{0.90--1.00}
\expandafter\def\csname v@x.view.complete.range\endcsname{0.03--0.87}
\expandafter\def\csname v@x.view.incomplete.range\endcsname{0.03--0.33}
\expandafter\def\csname v@x.view.share.range\endcsname{0.00--1.00}
\expandafter\def\csname v@x.qwen.8.complete.pct\endcsname{100.0\%}
\expandafter\def\csname v@x.qwen.8.hidden\endcsname{1.8}
\expandafter\def\csname v@x.qwen.8.complete\endcsname{1.00}
\expandafter\def\csname v@x.qwen.8.after\endcsname{0.85}
\expandafter\def\csname v@x.qwen.16.complete.pct\endcsname{80.4\%}
\expandafter\def\csname v@x.qwen.16.hidden\endcsname{3.5}
\expandafter\def\csname v@x.qwen.16.complete\endcsname{0.80}
\expandafter\def\csname v@x.qwen.16.after\endcsname{0.75}
\expandafter\def\csname v@x.qwen.32.complete.pct\endcsname{47.1\%}
\expandafter\def\csname v@x.qwen.32.hidden\endcsname{7.1}
\expandafter\def\csname v@x.qwen.32.complete\endcsname{0.47}
\expandafter\def\csname v@x.qwen.32.after\endcsname{0.45}
\expandafter\def\csname v@x.qwen.64.complete.pct\endcsname{13.2\%}
\expandafter\def\csname v@x.qwen.64.hidden\endcsname{14.2}
\expandafter\def\csname v@x.qwen.64.complete\endcsname{0.13}
\expandafter\def\csname v@x.qwen.64.after\endcsname{0.20}
\expandafter\def\csname v@x.olmo.8.complete.pct\endcsname{75.7\%}
\expandafter\def\csname v@x.olmo.8.hidden\endcsname{1.5}
\expandafter\def\csname v@x.olmo.8.complete\endcsname{0.76}
\expandafter\def\csname v@x.olmo.8.after\endcsname{0.87}
\expandafter\def\csname v@x.olmo.16.complete.pct\endcsname{38.4\%}
\expandafter\def\csname v@x.olmo.16.hidden\endcsname{2.9}
\expandafter\def\csname v@x.olmo.16.complete\endcsname{0.38}
\expandafter\def\csname v@x.olmo.16.after\endcsname{0.76}
\expandafter\def\csname v@x.olmo.32.complete.pct\endcsname{13.9\%}
\expandafter\def\csname v@x.olmo.32.hidden\endcsname{6.0}
\expandafter\def\csname v@x.olmo.32.complete\endcsname{0.14}
\expandafter\def\csname v@x.olmo.32.after\endcsname{0.39}
\expandafter\def\csname v@x.olmo.64.complete.pct\endcsname{0.4\%}
\expandafter\def\csname v@x.olmo.64.hidden\endcsname{11.2}
\expandafter\def\csname v@x.olmo.64.complete\endcsname{0.00}
\expandafter\def\csname v@x.olmo.64.after\endcsname{0.03}
\expandafter\def\csname v@d2.prov.cite.hb.qwen\endcsname{$-$0.01 [$-$0.04, 0.04]}
\expandafter\def\csname v@d2.prov.debate.hb.qwen\endcsname{$+$0.82 [0.69, 0.92]}
\expandafter\def\csname v@d2.cite.debate.hb.qwen\endcsname{$+$0.04 [$-$0.01, 0.11]}
\expandafter\def\csname v@d2.declared.hb.qwen\endcsname{0.79}
\expandafter\def\csname v@d2.lexical.hb.qwen\endcsname{0.94}
\expandafter\def\csname v@d2.final.prov.hb.qwen\endcsname{0.95}
\expandafter\def\csname v@d2.final.cite.hb.qwen\endcsname{0.91}
\expandafter\def\csname v@d2.debate.hb.qwen.16\endcsname{0.08}
\expandafter\def\csname v@d2.debate_cite.hb.qwen.16\endcsname{0.15}
\expandafter\def\csname v@d2.pool_evpc.hb.qwen.16\endcsname{0.88}
\expandafter\def\csname v@d2.pool_evpc_cite.hb.qwen.16\endcsname{0.93}
\expandafter\def\csname v@d2.pool_prov.hb.qwen.16\endcsname{0.92}
\expandafter\def\csname v@d2.debate.hb.qwen.32\endcsname{0.05}
\expandafter\def\csname v@d2.debate_cite.hb.qwen.32\endcsname{0.07}
\expandafter\def\csname v@d2.pool_evpc.hb.qwen.32\endcsname{0.91}
\expandafter\def\csname v@d2.pool_evpc_cite.hb.qwen.32\endcsname{0.94}
\expandafter\def\csname v@d2.pool_prov.hb.qwen.32\endcsname{0.94}
\expandafter\def\csname v@d2.prov.cite.hb.olmo\endcsname{$-$0.01 [$-$0.04, 0.03]}
\expandafter\def\csname v@d2.prov.debate.hb.olmo\endcsname{$+$0.76 [0.57, 0.90]}
\expandafter\def\csname v@d2.cite.debate.hb.olmo\endcsname{$-$0.04 [$-$0.13, 0.04]}
\expandafter\def\csname v@d2.declared.hb.olmo\endcsname{0.69}
\expandafter\def\csname v@d2.lexical.hb.olmo\endcsname{0.72}
\expandafter\def\csname v@d2.final.prov.hb.olmo\endcsname{0.91}
\expandafter\def\csname v@d2.final.cite.hb.olmo\endcsname{0.81}
\expandafter\def\csname v@d2.debate.hb.olmo.16\endcsname{0.21}
\expandafter\def\csname v@d2.debate_cite.hb.olmo.16\endcsname{0.21}
\expandafter\def\csname v@d2.pool_evpc.hb.olmo.16\endcsname{0.88}
\expandafter\def\csname v@d2.pool_evpc_cite.hb.olmo.16\endcsname{0.96}
\expandafter\def\csname v@d2.pool_prov.hb.olmo.16\endcsname{0.93}
\expandafter\def\csname v@d2.debate.hb.olmo.32\endcsname{0.21}
\expandafter\def\csname v@d2.debate_cite.hb.olmo.32\endcsname{0.12}
\expandafter\def\csname v@d2.pool_evpc.hb.olmo.32\endcsname{0.92}
\expandafter\def\csname v@d2.pool_evpc_cite.hb.olmo.32\endcsname{0.91}
\expandafter\def\csname v@d2.pool_prov.hb.olmo.32\endcsname{0.93}
\expandafter\def\csname v@d2.prov.cite.pz.qwen\endcsname{$-$0.16 [$-$0.20, $-$0.12]}
\expandafter\def\csname v@d2.prov.debate.pz.qwen\endcsname{$+$0.61 [0.54, 0.68]}
\expandafter\def\csname v@d2.cite.debate.pz.qwen\endcsname{$+$0.02 [$-$0.02, 0.05]}
\expandafter\def\csname v@d2.declared.pz.qwen\endcsname{0.40}
\expandafter\def\csname v@d2.lexical.pz.qwen\endcsname{0.57}
\expandafter\def\csname v@d2.final.prov.pz.qwen\endcsname{0.70}
\expandafter\def\csname v@d2.final.cite.pz.qwen\endcsname{0.85}
\expandafter\def\csname v@d2.debate.pz.qwen.16\endcsname{0.10}
\expandafter\def\csname v@d2.debate_cite.pz.qwen.16\endcsname{0.13}
\expandafter\def\csname v@d2.pool_evpc.pz.qwen.16\endcsname{0.85}
\expandafter\def\csname v@d2.pool_evpc_cite.pz.qwen.16\endcsname{0.89}
\expandafter\def\csname v@d2.pool_prov.pz.qwen.16\endcsname{0.71}
\expandafter\def\csname v@d2.debate.pz.qwen.32\endcsname{0.11}
\expandafter\def\csname v@d2.debate_cite.pz.qwen.32\endcsname{0.12}
\expandafter\def\csname v@d2.pool_evpc.pz.qwen.32\endcsname{0.88}
\expandafter\def\csname v@d2.pool_evpc_cite.pz.qwen.32\endcsname{0.90}
\expandafter\def\csname v@d2.pool_prov.pz.qwen.32\endcsname{0.75}
\expandafter\def\csname v@d2.prov.cite.pz.olmo\endcsname{$-$0.15 [$-$0.20, $-$0.10]}
\expandafter\def\csname v@d2.prov.debate.pz.olmo\endcsname{$+$0.50 [0.44, 0.57]}
\expandafter\def\csname v@d2.cite.debate.pz.olmo\endcsname{$-$0.02 [$-$0.04, 0.01]}
\expandafter\def\csname v@d2.declared.pz.olmo\endcsname{0.43}
\expandafter\def\csname v@d2.lexical.pz.olmo\endcsname{0.63}
\expandafter\def\csname v@d2.final.prov.pz.olmo\endcsname{0.70}
\expandafter\def\csname v@d2.final.cite.pz.olmo\endcsname{0.79}
\expandafter\def\csname v@d2.debate.pz.olmo.16\endcsname{0.08}
\expandafter\def\csname v@d2.debate_cite.pz.olmo.16\endcsname{0.06}
\expandafter\def\csname v@d2.pool_evpc.pz.olmo.16\endcsname{0.75}
\expandafter\def\csname v@d2.pool_evpc_cite.pz.olmo.16\endcsname{0.70}
\expandafter\def\csname v@d2.pool_prov.pz.olmo.16\endcsname{0.59}
\expandafter\def\csname v@d2.debate.pz.olmo.32\endcsname{0.09}
\expandafter\def\csname v@d2.debate_cite.pz.olmo.32\endcsname{0.07}
\expandafter\def\csname v@d2.pool_evpc.pz.olmo.32\endcsname{0.78}
\expandafter\def\csname v@d2.pool_evpc_cite.pz.olmo.32\endcsname{0.74}
\expandafter\def\csname v@d2.pool_prov.pz.olmo.32\endcsname{0.55}
\expandafter\def\csname v@d2.prov.cite.mq.qwen\endcsname{$-$0.03 [$-$0.09, 0.02]}
\expandafter\def\csname v@d2.prov.debate.mq.qwen\endcsname{$+$0.44 [0.37, 0.50]}
\expandafter\def\csname v@d2.cite.debate.mq.qwen\endcsname{$-$0.01 [$-$0.05, 0.02]}
\expandafter\def\csname v@d2.declared.mq.qwen\endcsname{0.41}
\expandafter\def\csname v@d2.lexical.mq.qwen\endcsname{0.57}
\expandafter\def\csname v@d2.final.prov.mq.qwen\endcsname{0.80}
\expandafter\def\csname v@d2.final.cite.mq.qwen\endcsname{0.34}
\expandafter\def\csname v@d2.debate.mq.qwen.16\endcsname{0.12}
\expandafter\def\csname v@d2.debate_cite.mq.qwen.16\endcsname{0.10}
\expandafter\def\csname v@d2.pool_evpc.mq.qwen.16\endcsname{0.57}
\expandafter\def\csname v@d2.pool_evpc_cite.mq.qwen.16\endcsname{0.57}
\expandafter\def\csname v@d2.pool_prov.mq.qwen.16\endcsname{0.54}
\expandafter\def\csname v@d2.prov.cite.mq.olmo\endcsname{$-$0.04 [$-$0.10, 0.02]}
\expandafter\def\csname v@d2.prov.debate.mq.olmo\endcsname{$+$0.39 [0.32, 0.46]}
\expandafter\def\csname v@d2.cite.debate.mq.olmo\endcsname{$+$0.03 [$-$0.02, 0.07]}
\expandafter\def\csname v@d2.declared.mq.olmo\endcsname{0.42}
\expandafter\def\csname v@d2.lexical.mq.olmo\endcsname{0.46}
\expandafter\def\csname v@d2.final.prov.mq.olmo\endcsname{0.73}
\expandafter\def\csname v@d2.final.cite.mq.olmo\endcsname{0.33}
\expandafter\def\csname v@d2.debate.mq.olmo.16\endcsname{0.09}
\expandafter\def\csname v@d2.debate_cite.mq.olmo.16\endcsname{0.12}
\expandafter\def\csname v@d2.pool_evpc.mq.olmo.16\endcsname{0.55}
\expandafter\def\csname v@d2.pool_evpc_cite.mq.olmo.16\endcsname{0.55}
\expandafter\def\csname v@d2.pool_prov.mq.olmo.16\endcsname{0.51}
\expandafter\def\csname v@d2.prov.cite.hb.range\endcsname{$-$0.01}
\expandafter\def\csname v@d2.declared.hb.range\endcsname{0.69--0.79}
\expandafter\def\csname v@d2.lexical.hb.range\endcsname{0.72--0.94}
\expandafter\def\csname v@d2.prov.cite.pz.range\endcsname{$-$0.16 to $-$0.15}
\expandafter\def\csname v@d2.declared.pz.range\endcsname{0.40--0.43}
\expandafter\def\csname v@d2.lexical.pz.range\endcsname{0.57--0.63}
\expandafter\def\csname v@d2.prov.cite.mq.range\endcsname{$-$0.04 to $-$0.03}
\expandafter\def\csname v@d2.declared.mq.range\endcsname{0.41--0.42}
\expandafter\def\csname v@d2.lexical.mq.range\endcsname{0.46--0.57}
\expandafter\def\csname v@d2.cite.prov.pz.range\endcsname{0.15--0.16}
\expandafter\def\csname v@d2.prov.debate.range\endcsname{$+$0.39 to $+$0.82}
\expandafter\def\csname v@d2.cite.debate.range\endcsname{$-$0.04 to $+$0.04}
\expandafter\def\csname v@d2.gate\endcsname{0.78}
\expandafter\def\csname v@d2.claim.1\endcsname{4}
\expandafter\def\csname v@d2.claim.1.mark\endcsname{\checkmark~(4 of 6)}
\expandafter\def\csname v@d2.claim.2\endcsname{6}
\expandafter\def\csname v@d2.claim.2.mark\endcsname{\checkmark~(6 of 6)}
\expandafter\def\csname v@pred.D2.in\endcsname{40}
\expandafter\def\csname v@pred.D2.n\endcsname{40}
\expandafter\def\csname v@d3.evpc.hist.hb.qwen\endcsname{$+$0.84 [0.72, 0.93]}
\expandafter\def\csname v@d3.evpc.digest.hb.qwen\endcsname{$+$0.85 [0.74, 0.94]}
\expandafter\def\csname v@d3.hist.debate.hb.qwen\endcsname{$-$0.01 [$-$0.06, 0.03]}
\expandafter\def\csname v@d3.digest.cap.hb.qwen\endcsname{$-$0.06 [$-$0.09, $-$0.03]}
\expandafter\def\csname v@d3.debate.hb.qwen.16\endcsname{0.08}
\expandafter\def\csname v@d3.debate_hist.hb.qwen.16\endcsname{0.05}
\expandafter\def\csname v@d3.pool_ev_cap.hb.qwen.16\endcsname{0.16}
\expandafter\def\csname v@d3.pool_digest.hb.qwen.16\endcsname{0.03}
\expandafter\def\csname v@d3.pool_evpc.hb.qwen.16\endcsname{0.88}
\expandafter\def\csname v@d3.debate.hb.qwen.32\endcsname{0.05}
\expandafter\def\csname v@d3.debate_hist.hb.qwen.32\endcsname{0.06}
\expandafter\def\csname v@d3.pool_ev_cap.hb.qwen.32\endcsname{0.04}
\expandafter\def\csname v@d3.pool_digest.hb.qwen.32\endcsname{0.05}
\expandafter\def\csname v@d3.pool_evpc.hb.qwen.32\endcsname{0.91}
\expandafter\def\csname v@d3.words.hb.qwen\endcsname{59}
\expandafter\def\csname v@d3.evpc.hist.hb.olmo\endcsname{$+$0.62 [0.46, 0.77]}
\expandafter\def\csname v@d3.evpc.digest.hb.olmo\endcsname{$+$0.75 [0.58, 0.89]}
\expandafter\def\csname v@d3.hist.debate.hb.olmo\endcsname{$+$0.06 [$-$0.03, 0.17]}
\expandafter\def\csname v@d3.digest.cap.hb.olmo\endcsname{$+$0.05 [0.00, 0.10]}
\expandafter\def\csname v@d3.debate.hb.olmo.16\endcsname{0.21}
\expandafter\def\csname v@d3.debate_hist.hb.olmo.16\endcsname{0.29}
\expandafter\def\csname v@d3.pool_ev_cap.hb.olmo.16\endcsname{0.12}
\expandafter\def\csname v@d3.pool_digest.hb.olmo.16\endcsname{0.15}
\expandafter\def\csname v@d3.pool_evpc.hb.olmo.16\endcsname{0.88}
\expandafter\def\csname v@d3.debate.hb.olmo.32\endcsname{0.21}
\expandafter\def\csname v@d3.debate_hist.hb.olmo.32\endcsname{0.26}
\expandafter\def\csname v@d3.pool_ev_cap.hb.olmo.32\endcsname{0.08}
\expandafter\def\csname v@d3.pool_digest.hb.olmo.32\endcsname{0.15}
\expandafter\def\csname v@d3.pool_evpc.hb.olmo.32\endcsname{0.92}
\expandafter\def\csname v@d3.words.hb.olmo\endcsname{48}
\expandafter\def\csname v@d3.evpc.hist.pz.qwen\endcsname{$+$0.76 [0.69, 0.82]}
\expandafter\def\csname v@d3.evpc.digest.pz.qwen\endcsname{$+$0.81 [0.74, 0.87]}
\expandafter\def\csname v@d3.hist.debate.pz.qwen\endcsname{0.00 [$-$0.02, 0.02]}
\expandafter\def\csname v@d3.digest.cap.pz.qwen\endcsname{$-$0.07 [$-$0.12, $-$0.03]}
\expandafter\def\csname v@d3.debate.pz.qwen.16\endcsname{0.10}
\expandafter\def\csname v@d3.debate_hist.pz.qwen.16\endcsname{0.11}
\expandafter\def\csname v@d3.pool_ev_cap.pz.qwen.16\endcsname{0.14}
\expandafter\def\csname v@d3.pool_digest.pz.qwen.16\endcsname{0.06}
\expandafter\def\csname v@d3.pool_evpc.pz.qwen.16\endcsname{0.85}
\expandafter\def\csname v@d3.debate.pz.qwen.32\endcsname{0.11}
\expandafter\def\csname v@d3.debate_hist.pz.qwen.32\endcsname{0.11}
\expandafter\def\csname v@d3.pool_ev_cap.pz.qwen.32\endcsname{0.13}
\expandafter\def\csname v@d3.pool_digest.pz.qwen.32\endcsname{0.06}
\expandafter\def\csname v@d3.pool_evpc.pz.qwen.32\endcsname{0.88}
\expandafter\def\csname v@d3.words.pz.qwen\endcsname{40}
\expandafter\def\csname v@d3.evpc.hist.pz.olmo\endcsname{$+$0.66 [0.59, 0.73]}
\expandafter\def\csname v@d3.evpc.digest.pz.olmo\endcsname{$+$0.72 [0.65, 0.78]}
\expandafter\def\csname v@d3.hist.debate.pz.olmo\endcsname{$+$0.02 [0.00, 0.04]}
\expandafter\def\csname v@d3.digest.cap.pz.olmo\endcsname{$+$0.01 [$-$0.01, 0.03]}
\expandafter\def\csname v@d3.debate.pz.olmo.16\endcsname{0.08}
\expandafter\def\csname v@d3.debate_hist.pz.olmo.16\endcsname{0.12}
\expandafter\def\csname v@d3.pool_ev_cap.pz.olmo.16\endcsname{0.04}
\expandafter\def\csname v@d3.pool_digest.pz.olmo.16\endcsname{0.04}
\expandafter\def\csname v@d3.pool_evpc.pz.olmo.16\endcsname{0.75}
\expandafter\def\csname v@d3.debate.pz.olmo.32\endcsname{0.09}
\expandafter\def\csname v@d3.debate_hist.pz.olmo.32\endcsname{0.09}
\expandafter\def\csname v@d3.pool_ev_cap.pz.olmo.32\endcsname{0.05}
\expandafter\def\csname v@d3.pool_digest.pz.olmo.32\endcsname{0.07}
\expandafter\def\csname v@d3.pool_evpc.pz.olmo.32\endcsname{0.78}
\expandafter\def\csname v@d3.words.pz.olmo\endcsname{20}
\expandafter\def\csname v@d3.evpc.hist.range\endcsname{$+$0.62 to $+$0.84}
\expandafter\def\csname v@d3.evpc.digest.range\endcsname{$+$0.72 to $+$0.85}
\expandafter\def\csname v@d3.hist.debate.range\endcsname{$-$0.01 to $+$0.06}
\expandafter\def\csname v@d3.digest.cap.range\endcsname{$-$0.07 to $+$0.05}
\expandafter\def\csname v@d3.hist.debate.max\endcsname{$+$0.06}
\expandafter\def\csname v@d3.hist.range\endcsname{0.05--0.29}
\expandafter\def\csname v@d3.digest.range\endcsname{0.03--0.15}
\expandafter\def\csname v@d3.hist.tok.ratio.range\endcsname{2.1--2.5$\times$}
\expandafter\def\csname v@d3.digest.tok.ratio.range\endcsname{1.0--1.1$\times$}
\expandafter\def\csname v@d3.words.range\endcsname{20--59}
\expandafter\def\csname v@d3.budget.range\endcsname{40--81}
\expandafter\def\csname v@d3.claim.1\endcsname{4}
\expandafter\def\csname v@d3.claim.1.mark\endcsname{\checkmark~(4 of 4)}
\expandafter\def\csname v@d3.claim.2\endcsname{4}
\expandafter\def\csname v@d3.claim.2.mark\endcsname{\checkmark~(4 of 4)}
\expandafter\def\csname v@pred.D3.in\endcsname{26}
\expandafter\def\csname v@pred.D3.n\endcsname{26}
\expandafter\def\csname v@d4.debate.qwen.4\endcsname{0.20}
\expandafter\def\csname v@d4.pool_evpc.qwen.4\endcsname{0.56}
\expandafter\def\csname v@d4.full_iter.qwen.4\endcsname{0.77}
\expandafter\def\csname v@d4.evpc.debate.qwen.4\endcsname{$+$0.36 [0.29, 0.42]}
\expandafter\def\csname v@d4.units.qwen.4\endcsname{2.2}
\expandafter\def\csname v@d4.debate.qwen.8\endcsname{0.13}
\expandafter\def\csname v@d4.pool_evpc.qwen.8\endcsname{0.47}
\expandafter\def\csname v@d4.full_iter.qwen.8\endcsname{0.78}
\expandafter\def\csname v@d4.evpc.debate.qwen.8\endcsname{$+$0.34 [0.27, 0.40]}
\expandafter\def\csname v@d4.units.qwen.8\endcsname{3.1}
\expandafter\def\csname v@d4.debate.qwen.16\endcsname{0.11}
\expandafter\def\csname v@d4.pool_evpc.qwen.16\endcsname{0.45}
\expandafter\def\csname v@d4.full_iter.qwen.16\endcsname{0.70}
\expandafter\def\csname v@d4.evpc.debate.qwen.16\endcsname{$+$0.34 [0.27, 0.40]}
\expandafter\def\csname v@d4.units.qwen.16\endcsname{4.9}
\expandafter\def\csname v@d4.private.qwen.16\endcsname{0.50}
\expandafter\def\csname v@d4.evpc.sag.qwen\endcsname{0.11}
\expandafter\def\csname v@d4.full.sag.qwen\endcsname{0.07}
\expandafter\def\csname v@d4.cross.debate.qwen\endcsname{0.00 [$-$0.03, 0.02]}
\expandafter\def\csname v@d4.cross.evpc.qwen\endcsname{$-$0.11 [$-$0.17, $-$0.06]}
\expandafter\def\csname v@d4.cross.full.qwen\endcsname{$-$0.11 [$-$0.16, $-$0.06]}
\expandafter\def\csname v@d4.debate.olmo.4\endcsname{0.19}
\expandafter\def\csname v@d4.pool_evpc.olmo.4\endcsname{0.47}
\expandafter\def\csname v@d4.full_iter.olmo.4\endcsname{0.65}
\expandafter\def\csname v@d4.evpc.debate.olmo.4\endcsname{$+$0.29 [0.22, 0.36]}
\expandafter\def\csname v@d4.units.olmo.4\endcsname{1.9}
\expandafter\def\csname v@d4.debate.olmo.8\endcsname{0.14}
\expandafter\def\csname v@d4.pool_evpc.olmo.8\endcsname{0.46}
\expandafter\def\csname v@d4.full_iter.olmo.8\endcsname{0.68}
\expandafter\def\csname v@d4.evpc.debate.olmo.8\endcsname{$+$0.32 [0.25, 0.39]}
\expandafter\def\csname v@d4.units.olmo.8\endcsname{2.8}
\expandafter\def\csname v@d4.debate.olmo.16\endcsname{0.12}
\expandafter\def\csname v@d4.pool_evpc.olmo.16\endcsname{0.37}
\expandafter\def\csname v@d4.full_iter.olmo.16\endcsname{0.64}
\expandafter\def\csname v@d4.evpc.debate.olmo.16\endcsname{$+$0.25 [0.18, 0.32]}
\expandafter\def\csname v@d4.units.olmo.16\endcsname{4.2}
\expandafter\def\csname v@d4.private.olmo.16\endcsname{0.56}
\expandafter\def\csname v@d4.evpc.sag.olmo\endcsname{0.10}
\expandafter\def\csname v@d4.full.sag.olmo\endcsname{0.01}
\expandafter\def\csname v@d4.cross.debate.olmo\endcsname{$+$0.03 [0.00, 0.07]}
\expandafter\def\csname v@d4.cross.evpc.olmo\endcsname{$-$0.17 [$-$0.24, $-$0.11]}
\expandafter\def\csname v@d4.cross.full.olmo\endcsname{$-$0.11 [$-$0.18, $-$0.05]}
\expandafter\def\csname v@d4.evpc.debate.range\endcsname{$+$0.25 to $+$0.36}
\expandafter\def\csname v@d4.claim.1\endcsname{holds}
\expandafter\def\csname v@d4.claim.1.mark\endcsname{\checkmark~(2 of 2)}
\expandafter\def\csname v@d4.claim.2\endcsname{holds}
\expandafter\def\csname v@d4.claim.2.mark\endcsname{\checkmark~(2 of 2)}
\expandafter\def\csname v@d4.claim.3\endcsname{holds}
\expandafter\def\csname v@d4.claim.3.mark\endcsname{\checkmark~(2 of 2)}
\expandafter\def\csname v@pred.D4.in\endcsname{24}
\expandafter\def\csname v@pred.D4.n\endcsname{24}
\expandafter\def\csname v@r.test.settings\endcsname{4}
\expandafter\def\csname v@r.test.facts.c0.hb.qwen\endcsname{0.78}
\expandafter\def\csname v@r.test.facts.c8.hb.qwen\endcsname{0.70}
\expandafter\def\csname v@r.test.calls.hb.qwen\endcsname{140}
\expandafter\def\csname v@r.test.facts.c0.pz.qwen\endcsname{0.75}
\expandafter\def\csname v@r.test.facts.c8.pz.qwen\endcsname{0.25}
\expandafter\def\csname v@r.test.calls.pz.qwen\endcsname{582}
\expandafter\def\csname v@r.test.facts.c0.hb.olmo\endcsname{0.67}
\expandafter\def\csname v@r.test.facts.c8.hb.olmo\endcsname{0.44}
\expandafter\def\csname v@r.test.calls.hb.olmo\endcsname{136}
\expandafter\def\csname v@r.test.facts.c0.pz.olmo\endcsname{0.54}
\expandafter\def\csname v@r.test.facts.c8.pz.olmo\endcsname{0.09}
\expandafter\def\csname v@r.test.calls.pz.olmo\endcsname{600}
\expandafter\def\csname v@r.test.facts.c0.range\endcsname{0.54--0.78}
\expandafter\def\csname v@r.test.facts.c8.range\endcsname{0.09--0.70}
\expandafter\def\csname v@r.test.argued.18.range\endcsname{0.11--0.35}
\expandafter\def\csname v@r.test.argued.81.range\endcsname{0.95--1.00}
\expandafter\def\csname v@r.test.facts.drop.range\endcsname{0.08--0.49}
\expandafter\def\csname v@r.test.count.wins.range\endcsname{0--7\%}
\expandafter\def\csname v@r.test.best.facts\endcsname{share}
\expandafter\def\csname v@r.test.best.labels\endcsname{log-ratio, share}
\expandafter\def\csname v@r.test.best.argued\endcsname{log-ratio}
\expandafter\def\csname v@r.test.claim.1\endcsname{4}
\expandafter\def\csname v@r.test.claim.1.holds\endcsname{holds}
\expandafter\def\csname v@r.test.claim.1.mark\endcsname{\checkmark~(4 of 4)}
\expandafter\def\csname v@r.test.claim.2\endcsname{3}
\expandafter\def\csname v@r.test.claim.2.holds\endcsname{holds}
\expandafter\def\csname v@r.test.claim.2.mark\endcsname{\checkmark~(3 of 4)}
\expandafter\def\csname v@r.test.claim.3\endcsname{4}
\expandafter\def\csname v@r.test.claim.3.holds\endcsname{holds}
\expandafter\def\csname v@r.test.claim.3.mark\endcsname{\checkmark~(4 of 4)}
\expandafter\def\csname v@r.test.claim.4\endcsname{4}
\expandafter\def\csname v@r.test.claim.4.holds\endcsname{holds}
\expandafter\def\csname v@r.test.claim.4.mark\endcsname{\checkmark~(4 of 4)}
\expandafter\def\csname v@pred.D1.in\endcsname{8}
\expandafter\def\csname v@pred.D1.n\endcsname{8}
\expandafter\def\csname v@gpuh.D.a30\endcsname{139}
\expandafter\def\csname v@search.peer.acc.qwen\endcsname{$+$0.19 [0.08, 0.29]}
\expandafter\def\csname v@search.peer.accp.qwen\endcsname{0.0005}
\expandafter\def\csname v@search.peer.tok.qwen\endcsname{0.67 [0.65, 0.69]}
\expandafter\def\csname v@search.peer.tokp.qwen\endcsname{0.0001}
\expandafter\def\csname v@search.lead.acc.qwen\endcsname{$+$0.13 [0.03, 0.22]}
\expandafter\def\csname v@search.lead.accp.qwen\endcsname{0.0048}
\expandafter\def\csname v@search.lead.tok.qwen\endcsname{0.91 [0.91, 0.92]}
\expandafter\def\csname v@search.lead.tokp.qwen\endcsname{0.0001}
\expandafter\def\csname v@search.peer.acc.olmo\endcsname{$+$0.32 [0.21, 0.43]}
\expandafter\def\csname v@search.peer.accp.olmo\endcsname{0.0001}
\expandafter\def\csname v@search.peer.tok.olmo\endcsname{0.78 [0.76, 0.80]}
\expandafter\def\csname v@search.peer.tokp.olmo\endcsname{0.0001}
\expandafter\def\csname v@search.lead.acc.olmo\endcsname{$+$0.23 [0.11, 0.35]}
\expandafter\def\csname v@search.lead.accp.olmo\endcsname{0.0001}
\expandafter\def\csname v@search.lead.tok.olmo\endcsname{0.93 [0.93, 0.94]}
\expandafter\def\csname v@search.lead.tokp.olmo\endcsname{0.0001}
\expandafter\def\csname v@search.peer.gain.range\endcsname{0.19--0.32}
\expandafter\def\csname v@search.peer.gain.pp.range\endcsname{19--32}
\expandafter\def\csname v@search.peer.router.range\endcsname{0.65--0.71}
\expandafter\def\csname v@search.peer.base.range\endcsname{0.39--0.46}
\expandafter\def\csname v@search.peer.savings.range\endcsname{22--33\%}
\expandafter\def\csname v@search.lead.gain.range\endcsname{0.13--0.23}
\expandafter\def\csname v@search.lead.gain.pp.range\endcsname{13--23}
\expandafter\def\csname v@search.lead.router.range\endcsname{0.50--0.51}
\expandafter\def\csname v@search.lead.base.range\endcsname{0.27--0.38}
\expandafter\def\csname v@search.lead.savings.range\endcsname{7--9\%}
\expandafter\def\csname v@search.pass\endcsname{8}
\expandafter\def\csname v@search.n\endcsname{8}
\expandafter\def\csname v@search.pred.in\endcsname{7}
\expandafter\def\csname v@search.pred.pass\endcsname{7}
\expandafter\def\csname v@search.cands\endcsname{2,417}
\expandafter\def\csname v@search.kept.qwen\endcsname{150}
\expandafter\def\csname v@search.dev.qwen\endcsname{52}
\expandafter\def\csname v@search.test.qwen\endcsname{98}
\expandafter\def\csname v@search.kept.olmo\endcsname{137}
\expandafter\def\csname v@search.dev.olmo\endcsname{60}
\expandafter\def\csname v@search.test.olmo\endcsname{77}
\expandafter\def\csname v@search.kept.pct\endcsname{6\%}
\expandafter\def\csname v@search.allheld.range\endcsname{59--71\%}
\expandafter\def\csname v@search.distinct.range\endcsname{6.5--8.6}
\expandafter\def\csname v@search.dev.gain.qwen\endcsname{$+$0.14 [0.04, 0.23]}
\expandafter\def\csname v@search.dev.copy.qwen\endcsname{$+$0.01 [$-$0.03, 0.05]}
\expandafter\def\csname v@search.dev.lead.qwen\endcsname{$+$0.12 [0.02, 0.22]}
\expandafter\def\csname v@search.dev.copytok.qwen\endcsname{3.5}
\expandafter\def\csname v@search.single.qwen\endcsname{$+$0.45 [0.27, 0.64]}
\expandafter\def\csname v@search.single.n.qwen\endcsname{22}
\expandafter\def\csname v@search.shared.qwen\endcsname{$+$0.11 [$-$0.01, 0.23]}
\expandafter\def\csname v@search.shared.n.qwen\endcsname{74}
\expandafter\def\csname v@search.dev.gain.olmo\endcsname{$+$0.15 [0.07, 0.24]}
\expandafter\def\csname v@search.dev.copy.olmo\endcsname{$+$0.03 [$-$0.04, 0.09]}
\expandafter\def\csname v@search.dev.lead.olmo\endcsname{0.00 [$-$0.09, 0.09]}
\expandafter\def\csname v@search.dev.copytok.olmo\endcsname{3.0}
\expandafter\def\csname v@search.single.olmo\endcsname{$+$0.44 [0.28, 0.59]}
\expandafter\def\csname v@search.single.n.olmo\endcsname{16}
\expandafter\def\csname v@search.shared.olmo\endcsname{$+$0.30 [0.16, 0.42]}
\expandafter\def\csname v@search.shared.n.olmo\endcsname{61}
\expandafter\def\csname v@search.dev.gain.range\endcsname{$+$0.14 to $+$0.15}
\expandafter\def\csname v@search.dev.copy.range\endcsname{$+$0.01 to $+$0.03}
\expandafter\def\csname v@search.dev.lead.range\endcsname{0.00 to $+$0.12}
\expandafter\def\csname v@search.single.range\endcsname{$+$0.44 to $+$0.45}
\expandafter\def\csname v@search.shared.range\endcsname{$+$0.11 to $+$0.30}
\expandafter\def\csname v@search.dev.gain.4.range\endcsname{$+$0.05 to $+$0.10}
\expandafter\def\csname v@search.dev.gain.8.range\endcsname{$+$0.09 to $+$0.17}
\expandafter\def\csname v@search.dev.gain.16.range\endcsname{$+$0.16 to $+$0.17}
\expandafter\def\csname v@search.dev.gain.32.range\endcsname{$+$0.19 to $+$0.25}
\expandafter\def\csname v@search.dev.copytok.range\endcsname{3.0--3.5}
\expandafter\def\csname v@gpuh.search\endcsname{18}
\expandafter\def\csname v@arch.pool.hb.qwen\endcsname{$+$0.80 [0.66, 0.91]}
\expandafter\def\csname v@arch.ow.hb.qwen\endcsname{$+$0.67 [0.57, 0.77]}
\expandafter\def\csname v@arch.hier.hb.qwen\endcsname{$+$0.60 [0.50, 0.69]}
\expandafter\def\csname v@arch.pool.pz.qwen\endcsname{$+$0.78 [0.74, 0.82]}
\expandafter\def\csname v@arch.ow.pz.qwen\endcsname{$+$0.49 [0.46, 0.53]}
\expandafter\def\csname v@arch.hier.pz.qwen\endcsname{$+$0.41 [0.38, 0.45]}
\expandafter\def\csname v@arch.pool.hb.olmo\endcsname{$+$0.56 [0.45, 0.66]}
\expandafter\def\csname v@arch.ow.hb.olmo\endcsname{$+$0.54 [0.42, 0.66]}
\expandafter\def\csname v@arch.hier.hb.olmo\endcsname{$+$0.53 [0.41, 0.64]}
\expandafter\def\csname v@arch.pool.pz.olmo\endcsname{$+$0.70 [0.65, 0.74]}
\expandafter\def\csname v@arch.ow.pz.olmo\endcsname{$+$0.45 [0.41, 0.49]}
\expandafter\def\csname v@arch.hier.pz.olmo\endcsname{$+$0.32 [0.29, 0.36]}
\expandafter\def\csname v@arch.pass\endcsname{12}
\expandafter\def\csname v@arch.n\endcsname{12}
\expandafter\def\csname v@arch.pool.gain.range\endcsname{$+$0.56 to $+$0.80}
\expandafter\def\csname v@arch.ow.gain.range\endcsname{$+$0.45 to $+$0.67}
\expandafter\def\csname v@arch.hier.gain.range\endcsname{$+$0.32 to $+$0.60}
\expandafter\def\csname v@arch.star.gain.range\endcsname{$+$0.60 to $+$0.69}
\expandafter\def\csname v@arch.decider.gain.min\endcsname{32}
\expandafter\def\csname v@arch.decider.gain.max\endcsname{67}
\expandafter\def\csname v@arch.decider.gain.range\endcsname{$+$0.32 to $+$0.67}
\expandafter\def\csname v@arch.decider.pass\endcsname{8}
\expandafter\def\csname v@arch.decider.n\endcsname{8}
\expandafter\def\csname v@arch.star.pass\endcsname{8}
\expandafter\def\csname v@arch.star.n\endcsname{8}
\expandafter\def\csname v@arch.raw.max\endcsname{0.05}
\expandafter\def\csname v@arch.evpc.range\endcsname{0.36--0.68}
\expandafter\def\csname v@arch.ow.raw.max\endcsname{0.05}
\expandafter\def\csname v@arch.ow.evpc.range\endcsname{0.50--0.68}
\expandafter\def\csname v@arch.hier.raw.max\endcsname{0.05}
\expandafter\def\csname v@arch.hier.evpc.range\endcsname{0.36--0.62}
\expandafter\def\csname v@arch.star.evpc.range\endcsname{0.65--0.72}
\expandafter\def\csname v@arch.star.raw.max\endcsname{0.06}
\expandafter\def\csname v@arch.peer.evpc.range\endcsname{0.79--0.90}
\expandafter\def\csname v@arch.peer.raw.range\endcsname{0.05--0.23}
\expandafter\def\csname v@arch.below.peer.range\endcsname{0.16--0.45}
\expandafter\def\csname v@arch.below.star.range\endcsname{0.12--0.24}
\expandafter\def\csname v@arch.dev.pool.gain.range\endcsname{$+$0.65 to $+$0.85}
\expandafter\def\csname v@arch.dev.ow.gain.range\endcsname{$+$0.42 to $+$0.80}
\expandafter\def\csname v@arch.dev.hier.gain.range\endcsname{$+$0.29 to $+$0.64}
\expandafter\def\csname v@arch.dev.star.gain.range\endcsname{$+$0.57 to $+$0.82}
\expandafter\def\csname v@arch.dev.below.peer.range\endcsname{0.06--0.47}
\expandafter\def\csname v@arch.wrong.ow.pz\endcsname{47--50\%}
\expandafter\def\csname v@arch.items.hb.qwen\endcsname{37}
\expandafter\def\csname v@arch.items.pz.qwen\endcsname{200}
\expandafter\def\csname v@arch.items.hb.olmo\endcsname{38}
\expandafter\def\csname v@arch.items.pz.olmo\endcsname{186}
\expandafter\def\csname v@arch.reach.ow.range\endcsname{0.81--0.91}
\expandafter\def\csname v@arch.reach.ow.raw.range\endcsname{0.51--0.70}
\expandafter\def\csname v@arch.reach.hier.range\endcsname{0.58--0.81}
\expandafter\def\csname v@arch.reach.hier.raw.range\endcsname{0.21--0.49}
\expandafter\def\csname v@arch.copies.raw.range\endcsname{2.6--76}
\expandafter\def\csname v@arch.copies.evpc.max\endcsname{0}
\expandafter\def\csname v@arch.tok.run.pool\endcsname{80\%}
\expandafter\def\csname v@arch.tok.run.pool.range\endcsname{0.17--0.23}
\expandafter\def\csname v@arch.tok.prompt.pool\endcsname{84\%}
\expandafter\def\csname v@arch.tok.prompt.pool.range\endcsname{0.12--0.19}
\expandafter\def\csname v@arch.tok.run.ow\endcsname{11\%}
\expandafter\def\csname v@arch.tok.run.ow.range\endcsname{0.87--0.92}
\expandafter\def\csname v@arch.tok.prompt.ow\endcsname{79\%}
\expandafter\def\csname v@arch.tok.prompt.ow.range\endcsname{0.16--0.27}
\expandafter\def\csname v@arch.tok.run.hier\endcsname{11\%}
\expandafter\def\csname v@arch.tok.run.hier.range\endcsname{0.88--0.91}
\expandafter\def\csname v@arch.tok.prompt.hier\endcsname{46\%}
\expandafter\def\csname v@arch.tok.prompt.hier.range\endcsname{0.50--0.58}
\expandafter\def\csname v@arch.tok.peer.every.ratio.range\endcsname{1.6 to 5.9}
\expandafter\def\csname v@arch.tok.decider.share.range\endcsname{4--18\%}
\expandafter\def\csname v@arch.planfail.max\endcsname{7.3\%}
\expandafter\def\csname v@arch.pred.in\endcsname{95}
\expandafter\def\csname v@arch.pred.n\endcsname{96}
\expandafter\def\csname v@arch.excl.hb.qwen.pool_evpc\endcsname{21}
\expandafter\def\csname v@arch.excl.hb.qwen.pool_raw\endcsname{9}
\expandafter\def\csname v@arch.exclitems.hb.qwen\endcsname{2}
\expandafter\def\csname v@arch.excl.pz.qwen.pool_raw\endcsname{1}
\expandafter\def\csname v@arch.excl.pz.qwen.pool_evpc\endcsname{14}
\expandafter\def\csname v@arch.exclitems.pz.qwen\endcsname{0}
\expandafter\def\csname v@arch.excl.hb.olmo.pool_evpc\endcsname{5}
\expandafter\def\csname v@arch.exclitems.hb.olmo\endcsname{3}
\expandafter\def\csname v@arch.excl.pz.olmo.pool_evpc\endcsname{247}
\expandafter\def\csname v@arch.excl.pz.olmo.pool_raw\endcsname{103}
\expandafter\def\csname v@arch.exclitems.pz.olmo\endcsname{0}
\expandafter\def\csname v@gpuh.Y1\endcsname{56}
\expandafter\def\csname v@gpuh.Y2\endcsname{140}
\expandafter\def\csname v@gpuh.test\endcsname{129}
\expandafter\def\csname v@gpuh.H\endcsname{19.5}
\expandafter\def\csname v@gpuh.H.a30\endcsname{17.0}
\expandafter\def\csname v@gpuh.I\endcsname{4.8}
\expandafter\def\csname v@gpuh.h100.total\endcsname{148}
\expandafter\def\csname v@gpuh.a30.total\endcsname{704}
\expandafter\def\csname v@gpuh.task.total\endcsname{73}
\expandafter\def\csname v@C1.hb.qwen.pc\endcsname{$+$75 [62, 86]}
\expandafter\def\csname v@C1p.hb.qwen.pc\endcsname{0.0}
\expandafter\def\csname v@C2.hb.qwen.pc\endcsname{$+$63 [52, 72]}
\expandafter\def\csname v@C2p.hb.qwen.pc\endcsname{0.0}
\expandafter\def\csname v@C3.hb.qwen.pc\endcsname{$+$11 [5, 18]}
\expandafter\def\csname v@C3p.hb.qwen.pc\endcsname{0.0}
\expandafter\def\csname v@C4p.hb.qwen.pc\endcsname{10}
\expandafter\def\csname v@C1.pz.qwen.pc\endcsname{$+$77 [72, 81]}
\expandafter\def\csname v@C1p.pz.qwen.pc\endcsname{0.0}
\expandafter\def\csname v@C2.pz.qwen.pc\endcsname{$+$74 [69, 78]}
\expandafter\def\csname v@C2p.pz.qwen.pc\endcsname{0.0}
\expandafter\def\csname v@C3.pz.qwen.pc\endcsname{$+$71 [65, 77]}
\expandafter\def\csname v@C3p.pz.qwen.pc\endcsname{0.0}
\expandafter\def\csname v@C1.mq.qwen.pc\endcsname{$+$41 [36, 45]}
\expandafter\def\csname v@C1p.mq.qwen.pc\endcsname{0.0}
\expandafter\def\csname v@C2.mq.qwen.pc\endcsname{$+$36 [31, 40]}
\expandafter\def\csname v@C2p.mq.qwen.pc\endcsname{0.0}
\expandafter\def\csname v@C3.mq.qwen.pc\endcsname{$+$8 [5, 11]}
\expandafter\def\csname v@C3p.mq.qwen.pc\endcsname{0.0}
\expandafter\def\csname v@C1.hb.olmo.pc\endcsname{$+$60 [50, 71]}
\expandafter\def\csname v@C1p.hb.olmo.pc\endcsname{0.0}
\expandafter\def\csname v@C2.hb.olmo.pc\endcsname{$+$44 [33, 54]}
\expandafter\def\csname v@C2p.hb.olmo.pc\endcsname{0.0}
\expandafter\def\csname v@C3.hb.olmo.pc\endcsname{$+$34 [22, 47]}
\expandafter\def\csname v@C3p.hb.olmo.pc\endcsname{0.0}
\expandafter\def\csname v@C4.hb.olmo.pc\endcsname{61 [13, 96]}
\expandafter\def\csname v@C4p.hb.olmo.pc\endcsname{1}
\expandafter\def\csname v@C1.pz.olmo.pc\endcsname{$+$70 [66, 74]}
\expandafter\def\csname v@C1p.pz.olmo.pc\endcsname{0.0}
\expandafter\def\csname v@C2.pz.olmo.pc\endcsname{$+$65 [61, 70]}
\expandafter\def\csname v@C2p.pz.olmo.pc\endcsname{0.0}
\expandafter\def\csname v@C3.pz.olmo.pc\endcsname{$+$71 [66, 76]}
\expandafter\def\csname v@C3p.pz.olmo.pc\endcsname{0.0}
\expandafter\def\csname v@C1.mq.olmo.pc\endcsname{$+$39 [34, 44]}
\expandafter\def\csname v@C1p.mq.olmo.pc\endcsname{0.0}
\expandafter\def\csname v@C2.mq.olmo.pc\endcsname{$+$39 [35, 44]}
\expandafter\def\csname v@C2p.mq.olmo.pc\endcsname{0.0}
\expandafter\def\csname v@C3.mq.olmo.pc\endcsname{$+$11 [7, 15]}
\expandafter\def\csname v@C3p.mq.olmo.pc\endcsname{0.0}
\expandafter\def\csname v@C1.min.pc\endcsname{39}
\expandafter\def\csname v@C1.max.pc\endcsname{77}
\expandafter\def\csname v@C2.min.pc\endcsname{36}
\expandafter\def\csname v@C2.max.pc\endcsname{74}
\expandafter\def\csname v@size.debate.range.pc\endcsname{6--16}
\expandafter\def\csname v@size.evpc.range.pc\endcsname{76--89}
\expandafter\def\csname v@size.gain.range.pc\endcsname{$+$49 to $+$83}
\expandafter\def\csname v@size.gain.lo.min.pc\endcsname{37}
\expandafter\def\csname v@size.tok.ratio.range.pc\endcsname{61--83}
\expandafter\def\csname v@fil.cap.gap.abs.range.pc\endcsname{68--86}
\expandafter\def\csname v@fil.ev.gap.range.pc\endcsname{6--15}
\expandafter\def\csname v@fil.nodedup.gap.range.pc\endcsname{31--74}
\expandafter\def\csname v@fil.hist.range.pc\endcsname{5--22}
\expandafter\def\csname v@fil.digest.range.pc\endcsname{2--9}
\expandafter\def\csname v@fil.cap.range.pc\endcsname{1--12}
\expandafter\def\csname v@fil.hist.debate.range.pc\endcsname{$-$1 to $+$12}
\expandafter\def\csname v@fil.ev.hb.range.pc\endcsname{70--75}
\expandafter\def\csname v@fil.evpc.hb.range.pc\endcsname{78--86}
\expandafter\def\csname v@fil.capdedup.hb.range.pc\endcsname{56--81}
\expandafter\def\csname v@fil.nodedup.hb.range.pc\endcsname{42--53}
\expandafter\def\csname v@fil.ev.pz.range.pc\endcsname{68--84}
\expandafter\def\csname v@fil.evpc.pz.range.pc\endcsname{80--91}
\expandafter\def\csname v@fil.capdedup.pz.range.pc\endcsname{25--48}
\expandafter\def\csname v@fil.nodedup.pz.range.pc\endcsname{7--24}
\expandafter\def\csname v@z5.own.range.pc\endcsname{$-$51 to $-$35}
\expandafter\def\csname v@z5.claims.range.pc\endcsname{$-$58 to $-$23}
\expandafter\def\csname v@z5.para.range.pc\endcsname{$-$2 to 0}
\expandafter\def\csname v@z5.para.own.range.pc\endcsname{$+$32 to $+$50}
\expandafter\def\csname v@z5.view.wrong.range.pc\endcsname{59--73}
\expandafter\def\csname v@z5.view.held.range.pc\endcsname{38--68}
\expandafter\def\csname v@g5.hb.N4.debate.pc\endcsname{39 [26, 53]}
\expandafter\def\csname v@g5.hb.N4.pool_evpc.pc\endcsname{94 [87, 99]}
\expandafter\def\csname v@g5.hb.N4.full_iter.pc\endcsname{94 [86, 99]}
\expandafter\def\csname v@g5.hb.N4.diff.pc\endcsname{$+$55 [41, 68]}
\expandafter\def\csname v@g5.hb.N8.debate.pc\endcsname{31 [20, 44]}
\expandafter\def\csname v@g5.hb.N8.pool_evpc.pc\endcsname{93 [85, 98]}
\expandafter\def\csname v@g5.hb.N8.full_iter.pc\endcsname{96 [91, 100]}
\expandafter\def\csname v@g5.hb.N8.diff.pc\endcsname{$+$61 [47, 74]}
\expandafter\def\csname v@g5.hb.N16.debate.pc\endcsname{27 [17, 38]}
\expandafter\def\csname v@g5.hb.N16.pool_evpc.pc\endcsname{92 [83, 98]}
\expandafter\def\csname v@g5.hb.N16.full_iter.pc\endcsname{96 [89, 100]}
\expandafter\def\csname v@g5.hb.N16.diff.pc\endcsname{$+$65 [52, 76]}
\expandafter\def\csname v@g5.hb.N32.debate.pc\endcsname{18 [9, 27]}
\expandafter\def\csname v@g5.hb.N32.pool_evpc.pc\endcsname{91 [82, 98]}
\expandafter\def\csname v@g5.hb.N32.full_iter.pc\endcsname{95 [86, 100]}
\expandafter\def\csname v@g5.hb.N32.diff.pc\endcsname{$+$74 [63, 83]}
\expandafter\def\csname v@g5.diff.pooled.pc\endcsname{$+$63 [53, 74]}
\expandafter\def\csname v@g5.diff.est.pc\endcsname{63}
\expandafter\def\csname v@g5.diff.p.pc\endcsname{0.1}
\expandafter\def\csname v@g5.allruns.pc\endcsname{$+$61 [50, 71] (41 tasks)}
\expandafter\def\csname v@g5.debate.span.pc\endcsname{39 to 18}
\expandafter\def\csname v@g5.pool_evpc.range.pc\endcsname{91--94}
\expandafter\def\csname v@g5.full_iter.range.pc\endcsname{94--96}
\expandafter\def\csname v@g5.pilot.diff.pc\endcsname{$+$59 [45, 72]}
\expandafter\def\csname v@g5h.hb.N4.debate.pc\endcsname{48 [36, 61]}
\expandafter\def\csname v@g5h.hb.N4.pool_evpc.pc\endcsname{96 [89, 100]}
\expandafter\def\csname v@g5h.hb.N4.full_iter.pc\endcsname{91 [81, 98]}
\expandafter\def\csname v@g5h.hb.N4.diff.pc\endcsname{$+$48 [35, 60]}
\expandafter\def\csname v@g5h.hb.N8.debate.pc\endcsname{36 [24, 49]}
\expandafter\def\csname v@g5h.hb.N8.pool_evpc.pc\endcsname{97 [90, 100]}
\expandafter\def\csname v@g5h.hb.N8.full_iter.pc\endcsname{92 [83, 99]}
\expandafter\def\csname v@g5h.hb.N8.diff.pc\endcsname{$+$60 [47, 73]}
\expandafter\def\csname v@g5h.hb.N16.debate.pc\endcsname{34 [21, 47]}
\expandafter\def\csname v@g5h.hb.N16.pool_evpc.pc\endcsname{95 [88, 100]}
\expandafter\def\csname v@g5h.hb.N16.full_iter.pc\endcsname{92 [83, 99]}
\expandafter\def\csname v@g5h.hb.N16.diff.pc\endcsname{$+$61 [48, 75]}
\expandafter\def\csname v@g5h.hb.N32.debate.pc\endcsname{30 [17, 44]}
\expandafter\def\csname v@g5h.hb.N32.pool_evpc.pc\endcsname{95 [88, 100]}
\expandafter\def\csname v@g5h.hb.N32.full_iter.pc\endcsname{92 [83, 99]}
\expandafter\def\csname v@g5h.hb.N32.diff.pc\endcsname{$+$65 [51, 78]}
\expandafter\def\csname v@g5h.diff.pooled.pc\endcsname{$+$59 [47, 70]}
\expandafter\def\csname v@g5h.diff.est.pc\endcsname{59}
\expandafter\def\csname v@g5h.diff.p.pc\endcsname{0.1}
\expandafter\def\csname v@g5h.allruns.pc\endcsname{$+$53 [42, 65] (41 tasks)}
\expandafter\def\csname v@g5h.debate.span.pc\endcsname{48 to 30}
\expandafter\def\csname v@g5h.pool_evpc.range.pc\endcsname{95--97}
\expandafter\def\csname v@g5h.full_iter.range.pc\endcsname{91--92}
\expandafter\def\csname v@g5h.pilot.diff.pc\endcsname{$+$46 [31, 62]}
\expandafter\def\csname v@D.hb.gemma.balanced.16.pc\endcsname{43}
\expandafter\def\csname v@D.hb.gemma.debate.16.pc\endcsname{34}
\expandafter\def\csname v@D.hb.gemma.debate_evpc.16.pc\endcsname{94}
\expandafter\def\csname v@D.hb.gemma.full.16.pc\endcsname{94}
\expandafter\def\csname v@D.hb.gemma.full_iter.16.pc\endcsname{97}
\expandafter\def\csname v@D.hb.gemma.pool_evpc.16.pc\endcsname{96}
\expandafter\def\csname v@D.hb.gemma.solo.16.pc\endcsname{1}
\expandafter\def\csname v@D.hb.gemma.balanced.32.pc\endcsname{41}
\expandafter\def\csname v@D.hb.gemma.debate.32.pc\endcsname{28}
\expandafter\def\csname v@D.hb.gemma.debate_evpc.32.pc\endcsname{88}
\expandafter\def\csname v@D.hb.gemma.full.32.pc\endcsname{91}
\expandafter\def\csname v@D.hb.gemma.full_iter.32.pc\endcsname{95}
\expandafter\def\csname v@D.hb.gemma.pool_evpc.32.pc\endcsname{96}
\expandafter\def\csname v@D.hb.gemma.solo.32.pc\endcsname{2}
\expandafter\def\csname v@D.hb.gemma.balanced.4.pc\endcsname{56}
\expandafter\def\csname v@D.hb.gemma.debate.4.pc\endcsname{55}
\expandafter\def\csname v@D.hb.gemma.debate_evpc.4.pc\endcsname{96}
\expandafter\def\csname v@D.hb.gemma.full.4.pc\endcsname{91}
\expandafter\def\csname v@D.hb.gemma.full_iter.4.pc\endcsname{91}
\expandafter\def\csname v@D.hb.gemma.pool_evpc.4.pc\endcsname{95}
\expandafter\def\csname v@D.hb.gemma.solo.4.pc\endcsname{9}
\expandafter\def\csname v@D.hb.gemma.balanced.8.pc\endcsname{50}
\expandafter\def\csname v@D.hb.gemma.debate.8.pc\endcsname{39}
\expandafter\def\csname v@D.hb.gemma.debate_evpc.8.pc\endcsname{93}
\expandafter\def\csname v@D.hb.gemma.full.8.pc\endcsname{92}
\expandafter\def\csname v@D.hb.gemma.full_iter.8.pc\endcsname{94}
\expandafter\def\csname v@D.hb.gemma.pool_evpc.8.pc\endcsname{93}
\expandafter\def\csname v@D.hb.gemma.solo.8.pc\endcsname{1}
\expandafter\def\csname v@D.pz.gemma.balanced.16.pc\endcsname{32}
\expandafter\def\csname v@D.pz.gemma.debate.16.pc\endcsname{25}
\expandafter\def\csname v@D.pz.gemma.debate_evpc.16.pc\endcsname{82}
\expandafter\def\csname v@D.pz.gemma.full.16.pc\endcsname{100}
\expandafter\def\csname v@D.pz.gemma.full_iter.16.pc\endcsname{100}
\expandafter\def\csname v@D.pz.gemma.pool_evpc.16.pc\endcsname{99}
\expandafter\def\csname v@D.pz.gemma.solo.16.pc\endcsname{4}
\expandafter\def\csname v@D.pz.gemma.balanced.32.pc\endcsname{29}
\expandafter\def\csname v@D.pz.gemma.debate.32.pc\endcsname{21}
\expandafter\def\csname v@D.pz.gemma.debate_evpc.32.pc\endcsname{25}
\expandafter\def\csname v@D.pz.gemma.full.32.pc\endcsname{100}
\expandafter\def\csname v@D.pz.gemma.full_iter.32.pc\endcsname{100}
\expandafter\def\csname v@D.pz.gemma.pool_evpc.32.pc\endcsname{99}
\expandafter\def\csname v@D.pz.gemma.solo.32.pc\endcsname{4}
\expandafter\def\csname v@D.pz.gemma.balanced.4.pc\endcsname{41}
\expandafter\def\csname v@D.pz.gemma.debate.4.pc\endcsname{37}
\expandafter\def\csname v@D.pz.gemma.debate_evpc.4.pc\endcsname{100}
\expandafter\def\csname v@D.pz.gemma.full.4.pc\endcsname{100}
\expandafter\def\csname v@D.pz.gemma.full_iter.4.pc\endcsname{100}
\expandafter\def\csname v@D.pz.gemma.pool_evpc.4.pc\endcsname{99}
\expandafter\def\csname v@D.pz.gemma.solo.4.pc\endcsname{2}
\expandafter\def\csname v@D.pz.gemma.balanced.8.pc\endcsname{34}
\expandafter\def\csname v@D.pz.gemma.debate.8.pc\endcsname{31}
\expandafter\def\csname v@D.pz.gemma.debate_evpc.8.pc\endcsname{100}
\expandafter\def\csname v@D.pz.gemma.full.8.pc\endcsname{100}
\expandafter\def\csname v@D.pz.gemma.full_iter.8.pc\endcsname{100}
\expandafter\def\csname v@D.pz.gemma.pool_evpc.8.pc\endcsname{99}
\expandafter\def\csname v@D.pz.gemma.solo.8.pc\endcsname{3}
\expandafter\def\csname v@D.hb.mini.balanced.16.pc\endcsname{43}
\expandafter\def\csname v@D.hb.mini.debate.16.pc\endcsname{36}
\expandafter\def\csname v@D.hb.mini.debate_evpc.16.pc\endcsname{79}
\expandafter\def\csname v@D.hb.mini.full.16.pc\endcsname{74}
\expandafter\def\csname v@D.hb.mini.full_iter.16.pc\endcsname{70}
\expandafter\def\csname v@D.hb.mini.pool_evpc.16.pc\endcsname{86}
\expandafter\def\csname v@D.hb.mini.solo.16.pc\endcsname{0}
\expandafter\def\csname v@D.hb.mini.balanced.32.pc\endcsname{33}
\expandafter\def\csname v@D.hb.mini.debate.32.pc\endcsname{37}
\expandafter\def\csname v@D.hb.mini.debate_evpc.32.pc\endcsname{55}
\expandafter\def\csname v@D.hb.mini.full.32.pc\endcsname{74}
\expandafter\def\csname v@D.hb.mini.full_iter.32.pc\endcsname{80}
\expandafter\def\csname v@D.hb.mini.pool_evpc.32.pc\endcsname{87}
\expandafter\def\csname v@D.hb.mini.solo.32.pc\endcsname{1}
\expandafter\def\csname v@D.hb.mini.balanced.4.pc\endcsname{45}
\expandafter\def\csname v@D.hb.mini.debate.4.pc\endcsname{43}
\expandafter\def\csname v@D.hb.mini.debate_evpc.4.pc\endcsname{84}
\expandafter\def\csname v@D.hb.mini.full.4.pc\endcsname{69}
\expandafter\def\csname v@D.hb.mini.full_iter.4.pc\endcsname{66}
\expandafter\def\csname v@D.hb.mini.pool_evpc.4.pc\endcsname{84}
\expandafter\def\csname v@D.hb.mini.solo.4.pc\endcsname{9}
\expandafter\def\csname v@D.hb.mini.balanced.8.pc\endcsname{49}
\expandafter\def\csname v@D.hb.mini.debate.8.pc\endcsname{40}
\expandafter\def\csname v@D.hb.mini.debate_evpc.8.pc\endcsname{84}
\expandafter\def\csname v@D.hb.mini.full.8.pc\endcsname{74}
\expandafter\def\csname v@D.hb.mini.full_iter.8.pc\endcsname{67}
\expandafter\def\csname v@D.hb.mini.pool_evpc.8.pc\endcsname{82}
\expandafter\def\csname v@D.hb.mini.solo.8.pc\endcsname{2}
\expandafter\def\csname v@D.pz.mini.balanced.16.pc\endcsname{9}
\expandafter\def\csname v@D.pz.mini.debate.16.pc\endcsname{9}
\expandafter\def\csname v@D.pz.mini.debate_evpc.16.pc\endcsname{67}
\expandafter\def\csname v@D.pz.mini.full.16.pc\endcsname{99}
\expandafter\def\csname v@D.pz.mini.full_iter.16.pc\endcsname{98}
\expandafter\def\csname v@D.pz.mini.pool_evpc.16.pc\endcsname{96}
\expandafter\def\csname v@D.pz.mini.solo.16.pc\endcsname{0}
\expandafter\def\csname v@D.pz.mini.balanced.32.pc\endcsname{11}
\expandafter\def\csname v@D.pz.mini.debate.32.pc\endcsname{7}
\expandafter\def\csname v@D.pz.mini.debate_evpc.32.pc\endcsname{4}
\expandafter\def\csname v@D.pz.mini.full.32.pc\endcsname{100}
\expandafter\def\csname v@D.pz.mini.full_iter.32.pc\endcsname{99}
\expandafter\def\csname v@D.pz.mini.pool_evpc.32.pc\endcsname{98}
\expandafter\def\csname v@D.pz.mini.solo.32.pc\endcsname{0}
\expandafter\def\csname v@D.pz.mini.balanced.4.pc\endcsname{16}
\expandafter\def\csname v@D.pz.mini.debate.4.pc\endcsname{11}
\expandafter\def\csname v@D.pz.mini.debate_evpc.4.pc\endcsname{93}
\expandafter\def\csname v@D.pz.mini.full.4.pc\endcsname{95}
\expandafter\def\csname v@D.pz.mini.full_iter.4.pc\endcsname{94}
\expandafter\def\csname v@D.pz.mini.pool_evpc.4.pc\endcsname{95}
\expandafter\def\csname v@D.pz.mini.solo.4.pc\endcsname{1}
\expandafter\def\csname v@D.pz.mini.balanced.8.pc\endcsname{12}
\expandafter\def\csname v@D.pz.mini.debate.8.pc\endcsname{13}
\expandafter\def\csname v@D.pz.mini.debate_evpc.8.pc\endcsname{93}
\expandafter\def\csname v@D.pz.mini.full.8.pc\endcsname{99}
\expandafter\def\csname v@D.pz.mini.full_iter.8.pc\endcsname{96}
\expandafter\def\csname v@D.pz.mini.pool_evpc.8.pc\endcsname{95}
\expandafter\def\csname v@D.pz.mini.solo.8.pc\endcsname{0}
\expandafter\def\csname v@D.hb.olmo.balanced.16.pc\endcsname{20}
\expandafter\def\csname v@D.hb.olmo.countblind.16.pc\endcsname{27}
\expandafter\def\csname v@D.hb.olmo.debate.16.pc\endcsname{16}
\expandafter\def\csname v@D.hb.olmo.debate_anon.16.pc\endcsname{13}
\expandafter\def\csname v@D.hb.olmo.debate_evpc.16.pc\endcsname{71}
\expandafter\def\csname v@D.hb.olmo.exchange.16.pc\endcsname{31}
\expandafter\def\csname v@D.hb.olmo.full.16.pc\endcsname{48}
\expandafter\def\csname v@D.hb.olmo.full_iter.16.pc\endcsname{67}
\expandafter\def\csname v@D.hb.olmo.pool_evpc.16.pc\endcsname{79}
\expandafter\def\csname v@D.hb.olmo.shareall.16.pc\endcsname{5}
\expandafter\def\csname v@D.hb.olmo.solo.16.pc\endcsname{0}
\expandafter\def\csname v@D.hb.olmo.balanced.32.pc\endcsname{22}
\expandafter\def\csname v@D.hb.olmo.countblind.32.pc\endcsname{31}
\expandafter\def\csname v@D.hb.olmo.debate.32.pc\endcsname{10}
\expandafter\def\csname v@D.hb.olmo.debate_anon.32.pc\endcsname{12}
\expandafter\def\csname v@D.hb.olmo.debate_evpc.32.pc\endcsname{43}
\expandafter\def\csname v@D.hb.olmo.exchange.32.pc\endcsname{21}
\expandafter\def\csname v@D.hb.olmo.full.32.pc\endcsname{57}
\expandafter\def\csname v@D.hb.olmo.full_iter.32.pc\endcsname{70}
\expandafter\def\csname v@D.hb.olmo.pool_evpc.32.pc\endcsname{78}
\expandafter\def\csname v@D.hb.olmo.shareall.32.pc\endcsname{8}
\expandafter\def\csname v@D.hb.olmo.solo.32.pc\endcsname{0}
\expandafter\def\csname v@D.hb.olmo.balanced.4.pc\endcsname{28}
\expandafter\def\csname v@D.hb.olmo.countblind.4.pc\endcsname{29}
\expandafter\def\csname v@D.hb.olmo.debate.4.pc\endcsname{23}
\expandafter\def\csname v@D.hb.olmo.debate_anon.4.pc\endcsname{24}
\expandafter\def\csname v@D.hb.olmo.debate_evpc.4.pc\endcsname{75}
\expandafter\def\csname v@D.hb.olmo.exchange.4.pc\endcsname{44}
\expandafter\def\csname v@D.hb.olmo.full.4.pc\endcsname{52}
\expandafter\def\csname v@D.hb.olmo.full_iter.4.pc\endcsname{63}
\expandafter\def\csname v@D.hb.olmo.pool_evpc.4.pc\endcsname{74}
\expandafter\def\csname v@D.hb.olmo.shareall.4.pc\endcsname{18}
\expandafter\def\csname v@D.hb.olmo.solo.4.pc\endcsname{5}
\expandafter\def\csname v@D.hb.olmo.debate.64.pc\endcsname{8}
\expandafter\def\csname v@D.hb.olmo.full.64.pc\endcsname{60}
\expandafter\def\csname v@D.hb.olmo.full_iter.64.pc\endcsname{70}
\expandafter\def\csname v@D.hb.olmo.pool_evpc.64.pc\endcsname{80}
\expandafter\def\csname v@D.hb.olmo.solo.64.pc\endcsname{0}
\expandafter\def\csname v@D.hb.olmo.balanced.8.pc\endcsname{29}
\expandafter\def\csname v@D.hb.olmo.countblind.8.pc\endcsname{23}
\expandafter\def\csname v@D.hb.olmo.debate.8.pc\endcsname{16}
\expandafter\def\csname v@D.hb.olmo.debate_anon.8.pc\endcsname{15}
\expandafter\def\csname v@D.hb.olmo.debate_evpc.8.pc\endcsname{76}
\expandafter\def\csname v@D.hb.olmo.exchange.8.pc\endcsname{39}
\expandafter\def\csname v@D.hb.olmo.full.8.pc\endcsname{50}
\expandafter\def\csname v@D.hb.olmo.full_iter.8.pc\endcsname{65}
\expandafter\def\csname v@D.hb.olmo.pool_evpc.8.pc\endcsname{79}
\expandafter\def\csname v@D.hb.olmo.shareall.8.pc\endcsname{10}
\expandafter\def\csname v@D.hb.olmo.solo.8.pc\endcsname{0}
\expandafter\def\csname v@D.mq.olmo.balanced.16.pc\endcsname{10}
\expandafter\def\csname v@D.mq.olmo.countblind.16.pc\endcsname{9}
\expandafter\def\csname v@D.mq.olmo.debate.16.pc\endcsname{10}
\expandafter\def\csname v@D.mq.olmo.debate_evpc.16.pc\endcsname{45}
\expandafter\def\csname v@D.mq.olmo.full.16.pc\endcsname{75}
\expandafter\def\csname v@D.mq.olmo.full_iter.16.pc\endcsname{76}
\expandafter\def\csname v@D.mq.olmo.pool_evpc.16.pc\endcsname{56}
\expandafter\def\csname v@D.mq.olmo.solo.16.pc\endcsname{2}
\expandafter\def\csname v@D.mq.olmo.balanced.4.pc\endcsname{19}
\expandafter\def\csname v@D.mq.olmo.countblind.4.pc\endcsname{16}
\expandafter\def\csname v@D.mq.olmo.debate.4.pc\endcsname{17}
\expandafter\def\csname v@D.mq.olmo.debate_evpc.4.pc\endcsname{50}
\expandafter\def\csname v@D.mq.olmo.full.4.pc\endcsname{60}
\expandafter\def\csname v@D.mq.olmo.full_iter.4.pc\endcsname{66}
\expandafter\def\csname v@D.mq.olmo.pool_evpc.4.pc\endcsname{49}
\expandafter\def\csname v@D.mq.olmo.solo.4.pc\endcsname{7}
\expandafter\def\csname v@D.mq.olmo.balanced.8.pc\endcsname{13}
\expandafter\def\csname v@D.mq.olmo.countblind.8.pc\endcsname{14}
\expandafter\def\csname v@D.mq.olmo.debate.8.pc\endcsname{14}
\expandafter\def\csname v@D.mq.olmo.debate_evpc.8.pc\endcsname{50}
\expandafter\def\csname v@D.mq.olmo.full.8.pc\endcsname{68}
\expandafter\def\csname v@D.mq.olmo.full_iter.8.pc\endcsname{75}
\expandafter\def\csname v@D.mq.olmo.pool_evpc.8.pc\endcsname{52}
\expandafter\def\csname v@D.mq.olmo.solo.8.pc\endcsname{4}
\expandafter\def\csname v@D.pz.olmo.balanced.16.pc\endcsname{16}
\expandafter\def\csname v@D.pz.olmo.countblind.16.pc\endcsname{17}
\expandafter\def\csname v@D.pz.olmo.debate.16.pc\endcsname{10}
\expandafter\def\csname v@D.pz.olmo.debate_evpc.16.pc\endcsname{52}
\expandafter\def\csname v@D.pz.olmo.full.16.pc\endcsname{94}
\expandafter\def\csname v@D.pz.olmo.full_iter.16.pc\endcsname{93}
\expandafter\def\csname v@D.pz.olmo.pool_evpc.16.pc\endcsname{81}
\expandafter\def\csname v@D.pz.olmo.solo.16.pc\endcsname{3}
\expandafter\def\csname v@D.pz.olmo.balanced.32.pc\endcsname{20}
\expandafter\def\csname v@D.pz.olmo.countblind.32.pc\endcsname{15}
\expandafter\def\csname v@D.pz.olmo.debate.32.pc\endcsname{10}
\expandafter\def\csname v@D.pz.olmo.debate_evpc.32.pc\endcsname{11}
\expandafter\def\csname v@D.pz.olmo.full.32.pc\endcsname{96}
\expandafter\def\csname v@D.pz.olmo.full_iter.32.pc\endcsname{93}
\expandafter\def\csname v@D.pz.olmo.pool_evpc.32.pc\endcsname{80}
\expandafter\def\csname v@D.pz.olmo.solo.32.pc\endcsname{2}
\expandafter\def\csname v@D.pz.olmo.balanced.4.pc\endcsname{16}
\expandafter\def\csname v@D.pz.olmo.countblind.4.pc\endcsname{18}
\expandafter\def\csname v@D.pz.olmo.debate.4.pc\endcsname{15}
\expandafter\def\csname v@D.pz.olmo.debate_evpc.4.pc\endcsname{80}
\expandafter\def\csname v@D.pz.olmo.full.4.pc\endcsname{91}
\expandafter\def\csname v@D.pz.olmo.full_iter.4.pc\endcsname{85}
\expandafter\def\csname v@D.pz.olmo.pool_evpc.4.pc\endcsname{79}
\expandafter\def\csname v@D.pz.olmo.solo.4.pc\endcsname{5}
\expandafter\def\csname v@D.pz.olmo.balanced.8.pc\endcsname{17}
\expandafter\def\csname v@D.pz.olmo.countblind.8.pc\endcsname{18}
\expandafter\def\csname v@D.pz.olmo.debate.8.pc\endcsname{14}
\expandafter\def\csname v@D.pz.olmo.debate_evpc.8.pc\endcsname{80}
\expandafter\def\csname v@D.pz.olmo.full.8.pc\endcsname{90}
\expandafter\def\csname v@D.pz.olmo.full_iter.8.pc\endcsname{89}
\expandafter\def\csname v@D.pz.olmo.pool_evpc.8.pc\endcsname{82}
\expandafter\def\csname v@D.pz.olmo.solo.8.pc\endcsname{1}
\expandafter\def\csname v@D.hb.q32.debate.16.pc\endcsname{16}
\expandafter\def\csname v@D.hb.q32.full.16.pc\endcsname{83}
\expandafter\def\csname v@D.hb.q32.full_iter.16.pc\endcsname{92}
\expandafter\def\csname v@D.hb.q32.pool_evpc.16.pc\endcsname{89}
\expandafter\def\csname v@D.hb.q32.solo.16.pc\endcsname{3}
\expandafter\def\csname v@D.hb.q32.debate.32.pc\endcsname{13}
\expandafter\def\csname v@D.hb.q32.full.32.pc\endcsname{85}
\expandafter\def\csname v@D.hb.q32.full_iter.32.pc\endcsname{92}
\expandafter\def\csname v@D.hb.q32.pool_evpc.32.pc\endcsname{89}
\expandafter\def\csname v@D.hb.q32.solo.32.pc\endcsname{3}
\expandafter\def\csname v@D.hb.q32.debate.4.pc\endcsname{31}
\expandafter\def\csname v@D.hb.q32.full.4.pc\endcsname{83}
\expandafter\def\csname v@D.hb.q32.full_iter.4.pc\endcsname{89}
\expandafter\def\csname v@D.hb.q32.pool_evpc.4.pc\endcsname{91}
\expandafter\def\csname v@D.hb.q32.solo.4.pc\endcsname{6}
\expandafter\def\csname v@D.hb.q32.debate.8.pc\endcsname{20}
\expandafter\def\csname v@D.hb.q32.full.8.pc\endcsname{83}
\expandafter\def\csname v@D.hb.q32.full_iter.8.pc\endcsname{91}
\expandafter\def\csname v@D.hb.q32.pool_evpc.8.pc\endcsname{90}
\expandafter\def\csname v@D.hb.q32.solo.8.pc\endcsname{2}
\expandafter\def\csname v@D.hb.qwen.balanced.16.pc\endcsname{13}
\expandafter\def\csname v@D.hb.qwen.countblind.16.pc\endcsname{19}
\expandafter\def\csname v@D.hb.qwen.debate.16.pc\endcsname{6}
\expandafter\def\csname v@D.hb.qwen.debate_anon.16.pc\endcsname{8}
\expandafter\def\csname v@D.hb.qwen.debate_evpc.16.pc\endcsname{82}
\expandafter\def\csname v@D.hb.qwen.exchange.16.pc\endcsname{21}
\expandafter\def\csname v@D.hb.qwen.full.16.pc\endcsname{61}
\expandafter\def\csname v@D.hb.qwen.full_iter.16.pc\endcsname{74}
\expandafter\def\csname v@D.hb.qwen.pool_evpc.16.pc\endcsname{86}
\expandafter\def\csname v@D.hb.qwen.shareall.16.pc\endcsname{6}
\expandafter\def\csname v@D.hb.qwen.solo.16.pc\endcsname{3}
\expandafter\def\csname v@D.hb.qwen.balanced.32.pc\endcsname{13}
\expandafter\def\csname v@D.hb.qwen.countblind.32.pc\endcsname{18}
\expandafter\def\csname v@D.hb.qwen.debate.32.pc\endcsname{7}
\expandafter\def\csname v@D.hb.qwen.debate_anon.32.pc\endcsname{4}
\expandafter\def\csname v@D.hb.qwen.debate_evpc.32.pc\endcsname{75}
\expandafter\def\csname v@D.hb.qwen.exchange.32.pc\endcsname{12}
\expandafter\def\csname v@D.hb.qwen.full.32.pc\endcsname{60}
\expandafter\def\csname v@D.hb.qwen.full_iter.32.pc\endcsname{73}
\expandafter\def\csname v@D.hb.qwen.pool_evpc.32.pc\endcsname{84}
\expandafter\def\csname v@D.hb.qwen.shareall.32.pc\endcsname{3}
\expandafter\def\csname v@D.hb.qwen.solo.32.pc\endcsname{3}
\expandafter\def\csname v@D.hb.qwen.balanced.4.pc\endcsname{18}
\expandafter\def\csname v@D.hb.qwen.countblind.4.pc\endcsname{23}
\expandafter\def\csname v@D.hb.qwen.debate.4.pc\endcsname{11}
\expandafter\def\csname v@D.hb.qwen.debate_anon.4.pc\endcsname{17}
\expandafter\def\csname v@D.hb.qwen.debate_evpc.4.pc\endcsname{81}
\expandafter\def\csname v@D.hb.qwen.exchange.4.pc\endcsname{34}
\expandafter\def\csname v@D.hb.qwen.full.4.pc\endcsname{63}
\expandafter\def\csname v@D.hb.qwen.full_iter.4.pc\endcsname{70}
\expandafter\def\csname v@D.hb.qwen.pool_evpc.4.pc\endcsname{80}
\expandafter\def\csname v@D.hb.qwen.shareall.4.pc\endcsname{20}
\expandafter\def\csname v@D.hb.qwen.solo.4.pc\endcsname{5}
\expandafter\def\csname v@D.hb.qwen.debate.64.pc\endcsname{8}
\expandafter\def\csname v@D.hb.qwen.full.64.pc\endcsname{60}
\expandafter\def\csname v@D.hb.qwen.full_iter.64.pc\endcsname{73}
\expandafter\def\csname v@D.hb.qwen.pool_evpc.64.pc\endcsname{84}
\expandafter\def\csname v@D.hb.qwen.solo.64.pc\endcsname{3}
\expandafter\def\csname v@D.hb.qwen.balanced.8.pc\endcsname{15}
\expandafter\def\csname v@D.hb.qwen.countblind.8.pc\endcsname{19}
\expandafter\def\csname v@D.hb.qwen.debate.8.pc\endcsname{6}
\expandafter\def\csname v@D.hb.qwen.debate_anon.8.pc\endcsname{7}
\expandafter\def\csname v@D.hb.qwen.debate_evpc.8.pc\endcsname{85}
\expandafter\def\csname v@D.hb.qwen.exchange.8.pc\endcsname{24}
\expandafter\def\csname v@D.hb.qwen.full.8.pc\endcsname{59}
\expandafter\def\csname v@D.hb.qwen.full_iter.8.pc\endcsname{71}
\expandafter\def\csname v@D.hb.qwen.pool_evpc.8.pc\endcsname{82}
\expandafter\def\csname v@D.hb.qwen.shareall.8.pc\endcsname{7}
\expandafter\def\csname v@D.hb.qwen.solo.8.pc\endcsname{2}
\expandafter\def\csname v@D.mq.qwen.balanced.16.pc\endcsname{12}
\expandafter\def\csname v@D.mq.qwen.countblind.16.pc\endcsname{14}
\expandafter\def\csname v@D.mq.qwen.debate.16.pc\endcsname{11}
\expandafter\def\csname v@D.mq.qwen.debate_evpc.16.pc\endcsname{46}
\expandafter\def\csname v@D.mq.qwen.full.16.pc\endcsname{88}
\expandafter\def\csname v@D.mq.qwen.full_iter.16.pc\endcsname{89}
\expandafter\def\csname v@D.mq.qwen.pool_evpc.16.pc\endcsname{54}
\expandafter\def\csname v@D.mq.qwen.solo.16.pc\endcsname{3}
\expandafter\def\csname v@D.mq.qwen.balanced.4.pc\endcsname{20}
\expandafter\def\csname v@D.mq.qwen.countblind.4.pc\endcsname{26}
\expandafter\def\csname v@D.mq.qwen.debate.4.pc\endcsname{21}
\expandafter\def\csname v@D.mq.qwen.debate_evpc.4.pc\endcsname{56}
\expandafter\def\csname v@D.mq.qwen.full.4.pc\endcsname{84}
\expandafter\def\csname v@D.mq.qwen.full_iter.4.pc\endcsname{85}
\expandafter\def\csname v@D.mq.qwen.pool_evpc.4.pc\endcsname{57}
\expandafter\def\csname v@D.mq.qwen.solo.4.pc\endcsname{8}
\expandafter\def\csname v@D.mq.qwen.balanced.8.pc\endcsname{19}
\expandafter\def\csname v@D.mq.qwen.countblind.8.pc\endcsname{20}
\expandafter\def\csname v@D.mq.qwen.debate.8.pc\endcsname{13}
\expandafter\def\csname v@D.mq.qwen.debate_evpc.8.pc\endcsname{56}
\expandafter\def\csname v@D.mq.qwen.full.8.pc\endcsname{86}
\expandafter\def\csname v@D.mq.qwen.full_iter.8.pc\endcsname{88}
\expandafter\def\csname v@D.mq.qwen.pool_evpc.8.pc\endcsname{56}
\expandafter\def\csname v@D.mq.qwen.solo.8.pc\endcsname{4}
\expandafter\def\csname v@D.pz.qwen.balanced.16.pc\endcsname{16}
\expandafter\def\csname v@D.pz.qwen.countblind.16.pc\endcsname{16}
\expandafter\def\csname v@D.pz.qwen.debate.16.pc\endcsname{11}
\expandafter\def\csname v@D.pz.qwen.debate_evpc.16.pc\endcsname{69}
\expandafter\def\csname v@D.pz.qwen.full.16.pc\endcsname{94}
\expandafter\def\csname v@D.pz.qwen.full_iter.16.pc\endcsname{95}
\expandafter\def\csname v@D.pz.qwen.pool_evpc.16.pc\endcsname{90}
\expandafter\def\csname v@D.pz.qwen.solo.16.pc\endcsname{12}
\expandafter\def\csname v@D.pz.qwen.balanced.32.pc\endcsname{17}
\expandafter\def\csname v@D.pz.qwen.countblind.32.pc\endcsname{16}
\expandafter\def\csname v@D.pz.qwen.debate.32.pc\endcsname{10}
\expandafter\def\csname v@D.pz.qwen.debate_evpc.32.pc\endcsname{20}
\expandafter\def\csname v@D.pz.qwen.full.32.pc\endcsname{96}
\expandafter\def\csname v@D.pz.qwen.full_iter.32.pc\endcsname{96}
\expandafter\def\csname v@D.pz.qwen.pool_evpc.32.pc\endcsname{91}
\expandafter\def\csname v@D.pz.qwen.solo.32.pc\endcsname{12}
\expandafter\def\csname v@D.pz.qwen.balanced.4.pc\endcsname{15}
\expandafter\def\csname v@D.pz.qwen.countblind.4.pc\endcsname{21}
\expandafter\def\csname v@D.pz.qwen.debate.4.pc\endcsname{17}
\expandafter\def\csname v@D.pz.qwen.debate_evpc.4.pc\endcsname{85}
\expandafter\def\csname v@D.pz.qwen.full.4.pc\endcsname{89}
\expandafter\def\csname v@D.pz.qwen.full_iter.4.pc\endcsname{93}
\expandafter\def\csname v@D.pz.qwen.pool_evpc.4.pc\endcsname{87}
\expandafter\def\csname v@D.pz.qwen.solo.4.pc\endcsname{11}
\expandafter\def\csname v@D.pz.qwen.balanced.8.pc\endcsname{14}
\expandafter\def\csname v@D.pz.qwen.countblind.8.pc\endcsname{18}
\expandafter\def\csname v@D.pz.qwen.debate.8.pc\endcsname{12}
\expandafter\def\csname v@D.pz.qwen.debate_evpc.8.pc\endcsname{88}
\expandafter\def\csname v@D.pz.qwen.full.8.pc\endcsname{92}
\expandafter\def\csname v@D.pz.qwen.full_iter.8.pc\endcsname{94}
\expandafter\def\csname v@D.pz.qwen.pool_evpc.8.pc\endcsname{88}
\expandafter\def\csname v@D.pz.qwen.solo.8.pc\endcsname{12}
\expandafter\def\csname v@D.max.debate.pc\endcsname{23}
\expandafter\def\csname v@debsolo.max.pc\endcsname{8}
\expandafter\def\csname v@Dr.hb.qwen.pool_evpc.pc\endcsname{80--86}
\expandafter\def\csname v@Dr.hb.qwen.debate.pc\endcsname{6--11}
\expandafter\def\csname v@Dr.hb.qwen.debate_evpc.pc\endcsname{75--85}
\expandafter\def\csname v@Dr.hb.qwen.full_iter.pc\endcsname{70--74}
\expandafter\def\csname v@Dr.hb.qwen.balanced.pc\endcsname{13--18}
\expandafter\def\csname v@Dr.pz.qwen.pool_evpc.pc\endcsname{87--91}
\expandafter\def\csname v@Dr.pz.qwen.debate.pc\endcsname{10--17}
\expandafter\def\csname v@Dr.pz.qwen.debate_evpc.pc\endcsname{20--88}
\expandafter\def\csname v@Dr.pz.qwen.full_iter.pc\endcsname{93--96}
\expandafter\def\csname v@Dr.pz.qwen.balanced.pc\endcsname{14--17}
\expandafter\def\csname v@Dr.mq.qwen.pool_evpc.pc\endcsname{54--57}
\expandafter\def\csname v@Dr.mq.qwen.debate.pc\endcsname{11--21}
\expandafter\def\csname v@Dr.mq.qwen.debate_evpc.pc\endcsname{46--56}
\expandafter\def\csname v@Dr.mq.qwen.full_iter.pc\endcsname{85--89}
\expandafter\def\csname v@Dr.mq.qwen.balanced.pc\endcsname{12--20}
\expandafter\def\csname v@Dr.hb.olmo.pool_evpc.pc\endcsname{74--80}
\expandafter\def\csname v@Dr.hb.olmo.debate.pc\endcsname{8--23}
\expandafter\def\csname v@Dr.hb.olmo.debate_evpc.pc\endcsname{43--76}
\expandafter\def\csname v@Dr.hb.olmo.full_iter.pc\endcsname{63--70}
\expandafter\def\csname v@Dr.hb.olmo.balanced.pc\endcsname{20--29}
\expandafter\def\csname v@Dr.pz.olmo.pool_evpc.pc\endcsname{79--82}
\expandafter\def\csname v@Dr.pz.olmo.debate.pc\endcsname{10--15}
\expandafter\def\csname v@Dr.pz.olmo.debate_evpc.pc\endcsname{11--80}
\expandafter\def\csname v@Dr.pz.olmo.full_iter.pc\endcsname{85--93}
\expandafter\def\csname v@Dr.pz.olmo.balanced.pc\endcsname{16--20}
\expandafter\def\csname v@Dr.mq.olmo.pool_evpc.pc\endcsname{49--56}
\expandafter\def\csname v@Dr.mq.olmo.debate.pc\endcsname{10--17}
\expandafter\def\csname v@Dr.mq.olmo.debate_evpc.pc\endcsname{45--50}
\expandafter\def\csname v@Dr.mq.olmo.full_iter.pc\endcsname{66--76}
\expandafter\def\csname v@Dr.mq.olmo.balanced.pc\endcsname{10--19}
\expandafter\def\csname v@Dr.hb.mini.pool_evpc.pc\endcsname{82--87}
\expandafter\def\csname v@Dr.hb.mini.debate.pc\endcsname{36--43}
\expandafter\def\csname v@Dr.hb.mini.debate_evpc.pc\endcsname{55--84}
\expandafter\def\csname v@Dr.hb.mini.full_iter.pc\endcsname{66--80}
\expandafter\def\csname v@Dr.hb.mini.balanced.pc\endcsname{33--49}
\expandafter\def\csname v@Dr.pz.mini.pool_evpc.pc\endcsname{95--98}
\expandafter\def\csname v@Dr.pz.mini.debate.pc\endcsname{7--13}
\expandafter\def\csname v@Dr.pz.mini.debate_evpc.pc\endcsname{4--93}
\expandafter\def\csname v@Dr.pz.mini.full_iter.pc\endcsname{94--99}
\expandafter\def\csname v@Dr.pz.mini.balanced.pc\endcsname{9--16}
\expandafter\def\csname v@Dr.hb.gemma.pool_evpc.pc\endcsname{93--96}
\expandafter\def\csname v@Dr.hb.gemma.debate.pc\endcsname{28--55}
\expandafter\def\csname v@Dr.hb.gemma.debate_evpc.pc\endcsname{88--96}
\expandafter\def\csname v@Dr.hb.gemma.full_iter.pc\endcsname{91--97}
\expandafter\def\csname v@Dr.hb.gemma.balanced.pc\endcsname{41--56}
\expandafter\def\csname v@Dr.pz.gemma.pool_evpc.pc\endcsname{99--99}
\expandafter\def\csname v@Dr.pz.gemma.debate.pc\endcsname{21--37}
\expandafter\def\csname v@Dr.pz.gemma.debate_evpc.pc\endcsname{25--100}
\expandafter\def\csname v@Dr.pz.gemma.full_iter.pc\endcsname{100--100}
\expandafter\def\csname v@Dr.pz.gemma.balanced.pc\endcsname{29--41}
\expandafter\def\csname v@Dr.hb.q32.pool_evpc.pc\endcsname{89--91}
\expandafter\def\csname v@Dr.hb.q32.debate.pc\endcsname{13--31}
\expandafter\def\csname v@Dr.hb.q32.full_iter.pc\endcsname{89--92}
\expandafter\def\csname v@K.hb.qwen.fi.pc\endcsname{$+$12 [1, 23]}
\expandafter\def\csname v@K.hb.qwen.fi.pt.pc\endcsname{$+$12}
\expandafter\def\csname v@K.pz.qwen.fi.pc\endcsname{$-$6 [$-$8, $-$4]}
\expandafter\def\csname v@K.pz.qwen.fi.pt.pc\endcsname{$-$6}
\expandafter\def\csname v@K.mq.qwen.fi.pc\endcsname{$-$32 [$-$36, $-$27]}
\expandafter\def\csname v@K.mq.qwen.fi.pt.pc\endcsname{$-$32}
\expandafter\def\csname v@K.hb.olmo.fi.pc\endcsname{$+$10 [$-$1, 21]}
\expandafter\def\csname v@K.hb.olmo.fi.pt.pc\endcsname{$+$10}
\expandafter\def\csname v@K.pz.olmo.fi.pc\endcsname{$-$10 [$-$13, $-$7]}
\expandafter\def\csname v@K.pz.olmo.fi.pt.pc\endcsname{$-$10}
\expandafter\def\csname v@K.mq.olmo.fi.pc\endcsname{$-$20 [$-$24, $-$16]}
\expandafter\def\csname v@K.mq.olmo.fi.pt.pc\endcsname{$-$20}
\expandafter\def\csname v@K.below.range.pc\endcsname{$-$6 to $-$32}
\expandafter\def\csname v@tok.hb.ratio.pc\endcsname{61--65}
\expandafter\def\csname v@tok.pz.ratio.pc\endcsname{55--58}
\expandafter\def\csname v@tok.mq.ratio.pc\endcsname{68--74}
\expandafter\def\csname v@tok.range.pc\endcsname{55--74}
\expandafter\def\csname v@late.qwen.debate.K3.pc\endcsname{0}
\expandafter\def\csname v@late.qwen.debate.K0.pc\endcsname{7}
\expandafter\def\csname v@late.qwen.pool.K0.pc\endcsname{75}
\expandafter\def\csname v@late.qwen.pool.K3.pc\endcsname{69}
\expandafter\def\csname v@late.olmo.debate.K3.pc\endcsname{6}
\expandafter\def\csname v@late.olmo.debate.K0.pc\endcsname{18}
\expandafter\def\csname v@late.olmo.pool.K0.pc\endcsname{68}
\expandafter\def\csname v@late.olmo.pool.K3.pc\endcsname{61}
\expandafter\def\csname v@q32gap.min.pc\endcsname{49}
\expandafter\def\csname v@D16.range.debate.pc\endcsname{6--16}
\expandafter\def\csname v@D16.range.balanced.pc\endcsname{10--20}
\expandafter\def\csname v@D16.range.pool_evpc.pc\endcsname{54--90}
\expandafter\def\csname v@D16.range.debate_evpc.pc\endcsname{45--82}
\expandafter\def\csname v@D16.range.full_iter.pc\endcsname{67--95}
\expandafter\def\csname v@D16.range.solo.pc\endcsname{0--12}
\expandafter\def\csname v@q32gap.max.pc\endcsname{71}
\expandafter\def\csname v@pred.Texp.rho.maxdiff.pc\endcsname{2.3}
\expandafter\def\csname v@sel.ranked.pc\endcsname{80}
\expandafter\def\csname v@sel.pool.pc\endcsname{79}
\expandafter\def\csname v@gain.debate.range.pc\endcsname{83--97}
\expandafter\def\csname v@gain.matched.range.pc\endcsname{28--75}
\expandafter\def\csname v@gain.pool.range.pc\endcsname{50--73}
\expandafter\def\csname v@debate.drift.max.pc\endcsname{15}
\expandafter\def\csname v@gain.full.range.pc\endcsname{57--86}
\expandafter\def\csname v@sw.qwen.debate.4.pc\endcsname{11}
\expandafter\def\csname v@sw.qwen.debate.8.pc\endcsname{6}
\expandafter\def\csname v@sw.qwen.debate.16.pc\endcsname{5}
\expandafter\def\csname v@sw.qwen.debate.32.pc\endcsname{4}
\expandafter\def\csname v@swaway.qwen.debate.pc\endcsname{0--2}
\expandafter\def\csname v@sw.qwen.matched.4.pc\endcsname{19}
\expandafter\def\csname v@sw.qwen.matched.8.pc\endcsname{19}
\expandafter\def\csname v@sw.qwen.matched.16.pc\endcsname{20}
\expandafter\def\csname v@sw.qwen.matched.32.pc\endcsname{19}
\expandafter\def\csname v@swaway.qwen.matched.pc\endcsname{2--5}
\expandafter\def\csname v@sw.olmo.debate.4.pc\endcsname{35}
\expandafter\def\csname v@sw.olmo.debate.8.pc\endcsname{22}
\expandafter\def\csname v@sw.olmo.debate.16.pc\endcsname{15}
\expandafter\def\csname v@sw.olmo.debate.32.pc\endcsname{11}
\expandafter\def\csname v@swaway.olmo.debate.pc\endcsname{0--9}
\expandafter\def\csname v@sw.olmo.matched.4.pc\endcsname{58}
\expandafter\def\csname v@sw.olmo.matched.8.pc\endcsname{63}
\expandafter\def\csname v@sw.olmo.matched.16.pc\endcsname{74}
\expandafter\def\csname v@sw.olmo.matched.32.pc\endcsname{71}
\expandafter\def\csname v@swaway.olmo.matched.pc\endcsname{18--28}
\expandafter\def\csname v@swto.hb.qwen.debate.16.pc\endcsname{4}
\expandafter\def\csname v@swaway.hb.qwen.debate.16.pc\endcsname{1}
\expandafter\def\csname v@swto.hb.qwen.balanced.16.pc\endcsname{15}
\expandafter\def\csname v@swaway.hb.qwen.balanced.16.pc\endcsname{5}
\expandafter\def\csname v@swto.hb.qwen.pool_evpc.16.pc\endcsname{82}
\expandafter\def\csname v@swaway.hb.qwen.pool_evpc.16.pc\endcsname{6}
\expandafter\def\csname v@swto.pz.qwen.debate.16.pc\endcsname{4}
\expandafter\def\csname v@swaway.pz.qwen.debate.16.pc\endcsname{4}
\expandafter\def\csname v@swto.pz.qwen.balanced.16.pc\endcsname{11}
\expandafter\def\csname v@swaway.pz.qwen.balanced.16.pc\endcsname{6}
\expandafter\def\csname v@swto.pz.qwen.pool_evpc.16.pc\endcsname{83}
\expandafter\def\csname v@swaway.pz.qwen.pool_evpc.16.pc\endcsname{4}
\expandafter\def\csname v@swto.mq.qwen.debate.16.pc\endcsname{11}
\expandafter\def\csname v@swaway.mq.qwen.debate.16.pc\endcsname{1}
\expandafter\def\csname v@swto.mq.qwen.balanced.16.pc\endcsname{14}
\expandafter\def\csname v@swaway.mq.qwen.balanced.16.pc\endcsname{4}
\expandafter\def\csname v@swto.mq.qwen.pool_evpc.16.pc\endcsname{54}
\expandafter\def\csname v@swaway.mq.qwen.pool_evpc.16.pc\endcsname{3}
\expandafter\def\csname v@swto.hb.olmo.debate.16.pc\endcsname{15}
\expandafter\def\csname v@swaway.hb.olmo.debate.16.pc\endcsname{0}
\expandafter\def\csname v@swto.hb.olmo.balanced.16.pc\endcsname{46}
\expandafter\def\csname v@swaway.hb.olmo.balanced.16.pc\endcsname{28}
\expandafter\def\csname v@swto.hb.olmo.pool_evpc.16.pc\endcsname{80}
\expandafter\def\csname v@swaway.hb.olmo.pool_evpc.16.pc\endcsname{7}
\expandafter\def\csname v@swto.pz.olmo.debate.16.pc\endcsname{8}
\expandafter\def\csname v@swaway.pz.olmo.debate.16.pc\endcsname{3}
\expandafter\def\csname v@swto.pz.olmo.balanced.16.pc\endcsname{33}
\expandafter\def\csname v@swaway.pz.olmo.balanced.16.pc\endcsname{23}
\expandafter\def\csname v@swto.pz.olmo.pool_evpc.16.pc\endcsname{78}
\expandafter\def\csname v@swaway.pz.olmo.pool_evpc.16.pc\endcsname{4}
\expandafter\def\csname v@swto.mq.olmo.debate.16.pc\endcsname{12}
\expandafter\def\csname v@swaway.mq.olmo.debate.16.pc\endcsname{4}
\expandafter\def\csname v@swto.mq.olmo.balanced.16.pc\endcsname{18}
\expandafter\def\csname v@swaway.mq.olmo.balanced.16.pc\endcsname{11}
\expandafter\def\csname v@swto.mq.olmo.pool_evpc.16.pc\endcsname{61}
\expandafter\def\csname v@swaway.mq.olmo.pool_evpc.16.pc\endcsname{10}
\expandafter\def\csname v@swto.range.debate.pc\endcsname{4--15}
\expandafter\def\csname v@swaway.range.debate.pc\endcsname{0--4}
\expandafter\def\csname v@swto.range.balanced.pc\endcsname{11--46}
\expandafter\def\csname v@swaway.range.balanced.pc\endcsname{4--28}
\expandafter\def\csname v@swto.range.pool_evpc.pc\endcsname{54--83}
\expandafter\def\csname v@swaway.range.pool_evpc.pc\endcsname{3--10}
\expandafter\def\csname v@mod.strong.range.pc\endcsname{74--94}
\expandafter\def\csname v@mod.weak.range.pc\endcsname{54--68}
\expandafter\def\csname v@rd.hb.qwen.r1.pc\endcsname{70}
\expandafter\def\csname v@rd.hb.qwen.r6.pc\endcsname{80}
\expandafter\def\csname v@rd.hb.qwen.D.pc\endcsname{86}
\expandafter\def\csname v@rd.hb.olmo.r1.pc\endcsname{56}
\expandafter\def\csname v@rd.hb.olmo.r6.pc\endcsname{73}
\expandafter\def\csname v@rd.hb.olmo.D.pc\endcsname{79}
\expandafter\def\csname v@rd.pz.qwen.r1.pc\endcsname{36}
\expandafter\def\csname v@rd.pz.qwen.r6.pc\endcsname{77}
\expandafter\def\csname v@rd.pz.qwen.D.pc\endcsname{90}
\expandafter\def\csname v@rd.pz.olmo.r1.pc\endcsname{35}
\expandafter\def\csname v@rd.pz.olmo.r6.pc\endcsname{70}
\expandafter\def\csname v@rd.pz.olmo.D.pc\endcsname{81}
\expandafter\def\csname v@rd.mq.qwen.r1.pc\endcsname{19}
\expandafter\def\csname v@rd.mq.qwen.r6.pc\endcsname{44}
\expandafter\def\csname v@rd.mq.qwen.D.pc\endcsname{54}
\expandafter\def\csname v@rd.mq.olmo.r1.pc\endcsname{15}
\expandafter\def\csname v@rd.mq.olmo.r6.pc\endcsname{40}
\expandafter\def\csname v@rd.mq.olmo.D.pc\endcsname{56}
\expandafter\def\csname v@rd.gap.range.pc\endcsname{17--53}
\expandafter\def\csname v@rd.rounds.range.pc\endcsname{10--41}
\expandafter\def\csname v@rd.plur.range.pc\endcsname{6--15}
\expandafter\def\csname v@cite.local.disc.pc\endcsname{4--7}
\expandafter\def\csname v@g1.argued.hb.qwen.pc\endcsname{$-$44 [$-$59, $-$30]}
\expandafter\def\csname v@g1.argued.pz.qwen.pc\endcsname{$-$64 [$-$71, $-$55]}
\expandafter\def\csname v@g1.nodedup.qwen.4.pc\endcsname{81}
\expandafter\def\csname v@g1.nodedup.qwen.32.pc\endcsname{60}
\expandafter\def\csname v@g1.nodedup.gap.qwen.pc\endcsname{$-$31 [$-$45, $-$18]}
\expandafter\def\csname v@g1.ev4.qwen.pc\endcsname{$-$1 [$-$6, 3]}
\expandafter\def\csname v@g1.rep.qwen.pc\endcsname{0 [0, 0]}
\expandafter\def\csname v@g1.pull.qwen.label.pc\endcsname{$-$59}
\expandafter\def\csname v@g1.pull.qwen.hidden.pc\endcsname{$-$63}
\expandafter\def\csname v@g1.pull.qwen.labeldiff.pc\endcsname{$+$4}
\expandafter\def\csname v@g1.argued.hb.olmo.pc\endcsname{$-$33 [$-$49, $-$17]}
\expandafter\def\csname v@g1.argued.pz.olmo.pc\endcsname{$-$62 [$-$70, $-$55]}
\expandafter\def\csname v@g1.nodedup.olmo.4.pc\endcsname{73}
\expandafter\def\csname v@g1.nodedup.olmo.32.pc\endcsname{55}
\expandafter\def\csname v@g1.nodedup.gap.olmo.pc\endcsname{$-$36 [$-$55, $-$17]}
\expandafter\def\csname v@g1.ev4.olmo.pc\endcsname{$-$6 [$-$15, 2]}
\expandafter\def\csname v@g1.rep.olmo.pc\endcsname{$-$5 [$-$10, 0]}
\expandafter\def\csname v@g1.pull.olmo.label.pc\endcsname{88}
\expandafter\def\csname v@g1.pull.olmo.hidden.pc\endcsname{33}
\expandafter\def\csname v@g1.pull.olmo.labeldiff.pc\endcsname{$+$55}
\expandafter\def\csname v@g1f.olmo.16.pc\endcsname{93 [87, 100]}
\expandafter\def\csname v@g1f.olmo.32.pc\endcsname{89 [85, 94]}
\expandafter\def\csname v@g1.raw.range.pc\endcsname{$-$6 to $+$7}
\expandafter\def\csname v@g2.qwen.pool.pc\endcsname{88}
\expandafter\def\csname v@g2.qwen.hb_release.pc\endcsname{15}
\expandafter\def\csname v@g2.qwen.exchange_hb.pc\endcsname{54}
\expandafter\def\csname v@g2.qwen.revealall.pc\endcsname{54}
\expandafter\def\csname v@g2.qwen.dar.pc\endcsname{8}
\expandafter\def\csname v@g2.qwen.dar2.pc\endcsname{14}
\expandafter\def\csname v@g2.qwen.w300.pc\endcsname{$+$25 [14, 38]}
\expandafter\def\csname v@g2.olmo.pool.pc\endcsname{88}
\expandafter\def\csname v@g2.olmo.hb_release.pc\endcsname{22}
\expandafter\def\csname v@g2.olmo.exchange_hb.pc\endcsname{37}
\expandafter\def\csname v@g2.olmo.revealall.pc\endcsname{26}
\expandafter\def\csname v@g2.olmo.dar.pc\endcsname{11}
\expandafter\def\csname v@g2.olmo.dar2.pc\endcsname{9}
\expandafter\def\csname v@g2.olmo.w300.pc\endcsname{0 [$-$10, 10]}
\expandafter\def\csname v@g2.gap.range.pc\endcsname{34--80}
\expandafter\def\csname v@g3.hb.qwen.pc\endcsname{$+$2 [0, 4]}
\expandafter\def\csname v@g3.hb.olmo.pc\endcsname{$+$7 [4, 11]}
\expandafter\def\csname v@g3.pz.qwen.pc\endcsname{$-$1 [$-$2, 0]}
\expandafter\def\csname v@g3.pz.olmo.pc\endcsname{$+$3 [1, 4]}
\expandafter\def\csname v@g3.mq.qwen.pc\endcsname{0 [0, 1]}
\expandafter\def\csname v@g3.mq.olmo.pc\endcsname{$+$5 [3, 6]}
\expandafter\def\csname v@g3.other.range.pc\endcsname{$-$1 to $+$5}
\expandafter\def\csname v@g4.qwen.pool_evpc.K0.pc\endcsname{$+$78 [64, 90]}
\expandafter\def\csname v@g4.qwen.pool_evpc.K1.pc\endcsname{$+$76 [61, 90]}
\expandafter\def\csname v@g4.qwen.pool_evpc.K2.pc\endcsname{$+$76 [62, 89]}
\expandafter\def\csname v@g4.qwen.pool_evpc.K3.pc\endcsname{$+$77 [63, 89]}
\expandafter\def\csname v@g4.qwen.pool_evpc.K4.pc\endcsname{$+$73 [59, 86]}
\expandafter\def\csname v@g4.qwen.pool_evpc.K5.pc\endcsname{$+$72 [57, 85]}
\expandafter\def\csname v@g4.qwen.pool_evpc.r4.pc\endcsname{94 [84, 105]}
\expandafter\def\csname v@g4.qwen.pool_evpc.r5.pc\endcsname{92 [83, 101]}
\expandafter\def\csname v@g4.qwen.balanced.K0.pc\endcsname{$+$20 [11, 30]}
\expandafter\def\csname v@g4.qwen.balanced.K1.pc\endcsname{$+$5 [$-$5, 16]}
\expandafter\def\csname v@g4.qwen.balanced.K2.pc\endcsname{$-$4 [$-$15, 5]}
\expandafter\def\csname v@g4.qwen.balanced.K3.pc\endcsname{$-$4 [$-$14, 5]}
\expandafter\def\csname v@g4.qwen.balanced.K4.pc\endcsname{$-$4 [$-$13, 3]}
\expandafter\def\csname v@g4.qwen.balanced.K5.pc\endcsname{$-$2 [$-$12, 7]}
\expandafter\def\csname v@g4.qwen.balanced.r1.pc\endcsname{25 [$-$28, 88]}
\expandafter\def\csname v@g4.qwen.balanced.r2.pc\endcsname{$-$22 [$-$93, 28]}
\expandafter\def\csname v@g4.qwen.balanced.r3.pc\endcsname{$-$19 [$-$84, 29]}
\expandafter\def\csname v@g4.qwen.balanced.r4.pc\endcsname{$-$22 [$-$86, 15]}
\expandafter\def\csname v@g4.qwen.balanced.r5.pc\endcsname{$-$12 [$-$72, 34]}
\expandafter\def\csname v@g4.qwen.exchange.K0.pc\endcsname{$+$24 [14, 36]}
\expandafter\def\csname v@g4.qwen.exchange.K1.pc\endcsname{$+$29 [17, 42]}
\expandafter\def\csname v@g4.qwen.exchange.K2.pc\endcsname{$+$24 [14, 36]}
\expandafter\def\csname v@g4.qwen.exchange.K3.pc\endcsname{$+$21 [11, 32]}
\expandafter\def\csname v@g4.qwen.exchange.K4.pc\endcsname{$+$20 [9, 31]}
\expandafter\def\csname v@g4.qwen.exchange.K5.pc\endcsname{$+$20 [8, 33]}
\expandafter\def\csname v@g4.qwen.shareall.K0.pc\endcsname{$+$18 [9, 28]}
\expandafter\def\csname v@g4.qwen.shareall.K1.pc\endcsname{$-$1 [$-$10, 8]}
\expandafter\def\csname v@g4.qwen.shareall.K2.pc\endcsname{$-$1 [$-$9, 7]}
\expandafter\def\csname v@g4.qwen.shareall.K3.pc\endcsname{$+$1 [$-$7, 9]}
\expandafter\def\csname v@g4.qwen.shareall.K4.pc\endcsname{0 [$-$7, 7]}
\expandafter\def\csname v@g4.qwen.shareall.K5.pc\endcsname{$-$3 [$-$8, 2]}
\expandafter\def\csname v@g4.qwen.shareall.r1.pc\endcsname{$-$4 [$-$53, 65]}
\expandafter\def\csname v@g4.qwen.shareall.r2.pc\endcsname{$-$7 [$-$62, 47]}
\expandafter\def\csname v@g4.qwen.shareall.r3.pc\endcsname{7 [$-$39, 61]}
\expandafter\def\csname v@g4.qwen.shareall.r4.pc\endcsname{1 [$-$44, 51]}
\expandafter\def\csname v@g4.qwen.shareall.r5.pc\endcsname{$-$16 [$-$61, 14]}
\expandafter\def\csname v@g4.qwen.debate.K0.pc\endcsname{$+$10 [3, 17]}
\expandafter\def\csname v@g4.qwen.debate.K1.pc\endcsname{$+$1 [$-$7, 9]}
\expandafter\def\csname v@g4.qwen.debate.K2.pc\endcsname{$-$1 [$-$12, 11]}
\expandafter\def\csname v@g4.qwen.debate.K3.pc\endcsname{$-$4 [$-$10, 1]}
\expandafter\def\csname v@g4.qwen.debate.K4.pc\endcsname{$-$5 [$-$11, 1]}
\expandafter\def\csname v@g4.qwen.debate.K5.pc\endcsname{$-$6 [$-$13, 0]}
\expandafter\def\csname v@g4.olmo.pool_evpc.K0.pc\endcsname{$+$70 [54, 83]}
\expandafter\def\csname v@g4.olmo.pool_evpc.K1.pc\endcsname{$+$63 [44, 79]}
\expandafter\def\csname v@g4.olmo.pool_evpc.K2.pc\endcsname{$+$65 [50, 78]}
\expandafter\def\csname v@g4.olmo.pool_evpc.K3.pc\endcsname{$+$61 [46, 75]}
\expandafter\def\csname v@g4.olmo.pool_evpc.K4.pc\endcsname{$+$60 [43, 75]}
\expandafter\def\csname v@g4.olmo.pool_evpc.K5.pc\endcsname{$+$49 [32, 65]}
\expandafter\def\csname v@g4.olmo.pool_evpc.r1.pc\endcsname{90 [76, 103]}
\expandafter\def\csname v@g4.olmo.pool_evpc.r3.pc\endcsname{88 [77, 99]}
\expandafter\def\csname v@g4.olmo.pool_evpc.r4.pc\endcsname{86 [69, 101]}
\expandafter\def\csname v@g4.olmo.pool_evpc.r5.pc\endcsname{70 [53, 87]}
\expandafter\def\csname v@g4.olmo.balanced.K0.pc\endcsname{$+$29 [17, 42]}
\expandafter\def\csname v@g4.olmo.balanced.K1.pc\endcsname{$+$16 [2, 31]}
\expandafter\def\csname v@g4.olmo.balanced.K2.pc\endcsname{$+$11 [$-$3, 25]}
\expandafter\def\csname v@g4.olmo.balanced.K3.pc\endcsname{$+$8 [$-$3, 20]}
\expandafter\def\csname v@g4.olmo.balanced.K4.pc\endcsname{$+$2 [$-$9, 13]}
\expandafter\def\csname v@g4.olmo.balanced.K5.pc\endcsname{$-$4 [$-$14, 7]}
\expandafter\def\csname v@g4.olmo.balanced.r2.pc\endcsname{38 [$-$9, 100]}
\expandafter\def\csname v@g4.olmo.balanced.r3.pc\endcsname{29 [$-$9, 77]}
\expandafter\def\csname v@g4.olmo.balanced.r4.pc\endcsname{7 [$-$39, 42]}
\expandafter\def\csname v@g4.olmo.balanced.r5.pc\endcsname{$-$12 [$-$57, 25]}
\expandafter\def\csname v@g4.olmo.exchange.K0.pc\endcsname{$+$30 [10, 48]}
\expandafter\def\csname v@g4.olmo.exchange.K1.pc\endcsname{$+$28 [6, 48]}
\expandafter\def\csname v@g4.olmo.exchange.K2.pc\endcsname{$+$30 [16, 44]}
\expandafter\def\csname v@g4.olmo.exchange.K3.pc\endcsname{$+$34 [15, 53]}
\expandafter\def\csname v@g4.olmo.exchange.K4.pc\endcsname{$+$30 [14, 46]}
\expandafter\def\csname v@g4.olmo.exchange.K5.pc\endcsname{$+$20 [2, 38]}
\expandafter\def\csname v@g4.olmo.exchange.r5.pc\endcsname{67 [13, 105]}
\expandafter\def\csname v@g4.olmo.shareall.K0.pc\endcsname{$+$18 [8, 28]}
\expandafter\def\csname v@g4.olmo.shareall.K1.pc\endcsname{$+$5 [$-$9, 19]}
\expandafter\def\csname v@g4.olmo.shareall.K2.pc\endcsname{$+$2 [$-$11, 13]}
\expandafter\def\csname v@g4.olmo.shareall.K3.pc\endcsname{$-$7 [$-$20, 5]}
\expandafter\def\csname v@g4.olmo.shareall.K4.pc\endcsname{$-$8 [$-$20, 2]}
\expandafter\def\csname v@g4.olmo.shareall.K5.pc\endcsname{$-$11 [$-$21, $-$2]}
\expandafter\def\csname v@g4.olmo.shareall.r2.pc\endcsname{9 [$-$95, 65]}
\expandafter\def\csname v@g4.olmo.debate.K0.pc\endcsname{$+$19 [9, 30]}
\expandafter\def\csname v@g4.olmo.debate.K1.pc\endcsname{$+$14 [5, 22]}
\expandafter\def\csname v@g4.olmo.debate.K2.pc\endcsname{$+$6 [$-$4, 15]}
\expandafter\def\csname v@g4.olmo.debate.K3.pc\endcsname{$+$5 [$-$6, 16]}
\expandafter\def\csname v@g4.olmo.debate.K4.pc\endcsname{$+$5 [$-$3, 14]}
\expandafter\def\csname v@g4.olmo.debate.K5.pc\endcsname{0 [$-$9, 9]}
\expandafter\def\csname v@g4.olmo.debate.r4.pc\endcsname{28 [$-$17, 103]}
\expandafter\def\csname v@g4.olmo.debate.r5.pc\endcsname{0 [$-$60, 59]}
\expandafter\def\csname v@g4.mini.pool_evpc.K0.pc\endcsname{$+$82 [72, 91]}
\expandafter\def\csname v@g4.mini.pool_evpc.K1.pc\endcsname{$+$64 [49, 78]}
\expandafter\def\csname v@g4.mini.pool_evpc.K2.pc\endcsname{$+$70 [57, 82]}
\expandafter\def\csname v@g4.mini.pool_evpc.K3.pc\endcsname{$+$66 [54, 77]}
\expandafter\def\csname v@g4.mini.pool_evpc.K4.pc\endcsname{$+$61 [50, 71]}
\expandafter\def\csname v@g4.mini.pool_evpc.K5.pc\endcsname{$+$67 [52, 80]}
\expandafter\def\csname v@g4.mini.pool_evpc.r1.pc\endcsname{78 [65, 89]}
\expandafter\def\csname v@g4.mini.pool_evpc.r2.pc\endcsname{85 [75, 95]}
\expandafter\def\csname v@g4.mini.pool_evpc.r3.pc\endcsname{81 [70, 91]}
\expandafter\def\csname v@g4.mini.pool_evpc.r4.pc\endcsname{74 [65, 83]}
\expandafter\def\csname v@g4.mini.pool_evpc.r5.pc\endcsname{81 [69, 92]}
\expandafter\def\csname v@g4.mini.balanced.K0.pc\endcsname{$+$40 [30, 49]}
\expandafter\def\csname v@g4.mini.balanced.K1.pc\endcsname{$+$30 [18, 41]}
\expandafter\def\csname v@g4.mini.balanced.K2.pc\endcsname{$+$23 [11, 35]}
\expandafter\def\csname v@g4.mini.balanced.K3.pc\endcsname{$+$17 [3, 31]}
\expandafter\def\csname v@g4.mini.balanced.K4.pc\endcsname{$+$14 [1, 26]}
\expandafter\def\csname v@g4.mini.balanced.K5.pc\endcsname{$+$5 [$-$9, 20]}
\expandafter\def\csname v@g4.mini.balanced.r2.pc\endcsname{59 [31, 92]}
\expandafter\def\csname v@g4.mini.balanced.r3.pc\endcsname{42 [9, 75]}
\expandafter\def\csname v@g4.mini.balanced.r4.pc\endcsname{35 [3, 71]}
\expandafter\def\csname v@g4.mini.balanced.r5.pc\endcsname{13 [$-$24, 54]}
\expandafter\def\csname v@g4.mini.exchange.K0.pc\endcsname{$+$48 [31, 64]}
\expandafter\def\csname v@g4.mini.exchange.K1.pc\endcsname{$+$51 [37, 64]}
\expandafter\def\csname v@g4.mini.exchange.K2.pc\endcsname{$+$41 [27, 55]}
\expandafter\def\csname v@g4.mini.exchange.K3.pc\endcsname{$+$30 [14, 46]}
\expandafter\def\csname v@g4.mini.exchange.K4.pc\endcsname{$+$30 [13, 46]}
\expandafter\def\csname v@g4.mini.exchange.K5.pc\endcsname{$+$17 [2, 33]}
\expandafter\def\csname v@g4.mini.exchange.r3.pc\endcsname{63 [39, 87]}
\expandafter\def\csname v@g4.mini.exchange.r4.pc\endcsname{62 [32, 91]}
\expandafter\def\csname v@g4.mini.exchange.r5.pc\endcsname{36 [4, 62]}
\expandafter\def\csname v@g4.mini.shareall.K0.pc\endcsname{$+$39 [27, 52]}
\expandafter\def\csname v@g4.mini.shareall.K1.pc\endcsname{$+$20 [9, 31]}
\expandafter\def\csname v@g4.mini.shareall.K2.pc\endcsname{$+$4 [$-$9, 16]}
\expandafter\def\csname v@g4.mini.shareall.K3.pc\endcsname{$+$2 [$-$9, 12]}
\expandafter\def\csname v@g4.mini.shareall.K4.pc\endcsname{$-$5 [$-$16, 5]}
\expandafter\def\csname v@g4.mini.shareall.K5.pc\endcsname{$-$4 [$-$15, 8]}
\expandafter\def\csname v@g4.mini.shareall.r1.pc\endcsname{53 [28, 80]}
\expandafter\def\csname v@g4.mini.shareall.r2.pc\endcsname{9 [$-$28, 39]}
\expandafter\def\csname v@g4.mini.shareall.r3.pc\endcsname{4 [$-$26, 29]}
\expandafter\def\csname v@g4.mini.shareall.r4.pc\endcsname{$-$14 [$-$50, 12]}
\expandafter\def\csname v@g4.mini.shareall.r5.pc\endcsname{$-$9 [$-$45, 19]}
\expandafter\def\csname v@g4.mini.debate.K0.pc\endcsname{$+$31 [18, 43]}
\expandafter\def\csname v@g4.mini.debate.K1.pc\endcsname{$+$11 [$-$1, 22]}
\expandafter\def\csname v@g4.mini.debate.K2.pc\endcsname{$+$11 [$-$3, 25]}
\expandafter\def\csname v@g4.mini.debate.K3.pc\endcsname{$+$8 [$-$3, 18]}
\expandafter\def\csname v@g4.mini.debate.K4.pc\endcsname{$+$5 [$-$7, 18]}
\expandafter\def\csname v@g4.mini.debate.K5.pc\endcsname{$+$2 [$-$10, 15]}
\expandafter\def\csname v@g4.mini.debate.r1.pc\endcsname{36 [$-$3, 71]}
\expandafter\def\csname v@g4.mini.debate.r2.pc\endcsname{37 [$-$9, 77]}
\expandafter\def\csname v@g4.mini.debate.r3.pc\endcsname{25 [$-$9, 66]}
\expandafter\def\csname v@g4.mini.debate.r4.pc\endcsname{17 [$-$27, 60]}
\expandafter\def\csname v@g4.mini.debate.r5.pc\endcsname{5 [$-$37, 50]}
\expandafter\def\csname v@g4.gemma.pool_evpc.K0.pc\endcsname{$+$91 [82, 97]}
\expandafter\def\csname v@g4.gemma.pool_evpc.K1.pc\endcsname{$+$91 [84, 97]}
\expandafter\def\csname v@g4.gemma.pool_evpc.K2.pc\endcsname{$+$90 [82, 96]}
\expandafter\def\csname v@g4.gemma.pool_evpc.K3.pc\endcsname{$+$91 [83, 97]}
\expandafter\def\csname v@g4.gemma.pool_evpc.K4.pc\endcsname{$+$89 [81, 95]}
\expandafter\def\csname v@g4.gemma.pool_evpc.K5.pc\endcsname{$+$83 [73, 92]}
\expandafter\def\csname v@g4.gemma.pool_evpc.r4.pc\endcsname{97 [90, 105]}
\expandafter\def\csname v@g4.gemma.pool_evpc.r5.pc\endcsname{91 [81, 101]}
\expandafter\def\csname v@g4.gemma.balanced.K0.pc\endcsname{$+$42 [30, 55]}
\expandafter\def\csname v@g4.gemma.balanced.K1.pc\endcsname{$+$44 [29, 59]}
\expandafter\def\csname v@g4.gemma.balanced.K2.pc\endcsname{$+$43 [29, 57]}
\expandafter\def\csname v@g4.gemma.balanced.K3.pc\endcsname{$+$30 [16, 45]}
\expandafter\def\csname v@g4.gemma.balanced.K4.pc\endcsname{$+$41 [27, 55]}
\expandafter\def\csname v@g4.gemma.balanced.K5.pc\endcsname{$+$21 [9, 34]}
\expandafter\def\csname v@g4.gemma.balanced.r3.pc\endcsname{72 [42, 100]}
\expandafter\def\csname v@g4.gemma.balanced.r5.pc\endcsname{51 [22, 87]}
\expandafter\def\csname v@g4.gemma.exchange.K0.pc\endcsname{$+$72 [57, 85]}
\expandafter\def\csname v@g4.gemma.exchange.K1.pc\endcsname{$+$76 [65, 85]}
\expandafter\def\csname v@g4.gemma.exchange.K2.pc\endcsname{$+$66 [53, 78]}
\expandafter\def\csname v@g4.gemma.exchange.K3.pc\endcsname{$+$60 [47, 73]}
\expandafter\def\csname v@g4.gemma.exchange.K4.pc\endcsname{$+$54 [42, 66]}
\expandafter\def\csname v@g4.gemma.exchange.K5.pc\endcsname{$+$45 [32, 57]}
\expandafter\def\csname v@g4.gemma.exchange.r4.pc\endcsname{75 [60, 94]}
\expandafter\def\csname v@g4.gemma.exchange.r5.pc\endcsname{62 [46, 82]}
\expandafter\def\csname v@g4.gemma.shareall.K0.pc\endcsname{$+$51 [39, 64]}
\expandafter\def\csname v@g4.gemma.shareall.K1.pc\endcsname{$+$45 [29, 62]}
\expandafter\def\csname v@g4.gemma.shareall.K2.pc\endcsname{$+$40 [23, 56]}
\expandafter\def\csname v@g4.gemma.shareall.K3.pc\endcsname{$+$32 [17, 47]}
\expandafter\def\csname v@g4.gemma.shareall.K4.pc\endcsname{$+$32 [17, 47]}
\expandafter\def\csname v@g4.gemma.shareall.K5.pc\endcsname{$+$17 [5, 29]}
\expandafter\def\csname v@g4.gemma.shareall.r2.pc\endcsname{78 [52, 102]}
\expandafter\def\csname v@g4.gemma.shareall.r3.pc\endcsname{63 [38, 89]}
\expandafter\def\csname v@g4.gemma.shareall.r4.pc\endcsname{62 [36, 87]}
\expandafter\def\csname v@g4.gemma.shareall.r5.pc\endcsname{33 [10, 55]}
\expandafter\def\csname v@g4.gemma.debate.K0.pc\endcsname{$+$46 [33, 60]}
\expandafter\def\csname v@g4.gemma.debate.K1.pc\endcsname{$+$44 [30, 58]}
\expandafter\def\csname v@g4.gemma.debate.K2.pc\endcsname{$+$33 [19, 47]}
\expandafter\def\csname v@g4.gemma.debate.K3.pc\endcsname{$+$31 [16, 45]}
\expandafter\def\csname v@g4.gemma.debate.K4.pc\endcsname{$+$34 [20, 47]}
\expandafter\def\csname v@g4.gemma.debate.K5.pc\endcsname{$+$17 [5, 31]}
\expandafter\def\csname v@g4.gemma.debate.r2.pc\endcsname{73 [46, 102]}
\expandafter\def\csname v@g4.gemma.debate.r3.pc\endcsname{67 [38, 98]}
\expandafter\def\csname v@g4.gemma.debate.r4.pc\endcsname{74 [47, 104]}
\expandafter\def\csname v@g4.gemma.debate.r5.pc\endcsname{38 [11, 63]}
\expandafter\def\csname v@lk.qwen.debate.K0.pc\endcsname{$+$7 [2, 13]}
\expandafter\def\csname v@lk.qwen.debate.K3.pc\endcsname{0 [$-$10, 8]}
\expandafter\def\csname v@lk.qwen.pool.K0.pc\endcsname{$+$75 [62, 85]}
\expandafter\def\csname v@lk.qwen.pool.K3.pc\endcsname{$+$69 [57, 80]}
\expandafter\def\csname v@lk.olmo.debate.K0.pc\endcsname{$+$18 [10, 25]}
\expandafter\def\csname v@lk.olmo.debate.K3.pc\endcsname{$+$6 [0, 13]}
\expandafter\def\csname v@lk.olmo.pool.K0.pc\endcsname{$+$68 [56, 80]}
\expandafter\def\csname v@lk.olmo.pool.K3.pc\endcsname{$+$61 [50, 71]}
\expandafter\def\csname v@g4.pool.kept.pc\endcsname{74--100}
\expandafter\def\csname v@h1.hb16.qwen.pool_evpc.pc\endcsname{88}
\expandafter\def\csname v@h1.hb16.qwen.debate_evpc.pc\endcsname{88}
\expandafter\def\csname v@h1.hb16.qwen.balanced_evpc_rand.pc\endcsname{88}
\expandafter\def\csname v@h1.hb16.qwen.balanced_evpc.pc\endcsname{88}
\expandafter\def\csname v@h1.hb16.qwen.pool_evpc_raw.pc\endcsname{44}
\expandafter\def\csname v@h1.hb16.qwen.debate_evpc_raw.pc\endcsname{43}
\expandafter\def\csname v@h1.hb16.qwen.balanced_nu05_evpc_raw.pc\endcsname{62}
\expandafter\def\csname v@h1.hb16.qwen.balanced_evpc_rand_raw.pc\endcsname{66}
\expandafter\def\csname v@h1.hb16.qwen.balanced_evpc_raw.pc\endcsname{71}
\expandafter\def\csname v@h1.hb16.qwen.debate.pc\endcsname{8}
\expandafter\def\csname v@h1.hb16.qwen.balanced.pc\endcsname{20}
\expandafter\def\csname v@h1.hb16.qwen.solo.pc\endcsname{2}
\expandafter\def\csname v@h1.hb16.qwen.bal_vs_pool.pc\endcsname{$+$28 [17, 40]}
\expandafter\def\csname v@h1.hb16.qwen.bal_vs_pool.pt.pc\endcsname{$+$28}
\expandafter\def\csname v@h1.hb16.qwen.rand_vs_deb.pc\endcsname{$+$23 [11, 35]}
\expandafter\def\csname v@h1.hb16.qwen.rand_vs_deb.pt.pc\endcsname{$+$23}
\expandafter\def\csname v@h1.hb16.qwen.rank.pc\endcsname{$+$5 [$-$3, 13]}
\expandafter\def\csname v@h1.hb16.qwen.rank.pt.pc\endcsname{$+$5}
\expandafter\def\csname v@h1.hb16.qwen.cost.pc\endcsname{$+$16}
\expandafter\def\csname v@h1.hb16.qwen.costpool.pc\endcsname{$+$44}
\expandafter\def\csname v@h1.hb16.qwen.swto.pool_evpc_raw.pc\endcsname{42}
\expandafter\def\csname v@h1.hb16.qwen.swaway.pool_evpc_raw.pc\endcsname{0}
\expandafter\def\csname v@h1.hb16.qwen.gain.balanced_evpc_raw.pc\endcsname{67 [50, 82]}
\expandafter\def\csname v@h1.hb16.qwen.gain.balanced_evpc_raw.pt.pc\endcsname{67}
\expandafter\def\csname v@h1.hb16.qwen.swto.balanced_evpc_raw.pc\endcsname{71}
\expandafter\def\csname v@h1.hb16.qwen.swaway.balanced_evpc_raw.pc\endcsname{2}
\expandafter\def\csname v@h1.hb16.qwen.gain.debate_evpc_raw.pt.pc\endcsname{97}
\expandafter\def\csname v@h1.hb16.qwen.swto.debate_evpc_raw.pc\endcsname{44}
\expandafter\def\csname v@h1.hb16.qwen.swaway.debate_evpc_raw.pc\endcsname{3}
\expandafter\def\csname v@h1.hb16.qwen.gain.balanced_evpc_rand_raw.pc\endcsname{61 [45, 77]}
\expandafter\def\csname v@h1.hb16.qwen.gain.balanced_evpc_rand_raw.pt.pc\endcsname{61}
\expandafter\def\csname v@h1.hb16.qwen.swto.balanced_evpc_rand_raw.pc\endcsname{67}
\expandafter\def\csname v@h1.hb16.qwen.swaway.balanced_evpc_rand_raw.pc\endcsname{3}
\expandafter\def\csname v@h1.hb32.qwen.pool_evpc.pc\endcsname{91}
\expandafter\def\csname v@h1.hb32.qwen.debate_evpc.pc\endcsname{79}
\expandafter\def\csname v@h1.hb32.qwen.balanced_evpc_rand.pc\endcsname{85}
\expandafter\def\csname v@h1.hb32.qwen.balanced_evpc.pc\endcsname{91}
\expandafter\def\csname v@h1.hb32.qwen.pool_evpc_raw.pc\endcsname{31}
\expandafter\def\csname v@h1.hb32.qwen.debate_evpc_raw.pc\endcsname{34}
\expandafter\def\csname v@h1.hb32.qwen.balanced_evpc_rand_raw.pc\endcsname{57}
\expandafter\def\csname v@h1.hb32.qwen.balanced_evpc_raw.pc\endcsname{78}
\expandafter\def\csname v@h1.hb32.qwen.debate.pc\endcsname{5}
\expandafter\def\csname v@h1.hb32.qwen.balanced.pc\endcsname{22}
\expandafter\def\csname v@h1.hb32.qwen.solo.pc\endcsname{1}
\expandafter\def\csname v@h1.hb32.qwen.bal_vs_pool.pc\endcsname{$+$47 [31, 62]}
\expandafter\def\csname v@h1.hb32.qwen.bal_vs_pool.pt.pc\endcsname{$+$47}
\expandafter\def\csname v@h1.hb32.qwen.rand_vs_deb.pc\endcsname{$+$22 [13, 32]}
\expandafter\def\csname v@h1.hb32.qwen.rand_vs_deb.pt.pc\endcsname{$+$22}
\expandafter\def\csname v@h1.hb32.qwen.rank.pc\endcsname{$+$21 [11, 32]}
\expandafter\def\csname v@h1.hb32.qwen.rank.pt.pc\endcsname{$+$21}
\expandafter\def\csname v@h1.hb32.qwen.cost.pc\endcsname{$+$13}
\expandafter\def\csname v@h1.hb32.qwen.costpool.pc\endcsname{$+$59}
\expandafter\def\csname v@h1.hb32.qwen.swto.pool_evpc_raw.pc\endcsname{31}
\expandafter\def\csname v@h1.hb32.qwen.swaway.pool_evpc_raw.pc\endcsname{1}
\expandafter\def\csname v@h1.hb32.qwen.gain.balanced_evpc_raw.pc\endcsname{60 [48, 72]}
\expandafter\def\csname v@h1.hb32.qwen.gain.balanced_evpc_raw.pt.pc\endcsname{60}
\expandafter\def\csname v@h1.hb32.qwen.swto.balanced_evpc_raw.pc\endcsname{79}
\expandafter\def\csname v@h1.hb32.qwen.swaway.balanced_evpc_raw.pc\endcsname{2}
\expandafter\def\csname v@h1.hb32.qwen.gain.debate_evpc_raw.pt.pc\endcsname{97}
\expandafter\def\csname v@h1.hb32.qwen.swto.debate_evpc_raw.pc\endcsname{34}
\expandafter\def\csname v@h1.hb32.qwen.swaway.debate_evpc_raw.pc\endcsname{1}
\expandafter\def\csname v@h1.hb32.qwen.gain.balanced_evpc_rand_raw.pc\endcsname{72 [62, 82]}
\expandafter\def\csname v@h1.hb32.qwen.gain.balanced_evpc_rand_raw.pt.pc\endcsname{72}
\expandafter\def\csname v@h1.hb32.qwen.swto.balanced_evpc_rand_raw.pc\endcsname{64}
\expandafter\def\csname v@h1.hb32.qwen.swaway.balanced_evpc_rand_raw.pc\endcsname{9}
\expandafter\def\csname v@h1.pz16.qwen.pool_evpc.pc\endcsname{85}
\expandafter\def\csname v@h1.pz16.qwen.debate_evpc.pc\endcsname{60}
\expandafter\def\csname v@h1.pz16.qwen.balanced_evpc_rand.pc\endcsname{56}
\expandafter\def\csname v@h1.pz16.qwen.balanced_evpc.pc\endcsname{83}
\expandafter\def\csname v@h1.pz16.qwen.pool_evpc_raw.pc\endcsname{22}
\expandafter\def\csname v@h1.pz16.qwen.debate_evpc_raw.pc\endcsname{28}
\expandafter\def\csname v@h1.pz16.qwen.balanced_nu05_evpc_raw.pc\endcsname{27}
\expandafter\def\csname v@h1.pz16.qwen.balanced_evpc_rand_raw.pc\endcsname{26}
\expandafter\def\csname v@h1.pz16.qwen.balanced_evpc_raw.pc\endcsname{34}
\expandafter\def\csname v@h1.pz16.qwen.debate.pc\endcsname{10}
\expandafter\def\csname v@h1.pz16.qwen.balanced.pc\endcsname{15}
\expandafter\def\csname v@h1.pz16.qwen.solo.pc\endcsname{8}
\expandafter\def\csname v@h1.pz16.qwen.bal_vs_pool.pc\endcsname{$+$12 [6, 17]}
\expandafter\def\csname v@h1.pz16.qwen.bal_vs_pool.pt.pc\endcsname{$+$12}
\expandafter\def\csname v@h1.pz16.qwen.rand_vs_deb.pc\endcsname{$-$1 [$-$5, 2]}
\expandafter\def\csname v@h1.pz16.qwen.rand_vs_deb.pt.pc\endcsname{$-$1}
\expandafter\def\csname v@h1.pz16.qwen.rank.pc\endcsname{$+$7 [3, 12]}
\expandafter\def\csname v@h1.pz16.qwen.rank.pt.pc\endcsname{$+$7}
\expandafter\def\csname v@h1.pz16.qwen.cost.pc\endcsname{$+$49}
\expandafter\def\csname v@h1.pz16.qwen.costpool.pc\endcsname{$+$64}
\expandafter\def\csname v@h1.pz16.qwen.gain.pool_evpc_raw.pc\endcsname{85 [64, 104]}
\expandafter\def\csname v@h1.pz16.qwen.gain.pool_evpc_raw.pt.pc\endcsname{85}
\expandafter\def\csname v@h1.pz16.qwen.swto.pool_evpc_raw.pc\endcsname{16}
\expandafter\def\csname v@h1.pz16.qwen.swaway.pool_evpc_raw.pc\endcsname{1}
\expandafter\def\csname v@h1.pz16.qwen.gain.balanced_evpc_raw.pc\endcsname{44 [30, 56]}
\expandafter\def\csname v@h1.pz16.qwen.gain.balanced_evpc_raw.pt.pc\endcsname{44}
\expandafter\def\csname v@h1.pz16.qwen.swto.balanced_evpc_raw.pc\endcsname{35}
\expandafter\def\csname v@h1.pz16.qwen.swaway.balanced_evpc_raw.pc\endcsname{8}
\expandafter\def\csname v@h1.pz16.qwen.gain.debate_evpc_raw.pc\endcsname{92 [81, 103]}
\expandafter\def\csname v@h1.pz16.qwen.gain.debate_evpc_raw.pt.pc\endcsname{92}
\expandafter\def\csname v@h1.pz16.qwen.swto.debate_evpc_raw.pc\endcsname{22}
\expandafter\def\csname v@h1.pz16.qwen.swaway.debate_evpc_raw.pc\endcsname{1}
\expandafter\def\csname v@h1.pz16.qwen.gain.balanced_evpc_rand_raw.pc\endcsname{63 [50, 75]}
\expandafter\def\csname v@h1.pz16.qwen.gain.balanced_evpc_rand_raw.pt.pc\endcsname{63}
\expandafter\def\csname v@h1.pz16.qwen.swto.balanced_evpc_rand_raw.pc\endcsname{28}
\expandafter\def\csname v@h1.pz16.qwen.swaway.balanced_evpc_rand_raw.pc\endcsname{8}
\expandafter\def\csname v@h1.late16.qwen.pool_evpc.pc\endcsname{90}
\expandafter\def\csname v@h1.late16.qwen.debate_evpc.pc\endcsname{80}
\expandafter\def\csname v@h1.late16.qwen.balanced_evpc.pc\endcsname{90}
\expandafter\def\csname v@h1.late16.qwen.pool_evpc_raw.pc\endcsname{9}
\expandafter\def\csname v@h1.late16.qwen.debate_evpc_raw.pc\endcsname{11}
\expandafter\def\csname v@h1.late16.qwen.balanced_evpc_rand_raw.pc\endcsname{27}
\expandafter\def\csname v@h1.late16.qwen.balanced_evpc_raw.pc\endcsname{28}
\expandafter\def\csname v@h1.late16.qwen.debate.pc\endcsname{2}
\expandafter\def\csname v@h1.late16.qwen.solo.pc\endcsname{2}
\expandafter\def\csname v@h1.late16.qwen.bal_vs_pool.pc\endcsname{$+$19 [8, 30]}
\expandafter\def\csname v@h1.late16.qwen.bal_vs_pool.pt.pc\endcsname{$+$19}
\expandafter\def\csname v@h1.late16.qwen.rand_vs_deb.pc\endcsname{$+$15 [7, 24]}
\expandafter\def\csname v@h1.late16.qwen.rand_vs_deb.pt.pc\endcsname{$+$15}
\expandafter\def\csname v@h1.late16.qwen.rank.pc\endcsname{$+$2 [$-$7, 10]}
\expandafter\def\csname v@h1.late16.qwen.rank.pt.pc\endcsname{$+$2}
\expandafter\def\csname v@h1.late16.qwen.cost.pc\endcsname{$+$61}
\expandafter\def\csname v@h1.late16.qwen.costpool.pc\endcsname{$+$81}
\expandafter\def\csname v@h1.late16.qwen.swto.pool_evpc_raw.pc\endcsname{9}
\expandafter\def\csname v@h1.late16.qwen.swaway.pool_evpc_raw.pc\endcsname{0}
\expandafter\def\csname v@h1.late16.qwen.gain.balanced_evpc_raw.pc\endcsname{64 [29, 90]}
\expandafter\def\csname v@h1.late16.qwen.gain.balanced_evpc_raw.pt.pc\endcsname{64}
\expandafter\def\csname v@h1.late16.qwen.swto.balanced_evpc_raw.pc\endcsname{28}
\expandafter\def\csname v@h1.late16.qwen.swaway.balanced_evpc_raw.pc\endcsname{0}
\expandafter\def\csname v@h1.late16.qwen.swto.debate_evpc_raw.pc\endcsname{12}
\expandafter\def\csname v@h1.late16.qwen.swaway.debate_evpc_raw.pc\endcsname{0}
\expandafter\def\csname v@h1.late16.qwen.gain.balanced_evpc_rand_raw.pt.pc\endcsname{87}
\expandafter\def\csname v@h1.late16.qwen.swto.balanced_evpc_rand_raw.pc\endcsname{28}
\expandafter\def\csname v@h1.late16.qwen.swaway.balanced_evpc_rand_raw.pc\endcsname{2}
\expandafter\def\csname v@h1.kernel.qwen.pool_evpc.K0.pc\endcsname{$+$86 [75, 96]}
\expandafter\def\csname v@h1.kernel.qwen.pool_evpc.K0.pt.pc\endcsname{$+$86}
\expandafter\def\csname v@h1.kernel.qwen.pool_evpc.K.pc\endcsname{$+$89 [79, 96]}
\expandafter\def\csname v@h1.kernel.qwen.pool_evpc.K.pt.pc\endcsname{$+$89}
\expandafter\def\csname v@h1.kernel.qwen.pool_evpc.ratio.pt.pc\endcsname{102}
\expandafter\def\csname v@h1.kernel.qwen.pool_evpc_raw.K0.pc\endcsname{$+$42 [29, 56]}
\expandafter\def\csname v@h1.kernel.qwen.pool_evpc_raw.K0.pt.pc\endcsname{$+$42}
\expandafter\def\csname v@h1.kernel.qwen.pool_evpc_raw.K.pc\endcsname{$+$8 [2, 16]}
\expandafter\def\csname v@h1.kernel.qwen.pool_evpc_raw.K.pt.pc\endcsname{$+$8}
\expandafter\def\csname v@h1.kernel.qwen.pool_evpc_raw.ratio.pc\endcsname{19 [4, 39]}
\expandafter\def\csname v@h1.kernel.qwen.pool_evpc_raw.ratio.pt.pc\endcsname{19}
\expandafter\def\csname v@h1.kernel.qwen.balanced_evpc_raw.K0.pc\endcsname{$+$70 [56, 83]}
\expandafter\def\csname v@h1.kernel.qwen.balanced_evpc_raw.K0.pt.pc\endcsname{$+$70}
\expandafter\def\csname v@h1.kernel.qwen.balanced_evpc_raw.K.pc\endcsname{$+$27 [13, 41]}
\expandafter\def\csname v@h1.kernel.qwen.balanced_evpc_raw.K.pt.pc\endcsname{$+$27}
\expandafter\def\csname v@h1.kernel.qwen.balanced_evpc_raw.ratio.pc\endcsname{38 [20, 56]}
\expandafter\def\csname v@h1.kernel.qwen.balanced_evpc_raw.ratio.pt.pc\endcsname{38}
\expandafter\def\csname v@h1.kernel.qwen.debate_evpc_raw.K0.pc\endcsname{$+$42 [27, 57]}
\expandafter\def\csname v@h1.kernel.qwen.debate_evpc_raw.K0.pt.pc\endcsname{$+$42}
\expandafter\def\csname v@h1.kernel.qwen.debate_evpc_raw.K.pc\endcsname{$+$10 [4, 17]}
\expandafter\def\csname v@h1.kernel.qwen.debate_evpc_raw.K.pt.pc\endcsname{$+$10}
\expandafter\def\csname v@h1.kernel.qwen.debate_evpc_raw.ratio.pc\endcsname{24 [11, 39]}
\expandafter\def\csname v@h1.kernel.qwen.debate_evpc_raw.ratio.pt.pc\endcsname{24}
\expandafter\def\csname v@h1.kernel.qwen.balanced_evpc_rand_raw.K0.pc\endcsname{$+$65 [50, 79]}
\expandafter\def\csname v@h1.kernel.qwen.balanced_evpc_rand_raw.K0.pt.pc\endcsname{$+$65}
\expandafter\def\csname v@h1.kernel.qwen.balanced_evpc_rand_raw.K.pc\endcsname{$+$25 [14, 38]}
\expandafter\def\csname v@h1.kernel.qwen.balanced_evpc_rand_raw.K.pt.pc\endcsname{$+$25}
\expandafter\def\csname v@h1.kernel.qwen.balanced_evpc_rand_raw.ratio.pc\endcsname{39 [24, 55]}
\expandafter\def\csname v@h1.kernel.qwen.balanced_evpc_rand_raw.ratio.pt.pc\endcsname{39}
\expandafter\def\csname v@h1.hb16.olmo.pool_evpc.pc\endcsname{88}
\expandafter\def\csname v@h1.hb16.olmo.debate_evpc.pc\endcsname{81}
\expandafter\def\csname v@h1.hb16.olmo.balanced_evpc_rand.pc\endcsname{80}
\expandafter\def\csname v@h1.hb16.olmo.balanced_evpc.pc\endcsname{90}
\expandafter\def\csname v@h1.hb16.olmo.pool_evpc_raw.pc\endcsname{54}
\expandafter\def\csname v@h1.hb16.olmo.debate_evpc_raw.pc\endcsname{46}
\expandafter\def\csname v@h1.hb16.olmo.balanced_nu05_evpc_raw.pc\endcsname{52}
\expandafter\def\csname v@h1.hb16.olmo.balanced_evpc_rand_raw.pc\endcsname{54}
\expandafter\def\csname v@h1.hb16.olmo.balanced_evpc_raw.pc\endcsname{63}
\expandafter\def\csname v@h1.hb16.olmo.debate.pc\endcsname{21}
\expandafter\def\csname v@h1.hb16.olmo.balanced.pc\endcsname{31}
\expandafter\def\csname v@h1.hb16.olmo.solo.pc\endcsname{11}
\expandafter\def\csname v@h1.hb16.olmo.bal_vs_pool.pc\endcsname{$+$9 [$-$2, 19]}
\expandafter\def\csname v@h1.hb16.olmo.bal_vs_pool.pt.pc\endcsname{$+$9}
\expandafter\def\csname v@h1.hb16.olmo.rand_vs_deb.pc\endcsname{$+$8 [$-$7, 23]}
\expandafter\def\csname v@h1.hb16.olmo.rand_vs_deb.pt.pc\endcsname{$+$8}
\expandafter\def\csname v@h1.hb16.olmo.rank.pc\endcsname{$+$9 [$-$4, 21]}
\expandafter\def\csname v@h1.hb16.olmo.rank.pt.pc\endcsname{$+$9}
\expandafter\def\csname v@h1.hb16.olmo.cost.pc\endcsname{$+$27}
\expandafter\def\csname v@h1.hb16.olmo.costpool.pc\endcsname{$+$33}
\expandafter\def\csname v@h1.hb16.olmo.gain.pool_evpc_raw.pc\endcsname{83 [76, 91]}
\expandafter\def\csname v@h1.hb16.olmo.gain.pool_evpc_raw.pt.pc\endcsname{83}
\expandafter\def\csname v@h1.hb16.olmo.swto.pool_evpc_raw.pc\endcsname{48}
\expandafter\def\csname v@h1.hb16.olmo.swaway.pool_evpc_raw.pc\endcsname{6}
\expandafter\def\csname v@h1.hb16.olmo.gain.balanced_evpc_raw.pc\endcsname{19 [7, 30]}
\expandafter\def\csname v@h1.hb16.olmo.gain.balanced_evpc_raw.pt.pc\endcsname{19}
\expandafter\def\csname v@h1.hb16.olmo.swto.balanced_evpc_raw.pc\endcsname{104}
\expandafter\def\csname v@h1.hb16.olmo.swaway.balanced_evpc_raw.pc\endcsname{55}
\expandafter\def\csname v@h1.hb16.olmo.gain.debate_evpc_raw.pc\endcsname{87 [75, 95]}
\expandafter\def\csname v@h1.hb16.olmo.gain.debate_evpc_raw.pt.pc\endcsname{87}
\expandafter\def\csname v@h1.hb16.olmo.swto.debate_evpc_raw.pc\endcsname{43}
\expandafter\def\csname v@h1.hb16.olmo.swaway.debate_evpc_raw.pc\endcsname{9}
\expandafter\def\csname v@h1.hb16.olmo.gain.balanced_evpc_rand_raw.pc\endcsname{37 [26, 47]}
\expandafter\def\csname v@h1.hb16.olmo.gain.balanced_evpc_rand_raw.pt.pc\endcsname{37}
\expandafter\def\csname v@h1.hb16.olmo.swto.balanced_evpc_rand_raw.pc\endcsname{83}
\expandafter\def\csname v@h1.hb16.olmo.swaway.balanced_evpc_rand_raw.pc\endcsname{43}
\expandafter\def\csname v@h1.hb32.olmo.pool_evpc.pc\endcsname{92}
\expandafter\def\csname v@h1.hb32.olmo.debate_evpc.pc\endcsname{57}
\expandafter\def\csname v@h1.hb32.olmo.balanced_evpc_rand.pc\endcsname{62}
\expandafter\def\csname v@h1.hb32.olmo.balanced_evpc.pc\endcsname{90}
\expandafter\def\csname v@h1.hb32.olmo.pool_evpc_raw.pc\endcsname{48}
\expandafter\def\csname v@h1.hb32.olmo.debate_evpc_raw.pc\endcsname{41}
\expandafter\def\csname v@h1.hb32.olmo.balanced_evpc_rand_raw.pc\endcsname{46}
\expandafter\def\csname v@h1.hb32.olmo.balanced_evpc_raw.pc\endcsname{72}
\expandafter\def\csname v@h1.hb32.olmo.debate.pc\endcsname{21}
\expandafter\def\csname v@h1.hb32.olmo.balanced.pc\endcsname{34}
\expandafter\def\csname v@h1.hb32.olmo.solo.pc\endcsname{9}
\expandafter\def\csname v@h1.hb32.olmo.bal_vs_pool.pc\endcsname{$+$23 [11, 36]}
\expandafter\def\csname v@h1.hb32.olmo.bal_vs_pool.pt.pc\endcsname{$+$23}
\expandafter\def\csname v@h1.hb32.olmo.rand_vs_deb.pc\endcsname{$+$5 [$-$6, 17]}
\expandafter\def\csname v@h1.hb32.olmo.rand_vs_deb.pt.pc\endcsname{$+$5}
\expandafter\def\csname v@h1.hb32.olmo.rank.pc\endcsname{$+$26 [13, 39]}
\expandafter\def\csname v@h1.hb32.olmo.rank.pt.pc\endcsname{$+$26}
\expandafter\def\csname v@h1.hb32.olmo.cost.pc\endcsname{$+$18}
\expandafter\def\csname v@h1.hb32.olmo.costpool.pc\endcsname{$+$43}
\expandafter\def\csname v@h1.hb32.olmo.gain.pool_evpc_raw.pc\endcsname{66 [56, 77]}
\expandafter\def\csname v@h1.hb32.olmo.gain.pool_evpc_raw.pt.pc\endcsname{66}
\expandafter\def\csname v@h1.hb32.olmo.swto.pool_evpc_raw.pc\endcsname{49}
\expandafter\def\csname v@h1.hb32.olmo.swaway.pool_evpc_raw.pc\endcsname{10}
\expandafter\def\csname v@h1.hb32.olmo.gain.balanced_evpc_raw.pc\endcsname{20 [9, 29]}
\expandafter\def\csname v@h1.hb32.olmo.gain.balanced_evpc_raw.pt.pc\endcsname{20}
\expandafter\def\csname v@h1.hb32.olmo.swaway.balanced_evpc_raw.pc\endcsname{47}
\expandafter\def\csname v@h1.hb32.olmo.gain.debate_evpc_raw.pc\endcsname{92 [82, 103]}
\expandafter\def\csname v@h1.hb32.olmo.gain.debate_evpc_raw.pt.pc\endcsname{92}
\expandafter\def\csname v@h1.hb32.olmo.swto.debate_evpc_raw.pc\endcsname{38}
\expandafter\def\csname v@h1.hb32.olmo.swaway.debate_evpc_raw.pc\endcsname{6}
\expandafter\def\csname v@h1.hb32.olmo.gain.balanced_evpc_rand_raw.pc\endcsname{41 [29, 51]}
\expandafter\def\csname v@h1.hb32.olmo.gain.balanced_evpc_rand_raw.pt.pc\endcsname{41}
\expandafter\def\csname v@h1.hb32.olmo.swto.balanced_evpc_rand_raw.pc\endcsname{97}
\expandafter\def\csname v@h1.hb32.olmo.swaway.balanced_evpc_rand_raw.pc\endcsname{65}
\expandafter\def\csname v@h1.pz16.olmo.pool_evpc.pc\endcsname{75}
\expandafter\def\csname v@h1.pz16.olmo.debate_evpc.pc\endcsname{44}
\expandafter\def\csname v@h1.pz16.olmo.balanced_evpc_rand.pc\endcsname{36}
\expandafter\def\csname v@h1.pz16.olmo.balanced_evpc.pc\endcsname{70}
\expandafter\def\csname v@h1.pz16.olmo.pool_evpc_raw.pc\endcsname{13}
\expandafter\def\csname v@h1.pz16.olmo.debate_evpc_raw.pc\endcsname{15}
\expandafter\def\csname v@h1.pz16.olmo.balanced_nu05_evpc_raw.pc\endcsname{17}
\expandafter\def\csname v@h1.pz16.olmo.balanced_evpc_rand_raw.pc\endcsname{19}
\expandafter\def\csname v@h1.pz16.olmo.balanced_evpc_raw.pc\endcsname{19}
\expandafter\def\csname v@h1.pz16.olmo.debate.pc\endcsname{8}
\expandafter\def\csname v@h1.pz16.olmo.balanced.pc\endcsname{14}
\expandafter\def\csname v@h1.pz16.olmo.solo.pc\endcsname{4}
\expandafter\def\csname v@h1.pz16.olmo.bal_vs_pool.pc\endcsname{$+$7 [3, 11]}
\expandafter\def\csname v@h1.pz16.olmo.bal_vs_pool.pt.pc\endcsname{$+$7}
\expandafter\def\csname v@h1.pz16.olmo.rand_vs_deb.pc\endcsname{$+$5 [$-$1, 10]}
\expandafter\def\csname v@h1.pz16.olmo.rand_vs_deb.pt.pc\endcsname{$+$5}
\expandafter\def\csname v@h1.pz16.olmo.rank.pc\endcsname{0 [$-$4, 4]}
\expandafter\def\csname v@h1.pz16.olmo.rank.pt.pc\endcsname{0}
\expandafter\def\csname v@h1.pz16.olmo.cost.pc\endcsname{$+$51}
\expandafter\def\csname v@h1.pz16.olmo.costpool.pc\endcsname{$+$62}
\expandafter\def\csname v@h1.pz16.olmo.gain.pool_evpc_raw.pc\endcsname{84 [64, 99]}
\expandafter\def\csname v@h1.pz16.olmo.gain.pool_evpc_raw.pt.pc\endcsname{84}
\expandafter\def\csname v@h1.pz16.olmo.swto.pool_evpc_raw.pc\endcsname{11}
\expandafter\def\csname v@h1.pz16.olmo.swaway.pool_evpc_raw.pc\endcsname{1}
\expandafter\def\csname v@h1.pz16.olmo.gain.balanced_evpc_raw.pc\endcsname{20 [8, 29]}
\expandafter\def\csname v@h1.pz16.olmo.gain.balanced_evpc_raw.pt.pc\endcsname{20}
\expandafter\def\csname v@h1.pz16.olmo.swto.balanced_evpc_raw.pc\endcsname{40}
\expandafter\def\csname v@h1.pz16.olmo.swaway.balanced_evpc_raw.pc\endcsname{25}
\expandafter\def\csname v@h1.pz16.olmo.gain.debate_evpc_raw.pc\endcsname{92 [78, 105]}
\expandafter\def\csname v@h1.pz16.olmo.gain.debate_evpc_raw.pt.pc\endcsname{92}
\expandafter\def\csname v@h1.pz16.olmo.swto.debate_evpc_raw.pc\endcsname{13}
\expandafter\def\csname v@h1.pz16.olmo.swaway.debate_evpc_raw.pc\endcsname{2}
\expandafter\def\csname v@h1.pz16.olmo.gain.balanced_evpc_rand_raw.pc\endcsname{35 [21, 47]}
\expandafter\def\csname v@h1.pz16.olmo.gain.balanced_evpc_rand_raw.pt.pc\endcsname{35}
\expandafter\def\csname v@h1.pz16.olmo.swto.balanced_evpc_rand_raw.pc\endcsname{35}
\expandafter\def\csname v@h1.pz16.olmo.swaway.balanced_evpc_rand_raw.pc\endcsname{20}
\expandafter\def\csname v@h1.late16.olmo.pool_evpc.pc\endcsname{85}
\expandafter\def\csname v@h1.late16.olmo.debate_evpc.pc\endcsname{71}
\expandafter\def\csname v@h1.late16.olmo.balanced_evpc.pc\endcsname{87}
\expandafter\def\csname v@h1.late16.olmo.pool_evpc_raw.pc\endcsname{23}
\expandafter\def\csname v@h1.late16.olmo.debate_evpc_raw.pc\endcsname{20}
\expandafter\def\csname v@h1.late16.olmo.balanced_evpc_rand_raw.pc\endcsname{40}
\expandafter\def\csname v@h1.late16.olmo.balanced_evpc_raw.pc\endcsname{44}
\expandafter\def\csname v@h1.late16.olmo.debate.pc\endcsname{16}
\expandafter\def\csname v@h1.late16.olmo.solo.pc\endcsname{10}
\expandafter\def\csname v@h1.late16.olmo.bal_vs_pool.pc\endcsname{$+$21 [9, 32]}
\expandafter\def\csname v@h1.late16.olmo.bal_vs_pool.pt.pc\endcsname{$+$21}
\expandafter\def\csname v@h1.late16.olmo.rand_vs_deb.pc\endcsname{$+$20 [7, 34]}
\expandafter\def\csname v@h1.late16.olmo.rand_vs_deb.pt.pc\endcsname{$+$20}
\expandafter\def\csname v@h1.late16.olmo.rank.pc\endcsname{$+$4 [$-$6, 14]}
\expandafter\def\csname v@h1.late16.olmo.rank.pt.pc\endcsname{$+$4}
\expandafter\def\csname v@h1.late16.olmo.cost.pc\endcsname{$+$43}
\expandafter\def\csname v@h1.late16.olmo.costpool.pc\endcsname{$+$61}
\expandafter\def\csname v@h1.late16.olmo.gain.pool_evpc_raw.pc\endcsname{75 [50, 94]}
\expandafter\def\csname v@h1.late16.olmo.gain.pool_evpc_raw.pt.pc\endcsname{75}
\expandafter\def\csname v@h1.late16.olmo.swto.pool_evpc_raw.pc\endcsname{22}
\expandafter\def\csname v@h1.late16.olmo.swaway.pool_evpc_raw.pc\endcsname{8}
\expandafter\def\csname v@h1.late16.olmo.gain.balanced_evpc_raw.pc\endcsname{44 [29, 57]}
\expandafter\def\csname v@h1.late16.olmo.gain.balanced_evpc_raw.pt.pc\endcsname{44}
\expandafter\def\csname v@h1.late16.olmo.swto.balanced_evpc_raw.pc\endcsname{62}
\expandafter\def\csname v@h1.late16.olmo.swaway.balanced_evpc_raw.pc\endcsname{30}
\expandafter\def\csname v@h1.late16.olmo.gain.debate_evpc_raw.pt.pc\endcsname{95}
\expandafter\def\csname v@h1.late16.olmo.swto.debate_evpc_raw.pc\endcsname{19}
\expandafter\def\csname v@h1.late16.olmo.swaway.debate_evpc_raw.pc\endcsname{8}
\expandafter\def\csname v@h1.late16.olmo.gain.balanced_evpc_rand_raw.pc\endcsname{37 [24, 49]}
\expandafter\def\csname v@h1.late16.olmo.gain.balanced_evpc_rand_raw.pt.pc\endcsname{37}
\expandafter\def\csname v@h1.late16.olmo.swto.balanced_evpc_rand_raw.pc\endcsname{66}
\expandafter\def\csname v@h1.late16.olmo.swaway.balanced_evpc_rand_raw.pc\endcsname{36}
\expandafter\def\csname v@h1.kernel.olmo.pool_evpc.K0.pc\endcsname{$+$77 [59, 92]}
\expandafter\def\csname v@h1.kernel.olmo.pool_evpc.K0.pt.pc\endcsname{$+$77}
\expandafter\def\csname v@h1.kernel.olmo.pool_evpc.K.pc\endcsname{$+$74 [56, 90]}
\expandafter\def\csname v@h1.kernel.olmo.pool_evpc.K.pt.pc\endcsname{$+$74}
\expandafter\def\csname v@h1.kernel.olmo.pool_evpc.ratio.pc\endcsname{97 [90, 104]}
\expandafter\def\csname v@h1.kernel.olmo.pool_evpc.ratio.pt.pc\endcsname{97}
\expandafter\def\csname v@h1.kernel.olmo.pool_evpc_raw.K0.pc\endcsname{$+$43 [26, 60]}
\expandafter\def\csname v@h1.kernel.olmo.pool_evpc_raw.K0.pt.pc\endcsname{$+$43}
\expandafter\def\csname v@h1.kernel.olmo.pool_evpc_raw.K.pc\endcsname{$+$13 [2, 24]}
\expandafter\def\csname v@h1.kernel.olmo.pool_evpc_raw.K.pt.pc\endcsname{$+$13}
\expandafter\def\csname v@h1.kernel.olmo.pool_evpc_raw.ratio.pc\endcsname{30 [7, 49]}
\expandafter\def\csname v@h1.kernel.olmo.pool_evpc_raw.ratio.pt.pc\endcsname{30}
\expandafter\def\csname v@h1.kernel.olmo.balanced_evpc_raw.K0.pc\endcsname{$+$52 [33, 69]}
\expandafter\def\csname v@h1.kernel.olmo.balanced_evpc_raw.K0.pt.pc\endcsname{$+$52}
\expandafter\def\csname v@h1.kernel.olmo.balanced_evpc_raw.K.pc\endcsname{$+$34 [19, 48]}
\expandafter\def\csname v@h1.kernel.olmo.balanced_evpc_raw.K.pt.pc\endcsname{$+$34}
\expandafter\def\csname v@h1.kernel.olmo.balanced_evpc_raw.ratio.pc\endcsname{65 [43, 90]}
\expandafter\def\csname v@h1.kernel.olmo.balanced_evpc_raw.ratio.pt.pc\endcsname{65}
\expandafter\def\csname v@h1.kernel.olmo.debate_evpc_raw.K0.pc\endcsname{$+$35 [21, 49]}
\expandafter\def\csname v@h1.kernel.olmo.debate_evpc_raw.K0.pt.pc\endcsname{$+$35}
\expandafter\def\csname v@h1.kernel.olmo.debate_evpc_raw.K.pc\endcsname{$+$9 [0, 20]}
\expandafter\def\csname v@h1.kernel.olmo.debate_evpc_raw.K.pt.pc\endcsname{$+$9}
\expandafter\def\csname v@h1.kernel.olmo.debate_evpc_raw.ratio.pc\endcsname{27 [0, 55]}
\expandafter\def\csname v@h1.kernel.olmo.debate_evpc_raw.ratio.pt.pc\endcsname{27}
\expandafter\def\csname v@h1.kernel.olmo.balanced_evpc_rand_raw.K0.pc\endcsname{$+$43 [24, 60]}
\expandafter\def\csname v@h1.kernel.olmo.balanced_evpc_rand_raw.K0.pt.pc\endcsname{$+$43}
\expandafter\def\csname v@h1.kernel.olmo.balanced_evpc_rand_raw.K.pc\endcsname{$+$30 [14, 45]}
\expandafter\def\csname v@h1.kernel.olmo.balanced_evpc_rand_raw.K.pt.pc\endcsname{$+$30}
\expandafter\def\csname v@h1.kernel.olmo.balanced_evpc_rand_raw.ratio.pc\endcsname{69 [42, 98]}
\expandafter\def\csname v@h1.kernel.olmo.balanced_evpc_rand_raw.ratio.pt.pc\endcsname{69}
\expandafter\def\csname v@h1.hb16.gemma.rand_vs_deb.pt.pc\endcsname{$+$11}
\expandafter\def\csname v@h1.hb16.gemma.rank.pt.pc\endcsname{$+$8}
\expandafter\def\csname v@h1.hb16.gemma.bal_vs_pool.pc\endcsname{$+$14 [1, 28]}
\expandafter\def\csname v@h1.hb16.gemma.bal_vs_pool.pt.pc\endcsname{$+$14}
\expandafter\def\csname v@h1.hb16.mini.rand_vs_deb.pt.pc\endcsname{$-$4}
\expandafter\def\csname v@h1.hb16.mini.rank.pt.pc\endcsname{$-$5}
\expandafter\def\csname v@h1.hb16.mini.bal_vs_pool.pc\endcsname{$-$12 [$-$23, $-$1]}
\expandafter\def\csname v@h1.hb16.mini.bal_vs_pool.pt.pc\endcsname{$-$12}
\expandafter\def\csname v@h1.bal_vs_pool.range.pc\endcsname{7 to 47}
\expandafter\def\csname v@h1.rand_vs_deb.range.pc\endcsname{$-$1 to $+$23}
\expandafter\def\csname v@h1.costratio.range.pc\endcsname{22 to 82}
\expandafter\def\csname v@h1.rank.range.pc\endcsname{0 to 26}
\expandafter\def\csname v@h1.size.range.pc\endcsname{$-$8 to $+$6}
\expandafter\def\csname v@h2.hb.qwen.pool_evpc.pc\endcsname{88}
\expandafter\def\csname v@h2.hb.qwen.pool_evpc_cs1.pc\endcsname{85}
\expandafter\def\csname v@h2.hb.qwen.pool_evpc_cs1.d.pc\endcsname{$-$3}
\expandafter\def\csname v@h2.hb.qwen.pool_evpc_cs2.pc\endcsname{86}
\expandafter\def\csname v@h2.hb.qwen.pool_evpc_cs2.d.pc\endcsname{$-$2}
\expandafter\def\csname v@h2.hb.qwen.pool_evpc_cs4.pc\endcsname{82}
\expandafter\def\csname v@h2.hb.qwen.pool_evpc_cs4.d.pc\endcsname{$-$6}
\expandafter\def\csname v@h2.hb.qwen.pool_evpc_ch4.pc\endcsname{89}
\expandafter\def\csname v@h2.hb.qwen.pool_evpc_ch4.d.pc\endcsname{$+$1}
\expandafter\def\csname v@h2.hb.qwen.pool_evpc_cs4ch4.pc\endcsname{82}
\expandafter\def\csname v@h2.hb.qwen.pool_evpc_cs4ch4.d.pc\endcsname{$-$6}
\expandafter\def\csname v@h2.hb.qwen.pool_ev.pc\endcsname{85}
\expandafter\def\csname v@h2.hb.qwen.pool_ev.d.pc\endcsname{$-$3}
\expandafter\def\csname v@h2.hb.qwen.pool_ev_nodedup.pc\endcsname{70}
\expandafter\def\csname v@h2.hb.qwen.pool_ev_nodedup.d.pc\endcsname{$-$18}
\expandafter\def\csname v@h2.pzs.qwen.pool_evpc.pc\endcsname{87}
\expandafter\def\csname v@h2.pzs.qwen.pool_evpc_cs1.pc\endcsname{69}
\expandafter\def\csname v@h2.pzs.qwen.pool_evpc_cs1.d.pc\endcsname{$-$18}
\expandafter\def\csname v@h2.pzs.qwen.pool_evpc_cs2.pc\endcsname{53}
\expandafter\def\csname v@h2.pzs.qwen.pool_evpc_cs2.d.pc\endcsname{$-$34}
\expandafter\def\csname v@h2.pzs.qwen.pool_evpc_cs4.pc\endcsname{16}
\expandafter\def\csname v@h2.pzs.qwen.pool_evpc_cs4.d.pc\endcsname{$-$71}
\expandafter\def\csname v@h2.pzs.qwen.pool_evpc_ch4.pc\endcsname{88}
\expandafter\def\csname v@h2.pzs.qwen.pool_evpc_ch4.d.pc\endcsname{$+$1}
\expandafter\def\csname v@h2.pzs.qwen.pool_evpc_cs4ch4.pc\endcsname{73}
\expandafter\def\csname v@h2.pzs.qwen.pool_evpc_cs4ch4.d.pc\endcsname{$-$14}
\expandafter\def\csname v@h2.pzs.qwen.pool_ev.pc\endcsname{77}
\expandafter\def\csname v@h2.pzs.qwen.pool_ev.d.pc\endcsname{$-$10}
\expandafter\def\csname v@h2.pzs.qwen.pool_ev_nodedup.pc\endcsname{1}
\expandafter\def\csname v@h2.pzs.qwen.pool_ev_nodedup.d.pc\endcsname{$-$86}
\expandafter\def\csname v@h2.pzw.qwen.pool_evpc.pc\endcsname{84}
\expandafter\def\csname v@h2.pzw.qwen.pool_evpc_cs1.pc\endcsname{69}
\expandafter\def\csname v@h2.pzw.qwen.pool_evpc_cs1.d.pc\endcsname{$-$15}
\expandafter\def\csname v@h2.pzw.qwen.pool_evpc_cs2.pc\endcsname{62}
\expandafter\def\csname v@h2.pzw.qwen.pool_evpc_cs2.d.pc\endcsname{$-$21}
\expandafter\def\csname v@h2.pzw.qwen.pool_evpc_cs4.pc\endcsname{52}
\expandafter\def\csname v@h2.pzw.qwen.pool_evpc_cs4.d.pc\endcsname{$-$32}
\expandafter\def\csname v@h2.pzw.qwen.pool_evpc_ch4.pc\endcsname{90}
\expandafter\def\csname v@h2.pzw.qwen.pool_evpc_ch4.d.pc\endcsname{$+$7}
\expandafter\def\csname v@h2.pzw.qwen.pool_evpc_cs4ch4.pc\endcsname{71}
\expandafter\def\csname v@h2.pzw.qwen.pool_evpc_cs4ch4.d.pc\endcsname{$-$12}
\expandafter\def\csname v@h2.pzw.qwen.pool_ev.pc\endcsname{75}
\expandafter\def\csname v@h2.pzw.qwen.pool_ev.d.pc\endcsname{$-$9}
\expandafter\def\csname v@h2.pzw.qwen.pool_ev_nodedup.pc\endcsname{34}
\expandafter\def\csname v@h2.pzw.qwen.pool_ev_nodedup.d.pc\endcsname{$-$50}
\expandafter\def\csname v@h2.pz.qwen.pool_evpc.pc\endcsname{85}
\expandafter\def\csname v@h2.pz.qwen.pool_evpc_cs1.pc\endcsname{69}
\expandafter\def\csname v@h2.pz.qwen.pool_evpc_cs1.d.pc\endcsname{$-$17}
\expandafter\def\csname v@h2.pz.qwen.pool_evpc_cs2.pc\endcsname{58}
\expandafter\def\csname v@h2.pz.qwen.pool_evpc_cs2.d.pc\endcsname{$-$28}
\expandafter\def\csname v@h2.pz.qwen.pool_evpc_cs4.pc\endcsname{34}
\expandafter\def\csname v@h2.pz.qwen.pool_evpc_cs4.d.pc\endcsname{$-$51}
\expandafter\def\csname v@h2.pz.qwen.pool_evpc_ch4.pc\endcsname{89}
\expandafter\def\csname v@h2.pz.qwen.pool_evpc_ch4.d.pc\endcsname{$+$4}
\expandafter\def\csname v@h2.pz.qwen.pool_evpc_cs4ch4.pc\endcsname{72}
\expandafter\def\csname v@h2.pz.qwen.pool_evpc_cs4ch4.d.pc\endcsname{$-$13}
\expandafter\def\csname v@h2.pz.qwen.pool_ev.pc\endcsname{76}
\expandafter\def\csname v@h2.pz.qwen.pool_ev.d.pc\endcsname{$-$9}
\expandafter\def\csname v@h2.pz.qwen.pool_ev_nodedup.pc\endcsname{18}
\expandafter\def\csname v@h2.pz.qwen.pool_ev_nodedup.d.pc\endcsname{$-$68}
\expandafter\def\csname v@h2.hb.qwen.cs4.ci.pc\endcsname{$-$6 [$-$15, 1]}
\expandafter\def\csname v@h2.pz.qwen.cs4.ci.pc\endcsname{$-$51 [$-$59, $-$43]}
\expandafter\def\csname v@h2.hb.qwen.ch4.ci.pc\endcsname{$+$1 [$-$5, 7]}
\expandafter\def\csname v@h2.pz.qwen.ch4.ci.pc\endcsname{$+$4 [0, 7]}
\expandafter\def\csname v@h2.hb.qwen.offset.ci.pc\endcsname{0 [0, 0]}
\expandafter\def\csname v@h2.pz.qwen.offset.ci.pc\endcsname{$+$38 [30, 45]}
\expandafter\def\csname v@h2.hb.olmo.pool_evpc.pc\endcsname{88}
\expandafter\def\csname v@h2.hb.olmo.pool_evpc_cs1.pc\endcsname{89}
\expandafter\def\csname v@h2.hb.olmo.pool_evpc_cs1.d.pc\endcsname{$+$2}
\expandafter\def\csname v@h2.hb.olmo.pool_evpc_cs2.pc\endcsname{81}
\expandafter\def\csname v@h2.hb.olmo.pool_evpc_cs2.d.pc\endcsname{$-$6}
\expandafter\def\csname v@h2.hb.olmo.pool_evpc_cs4.pc\endcsname{73}
\expandafter\def\csname v@h2.hb.olmo.pool_evpc_cs4.d.pc\endcsname{$-$14}
\expandafter\def\csname v@h2.hb.olmo.pool_evpc_ch4.pc\endcsname{90}
\expandafter\def\csname v@h2.hb.olmo.pool_evpc_ch4.d.pc\endcsname{$+$2}
\expandafter\def\csname v@h2.hb.olmo.pool_evpc_cs4ch4.pc\endcsname{76}
\expandafter\def\csname v@h2.hb.olmo.pool_evpc_cs4ch4.d.pc\endcsname{$-$12}
\expandafter\def\csname v@h2.hb.olmo.pool_ev.pc\endcsname{86}
\expandafter\def\csname v@h2.hb.olmo.pool_ev.d.pc\endcsname{$-$2}
\expandafter\def\csname v@h2.hb.olmo.pool_ev_nodedup.pc\endcsname{54}
\expandafter\def\csname v@h2.hb.olmo.pool_ev_nodedup.d.pc\endcsname{$-$34}
\expandafter\def\csname v@h2.pzs.olmo.pool_evpc.pc\endcsname{76}
\expandafter\def\csname v@h2.pzs.olmo.pool_evpc_cs1.pc\endcsname{44}
\expandafter\def\csname v@h2.pzs.olmo.pool_evpc_cs1.d.pc\endcsname{$-$32}
\expandafter\def\csname v@h2.pzs.olmo.pool_evpc_cs2.pc\endcsname{17}
\expandafter\def\csname v@h2.pzs.olmo.pool_evpc_cs2.d.pc\endcsname{$-$59}
\expandafter\def\csname v@h2.pzs.olmo.pool_evpc_cs4.pc\endcsname{3}
\expandafter\def\csname v@h2.pzs.olmo.pool_evpc_cs4.d.pc\endcsname{$-$73}
\expandafter\def\csname v@h2.pzs.olmo.pool_evpc_ch4.pc\endcsname{81}
\expandafter\def\csname v@h2.pzs.olmo.pool_evpc_ch4.d.pc\endcsname{$+$5}
\expandafter\def\csname v@h2.pzs.olmo.pool_evpc_cs4ch4.pc\endcsname{55}
\expandafter\def\csname v@h2.pzs.olmo.pool_evpc_cs4ch4.d.pc\endcsname{$-$22}
\expandafter\def\csname v@h2.pzs.olmo.pool_ev.pc\endcsname{62}
\expandafter\def\csname v@h2.pzs.olmo.pool_ev.d.pc\endcsname{$-$14}
\expandafter\def\csname v@h2.pzs.olmo.pool_ev_nodedup.pc\endcsname{0}
\expandafter\def\csname v@h2.pzs.olmo.pool_ev_nodedup.d.pc\endcsname{$-$76}
\expandafter\def\csname v@h2.pzw.olmo.pool_evpc.pc\endcsname{74}
\expandafter\def\csname v@h2.pzw.olmo.pool_evpc_cs1.pc\endcsname{50}
\expandafter\def\csname v@h2.pzw.olmo.pool_evpc_cs1.d.pc\endcsname{$-$24}
\expandafter\def\csname v@h2.pzw.olmo.pool_evpc_cs2.pc\endcsname{37}
\expandafter\def\csname v@h2.pzw.olmo.pool_evpc_cs2.d.pc\endcsname{$-$37}
\expandafter\def\csname v@h2.pzw.olmo.pool_evpc_cs4.pc\endcsname{22}
\expandafter\def\csname v@h2.pzw.olmo.pool_evpc_cs4.d.pc\endcsname{$-$53}
\expandafter\def\csname v@h2.pzw.olmo.pool_evpc_ch4.pc\endcsname{73}
\expandafter\def\csname v@h2.pzw.olmo.pool_evpc_ch4.d.pc\endcsname{$-$2}
\expandafter\def\csname v@h2.pzw.olmo.pool_evpc_cs4ch4.pc\endcsname{57}
\expandafter\def\csname v@h2.pzw.olmo.pool_evpc_cs4ch4.d.pc\endcsname{$-$18}
\expandafter\def\csname v@h2.pzw.olmo.pool_ev.pc\endcsname{59}
\expandafter\def\csname v@h2.pzw.olmo.pool_ev.d.pc\endcsname{$-$15}
\expandafter\def\csname v@h2.pzw.olmo.pool_ev_nodedup.pc\endcsname{13}
\expandafter\def\csname v@h2.pzw.olmo.pool_ev_nodedup.d.pc\endcsname{$-$61}
\expandafter\def\csname v@h2.pz.olmo.pool_evpc.pc\endcsname{75}
\expandafter\def\csname v@h2.pz.olmo.pool_evpc_cs1.pc\endcsname{47}
\expandafter\def\csname v@h2.pz.olmo.pool_evpc_cs1.d.pc\endcsname{$-$28}
\expandafter\def\csname v@h2.pz.olmo.pool_evpc_cs2.pc\endcsname{27}
\expandafter\def\csname v@h2.pz.olmo.pool_evpc_cs2.d.pc\endcsname{$-$48}
\expandafter\def\csname v@h2.pz.olmo.pool_evpc_cs4.pc\endcsname{12}
\expandafter\def\csname v@h2.pz.olmo.pool_evpc_cs4.d.pc\endcsname{$-$63}
\expandafter\def\csname v@h2.pz.olmo.pool_evpc_ch4.pc\endcsname{77}
\expandafter\def\csname v@h2.pz.olmo.pool_evpc_ch4.d.pc\endcsname{$+$2}
\expandafter\def\csname v@h2.pz.olmo.pool_evpc_cs4ch4.pc\endcsname{56}
\expandafter\def\csname v@h2.pz.olmo.pool_evpc_cs4ch4.d.pc\endcsname{$-$20}
\expandafter\def\csname v@h2.pz.olmo.pool_ev.pc\endcsname{61}
\expandafter\def\csname v@h2.pz.olmo.pool_ev.d.pc\endcsname{$-$15}
\expandafter\def\csname v@h2.pz.olmo.pool_ev_nodedup.pc\endcsname{7}
\expandafter\def\csname v@h2.pz.olmo.pool_ev_nodedup.d.pc\endcsname{$-$68}
\expandafter\def\csname v@h2.hb.olmo.cs4.ci.pc\endcsname{$-$14 [$-$27, $-$2]}
\expandafter\def\csname v@h2.pz.olmo.cs4.ci.pc\endcsname{$-$63 [$-$71, $-$55]}
\expandafter\def\csname v@h2.hb.olmo.ch4.ci.pc\endcsname{$+$2 [$-$4, 9]}
\expandafter\def\csname v@h2.pz.olmo.ch4.ci.pc\endcsname{$+$2 [$-$3, 6]}
\expandafter\def\csname v@h2.hb.olmo.offset.ci.pc\endcsname{$+$2 [$-$5, 9]}
\expandafter\def\csname v@h2.pz.olmo.offset.ci.pc\endcsname{$+$43 [36, 51]}
\expandafter\def\csname v@h2.ch4.range.pc\endcsname{$+$1 to $+$4}
\expandafter\def\csname v@h2.offset.range.pc\endcsname{$+$38 to $+$43}
\expandafter\def\csname v@h3.qwen.pool.range.pc\endcsname{85--89}
\expandafter\def\csname v@h3.qwen.pool.4.pc\endcsname{89}
\expandafter\def\csname v@h3.qwen.nd.4.pc\endcsname{45}
\expandafter\def\csname v@h3.qwen.gap.4.pc\endcsname{$-$44 [$-$52, $-$36]}
\expandafter\def\csname v@h3.qwen.pool.8.pc\endcsname{87}
\expandafter\def\csname v@h3.qwen.nd.8.pc\endcsname{27}
\expandafter\def\csname v@h3.qwen.gap.8.pc\endcsname{$-$60 [$-$68, $-$52]}
\expandafter\def\csname v@h3.qwen.pool.16.pc\endcsname{85}
\expandafter\def\csname v@h3.qwen.nd.16.pc\endcsname{18}
\expandafter\def\csname v@h3.qwen.gap.16.pc\endcsname{$-$68 [$-$76, $-$60]}
\expandafter\def\csname v@h3.qwen.pool.32.pc\endcsname{88}
\expandafter\def\csname v@h3.qwen.nd.32.pc\endcsname{16}
\expandafter\def\csname v@h3.qwen.gap.32.pc\endcsname{$-$72 [$-$80, $-$65]}
\expandafter\def\csname v@h3.olmo.pool.range.pc\endcsname{75--78}
\expandafter\def\csname v@h3.olmo.pool.4.pc\endcsname{76}
\expandafter\def\csname v@h3.olmo.nd.4.pc\endcsname{20}
\expandafter\def\csname v@h3.olmo.gap.4.pc\endcsname{$-$56 [$-$63, $-$49]}
\expandafter\def\csname v@h3.olmo.pool.8.pc\endcsname{78}
\expandafter\def\csname v@h3.olmo.nd.8.pc\endcsname{11}
\expandafter\def\csname v@h3.olmo.gap.8.pc\endcsname{$-$66 [$-$74, $-$58]}
\expandafter\def\csname v@h3.olmo.pool.16.pc\endcsname{75}
\expandafter\def\csname v@h3.olmo.nd.16.pc\endcsname{7}
\expandafter\def\csname v@h3.olmo.gap.16.pc\endcsname{$-$68 [$-$76, $-$61]}
\expandafter\def\csname v@h3.olmo.pool.32.pc\endcsname{78}
\expandafter\def\csname v@h3.olmo.nd.32.pc\endcsname{5}
\expandafter\def\csname v@h3.olmo.gap.32.pc\endcsname{$-$74 [$-$81, $-$66]}
\expandafter\def\csname v@an.test.qwen.solo.pc\endcsname{80}
\expandafter\def\csname v@an.test.qwen.debate.pc\endcsname{87}
\expandafter\def\csname v@an.test.qwen.balanced.pc\endcsname{68}
\expandafter\def\csname v@an.test.qwen.pool_evpc.pc\endcsname{9}
\expandafter\def\csname v@an.test.qwen.full_iter.pc\endcsname{12}
\expandafter\def\csname v@an.pair.qwen.debate.pc\endcsname{$+$78 [65, 89]}
\expandafter\def\csname v@an.pair.qwen.full.pc\endcsname{$-$3 [$-$12, 5]}
\expandafter\def\csname v@an.pair.qwen.balanced.pc\endcsname{$+$59 [46, 72]}
\expandafter\def\csname v@an.test.olmo.solo.pc\endcsname{79}
\expandafter\def\csname v@an.test.olmo.debate.pc\endcsname{67}
\expandafter\def\csname v@an.test.olmo.balanced.pc\endcsname{32}
\expandafter\def\csname v@an.test.olmo.pool_evpc.pc\endcsname{10}
\expandafter\def\csname v@an.test.olmo.full_iter.pc\endcsname{10}
\expandafter\def\csname v@an.pair.olmo.debate.pc\endcsname{$+$57 [45, 68]}
\expandafter\def\csname v@an.pair.olmo.full.pc\endcsname{$+$1 [$-$7, 9]}
\expandafter\def\csname v@an.pair.olmo.balanced.pc\endcsname{$+$22 [10, 35]}
\expandafter\def\csname v@an.test.gemma.solo.pc\endcsname{82}
\expandafter\def\csname v@an.test.gemma.debate.pc\endcsname{60}
\expandafter\def\csname v@an.test.gemma.balanced.pc\endcsname{40}
\expandafter\def\csname v@an.test.gemma.pool_evpc.pc\endcsname{3}
\expandafter\def\csname v@an.test.gemma.full_iter.pc\endcsname{1}
\expandafter\def\csname v@an.test.mini.solo.pc\endcsname{70}
\expandafter\def\csname v@an.test.mini.debate.pc\endcsname{41}
\expandafter\def\csname v@an.test.mini.balanced.pc\endcsname{13}
\expandafter\def\csname v@an.test.mini.pool_evpc.pc\endcsname{4}
\expandafter\def\csname v@an.test.mini.full_iter.pc\endcsname{1}
\expandafter\def\csname v@an.hb.qwen.solo.pc\endcsname{74}
\expandafter\def\csname v@an.hb.qwen.debate.pc\endcsname{86}
\expandafter\def\csname v@an.hb.qwen.pool_evpc.pc\endcsname{7}
\expandafter\def\csname v@an.hb.qwen.pool_evpc_cs4.pc\endcsname{10}
\expandafter\def\csname v@an.hb.qwen.pool_evpc_raw.pc\endcsname{55}
\expandafter\def\csname v@an.hb.qwen.debate_evpc_raw.pc\endcsname{53}
\expandafter\def\csname v@an.hb.qwen.balanced_evpc_rand_raw.pc\endcsname{29}
\expandafter\def\csname v@an.hb.qwen.balanced_evpc_raw.pc\endcsname{25}
\expandafter\def\csname v@an.hb.olmo.solo.pc\endcsname{66}
\expandafter\def\csname v@an.hb.olmo.debate.pc\endcsname{54}
\expandafter\def\csname v@an.hb.olmo.pool_evpc.pc\endcsname{3}
\expandafter\def\csname v@an.hb.olmo.pool_evpc_cs4.pc\endcsname{11}
\expandafter\def\csname v@an.hb.olmo.pool_evpc_raw.pc\endcsname{34}
\expandafter\def\csname v@an.hb.olmo.debate_evpc_raw.pc\endcsname{34}
\expandafter\def\csname v@an.hb.olmo.balanced_evpc_rand_raw.pc\endcsname{18}
\expandafter\def\csname v@an.hb.olmo.balanced_evpc_raw.pc\endcsname{14}
\expandafter\def\csname v@an.hb.early.range.pc\endcsname{30--43}
\expandafter\def\csname v@red.hb.qwen.debate.0.pc\endcsname{84}
\expandafter\def\csname v@red.hb.qwen.debate.5.pc\endcsname{48}
\expandafter\def\csname v@red.hb.qwen.debate.drop.pc\endcsname{36}
\expandafter\def\csname v@red.hb.qwen.pool_evpc.0.pc\endcsname{84}
\expandafter\def\csname v@red.hb.qwen.pool_evpc.5.pc\endcsname{63}
\expandafter\def\csname v@red.hb.qwen.pool_evpc.drop.pc\endcsname{21}
\expandafter\def\csname v@red.hb.qwen.pool_evpc_raw.0.pc\endcsname{84}
\expandafter\def\csname v@red.hb.qwen.pool_evpc_raw.5.pc\endcsname{43}
\expandafter\def\csname v@red.hb.qwen.pool_evpc_raw.drop.pc\endcsname{41}
\expandafter\def\csname v@red.hb.qwen.debate_evpc_raw.0.pc\endcsname{84}
\expandafter\def\csname v@red.hb.qwen.debate_evpc_raw.5.pc\endcsname{49}
\expandafter\def\csname v@red.hb.qwen.debate_evpc_raw.drop.pc\endcsname{34}
\expandafter\def\csname v@red.hb.qwen.balanced_evpc_rand_raw.0.pc\endcsname{84}
\expandafter\def\csname v@red.hb.qwen.balanced_evpc_rand_raw.5.pc\endcsname{47}
\expandafter\def\csname v@red.hb.qwen.balanced_evpc_rand_raw.drop.pc\endcsname{37}
\expandafter\def\csname v@red.hb.qwen.balanced_evpc_raw.0.pc\endcsname{84}
\expandafter\def\csname v@red.hb.qwen.balanced_evpc_raw.5.pc\endcsname{54}
\expandafter\def\csname v@red.hb.qwen.balanced_evpc_raw.drop.pc\endcsname{30}
\expandafter\def\csname v@red.hb.olmo.debate.0.pc\endcsname{81}
\expandafter\def\csname v@red.hb.olmo.debate.5.pc\endcsname{37}
\expandafter\def\csname v@red.hb.olmo.debate.drop.pc\endcsname{44}
\expandafter\def\csname v@red.hb.olmo.pool_evpc.0.pc\endcsname{81}
\expandafter\def\csname v@red.hb.olmo.pool_evpc.5.pc\endcsname{55}
\expandafter\def\csname v@red.hb.olmo.pool_evpc.drop.pc\endcsname{26}
\expandafter\def\csname v@red.hb.olmo.pool_evpc_raw.0.pc\endcsname{72}
\expandafter\def\csname v@red.hb.olmo.pool_evpc_raw.5.pc\endcsname{32}
\expandafter\def\csname v@red.hb.olmo.pool_evpc_raw.drop.pc\endcsname{39}
\expandafter\def\csname v@red.hb.olmo.debate_evpc_raw.0.pc\endcsname{72}
\expandafter\def\csname v@red.hb.olmo.debate_evpc_raw.5.pc\endcsname{30}
\expandafter\def\csname v@red.hb.olmo.debate_evpc_raw.drop.pc\endcsname{42}
\expandafter\def\csname v@red.hb.olmo.balanced_evpc_rand_raw.0.pc\endcsname{72}
\expandafter\def\csname v@red.hb.olmo.balanced_evpc_rand_raw.5.pc\endcsname{33}
\expandafter\def\csname v@red.hb.olmo.balanced_evpc_rand_raw.drop.pc\endcsname{39}
\expandafter\def\csname v@red.hb.olmo.balanced_evpc_raw.0.pc\endcsname{72}
\expandafter\def\csname v@red.hb.olmo.balanced_evpc_raw.5.pc\endcsname{38}
\expandafter\def\csname v@red.hb.olmo.balanced_evpc_raw.drop.pc\endcsname{33}
\expandafter\def\csname v@an.pz.qwen.solo.pc\endcsname{64}
\expandafter\def\csname v@an.pz.qwen.debate.pc\endcsname{72}
\expandafter\def\csname v@an.pz.qwen.pool_evpc.pc\endcsname{5}
\expandafter\def\csname v@an.pz.qwen.pool_evpc_cs4.pc\endcsname{31}
\expandafter\def\csname v@an.pz.qwen.pool_evpc_raw.pc\endcsname{75}
\expandafter\def\csname v@an.pz.qwen.debate_evpc_raw.pc\endcsname{69}
\expandafter\def\csname v@an.pz.qwen.balanced_evpc_rand_raw.pc\endcsname{65}
\expandafter\def\csname v@an.pz.qwen.balanced_evpc_raw.pc\endcsname{60}
\expandafter\def\csname v@an.pz.olmo.solo.pc\endcsname{64}
\expandafter\def\csname v@an.pz.olmo.debate.pc\endcsname{84}
\expandafter\def\csname v@an.pz.olmo.pool_evpc.pc\endcsname{8}
\expandafter\def\csname v@an.pz.olmo.pool_evpc_cs4.pc\endcsname{49}
\expandafter\def\csname v@an.pz.olmo.pool_evpc_raw.pc\endcsname{82}
\expandafter\def\csname v@an.pz.olmo.debate_evpc_raw.pc\endcsname{77}
\expandafter\def\csname v@an.pz.olmo.balanced_evpc_rand_raw.pc\endcsname{44}
\expandafter\def\csname v@an.pz.olmo.balanced_evpc_raw.pc\endcsname{43}
\expandafter\def\csname v@an.pz.early.range.pc\endcsname{65--77}
\expandafter\def\csname v@red.pz.qwen.debate.0.pc\endcsname{60}
\expandafter\def\csname v@red.pz.qwen.debate.5.pc\endcsname{30}
\expandafter\def\csname v@red.pz.qwen.debate.drop.pc\endcsname{30}
\expandafter\def\csname v@red.pz.qwen.pool_evpc.0.pc\endcsname{60}
\expandafter\def\csname v@red.pz.qwen.pool_evpc.5.pc\endcsname{59}
\expandafter\def\csname v@red.pz.qwen.pool_evpc.drop.pc\endcsname{1}
\expandafter\def\csname v@red.pz.qwen.pool_evpc_raw.0.pc\endcsname{59}
\expandafter\def\csname v@red.pz.qwen.pool_evpc_raw.5.pc\endcsname{38}
\expandafter\def\csname v@red.pz.qwen.pool_evpc_raw.drop.pc\endcsname{22}
\expandafter\def\csname v@red.pz.qwen.debate_evpc_raw.0.pc\endcsname{59}
\expandafter\def\csname v@red.pz.qwen.debate_evpc_raw.5.pc\endcsname{40}
\expandafter\def\csname v@red.pz.qwen.debate_evpc_raw.drop.pc\endcsname{20}
\expandafter\def\csname v@red.pz.qwen.balanced_evpc_rand_raw.0.pc\endcsname{59}
\expandafter\def\csname v@red.pz.qwen.balanced_evpc_rand_raw.5.pc\endcsname{38}
\expandafter\def\csname v@red.pz.qwen.balanced_evpc_rand_raw.drop.pc\endcsname{21}
\expandafter\def\csname v@red.pz.qwen.balanced_evpc_raw.0.pc\endcsname{59}
\expandafter\def\csname v@red.pz.qwen.balanced_evpc_raw.5.pc\endcsname{42}
\expandafter\def\csname v@red.pz.qwen.balanced_evpc_raw.drop.pc\endcsname{17}
\expandafter\def\csname v@red.pz.olmo.debate.0.pc\endcsname{64}
\expandafter\def\csname v@red.pz.olmo.debate.5.pc\endcsname{11}
\expandafter\def\csname v@red.pz.olmo.debate.drop.pc\endcsname{53}
\expandafter\def\csname v@red.pz.olmo.pool_evpc.0.pc\endcsname{64}
\expandafter\def\csname v@red.pz.olmo.pool_evpc.5.pc\endcsname{39}
\expandafter\def\csname v@red.pz.olmo.pool_evpc.drop.pc\endcsname{24}
\expandafter\def\csname v@red.pz.olmo.pool_evpc_raw.0.pc\endcsname{64}
\expandafter\def\csname v@red.pz.olmo.pool_evpc_raw.5.pc\endcsname{15}
\expandafter\def\csname v@red.pz.olmo.pool_evpc_raw.drop.pc\endcsname{50}
\expandafter\def\csname v@red.pz.olmo.debate_evpc_raw.0.pc\endcsname{64}
\expandafter\def\csname v@red.pz.olmo.debate_evpc_raw.5.pc\endcsname{17}
\expandafter\def\csname v@red.pz.olmo.debate_evpc_raw.drop.pc\endcsname{48}
\expandafter\def\csname v@red.pz.olmo.balanced_evpc_rand_raw.0.pc\endcsname{64}
\expandafter\def\csname v@red.pz.olmo.balanced_evpc_rand_raw.5.pc\endcsname{17}
\expandafter\def\csname v@red.pz.olmo.balanced_evpc_rand_raw.drop.pc\endcsname{47}
\expandafter\def\csname v@red.pz.olmo.balanced_evpc_raw.0.pc\endcsname{64}
\expandafter\def\csname v@red.pz.olmo.balanced_evpc_raw.5.pc\endcsname{17}
\expandafter\def\csname v@red.pz.olmo.balanced_evpc_raw.drop.pc\endcsname{47}
\expandafter\def\csname v@gainchk.debate.fe.pc\endcsname{83--97}
\expandafter\def\csname v@gainchk.pool.fe.pc\endcsname{50--73}
\expandafter\def\csname v@gainchk.pool.po.pc\endcsname{77--97}
\expandafter\def\csname v@i.qwen.16.dev_mix_p2.solo.pc\endcsname{9}
\expandafter\def\csname v@i.qwen.16.dev_mix_p2.solo.D0.pc\endcsname{9}
\expandafter\def\csname v@i.olmo.16.dev_mix_p2.solo.pc\endcsname{18}
\expandafter\def\csname v@i.olmo.16.dev_mix_p2.solo.D0.pc\endcsname{18}
\expandafter\def\csname v@i.qwen.16.dev_mix_p2.debate.pc\endcsname{16}
\expandafter\def\csname v@i.qwen.16.dev_mix_p2.debate.D0.pc\endcsname{9}
\expandafter\def\csname v@i.olmo.16.dev_mix_p2.debate.pc\endcsname{32}
\expandafter\def\csname v@i.olmo.16.dev_mix_p2.debate.D0.pc\endcsname{18}
\expandafter\def\csname v@i.qwen.16.dev_mix_p2.debate_evpc.pc\endcsname{92}
\expandafter\def\csname v@i.qwen.16.dev_mix_p2.debate_evpc.D0.pc\endcsname{9}
\expandafter\def\csname v@i.olmo.16.dev_mix_p2.debate_evpc.pc\endcsname{80}
\expandafter\def\csname v@i.olmo.16.dev_mix_p2.debate_evpc.D0.pc\endcsname{18}
\expandafter\def\csname v@i.qwen.16.dev_mix_p2.pool_evpc.pc\endcsname{93}
\expandafter\def\csname v@i.qwen.16.dev_mix_p2.pool_evpc.D0.pc\endcsname{9}
\expandafter\def\csname v@i.olmo.16.dev_mix_p2.pool_evpc.pc\endcsname{94}
\expandafter\def\csname v@i.olmo.16.dev_mix_p2.pool_evpc.D0.pc\endcsname{18}
\expandafter\def\csname v@i.qwen.16.dev_mix_p2.full_iter.pc\endcsname{90}
\expandafter\def\csname v@i.qwen.16.dev_mix_p2.full_iter.D0.pc\endcsname{78}
\expandafter\def\csname v@i.olmo.16.dev_mix_p2.full_iter.pc\endcsname{89}
\expandafter\def\csname v@i.olmo.16.dev_mix_p2.full_iter.D0.pc\endcsname{70}
\expandafter\def\csname v@i.qwen.16.dev_swap_p2.pool_evpc.pc\endcsname{94}
\expandafter\def\csname v@i.qwen.16.dev_swap_p2.pool_evpc.D0.pc\endcsname{96}
\expandafter\def\csname v@i.olmo.16.dev_swap_p2.pool_evpc.pc\endcsname{84}
\expandafter\def\csname v@i.olmo.16.dev_swap_p2.pool_evpc.D0.pc\endcsname{97}
\expandafter\def\csname v@i.qwen.16.dev_swap_p2.pool_evpc_cs4.pc\endcsname{96}
\expandafter\def\csname v@i.qwen.16.dev_swap_p2.pool_evpc_cs4.D0.pc\endcsname{96}
\expandafter\def\csname v@i.olmo.16.dev_swap_p2.pool_evpc_cs4.pc\endcsname{95}
\expandafter\def\csname v@i.olmo.16.dev_swap_p2.pool_evpc_cs4.D0.pc\endcsname{97}
\expandafter\def\csname v@i.qwen.16.dev_swap_p2.pool_evpc_ch4.pc\endcsname{94}
\expandafter\def\csname v@i.qwen.16.dev_swap_p2.pool_evpc_ch4.D0.pc\endcsname{96}
\expandafter\def\csname v@i.olmo.16.dev_swap_p2.pool_evpc_ch4.pc\endcsname{73}
\expandafter\def\csname v@i.olmo.16.dev_swap_p2.pool_evpc_ch4.D0.pc\endcsname{97}
\expandafter\def\csname v@i.qwen.16.dev_replicate_p2.pool_evpc.pc\endcsname{96}
\expandafter\def\csname v@i.qwen.16.dev_replicate_p2.pool_evpc.D0.pc\endcsname{11}
\expandafter\def\csname v@i.olmo.16.dev_replicate_p2.pool_evpc.pc\endcsname{95}
\expandafter\def\csname v@i.olmo.16.dev_replicate_p2.pool_evpc.D0.pc\endcsname{16}
\expandafter\def\csname v@i.qwen.16.dev_replicate_p2.pool_ev.pc\endcsname{94}
\expandafter\def\csname v@i.qwen.16.dev_replicate_p2.pool_ev.D0.pc\endcsname{11}
\expandafter\def\csname v@i.olmo.16.dev_replicate_p2.pool_ev.pc\endcsname{95}
\expandafter\def\csname v@i.olmo.16.dev_replicate_p2.pool_ev.D0.pc\endcsname{16}
\expandafter\def\csname v@i.qwen.32.dev_replicate_p2.pool_evpc.pc\endcsname{96}
\expandafter\def\csname v@i.qwen.32.dev_replicate_p2.pool_evpc.D0.pc\endcsname{10}
\expandafter\def\csname v@i.olmo.32.dev_replicate_p2.pool_evpc.pc\endcsname{98}
\expandafter\def\csname v@i.olmo.32.dev_replicate_p2.pool_evpc.D0.pc\endcsname{16}
\expandafter\def\csname v@i.qwen.32.dev_replicate_p2.pool_ev.pc\endcsname{95}
\expandafter\def\csname v@i.qwen.32.dev_replicate_p2.pool_ev.D0.pc\endcsname{10}
\expandafter\def\csname v@i.olmo.32.dev_replicate_p2.pool_ev.pc\endcsname{94}
\expandafter\def\csname v@i.olmo.32.dev_replicate_p2.pool_ev.D0.pc\endcsname{16}
\expandafter\def\csname v@i.qwen.4.dev_p2_w300.debate.pc\endcsname{53}
\expandafter\def\csname v@i.qwen.4.dev_p2_w300.debate.D0.pc\endcsname{13}
\expandafter\def\csname v@i.olmo.4.dev_p2_w300.debate.pc\endcsname{40}
\expandafter\def\csname v@i.olmo.4.dev_p2_w300.debate.D0.pc\endcsname{16}
\expandafter\def\csname v@i.qwen.8.dev_p2_w300.debate.pc\endcsname{26}
\expandafter\def\csname v@i.qwen.8.dev_p2_w300.debate.D0.pc\endcsname{4}
\expandafter\def\csname v@i.olmo.8.dev_p2_w300.debate.pc\endcsname{26}
\expandafter\def\csname v@i.olmo.8.dev_p2_w300.debate.D0.pc\endcsname{7}
\expandafter\def\csname v@i.qwen.16.dev_p2_w300.debate.pc\endcsname{19}
\expandafter\def\csname v@i.qwen.16.dev_p2_w300.debate.D0.pc\endcsname{1}
\expandafter\def\csname v@i.olmo.16.dev_p2_w300.debate.pc\endcsname{23}
\expandafter\def\csname v@i.olmo.16.dev_p2_w300.debate.D0.pc\endcsname{4}
\expandafter\def\csname v@i.qwen.32.dev_p2_w300.debate.pc\endcsname{19}
\expandafter\def\csname v@i.qwen.32.dev_p2_w300.debate.D0.pc\endcsname{1}
\expandafter\def\csname v@i.olmo.32.dev_p2_w300.debate.pc\endcsname{11}
\expandafter\def\csname v@i.olmo.32.dev_p2_w300.debate.D0.pc\endcsname{4}
\expandafter\def\csname v@i.qwen.8.dev_p2_w300.pool_evpc.pc\endcsname{96}
\expandafter\def\csname v@i.qwen.8.dev_p2_w300.pool_evpc.D0.pc\endcsname{4}
\expandafter\def\csname v@i.olmo.8.dev_p2_w300.pool_evpc.pc\endcsname{93}
\expandafter\def\csname v@i.olmo.8.dev_p2_w300.pool_evpc.D0.pc\endcsname{7}
\expandafter\def\csname v@i.qwen.32.dev_p2_w300.pool_evpc.pc\endcsname{96}
\expandafter\def\csname v@i.qwen.32.dev_p2_w300.pool_evpc.D0.pc\endcsname{1}
\expandafter\def\csname v@i.olmo.32.dev_p2_w300.pool_evpc.pc\endcsname{95}
\expandafter\def\csname v@i.olmo.32.dev_p2_w300.pool_evpc.D0.pc\endcsname{4}
\expandafter\def\csname v@i.qwen.16.dev_mix_p2.pool_evpc-full_iter.pc\endcsname{$+$3 [$-$5, 14]}
\expandafter\def\csname v@i.qwen.16.dev_mix_p2.pool_evpc-full_iter.pt.pc\endcsname{$+$3}
\expandafter\def\csname v@i.qwen.16.dev_mix_p2.pool_evpc-debate.pc\endcsname{$+$78 [64, 90]}
\expandafter\def\csname v@i.qwen.16.dev_mix_p2.pool_evpc-debate.pt.pc\endcsname{$+$78}
\expandafter\def\csname v@i.qwen.16.dev_mix_p2.pool_evpc-debate_evpc.pc\endcsname{$+$1 [0, 3]}
\expandafter\def\csname v@i.qwen.16.dev_mix_p2.pool_evpc-debate_evpc.pt.pc\endcsname{$+$1}
\expandafter\def\csname v@i.qwen.16.dev_mix_p2.debate-solo.pc\endcsname{$+$7 [2, 11]}
\expandafter\def\csname v@i.qwen.16.dev_mix_p2.debate-solo.pt.pc\endcsname{$+$7}
\expandafter\def\csname v@i.olmo.16.dev_mix_p2.pool_evpc-full_iter.pc\endcsname{$+$5 [$-$5, 15]}
\expandafter\def\csname v@i.olmo.16.dev_mix_p2.pool_evpc-full_iter.pt.pc\endcsname{$+$5}
\expandafter\def\csname v@i.olmo.16.dev_mix_p2.pool_evpc-debate.pc\endcsname{$+$62 [43, 79]}
\expandafter\def\csname v@i.olmo.16.dev_mix_p2.pool_evpc-debate.pt.pc\endcsname{$+$62}
\expandafter\def\csname v@i.olmo.16.dev_mix_p2.pool_evpc-debate_evpc.pc\endcsname{$+$14 [6, 22]}
\expandafter\def\csname v@i.olmo.16.dev_mix_p2.pool_evpc-debate_evpc.pt.pc\endcsname{$+$14}
\expandafter\def\csname v@i.olmo.16.dev_mix_p2.debate-solo.pc\endcsname{$+$14 [2, 27]}
\expandafter\def\csname v@i.olmo.16.dev_mix_p2.debate-solo.pt.pc\endcsname{$+$14}
\expandafter\def\csname v@i.qwen.16.dev_swap_p2.pool_evpc_cs4-pool_evpc.pc\endcsname{$+$2 [0, 5]}
\expandafter\def\csname v@i.qwen.16.dev_swap_p2.pool_evpc_cs4-pool_evpc.pt.pc\endcsname{$+$2}
\expandafter\def\csname v@i.qwen.16.dev_swap_p2.pool_evpc_ch4-pool_evpc.pc\endcsname{0 [$-$4, 5]}
\expandafter\def\csname v@i.qwen.16.dev_swap_p2.pool_evpc_ch4-pool_evpc.pt.pc\endcsname{0}
\expandafter\def\csname v@i.qwen.16.dev_swap_p2.pool_evpc_ch4-pool_evpc_cs4.pc\endcsname{$-$2 [$-$5, 1]}
\expandafter\def\csname v@i.qwen.16.dev_swap_p2.pool_evpc_ch4-pool_evpc_cs4.pt.pc\endcsname{$-$2}
\expandafter\def\csname v@i.olmo.16.dev_swap_p2.pool_evpc_cs4-pool_evpc.pc\endcsname{$+$11 [0, 22]}
\expandafter\def\csname v@i.olmo.16.dev_swap_p2.pool_evpc_cs4-pool_evpc.pt.pc\endcsname{$+$11}
\expandafter\def\csname v@i.olmo.16.dev_swap_p2.pool_evpc_ch4-pool_evpc.pc\endcsname{$-$11 [$-$22, $-$2]}
\expandafter\def\csname v@i.olmo.16.dev_swap_p2.pool_evpc_ch4-pool_evpc.pt.pc\endcsname{$-$11}
\expandafter\def\csname v@i.olmo.16.dev_swap_p2.pool_evpc_ch4-pool_evpc_cs4.pc\endcsname{$-$22 [$-$36, $-$10]}
\expandafter\def\csname v@i.olmo.16.dev_swap_p2.pool_evpc_ch4-pool_evpc_cs4.pt.pc\endcsname{$-$22}
\expandafter\def\csname v@i.qwen.16.dev_p2.pool_evpc_cs4-pool_evpc.pc\endcsname{$-$6 [$-$15, 1]}
\expandafter\def\csname v@i.qwen.16.dev_p2.pool_evpc_cs4-pool_evpc.pt.pc\endcsname{$-$6}
\expandafter\def\csname v@i.qwen.16.dev_p2.pool_evpc_ch4-pool_evpc.pc\endcsname{$+$1 [$-$5, 7]}
\expandafter\def\csname v@i.qwen.16.dev_p2.pool_evpc_ch4-pool_evpc.pt.pc\endcsname{$+$1}
\expandafter\def\csname v@i.olmo.16.dev_p2.pool_evpc_cs4-pool_evpc.pc\endcsname{$-$14 [$-$27, $-$2]}
\expandafter\def\csname v@i.olmo.16.dev_p2.pool_evpc_cs4-pool_evpc.pt.pc\endcsname{$-$14}
\expandafter\def\csname v@i.olmo.16.dev_p2.pool_evpc_ch4-pool_evpc.pc\endcsname{$+$2 [$-$4, 9]}
\expandafter\def\csname v@i.olmo.16.dev_p2.pool_evpc_ch4-pool_evpc.pt.pc\endcsname{$+$2}
\expandafter\def\csname v@i.qwen.16.dev_replicate_p2.pool_ev-pool_evpc.pc\endcsname{$-$2 [$-$6, 0]}
\expandafter\def\csname v@i.qwen.16.dev_replicate_p2.pool_ev-pool_evpc.pt.pc\endcsname{$-$2}
\expandafter\def\csname v@i.qwen.16.dev_replicate_p2.pool_ev_nodedup-pool_evpc.pc\endcsname{$-$19 [$-$32, $-$8]}
\expandafter\def\csname v@i.qwen.16.dev_replicate_p2.pool_ev_nodedup-pool_evpc.pt.pc\endcsname{$-$19}
\expandafter\def\csname v@i.olmo.16.dev_replicate_p2.pool_ev-pool_evpc.pc\endcsname{$+$1 [$-$3, 6]}
\expandafter\def\csname v@i.olmo.16.dev_replicate_p2.pool_ev-pool_evpc.pt.pc\endcsname{$+$1}
\expandafter\def\csname v@i.olmo.16.dev_replicate_p2.pool_ev_nodedup-pool_evpc.pc\endcsname{$-$16 [$-$31, $-$1]}
\expandafter\def\csname v@i.olmo.16.dev_replicate_p2.pool_ev_nodedup-pool_evpc.pt.pc\endcsname{$-$16}
\expandafter\def\csname v@i.qwen.32.dev_replicate_p2.pool_ev-pool_evpc.pc\endcsname{$-$1 [$-$3, 0]}
\expandafter\def\csname v@i.qwen.32.dev_replicate_p2.pool_ev-pool_evpc.pt.pc\endcsname{$-$1}
\expandafter\def\csname v@i.qwen.32.dev_replicate_p2.pool_ev_nodedup-pool_evpc.pc\endcsname{$-$27 [$-$45, $-$11]}
\expandafter\def\csname v@i.qwen.32.dev_replicate_p2.pool_ev_nodedup-pool_evpc.pt.pc\endcsname{$-$27}
\expandafter\def\csname v@i.olmo.32.dev_replicate_p2.pool_ev-pool_evpc.pc\endcsname{$-$4 [$-$11, 1]}
\expandafter\def\csname v@i.olmo.32.dev_replicate_p2.pool_ev-pool_evpc.pt.pc\endcsname{$-$4}
\expandafter\def\csname v@i.olmo.32.dev_replicate_p2.pool_ev_nodedup-pool_evpc.pc\endcsname{$-$24 [$-$41, $-$9]}
\expandafter\def\csname v@i.olmo.32.dev_replicate_p2.pool_ev_nodedup-pool_evpc.pt.pc\endcsname{$-$24}
\expandafter\def\csname v@i.qwen.16.dev_p2.pool_ev-pool_evpc.pc\endcsname{$-$3 [$-$11, 3]}
\expandafter\def\csname v@i.qwen.16.dev_p2.pool_ev-pool_evpc.pt.pc\endcsname{$-$3}
\expandafter\def\csname v@i.qwen.16.dev_p2.pool_ev_nodedup-pool_evpc.pc\endcsname{$-$18 [$-$30, $-$7]}
\expandafter\def\csname v@i.qwen.16.dev_p2.pool_ev_nodedup-pool_evpc.pt.pc\endcsname{$-$18}
\expandafter\def\csname v@i.olmo.16.dev_p2.pool_ev-pool_evpc.pc\endcsname{$-$2 [$-$7, 5]}
\expandafter\def\csname v@i.olmo.16.dev_p2.pool_ev-pool_evpc.pt.pc\endcsname{$-$2}
\expandafter\def\csname v@i.olmo.16.dev_p2.pool_ev_nodedup-pool_evpc.pc\endcsname{$-$34 [$-$49, $-$19]}
\expandafter\def\csname v@i.olmo.16.dev_p2.pool_ev_nodedup-pool_evpc.pt.pc\endcsname{$-$34}
\expandafter\def\csname v@i.qwen.32.dev_p2.pool_ev-pool_evpc.pc\endcsname{$-$5 [$-$12, 0]}
\expandafter\def\csname v@i.qwen.32.dev_p2.pool_ev-pool_evpc.pt.pc\endcsname{$-$5}
\expandafter\def\csname v@i.qwen.32.dev_p2.pool_ev_nodedup-pool_evpc.pc\endcsname{$-$31 [$-$45, $-$18]}
\expandafter\def\csname v@i.qwen.32.dev_p2.pool_ev_nodedup-pool_evpc.pt.pc\endcsname{$-$31}
\expandafter\def\csname v@i.olmo.32.dev_p2.pool_ev-pool_evpc.pc\endcsname{$-$4 [$-$10, 1]}
\expandafter\def\csname v@i.olmo.32.dev_p2.pool_ev-pool_evpc.pt.pc\endcsname{$-$4}
\expandafter\def\csname v@i.olmo.32.dev_p2.pool_ev_nodedup-pool_evpc.pc\endcsname{$-$36 [$-$55, $-$17]}
\expandafter\def\csname v@i.olmo.32.dev_p2.pool_ev_nodedup-pool_evpc.pt.pc\endcsname{$-$36}
\expandafter\def\csname v@i.qwen.8.dev_p2_w300.pool_evpc-debate.pc\endcsname{$+$70 [54, 83]}
\expandafter\def\csname v@i.qwen.8.dev_p2_w300.pool_evpc-debate.pt.pc\endcsname{$+$70}
\expandafter\def\csname v@i.olmo.8.dev_p2_w300.pool_evpc-debate.pc\endcsname{$+$68 [56, 79]}
\expandafter\def\csname v@i.olmo.8.dev_p2_w300.pool_evpc-debate.pt.pc\endcsname{$+$68}
\expandafter\def\csname v@i.qwen.32.dev_p2_w300.pool_evpc-debate.pc\endcsname{$+$77 [62, 90]}
\expandafter\def\csname v@i.qwen.32.dev_p2_w300.pool_evpc-debate.pt.pc\endcsname{$+$77}
\expandafter\def\csname v@i.olmo.32.dev_p2_w300.pool_evpc-debate.pc\endcsname{$+$83 [68, 96]}
\expandafter\def\csname v@i.olmo.32.dev_p2_w300.pool_evpc-debate.pt.pc\endcsname{$+$83}
\expandafter\def\csname v@i.qwen.cross.I1.solo.16.pc\endcsname{$+$7 [0, 18]}
\expandafter\def\csname v@i.qwen.cross.I1.pool_evpc.16.pc\endcsname{$+$5 [$-$2, 14]}
\expandafter\def\csname v@i.qwen.cross.I4.debate.32.pc\endcsname{$+$14 [1, 28]}
\expandafter\def\csname v@i.olmo.cross.I1.solo.16.pc\endcsname{$+$7 [0, 18]}
\expandafter\def\csname v@i.olmo.cross.I1.pool_evpc.16.pc\endcsname{$+$6 [1, 12]}
\expandafter\def\csname v@i.olmo.cross.I4.debate.32.pc\endcsname{$-$9 [$-$24, 6]}
\expandafter\def\csname v@gsim.fe.bias.range.pc\endcsname{$-$5 to $-$4}
\expandafter\def\csname v@gsim.po.low.range.pc\endcsname{87--88}
\expandafter\def\csname v@gsim.0.2.fe.pc\endcsname{15}
\expandafter\def\csname v@gsim.0.2.po.pc\endcsname{87}
\expandafter\def\csname v@gsim.0.4.fe.pc\endcsname{36}
\expandafter\def\csname v@gsim.0.4.po.pc\endcsname{88}
\expandafter\def\csname v@gsim.0.6.fe.pc\endcsname{56}
\expandafter\def\csname v@gsim.0.6.po.pc\endcsname{87}
\expandafter\def\csname v@gsim.0.8.fe.pc\endcsname{75}
\expandafter\def\csname v@gsim.0.8.po.pc\endcsname{87}
\expandafter\def\csname v@gsim.1.0.fe.pc\endcsname{95}
\expandafter\def\csname v@gsim.1.0.po.pc\endcsname{100}
\expandafter\def\csname v@gsim.all.low.range.pc\endcsname{0--6}
\expandafter\def\csname v@gsim.floor.bias.range.pc\endcsname{$-$5 to $-$1}
\expandafter\def\csname v@gsim.floor.width.range.pc\endcsname{15--39}
\expandafter\def\csname v@gsim.small.bias.range.pc\endcsname{$-$6 to 0}
\expandafter\def\csname v@gsim.small.width.range.pc\endcsname{27--74}
\expandafter\def\csname v@dwell.test.qwen.debate.pc\endcsname{54}
\expandafter\def\csname v@dwell.test.qwen.debate.ci.pc\endcsname{54 [49, 59]}
\expandafter\def\csname v@dwell.test.qwen.debate.rev.pc\endcsname{0}
\expandafter\def\csname v@dwell.test.qwen.debate.rev.ci.pc\endcsname{0 [0, 0]}
\expandafter\def\csname v@dwell.test.qwen.debate.msw.pc\endcsname{21}
\expandafter\def\csname v@dwell.test.qwen.debate.msw.ci.pc\endcsname{21 [15, 27]}
\expandafter\def\csname v@dwell.test.qwen.debate.mst.pc\endcsname{88}
\expandafter\def\csname v@dwell.test.qwen.debate.mst.ci.pc\endcsname{88 [84, 92]}
\expandafter\def\csname v@dwell.test.qwen.debate.sw.pc\endcsname{27}
\expandafter\def\csname v@dwell.test.qwen.debate.sw.ci.pc\endcsname{27 [16, 39]}
\expandafter\def\csname v@dwell.test.qwen.balanced.pc\endcsname{62}
\expandafter\def\csname v@dwell.test.qwen.balanced.ci.pc\endcsname{62 [56, 67]}
\expandafter\def\csname v@dwell.test.qwen.balanced.rev.pc\endcsname{5}
\expandafter\def\csname v@dwell.test.qwen.balanced.rev.ci.pc\endcsname{5 [0, 10]}
\expandafter\def\csname v@dwell.test.qwen.balanced.msw.pc\endcsname{28}
\expandafter\def\csname v@dwell.test.qwen.balanced.msw.ci.pc\endcsname{28 [20, 37]}
\expandafter\def\csname v@dwell.test.qwen.balanced.mst.pc\endcsname{76}
\expandafter\def\csname v@dwell.test.qwen.balanced.mst.ci.pc\endcsname{76 [69, 82]}
\expandafter\def\csname v@dwell.test.qwen.balanced.sw.pc\endcsname{41}
\expandafter\def\csname v@dwell.test.qwen.balanced.sw.ci.pc\endcsname{41 [26, 57]}
\expandafter\def\csname v@dwell.test.qwen.pool_evpc.pc\endcsname{69}
\expandafter\def\csname v@dwell.test.qwen.pool_evpc.ci.pc\endcsname{69 [59, 79]}
\expandafter\def\csname v@dwell.test.qwen.pool_evpc.rev.pc\endcsname{17}
\expandafter\def\csname v@dwell.test.qwen.pool_evpc.rev.ci.pc\endcsname{17 [0, 29]}
\expandafter\def\csname v@dwell.test.qwen.pool_evpc.msw.pc\endcsname{23}
\expandafter\def\csname v@dwell.test.qwen.pool_evpc.msw.ci.pc\endcsname{23 [12, 36]}
\expandafter\def\csname v@dwell.test.qwen.pool_evpc.mst.pc\endcsname{72}
\expandafter\def\csname v@dwell.test.qwen.pool_evpc.mst.ci.pc\endcsname{72 [64, 80]}
\expandafter\def\csname v@dwell.test.qwen.pool_evpc.sw.pc\endcsname{26}
\expandafter\def\csname v@dwell.test.qwen.pool_evpc.sw.ci.pc\endcsname{26 [12, 42]}
\expandafter\def\csname v@dwell.dev.qwen.debate.pc\endcsname{52}
\expandafter\def\csname v@dwell.dev.qwen.debate.ci.pc\endcsname{52 [42, 61]}
\expandafter\def\csname v@dwell.dev.qwen.debate.rev.pc\endcsname{8}
\expandafter\def\csname v@dwell.dev.qwen.debate.rev.ci.pc\endcsname{8 [0, 16]}
\expandafter\def\csname v@dwell.dev.qwen.debate.msw.pc\endcsname{20}
\expandafter\def\csname v@dwell.dev.qwen.debate.msw.ci.pc\endcsname{20 [13, 26]}
\expandafter\def\csname v@dwell.dev.qwen.debate.mst.pc\endcsname{85}
\expandafter\def\csname v@dwell.dev.qwen.debate.mst.ci.pc\endcsname{85 [78, 90]}
\expandafter\def\csname v@dwell.dev.qwen.debate.sw.pc\endcsname{28}
\expandafter\def\csname v@dwell.dev.qwen.debate.sw.ci.pc\endcsname{28 [15, 43]}
\expandafter\def\csname v@dwell.dev.qwen.balanced.pc\endcsname{56}
\expandafter\def\csname v@dwell.dev.qwen.balanced.ci.pc\endcsname{56 [52, 62]}
\expandafter\def\csname v@dwell.dev.qwen.balanced.rev.pc\endcsname{5}
\expandafter\def\csname v@dwell.dev.qwen.balanced.rev.ci.pc\endcsname{5 [0, 12]}
\expandafter\def\csname v@dwell.dev.qwen.balanced.msw.pc\endcsname{21}
\expandafter\def\csname v@dwell.dev.qwen.balanced.msw.ci.pc\endcsname{21 [15, 28]}
\expandafter\def\csname v@dwell.dev.qwen.balanced.mst.pc\endcsname{68}
\expandafter\def\csname v@dwell.dev.qwen.balanced.mst.ci.pc\endcsname{68 [59, 76]}
\expandafter\def\csname v@dwell.dev.qwen.balanced.sw.pc\endcsname{69}
\expandafter\def\csname v@dwell.dev.qwen.balanced.sw.ci.pc\endcsname{69 [50, 88]}
\expandafter\def\csname v@dwell.dev.qwen.pool_evpc.pc\endcsname{41}
\expandafter\def\csname v@dwell.dev.qwen.pool_evpc.ci.pc\endcsname{41 [0, 46]}
\expandafter\def\csname v@dwell.dev.qwen.pool_evpc.rev.pc\endcsname{0}
\expandafter\def\csname v@dwell.dev.qwen.pool_evpc.rev.ci.pc\endcsname{0 [0, 0]}
\expandafter\def\csname v@dwell.dev.qwen.pool_evpc.msw.pc\endcsname{30}
\expandafter\def\csname v@dwell.dev.qwen.pool_evpc.msw.ci.pc\endcsname{30 [0, 78]}
\expandafter\def\csname v@dwell.dev.qwen.pool_evpc.mst.pc\endcsname{81}
\expandafter\def\csname v@dwell.dev.qwen.pool_evpc.mst.ci.pc\endcsname{81 [74, 87]}
\expandafter\def\csname v@dwell.dev.qwen.pool_evpc.sw.pc\endcsname{4}
\expandafter\def\csname v@dwell.dev.qwen.pool_evpc.sw.ci.pc\endcsname{4 [0, 9]}
\expandafter\def\csname v@dwell.dev.qwen.pool_evpc_raw.pc\endcsname{54}
\expandafter\def\csname v@dwell.dev.qwen.pool_evpc_raw.ci.pc\endcsname{54 [49, 59]}
\expandafter\def\csname v@dwell.dev.qwen.pool_evpc_raw.rev.pc\endcsname{0}
\expandafter\def\csname v@dwell.dev.qwen.pool_evpc_raw.rev.ci.pc\endcsname{0 [0, 0]}
\expandafter\def\csname v@dwell.dev.qwen.pool_evpc_raw.msw.pc\endcsname{21}
\expandafter\def\csname v@dwell.dev.qwen.pool_evpc_raw.msw.ci.pc\endcsname{21 [15, 28]}
\expandafter\def\csname v@dwell.dev.qwen.pool_evpc_raw.mst.pc\endcsname{84}
\expandafter\def\csname v@dwell.dev.qwen.pool_evpc_raw.mst.ci.pc\endcsname{84 [78, 89]}
\expandafter\def\csname v@dwell.dev.qwen.pool_evpc_raw.sw.pc\endcsname{39}
\expandafter\def\csname v@dwell.dev.qwen.pool_evpc_raw.sw.ci.pc\endcsname{39 [22, 58]}
\expandafter\def\csname v@dwell.dev.qwen.debate_evpc_raw.pc\endcsname{59}
\expandafter\def\csname v@dwell.dev.qwen.debate_evpc_raw.ci.pc\endcsname{59 [54, 65]}
\expandafter\def\csname v@dwell.dev.qwen.debate_evpc_raw.rev.pc\endcsname{6}
\expandafter\def\csname v@dwell.dev.qwen.debate_evpc_raw.rev.ci.pc\endcsname{6 [0, 14]}
\expandafter\def\csname v@dwell.dev.qwen.debate_evpc_raw.msw.pc\endcsname{20}
\expandafter\def\csname v@dwell.dev.qwen.debate_evpc_raw.msw.ci.pc\endcsname{20 [16, 24]}
\expandafter\def\csname v@dwell.dev.qwen.debate_evpc_raw.mst.pc\endcsname{76}
\expandafter\def\csname v@dwell.dev.qwen.debate_evpc_raw.mst.ci.pc\endcsname{76 [71, 82]}
\expandafter\def\csname v@dwell.dev.qwen.debate_evpc_raw.sw.pc\endcsname{75}
\expandafter\def\csname v@dwell.dev.qwen.debate_evpc_raw.sw.ci.pc\endcsname{75 [55, 97]}
\expandafter\def\csname v@dwell.dev.qwen.balanced_evpc_rand_raw.pc\endcsname{61}
\expandafter\def\csname v@dwell.dev.qwen.balanced_evpc_rand_raw.ci.pc\endcsname{61 [57, 66]}
\expandafter\def\csname v@dwell.dev.qwen.balanced_evpc_rand_raw.rev.pc\endcsname{1}
\expandafter\def\csname v@dwell.dev.qwen.balanced_evpc_rand_raw.rev.ci.pc\endcsname{1 [0, 5]}
\expandafter\def\csname v@dwell.dev.qwen.balanced_evpc_rand_raw.msw.pc\endcsname{28}
\expandafter\def\csname v@dwell.dev.qwen.balanced_evpc_rand_raw.msw.ci.pc\endcsname{28 [20, 37]}
\expandafter\def\csname v@dwell.dev.qwen.balanced_evpc_rand_raw.mst.pc\endcsname{75}
\expandafter\def\csname v@dwell.dev.qwen.balanced_evpc_rand_raw.mst.ci.pc\endcsname{75 [69, 81]}
\expandafter\def\csname v@dwell.dev.qwen.balanced_evpc_rand_raw.sw.pc\endcsname{75}
\expandafter\def\csname v@dwell.dev.qwen.balanced_evpc_rand_raw.sw.ci.pc\endcsname{75 [51, 100]}
\expandafter\def\csname v@dwell.dev.qwen.balanced_evpc_raw.pc\endcsname{60}
\expandafter\def\csname v@dwell.dev.qwen.balanced_evpc_raw.ci.pc\endcsname{60 [55, 66]}
\expandafter\def\csname v@dwell.dev.qwen.balanced_evpc_raw.rev.pc\endcsname{5}
\expandafter\def\csname v@dwell.dev.qwen.balanced_evpc_raw.rev.ci.pc\endcsname{5 [0, 11]}
\expandafter\def\csname v@dwell.dev.qwen.balanced_evpc_raw.msw.pc\endcsname{22}
\expandafter\def\csname v@dwell.dev.qwen.balanced_evpc_raw.msw.ci.pc\endcsname{22 [18, 26]}
\expandafter\def\csname v@dwell.dev.qwen.balanced_evpc_raw.mst.pc\endcsname{78}
\expandafter\def\csname v@dwell.dev.qwen.balanced_evpc_raw.mst.ci.pc\endcsname{78 [72, 83]}
\expandafter\def\csname v@dwell.dev.qwen.balanced_evpc_raw.sw.pc\endcsname{46}
\expandafter\def\csname v@dwell.dev.qwen.balanced_evpc_raw.sw.ci.pc\endcsname{46 [29, 64]}
\expandafter\def\csname v@dwell.test.olmo.debate.pc\endcsname{51}
\expandafter\def\csname v@dwell.test.olmo.debate.ci.pc\endcsname{51 [47, 54]}
\expandafter\def\csname v@dwell.test.olmo.debate.rev.pc\endcsname{0}
\expandafter\def\csname v@dwell.test.olmo.debate.rev.ci.pc\endcsname{0 [0, 0]}
\expandafter\def\csname v@dwell.test.olmo.debate.msw.pc\endcsname{24}
\expandafter\def\csname v@dwell.test.olmo.debate.msw.ci.pc\endcsname{24 [18, 30]}
\expandafter\def\csname v@dwell.test.olmo.debate.mst.pc\endcsname{73}
\expandafter\def\csname v@dwell.test.olmo.debate.mst.ci.pc\endcsname{73 [68, 78]}
\expandafter\def\csname v@dwell.test.olmo.debate.sw.pc\endcsname{52}
\expandafter\def\csname v@dwell.test.olmo.debate.sw.ci.pc\endcsname{52 [38, 65]}
\expandafter\def\csname v@dwell.test.olmo.balanced.pc\endcsname{91}
\expandafter\def\csname v@dwell.test.olmo.balanced.ci.pc\endcsname{91 [86, 96]}
\expandafter\def\csname v@dwell.test.olmo.balanced.rev.pc\endcsname{36}
\expandafter\def\csname v@dwell.test.olmo.balanced.rev.ci.pc\endcsname{36 [29, 43]}
\expandafter\def\csname v@dwell.test.olmo.balanced.msw.pc\endcsname{27}
\expandafter\def\csname v@dwell.test.olmo.balanced.msw.ci.pc\endcsname{27 [23, 33]}
\expandafter\def\csname v@dwell.test.olmo.balanced.mst.pc\endcsname{55}
\expandafter\def\csname v@dwell.test.olmo.balanced.mst.ci.pc\endcsname{55 [48, 64]}
\expandafter\def\csname v@dwell.test.olmo.pool_evpc.pc\endcsname{70}
\expandafter\def\csname v@dwell.test.olmo.pool_evpc.ci.pc\endcsname{70 [60, 80]}
\expandafter\def\csname v@dwell.test.olmo.pool_evpc.rev.pc\endcsname{14}
\expandafter\def\csname v@dwell.test.olmo.pool_evpc.rev.ci.pc\endcsname{14 [6, 24]}
\expandafter\def\csname v@dwell.test.olmo.pool_evpc.msw.pc\endcsname{19}
\expandafter\def\csname v@dwell.test.olmo.pool_evpc.msw.ci.pc\endcsname{19 [14, 23]}
\expandafter\def\csname v@dwell.test.olmo.pool_evpc.mst.pc\endcsname{63}
\expandafter\def\csname v@dwell.test.olmo.pool_evpc.mst.ci.pc\endcsname{63 [56, 69]}
\expandafter\def\csname v@dwell.test.olmo.pool_evpc.sw.pc\endcsname{55}
\expandafter\def\csname v@dwell.test.olmo.pool_evpc.sw.ci.pc\endcsname{55 [34, 79]}
\expandafter\def\csname v@dwell.dev.olmo.debate.pc\endcsname{58}
\expandafter\def\csname v@dwell.dev.olmo.debate.ci.pc\endcsname{58 [52, 65]}
\expandafter\def\csname v@dwell.dev.olmo.debate.rev.pc\endcsname{9}
\expandafter\def\csname v@dwell.dev.olmo.debate.rev.ci.pc\endcsname{9 [2, 17]}
\expandafter\def\csname v@dwell.dev.olmo.debate.msw.pc\endcsname{24}
\expandafter\def\csname v@dwell.dev.olmo.debate.msw.ci.pc\endcsname{24 [19, 28]}
\expandafter\def\csname v@dwell.dev.olmo.debate.mst.pc\endcsname{67}
\expandafter\def\csname v@dwell.dev.olmo.debate.mst.ci.pc\endcsname{67 [60, 75]}
\expandafter\def\csname v@dwell.dev.olmo.debate.sw.pc\endcsname{69}
\expandafter\def\csname v@dwell.dev.olmo.debate.sw.ci.pc\endcsname{69 [46, 94]}
\expandafter\def\csname v@dwell.dev.olmo.balanced.pc\endcsname{86}
\expandafter\def\csname v@dwell.dev.olmo.balanced.ci.pc\endcsname{86 [80, 92]}
\expandafter\def\csname v@dwell.dev.olmo.balanced.rev.pc\endcsname{24}
\expandafter\def\csname v@dwell.dev.olmo.balanced.rev.ci.pc\endcsname{24 [18, 31]}
\expandafter\def\csname v@dwell.dev.olmo.balanced.msw.pc\endcsname{23}
\expandafter\def\csname v@dwell.dev.olmo.balanced.msw.ci.pc\endcsname{23 [20, 27]}
\expandafter\def\csname v@dwell.dev.olmo.balanced.mst.pc\endcsname{49}
\expandafter\def\csname v@dwell.dev.olmo.balanced.mst.ci.pc\endcsname{49 [41, 57]}
\expandafter\def\csname v@dwell.dev.olmo.pool_evpc.pc\endcsname{61}
\expandafter\def\csname v@dwell.dev.olmo.pool_evpc.ci.pc\endcsname{61 [48, 73]}
\expandafter\def\csname v@dwell.dev.olmo.pool_evpc.rev.pc\endcsname{11}
\expandafter\def\csname v@dwell.dev.olmo.pool_evpc.rev.ci.pc\endcsname{11 [0, 20]}
\expandafter\def\csname v@dwell.dev.olmo.pool_evpc.msw.pc\endcsname{25}
\expandafter\def\csname v@dwell.dev.olmo.pool_evpc.msw.ci.pc\endcsname{25 [13, 38]}
\expandafter\def\csname v@dwell.dev.olmo.pool_evpc.mst.pc\endcsname{68}
\expandafter\def\csname v@dwell.dev.olmo.pool_evpc.mst.ci.pc\endcsname{68 [61, 75]}
\expandafter\def\csname v@dwell.dev.olmo.pool_evpc.sw.pc\endcsname{32}
\expandafter\def\csname v@dwell.dev.olmo.pool_evpc.sw.ci.pc\endcsname{32 [18, 50]}
\expandafter\def\csname v@dwell.dev.olmo.pool_evpc_raw.pc\endcsname{56}
\expandafter\def\csname v@dwell.dev.olmo.pool_evpc_raw.ci.pc\endcsname{56 [50, 61]}
\expandafter\def\csname v@dwell.dev.olmo.pool_evpc_raw.rev.pc\endcsname{7}
\expandafter\def\csname v@dwell.dev.olmo.pool_evpc_raw.rev.ci.pc\endcsname{7 [2, 13]}
\expandafter\def\csname v@dwell.dev.olmo.pool_evpc_raw.msw.pc\endcsname{28}
\expandafter\def\csname v@dwell.dev.olmo.pool_evpc_raw.msw.ci.pc\endcsname{28 [23, 35]}
\expandafter\def\csname v@dwell.dev.olmo.pool_evpc_raw.mst.pc\endcsname{68}
\expandafter\def\csname v@dwell.dev.olmo.pool_evpc_raw.mst.ci.pc\endcsname{68 [63, 73]}
\expandafter\def\csname v@dwell.dev.olmo.pool_evpc_raw.sw.pc\endcsname{79}
\expandafter\def\csname v@dwell.dev.olmo.pool_evpc_raw.sw.ci.pc\endcsname{79 [57, 101]}
\expandafter\def\csname v@dwell.dev.olmo.debate_evpc_raw.pc\endcsname{61}
\expandafter\def\csname v@dwell.dev.olmo.debate_evpc_raw.ci.pc\endcsname{61 [54, 67]}
\expandafter\def\csname v@dwell.dev.olmo.debate_evpc_raw.rev.pc\endcsname{10}
\expandafter\def\csname v@dwell.dev.olmo.debate_evpc_raw.rev.ci.pc\endcsname{10 [4, 16]}
\expandafter\def\csname v@dwell.dev.olmo.debate_evpc_raw.msw.pc\endcsname{19}
\expandafter\def\csname v@dwell.dev.olmo.debate_evpc_raw.msw.ci.pc\endcsname{19 [16, 21]}
\expandafter\def\csname v@dwell.dev.olmo.debate_evpc_raw.mst.pc\endcsname{65}
\expandafter\def\csname v@dwell.dev.olmo.debate_evpc_raw.mst.ci.pc\endcsname{65 [59, 71]}
\expandafter\def\csname v@dwell.dev.olmo.debate_evpc_raw.sw.pc\endcsname{98}
\expandafter\def\csname v@dwell.dev.olmo.balanced_evpc_rand_raw.pc\endcsname{81}
\expandafter\def\csname v@dwell.dev.olmo.balanced_evpc_rand_raw.ci.pc\endcsname{81 [75, 88]}
\expandafter\def\csname v@dwell.dev.olmo.balanced_evpc_rand_raw.rev.pc\endcsname{27}
\expandafter\def\csname v@dwell.dev.olmo.balanced_evpc_rand_raw.rev.ci.pc\endcsname{27 [17, 35]}
\expandafter\def\csname v@dwell.dev.olmo.balanced_evpc_rand_raw.msw.pc\endcsname{24}
\expandafter\def\csname v@dwell.dev.olmo.balanced_evpc_rand_raw.msw.ci.pc\endcsname{24 [21, 29]}
\expandafter\def\csname v@dwell.dev.olmo.balanced_evpc_rand_raw.mst.pc\endcsname{46}
\expandafter\def\csname v@dwell.dev.olmo.balanced_evpc_rand_raw.mst.ci.pc\endcsname{46 [39, 54]}
\expandafter\def\csname v@dwell.dev.olmo.balanced_evpc_raw.pc\endcsname{90}
\expandafter\def\csname v@dwell.dev.olmo.balanced_evpc_raw.ci.pc\endcsname{90 [84, 97]}
\expandafter\def\csname v@dwell.dev.olmo.balanced_evpc_raw.rev.pc\endcsname{32}
\expandafter\def\csname v@dwell.dev.olmo.balanced_evpc_raw.rev.ci.pc\endcsname{32 [25, 38]}
\expandafter\def\csname v@dwell.dev.olmo.balanced_evpc_raw.msw.pc\endcsname{25}
\expandafter\def\csname v@dwell.dev.olmo.balanced_evpc_raw.msw.ci.pc\endcsname{25 [21, 29]}
\expandafter\def\csname v@dwell.dev.olmo.balanced_evpc_raw.mst.pc\endcsname{47}
\expandafter\def\csname v@dwell.dev.olmo.balanced_evpc_raw.mst.ci.pc\endcsname{47 [41, 53]}
\expandafter\def\csname v@red.mix.qwen.debate.hidden.pc\endcsname{55}
\expandafter\def\csname v@red.mix.qwen.debate.moved.pc\endcsname{32}
\expandafter\def\csname v@red.mix.qwen.debate_evpc.hidden.pc\endcsname{64}
\expandafter\def\csname v@red.mix.qwen.debate_evpc.moved.pc\endcsname{37}
\expandafter\def\csname v@red.mix.qwen.pool_evpc.hidden.pc\endcsname{61}
\expandafter\def\csname v@red.mix.qwen.pool_evpc.moved.pc\endcsname{28}
\expandafter\def\csname v@red.mix.olmo.debate.hidden.pc\endcsname{40}
\expandafter\def\csname v@red.mix.olmo.debate.moved.pc\endcsname{23}
\expandafter\def\csname v@red.mix.olmo.debate_evpc.hidden.pc\endcsname{51}
\expandafter\def\csname v@red.mix.olmo.debate_evpc.moved.pc\endcsname{30}
\expandafter\def\csname v@red.mix.olmo.pool_evpc.hidden.pc\endcsname{50}
\expandafter\def\csname v@red.mix.olmo.pool_evpc.moved.pc\endcsname{24}
\expandafter\def\csname v@an.str.qwen.strong.pool_evpc.decoy.pc\endcsname{15}
\expandafter\def\csname v@an.str.qwen.strong.pool_evpc_cs4.decoy.pc\endcsname{66}
\expandafter\def\csname v@an.str.qwen.strong.pool_evpc_ch4.decoy.pc\endcsname{13}
\expandafter\def\csname v@an.str.qwen.weak.pool_evpc.decoy.pc\endcsname{13}
\expandafter\def\csname v@an.str.qwen.weak.pool_evpc_cs4.decoy.pc\endcsname{27}
\expandafter\def\csname v@an.str.qwen.weak.pool_evpc_ch4.decoy.pc\endcsname{11}
\expandafter\def\csname v@an.str.olmo.strong.pool_evpc.decoy.pc\endcsname{13}
\expandafter\def\csname v@an.str.olmo.strong.pool_evpc_cs4.decoy.pc\endcsname{77}
\expandafter\def\csname v@an.str.olmo.strong.pool_evpc_ch4.decoy.pc\endcsname{10}
\expandafter\def\csname v@an.str.olmo.weak.pool_evpc.decoy.pc\endcsname{16}
\expandafter\def\csname v@an.str.olmo.weak.pool_evpc_cs4.decoy.pc\endcsname{53}
\expandafter\def\csname v@an.str.olmo.weak.pool_evpc_ch4.decoy.pc\endcsname{14}
\expandafter\def\csname v@ei.qwen.pool.S.n.4.pc\endcsname{0}
\expandafter\def\csname v@ei.qwen.pool.H.E.4.pc\endcsname{84}
\expandafter\def\csname v@ei.qwen.pool.H.I.4.pc\endcsname{3}
\expandafter\def\csname v@ei.qwen.pool.H.n.4.pc\endcsname{81}
\expandafter\def\csname v@ei.qwen.pool.S.n.8.pc\endcsname{0}
\expandafter\def\csname v@ei.qwen.pool.H.E.8.pc\endcsname{84}
\expandafter\def\csname v@ei.qwen.pool.H.I.8.pc\endcsname{3}
\expandafter\def\csname v@ei.qwen.pool.H.n.8.pc\endcsname{80}
\expandafter\def\csname v@ei.qwen.pool.S.n.16.pc\endcsname{0}
\expandafter\def\csname v@ei.qwen.pool.H.E.16.pc\endcsname{83}
\expandafter\def\csname v@ei.qwen.pool.H.I.16.pc\endcsname{4}
\expandafter\def\csname v@ei.qwen.pool.H.n.16.pc\endcsname{80}
\expandafter\def\csname v@ei.qwen.pool.S.n.32.pc\endcsname{0}
\expandafter\def\csname v@ei.qwen.pool.H.E.32.pc\endcsname{84}
\expandafter\def\csname v@ei.qwen.pool.H.I.32.pc\endcsname{4}
\expandafter\def\csname v@ei.qwen.pool.H.n.32.pc\endcsname{80}
\expandafter\def\csname v@ei.qwen.pool.S.n.64.pc\endcsname{0}
\expandafter\def\csname v@ei.qwen.pool.H.E.64.pc\endcsname{84}
\expandafter\def\csname v@ei.qwen.pool.H.I.64.pc\endcsname{4}
\expandafter\def\csname v@ei.qwen.pool.H.n.64.pc\endcsname{81}
\expandafter\def\csname v@ei.qwen.six.S.n.4.pc\endcsname{0}
\expandafter\def\csname v@ei.qwen.six.H.E.4.pc\endcsname{84}
\expandafter\def\csname v@ei.qwen.six.H.I.4.pc\endcsname{3}
\expandafter\def\csname v@ei.qwen.six.H.n.4.pc\endcsname{81}
\expandafter\def\csname v@ei.qwen.six.S.n.8.pc\endcsname{0}
\expandafter\def\csname v@ei.qwen.six.H.E.8.pc\endcsname{72}
\expandafter\def\csname v@ei.qwen.six.H.I.8.pc\endcsname{3}
\expandafter\def\csname v@ei.qwen.six.H.n.8.pc\endcsname{70}
\expandafter\def\csname v@ei.qwen.six.S.n.16.pc\endcsname{0}
\expandafter\def\csname v@ei.qwen.six.H.E.16.pc\endcsname{34}
\expandafter\def\csname v@ei.qwen.six.H.I.16.pc\endcsname{1}
\expandafter\def\csname v@ei.qwen.six.H.n.16.pc\endcsname{33}
\expandafter\def\csname v@ei.qwen.six.S.n.32.pc\endcsname{0}
\expandafter\def\csname v@ei.qwen.six.H.E.32.pc\endcsname{17}
\expandafter\def\csname v@ei.qwen.six.H.I.32.pc\endcsname{0}
\expandafter\def\csname v@ei.qwen.six.H.n.32.pc\endcsname{17}
\expandafter\def\csname v@ei.qwen.debate.S.I.4.pc\endcsname{0}
\expandafter\def\csname v@ei.qwen.debate.H.E.4.pc\endcsname{55}
\expandafter\def\csname v@ei.qwen.debate.H.I.4.pc\endcsname{0}
\expandafter\def\csname v@ei.qwen.debate.H.n.4.pc\endcsname{55}
\expandafter\def\csname v@ei.qwen.debate.S.I.8.pc\endcsname{0}
\expandafter\def\csname v@ei.qwen.debate.H.E.8.pc\endcsname{43}
\expandafter\def\csname v@ei.qwen.debate.H.I.8.pc\endcsname{0}
\expandafter\def\csname v@ei.qwen.debate.H.n.8.pc\endcsname{43}
\expandafter\def\csname v@ei.qwen.debate.S.I.16.pc\endcsname{0}
\expandafter\def\csname v@ei.qwen.debate.H.E.16.pc\endcsname{20}
\expandafter\def\csname v@ei.qwen.debate.H.I.16.pc\endcsname{0}
\expandafter\def\csname v@ei.qwen.debate.H.n.16.pc\endcsname{20}
\expandafter\def\csname v@ei.qwen.debate.S.I.32.pc\endcsname{0}
\expandafter\def\csname v@ei.qwen.debate.H.E.32.pc\endcsname{10}
\expandafter\def\csname v@ei.qwen.debate.H.I.32.pc\endcsname{0}
\expandafter\def\csname v@ei.qwen.debate.H.n.32.pc\endcsname{10}
\expandafter\def\csname v@ei.qwen.debate.S.I.64.pc\endcsname{0}
\expandafter\def\csname v@ei.qwen.debate.H.E.64.pc\endcsname{5}
\expandafter\def\csname v@ei.qwen.debate.H.I.64.pc\endcsname{0}
\expandafter\def\csname v@ei.qwen.debate.H.n.64.pc\endcsname{5}
\expandafter\def\csname v@ei.qwen.balanced.S.I.4.pc\endcsname{0}
\expandafter\def\csname v@ei.qwen.balanced.H.I.4.pc\endcsname{0}
\expandafter\def\csname v@ei.qwen.balanced.S.I.8.pc\endcsname{0}
\expandafter\def\csname v@ei.qwen.balanced.H.E.8.pc\endcsname{49}
\expandafter\def\csname v@ei.qwen.balanced.H.I.8.pc\endcsname{0}
\expandafter\def\csname v@ei.qwen.balanced.H.n.8.pc\endcsname{49}
\expandafter\def\csname v@ei.qwen.balanced.S.I.16.pc\endcsname{0}
\expandafter\def\csname v@ei.qwen.balanced.H.E.16.pc\endcsname{26}
\expandafter\def\csname v@ei.qwen.balanced.H.I.16.pc\endcsname{0}
\expandafter\def\csname v@ei.qwen.balanced.H.n.16.pc\endcsname{26}
\expandafter\def\csname v@ei.qwen.balanced.S.I.32.pc\endcsname{0}
\expandafter\def\csname v@ei.qwen.balanced.H.E.32.pc\endcsname{17}
\expandafter\def\csname v@ei.qwen.balanced.H.I.32.pc\endcsname{0}
\expandafter\def\csname v@ei.qwen.balanced.H.n.32.pc\endcsname{17}
\expandafter\def\csname v@ei.qwen.pool.H.n.range.pc\endcsname{80--81}
\expandafter\def\csname v@ei.qwen.debate.H.n.seq.pc\endcsname{55 / 43 / 20 / 10 / 5}
\expandafter\def\csname v@ei.qwen.pz.pool.H.n.range.pc\endcsname{78--80}
\expandafter\def\csname v@ei.qwen.pz.debate.H.n.seq.pc\endcsname{46 / 34 / 15 / 7}
\expandafter\def\csname v@ei.qwen.mq.pool.H.n.range.pc\endcsname{27--29}
\expandafter\def\csname v@ei.qwen.mq.debate.H.n.seq.pc\endcsname{19 / 14 / 6}
\expandafter\def\csname v@ei.olmo.pool.S.n.4.pc\endcsname{0}
\expandafter\def\csname v@ei.olmo.pool.H.E.4.pc\endcsname{71}
\expandafter\def\csname v@ei.olmo.pool.H.I.4.pc\endcsname{2}
\expandafter\def\csname v@ei.olmo.pool.H.n.4.pc\endcsname{70}
\expandafter\def\csname v@ei.olmo.pool.S.n.8.pc\endcsname{0}
\expandafter\def\csname v@ei.olmo.pool.H.E.8.pc\endcsname{74}
\expandafter\def\csname v@ei.olmo.pool.H.I.8.pc\endcsname{3}
\expandafter\def\csname v@ei.olmo.pool.H.n.8.pc\endcsname{71}
\expandafter\def\csname v@ei.olmo.pool.S.n.16.pc\endcsname{0}
\expandafter\def\csname v@ei.olmo.pool.H.E.16.pc\endcsname{71}
\expandafter\def\csname v@ei.olmo.pool.H.I.16.pc\endcsname{3}
\expandafter\def\csname v@ei.olmo.pool.H.n.16.pc\endcsname{68}
\expandafter\def\csname v@ei.olmo.pool.S.n.32.pc\endcsname{0}
\expandafter\def\csname v@ei.olmo.pool.H.E.32.pc\endcsname{73}
\expandafter\def\csname v@ei.olmo.pool.H.I.32.pc\endcsname{3}
\expandafter\def\csname v@ei.olmo.pool.H.n.32.pc\endcsname{70}
\expandafter\def\csname v@ei.olmo.pool.S.n.64.pc\endcsname{0}
\expandafter\def\csname v@ei.olmo.pool.H.E.64.pc\endcsname{72}
\expandafter\def\csname v@ei.olmo.pool.H.I.64.pc\endcsname{3}
\expandafter\def\csname v@ei.olmo.pool.H.n.64.pc\endcsname{69}
\expandafter\def\csname v@ei.olmo.six.S.n.4.pc\endcsname{0}
\expandafter\def\csname v@ei.olmo.six.H.E.4.pc\endcsname{71}
\expandafter\def\csname v@ei.olmo.six.H.I.4.pc\endcsname{2}
\expandafter\def\csname v@ei.olmo.six.H.n.4.pc\endcsname{69}
\expandafter\def\csname v@ei.olmo.six.S.n.8.pc\endcsname{0}
\expandafter\def\csname v@ei.olmo.six.H.E.8.pc\endcsname{64}
\expandafter\def\csname v@ei.olmo.six.H.I.8.pc\endcsname{2}
\expandafter\def\csname v@ei.olmo.six.H.n.8.pc\endcsname{62}
\expandafter\def\csname v@ei.olmo.six.S.n.16.pc\endcsname{0}
\expandafter\def\csname v@ei.olmo.six.H.E.16.pc\endcsname{30}
\expandafter\def\csname v@ei.olmo.six.H.I.16.pc\endcsname{1}
\expandafter\def\csname v@ei.olmo.six.H.n.16.pc\endcsname{29}
\expandafter\def\csname v@ei.olmo.six.S.n.32.pc\endcsname{0}
\expandafter\def\csname v@ei.olmo.six.H.E.32.pc\endcsname{15}
\expandafter\def\csname v@ei.olmo.six.H.I.32.pc\endcsname{0}
\expandafter\def\csname v@ei.olmo.six.H.n.32.pc\endcsname{15}
\expandafter\def\csname v@ei.olmo.debate.S.E.4.pc\endcsname{80}
\expandafter\def\csname v@ei.olmo.debate.S.I.4.pc\endcsname{0}
\expandafter\def\csname v@ei.olmo.debate.S.n.4.pc\endcsname{80}
\expandafter\def\csname v@ei.olmo.debate.H.E.4.pc\endcsname{34}
\expandafter\def\csname v@ei.olmo.debate.H.I.4.pc\endcsname{0}
\expandafter\def\csname v@ei.olmo.debate.H.n.4.pc\endcsname{34}
\expandafter\def\csname v@ei.olmo.debate.S.I.8.pc\endcsname{0}
\expandafter\def\csname v@ei.olmo.debate.H.E.8.pc\endcsname{26}
\expandafter\def\csname v@ei.olmo.debate.H.I.8.pc\endcsname{0}
\expandafter\def\csname v@ei.olmo.debate.H.n.8.pc\endcsname{26}
\expandafter\def\csname v@ei.olmo.debate.S.I.16.pc\endcsname{0}
\expandafter\def\csname v@ei.olmo.debate.H.E.16.pc\endcsname{12}
\expandafter\def\csname v@ei.olmo.debate.H.I.16.pc\endcsname{0}
\expandafter\def\csname v@ei.olmo.debate.H.n.16.pc\endcsname{12}
\expandafter\def\csname v@ei.olmo.debate.S.I.32.pc\endcsname{0}
\expandafter\def\csname v@ei.olmo.debate.H.E.32.pc\endcsname{6}
\expandafter\def\csname v@ei.olmo.debate.H.I.32.pc\endcsname{0}
\expandafter\def\csname v@ei.olmo.debate.H.n.32.pc\endcsname{6}
\expandafter\def\csname v@ei.olmo.debate.S.I.64.pc\endcsname{0}
\expandafter\def\csname v@ei.olmo.debate.H.E.64.pc\endcsname{3}
\expandafter\def\csname v@ei.olmo.debate.H.I.64.pc\endcsname{0}
\expandafter\def\csname v@ei.olmo.debate.H.n.64.pc\endcsname{3}
\expandafter\def\csname v@ei.olmo.balanced.S.I.4.pc\endcsname{0}
\expandafter\def\csname v@ei.olmo.balanced.H.E.4.pc\endcsname{70}
\expandafter\def\csname v@ei.olmo.balanced.H.I.4.pc\endcsname{0}
\expandafter\def\csname v@ei.olmo.balanced.H.n.4.pc\endcsname{70}
\expandafter\def\csname v@ei.olmo.balanced.S.I.8.pc\endcsname{0}
\expandafter\def\csname v@ei.olmo.balanced.H.E.8.pc\endcsname{31}
\expandafter\def\csname v@ei.olmo.balanced.H.I.8.pc\endcsname{0}
\expandafter\def\csname v@ei.olmo.balanced.H.n.8.pc\endcsname{31}
\expandafter\def\csname v@ei.olmo.balanced.S.I.16.pc\endcsname{0}
\expandafter\def\csname v@ei.olmo.balanced.H.E.16.pc\endcsname{16}
\expandafter\def\csname v@ei.olmo.balanced.H.I.16.pc\endcsname{0}
\expandafter\def\csname v@ei.olmo.balanced.H.n.16.pc\endcsname{16}
\expandafter\def\csname v@ei.olmo.balanced.S.I.32.pc\endcsname{0}
\expandafter\def\csname v@ei.olmo.balanced.H.E.32.pc\endcsname{10}
\expandafter\def\csname v@ei.olmo.balanced.H.I.32.pc\endcsname{0}
\expandafter\def\csname v@ei.olmo.balanced.H.n.32.pc\endcsname{10}
\expandafter\def\csname v@ei.olmo.pool.H.n.range.pc\endcsname{68--71}
\expandafter\def\csname v@ei.olmo.debate.H.n.seq.pc\endcsname{34 / 26 / 12 / 6 / 3}
\expandafter\def\csname v@ei.olmo.balanced.H.n.seq.pc\endcsname{70 / 31 / 16 / 10}
\expandafter\def\csname v@ei.olmo.pz.pool.H.n.range.pc\endcsname{69--71}
\expandafter\def\csname v@ei.olmo.pz.debate.H.n.seq.pc\endcsname{24 / 18 / 7 / 4}
\expandafter\def\csname v@ei.olmo.mq.pool.H.n.range.pc\endcsname{26--27}
\expandafter\def\csname v@ei.olmo.mq.debate.H.n.seq.pc\endcsname{12 / 9 / 4}
\expandafter\def\csname v@ei.nu.debate.range.pc\endcsname{97--104}
\expandafter\def\csname v@ei.nu.equal.range.pc\endcsname{0}
\expandafter\def\csname v@ei.nu.half.range.pc\endcsname{57--66}
\expandafter\def\csname v@kf.cap.range.pc\endcsname{4--16}
\expandafter\def\csname v@kf.nodedup.hb.range.pc\endcsname{54--70}
\expandafter\def\csname v@kf.nodedup.pz.range.pc\endcsname{5--18}
\expandafter\def\csname v@kf.capdedup.hb.range.pc\endcsname{76--86}
\expandafter\def\csname v@kf.ev.hb.range.pc\endcsname{85--88}
\expandafter\def\csname v@kf.evpc.hb.range.pc\endcsname{88--92}
\expandafter\def\csname v@kf.evpc.hb.evcells.range.pc\endcsname{88--92}
\expandafter\def\csname v@kf.capdedup.pz.range.pc\endcsname{20--44}
\expandafter\def\csname v@kf.ev.pz.range.pc\endcsname{61--76}
\expandafter\def\csname v@kf.evpc.pz.range.pc\endcsname{75--88}
\expandafter\def\csname v@kf.evpc.pz.evcells.range.pc\endcsname{75--85}
\expandafter\def\csname v@kf.cap.gap.range.pc\endcsname{$+$71 to $+$86}
\expandafter\def\csname v@kf.cap.gap.abs.range.pc\endcsname{71--86}
\expandafter\def\csname v@kf.debate.range.pc\endcsname{5--21}
\expandafter\def\csname v@kf.delay.r2.range.pc\endcsname{5--61}
\expandafter\def\csname v@kf.evpc.r2.range.pc\endcsname{57--88}
\expandafter\def\csname v@kf.delay.r3.range.pc\endcsname{41--86}
\expandafter\def\csname v@kf.evpc.r3.range.pc\endcsname{65--89}
\expandafter\def\csname v@kf.delay.final.range.pc\endcsname{71--92}
\expandafter\def\csname v@kf.delay.evpc.range.pc\endcsname{75--92}
\expandafter\def\csname v@kf.delay.endgap.range.pc\endcsname{$-$1 to $+$7}
\expandafter\def\csname v@kf.delay.lag.range.pc\endcsname{$-$4 to $+$3}
\expandafter\def\csname v@kf.k3.facts.range.pc\endcsname{61--84}
\expandafter\def\csname v@kf.k3.factspert.range.pc\endcsname{58--84}
\expandafter\def\csname v@kf.k3.args.range.pc\endcsname{6--15}
\expandafter\def\csname v@kf.k3.argspert.range.pc\endcsname{0--16}
\expandafter\def\csname v@kf.k3.nofb.range.pc\endcsname{5--20}
\expandafter\def\csname v@kf.k3.factsdiff.range.pc\endcsname{$-$6 to $+$6}
\expandafter\def\csname v@kf.k3.alone.range.pc\endcsname{2--14}
\expandafter\def\csname v@J3a.hb.qwen.pc\endcsname{$+$69 [56, 81]}
\expandafter\def\csname v@J3b.hb.qwen.pc\endcsname{$+$52 [41, 64]}
\expandafter\def\csname v@J3a.pz.qwen.pc\endcsname{$+$60 [56, 64]}
\expandafter\def\csname v@J3b.pz.qwen.pc\endcsname{$+$46 [40, 52]}
\expandafter\def\csname v@J3a.hb.olmo.pc\endcsname{$+$63 [52, 74]}
\expandafter\def\csname v@J3b.hb.olmo.pc\endcsname{$+$59 [47, 71]}
\expandafter\def\csname v@J3a.pz.olmo.pc\endcsname{$+$61 [57, 65]}
\expandafter\def\csname v@J3b.pz.olmo.pc\endcsname{$+$59 [54, 64]}
\expandafter\def\csname v@J3a.min.pc\endcsname{60}
\expandafter\def\csname v@J3a.max.pc\endcsname{69}
\expandafter\def\csname v@J3b.min.pc\endcsname{46}
\expandafter\def\csname v@J3b.max.pc\endcsname{59}
\expandafter\def\csname v@j3.evpc.range.pc\endcsname{57--69}
\expandafter\def\csname v@j3.evpc.min.pc\endcsname{57}
\expandafter\def\csname v@j3.raw.range.pc\endcsname{1--7}
\expandafter\def\csname v@j3.raw.max.pc\endcsname{7}
\expandafter\def\csname v@j3.nodedup.range.pc\endcsname{6--43}
\expandafter\def\csname v@j3.alone.range.pc\endcsname{0--6}
\expandafter\def\csname v@j3.gain.allruns.range.pc\endcsname{58--65}
\expandafter\def\csname v@j3.peerA.range.pc\endcsname{67--78}
\expandafter\def\csname v@j3.full.hb.range.pc\endcsname{48--60}
\expandafter\def\csname v@j3.full.pz.range.pc\endcsname{76--81}
\expandafter\def\csname v@j3.hb.below.range.pc\endcsname{16--21}
\expandafter\def\csname v@pred.J3.width.pc\endcsname{43}
\expandafter\def\csname v@j3.r1.hb.range.pc\endcsname{33--62}
\expandafter\def\csname v@j3.evpc.hb.range.pc\endcsname{57--68}
\expandafter\def\csname v@j3.r1.pz.range.pc\endcsname{34--38}
\expandafter\def\csname v@j3.evpc.pz.range.pc\endcsname{65--69}
\expandafter\def\csname v@j3.full.share.hb.pc\endcsname{20--49}
\expandafter\def\csname v@j3.full.share.pz.pc\endcsname{7--17}
\expandafter\def\csname v@j3.full.raw.range.pc\endcsname{0--14}
\expandafter\def\csname v@j3.full.evpc.range.pc\endcsname{71--87}
\expandafter\def\csname v@j3.full.nodedup.range.pc\endcsname{1--51}
\expandafter\def\csname v@j3.full.raw.r1.range.pc\endcsname{0--11}
\expandafter\def\csname v@j3.full.evpc.r1.range.pc\endcsname{47--79}
\expandafter\def\csname v@j1.evpc.pc\endcsname{83}
\expandafter\def\csname v@j1.raw.pc\endcsname{5}
\expandafter\def\csname v@j1.weigh.pc\endcsname{6}
\expandafter\def\csname v@j1.weigh.diff.pc\endcsname{$+$1 [0, 3]}
\expandafter\def\csname v@j2.gain.range.pc\endcsname{57--82}
\expandafter\def\csname v@kv.miss.evpc.range.pc\endcsname{71--83}
\expandafter\def\csname v@kv.miss.cap.range.pc\endcsname{2--6}
\expandafter\def\csname v@kv.miss.capdedup.range.pc\endcsname{29--37}
\expandafter\def\csname v@k3.did.range.pc\endcsname{$-$14 to $+$5}
\expandafter\def\csname v@k3b.nofb.range.pc\endcsname{55--85}
\expandafter\def\csname v@k3b.diff.range.pc\endcsname{$-$9 to $+$7}
\expandafter\def\csname v@r.facts.c0.range.pc\endcsname{47--88}
\expandafter\def\csname v@r.facts.c8.range.pc\endcsname{8--72}
\expandafter\def\csname v@r.facts.c84.range.pc\endcsname{13--74}
\expandafter\def\csname v@r.facts.drop.range.pc\endcsname{16--48}
\expandafter\def\csname v@r.argued.18.range.pc\endcsname{15--41}
\expandafter\def\csname v@r.argued.81.range.pc\endcsname{98--100}
\expandafter\def\csname v@u.last.range.pc\endcsname{$-$15 to $-$2}
\expandafter\def\csname v@u.own.range.pc\endcsname{$-$49 to $-$29}
\expandafter\def\csname v@u.ownlast.range.pc\endcsname{$-$49 to $-$25}
\expandafter\def\csname v@u.ownlast.deb.range.pc\endcsname{$+$22 to $+$62}
\expandafter\def\csname v@u.ownnd.range.pc\endcsname{$+$24 to $+$59}
\expandafter\def\csname v@u.first.range.pc\endcsname{$-$13 to $-$1}
\expandafter\def\csname v@u.ownnd.deb.range.pc\endcsname{$-$4 to 0}
\expandafter\def\csname v@u.evpc.deb.range.pc\endcsname{$+$68 to $+$88}
\expandafter\def\csname v@u2.para.range.pc\endcsname{$-$9 to $-$1}
\expandafter\def\csname v@u2.para.hb.qwen.pc\endcsname{$-$1 [$-$3, 1]}
\expandafter\def\csname v@u2.para.pz.qwen.pc\endcsname{$-$1 [$-$3, 1]}
\expandafter\def\csname v@u2.para.hb.olmo.pc\endcsname{$-$9 [$-$16, $-$3]}
\expandafter\def\csname v@u2.para.pz.olmo.pc\endcsname{$-$3 [$-$6, 1]}
\expandafter\def\csname v@u2.para.own.range.pc\endcsname{$+$29 to $+$48}
\expandafter\def\csname v@u2.para.own.hb.qwen.pc\endcsname{$+$29 [18, 40]}
\expandafter\def\csname v@u2.para.own.pz.qwen.pc\endcsname{$+$48 [41, 55]}
\expandafter\def\csname v@u2.para.own.hb.olmo.pc\endcsname{$+$38 [25, 51]}
\expandafter\def\csname v@u2.para.own.pz.olmo.pc\endcsname{$+$44 [37, 51]}
\expandafter\def\csname v@u3.claims.own.range.pc\endcsname{$-$11 to $+$20}
\expandafter\def\csname v@u3.gap.range.pc\endcsname{$+$25 to $+$57}
\expandafter\def\csname v@u16.own.range.pc\endcsname{$-$49 to $-$29}
\expandafter\def\csname v@u16.last.range.pc\endcsname{$-$16 to $-$1}
\expandafter\def\csname v@u16.ownnd.range.pc\endcsname{$+$22 to $+$58}
\expandafter\def\csname v@u16.evpc.deb.range.pc\endcsname{$+$65 to $+$88}
\expandafter\def\csname v@u16.para.range.pc\endcsname{$-$7 to 0}
\expandafter\def\csname v@u16.para.own.range.pc\endcsname{$+$27 to $+$45}
\expandafter\def\csname v@u16.claims.own.range.pc\endcsname{$-$11 to $+$20}
\expandafter\def\csname v@u.repro.maxdiff.pc\endcsname{5}
\expandafter\def\csname v@u.qwen.hb.16.debate.pc\endcsname{3}
\expandafter\def\csname v@u.qwen.hb.16.pool_ev_own_nodedup.pc\endcsname{4}
\expandafter\def\csname v@u.qwen.hb.16.pool_evpc_ownfirst.pc\endcsname{45}
\expandafter\def\csname v@u.qwen.hb.16.pool_evpc_own.pc\endcsname{62}
\expandafter\def\csname v@u.qwen.hb.16.pool_evpc_own_last.pc\endcsname{68}
\expandafter\def\csname v@u.qwen.hb.16.pool_evpc_claims.pc\endcsname{66}
\expandafter\def\csname v@u.qwen.hb.16.pool_evpc_para.pc\endcsname{89}
\expandafter\def\csname v@u.qwen.hb.16.pool_evpc_last.pc\endcsname{88}
\expandafter\def\csname v@u.qwen.hb.16.pool_evpc.pc\endcsname{91}
\expandafter\def\csname v@u.qwen.hb.32.debate.pc\endcsname{4}
\expandafter\def\csname v@u.qwen.hb.32.pool_ev_own_nodedup.pc\endcsname{1}
\expandafter\def\csname v@u.qwen.hb.32.pool_evpc_ownfirst.pc\endcsname{52}
\expandafter\def\csname v@u.qwen.hb.32.pool_evpc_own.pc\endcsname{61}
\expandafter\def\csname v@u.qwen.hb.32.pool_evpc_own_last.pc\endcsname{64}
\expandafter\def\csname v@u.qwen.hb.32.pool_evpc_para.pc\endcsname{92}
\expandafter\def\csname v@u.qwen.hb.32.pool_evpc_last.pc\endcsname{88}
\expandafter\def\csname v@u.qwen.hb.32.pool_evpc.pc\endcsname{92}
\expandafter\def\csname v@u.qwen.pz.16.debate.pc\endcsname{11}
\expandafter\def\csname v@u.qwen.pz.16.pool_ev_own_nodedup.pc\endcsname{12}
\expandafter\def\csname v@u.qwen.pz.16.pool_evpc_ownfirst.pc\endcsname{37}
\expandafter\def\csname v@u.qwen.pz.16.pool_evpc_own.pc\endcsname{41}
\expandafter\def\csname v@u.qwen.pz.16.pool_evpc_own_last.pc\endcsname{37}
\expandafter\def\csname v@u.qwen.pz.16.pool_evpc_claims.pc\endcsname{49}
\expandafter\def\csname v@u.qwen.pz.16.pool_evpc_para.pc\endcsname{86}
\expandafter\def\csname v@u.qwen.pz.16.pool_evpc_last.pc\endcsname{85}
\expandafter\def\csname v@u.qwen.pz.16.pool_evpc.pc\endcsname{86}
\expandafter\def\csname v@u.qwen.pz.32.debate.pc\endcsname{9}
\expandafter\def\csname v@u.qwen.pz.32.pool_ev_own_nodedup.pc\endcsname{9}
\expandafter\def\csname v@u.qwen.pz.32.pool_evpc_ownfirst.pc\endcsname{39}
\expandafter\def\csname v@u.qwen.pz.32.pool_evpc_own.pc\endcsname{36}
\expandafter\def\csname v@u.qwen.pz.32.pool_evpc_own_last.pc\endcsname{40}
\expandafter\def\csname v@u.qwen.pz.32.pool_evpc_para.pc\endcsname{88}
\expandafter\def\csname v@u.qwen.pz.32.pool_evpc_last.pc\endcsname{85}
\expandafter\def\csname v@u.qwen.pz.32.pool_evpc.pc\endcsname{89}
\expandafter\def\csname v@u.olmo.hb.16.debate.pc\endcsname{24}
\expandafter\def\csname v@u.olmo.hb.16.pool_ev_own_nodedup.pc\endcsname{17}
\expandafter\def\csname v@u.olmo.hb.16.pool_evpc_ownfirst.pc\endcsname{33}
\expandafter\def\csname v@u.olmo.hb.16.pool_evpc_own.pc\endcsname{40}
\expandafter\def\csname v@u.olmo.hb.16.pool_evpc_own_last.pc\endcsname{49}
\expandafter\def\csname v@u.olmo.hb.16.pool_evpc_claims.pc\endcsname{60}
\expandafter\def\csname v@u.olmo.hb.16.pool_evpc_para.pc\endcsname{82}
\expandafter\def\csname v@u.olmo.hb.16.pool_evpc_last.pc\endcsname{85}
\expandafter\def\csname v@u.olmo.hb.16.pool_evpc.pc\endcsname{89}
\expandafter\def\csname v@u.olmo.hb.32.debate.pc\endcsname{21}
\expandafter\def\csname v@u.olmo.hb.32.pool_ev_own_nodedup.pc\endcsname{20}
\expandafter\def\csname v@u.olmo.hb.32.pool_evpc_ownfirst.pc\endcsname{45}
\expandafter\def\csname v@u.olmo.hb.32.pool_evpc_own.pc\endcsname{49}
\expandafter\def\csname v@u.olmo.hb.32.pool_evpc_own_last.pc\endcsname{49}
\expandafter\def\csname v@u.olmo.hb.32.pool_evpc_para.pc\endcsname{83}
\expandafter\def\csname v@u.olmo.hb.32.pool_evpc_last.pc\endcsname{83}
\expandafter\def\csname v@u.olmo.hb.32.pool_evpc.pc\endcsname{95}
\expandafter\def\csname v@u.olmo.pz.16.debate.pc\endcsname{10}
\expandafter\def\csname v@u.olmo.pz.16.pool_ev_own_nodedup.pc\endcsname{6}
\expandafter\def\csname v@u.olmo.pz.16.pool_evpc_ownfirst.pc\endcsname{21}
\expandafter\def\csname v@u.olmo.pz.16.pool_evpc_own.pc\endcsname{31}
\expandafter\def\csname v@u.olmo.pz.16.pool_evpc_own_last.pc\endcsname{31}
\expandafter\def\csname v@u.olmo.pz.16.pool_evpc_claims.pc\endcsname{19}
\expandafter\def\csname v@u.olmo.pz.16.pool_evpc_para.pc\endcsname{72}
\expandafter\def\csname v@u.olmo.pz.16.pool_evpc_last.pc\endcsname{61}
\expandafter\def\csname v@u.olmo.pz.16.pool_evpc.pc\endcsname{76}
\expandafter\def\csname v@u.olmo.pz.32.debate.pc\endcsname{9}
\expandafter\def\csname v@u.olmo.pz.32.pool_ev_own_nodedup.pc\endcsname{6}
\expandafter\def\csname v@u.olmo.pz.32.pool_evpc_ownfirst.pc\endcsname{21}
\expandafter\def\csname v@u.olmo.pz.32.pool_evpc_own.pc\endcsname{30}
\expandafter\def\csname v@u.olmo.pz.32.pool_evpc_own_last.pc\endcsname{30}
\expandafter\def\csname v@u.olmo.pz.32.pool_evpc_para.pc\endcsname{76}
\expandafter\def\csname v@u.olmo.pz.32.pool_evpc_last.pc\endcsname{64}
\expandafter\def\csname v@u.olmo.pz.32.pool_evpc.pc\endcsname{78}
\expandafter\def\csname v@u.view.wrong.range.pc\endcsname{50--73}
\expandafter\def\csname v@u.view.held.range.pc\endcsname{32--67}
\expandafter\def\csname v@u.view.conclusion.range.pc\endcsname{13--62}
\expandafter\def\csname v@x.qwen.screen.d2.full.pc\endcsname{98}
\expandafter\def\csname v@x.qwen.screen.d4.full.pc\endcsname{86}
\expandafter\def\csname v@x.qwen.screen.d8.full.pc\endcsname{66}
\expandafter\def\csname v@x.qwen.screen.d16.full.pc\endcsname{34}
\expandafter\def\csname v@x.olmo.screen.d2.full.pc\endcsname{88}
\expandafter\def\csname v@x.olmo.screen.d4.full.pc\endcsname{77}
\expandafter\def\csname v@x.olmo.screen.d8.full.pc\endcsname{57}
\expandafter\def\csname v@x.olmo.screen.d16.full.pc\endcsname{36}
\expandafter\def\csname v@x.qwen.8.debate.pc\endcsname{8}
\expandafter\def\csname v@x.qwen.8.pool_ev_nodedup.pc\endcsname{61}
\expandafter\def\csname v@x.qwen.8.pool_evpc.pc\endcsname{99}
\expandafter\def\csname v@x.qwen.8.full_iter.pc\endcsname{100}
\expandafter\def\csname v@x.qwen.16.debate.pc\endcsname{0}
\expandafter\def\csname v@x.qwen.16.pool_ev_nodedup.pc\endcsname{31}
\expandafter\def\csname v@x.qwen.16.pool_evpc.pc\endcsname{84}
\expandafter\def\csname v@x.qwen.16.full_iter.pc\endcsname{98}
\expandafter\def\csname v@x.qwen.32.debate.pc\endcsname{0}
\expandafter\def\csname v@x.qwen.32.pool_ev_nodedup.pc\endcsname{0}
\expandafter\def\csname v@x.qwen.32.pool_evpc.pc\endcsname{60}
\expandafter\def\csname v@x.qwen.32.full_iter.pc\endcsname{100}
\expandafter\def\csname v@x.qwen.64.debate.pc\endcsname{0}
\expandafter\def\csname v@x.qwen.64.pool_ev_nodedup.pc\endcsname{0}
\expandafter\def\csname v@x.qwen.64.pool_evpc.pc\endcsname{17}
\expandafter\def\csname v@x.qwen.64.full_iter.pc\endcsname{90}
\expandafter\def\csname v@x.qwen.debate.range.pc\endcsname{0--8}
\expandafter\def\csname v@x.qwen.pool_ev_nodedup.range.pc\endcsname{0--61}
\expandafter\def\csname v@x.qwen.pool_evpc.range.pc\endcsname{17--99}
\expandafter\def\csname v@x.qwen.full_iter.range.pc\endcsname{90--100}
\expandafter\def\csname v@x.olmo.8.debate.pc\endcsname{12}
\expandafter\def\csname v@x.olmo.8.pool_ev_nodedup.pc\endcsname{58}
\expandafter\def\csname v@x.olmo.8.pool_evpc.pc\endcsname{80}
\expandafter\def\csname v@x.olmo.8.full_iter.pc\endcsname{100}
\expandafter\def\csname v@x.olmo.16.debate.pc\endcsname{0}
\expandafter\def\csname v@x.olmo.16.pool_ev_nodedup.pc\endcsname{15}
\expandafter\def\csname v@x.olmo.16.pool_evpc.pc\endcsname{61}
\expandafter\def\csname v@x.olmo.16.full_iter.pc\endcsname{100}
\expandafter\def\csname v@x.olmo.32.debate.pc\endcsname{0}
\expandafter\def\csname v@x.olmo.32.pool_ev_nodedup.pc\endcsname{0}
\expandafter\def\csname v@x.olmo.32.pool_evpc.pc\endcsname{29}
\expandafter\def\csname v@x.olmo.32.full_iter.pc\endcsname{99}
\expandafter\def\csname v@x.olmo.64.debate.pc\endcsname{0}
\expandafter\def\csname v@x.olmo.64.pool_ev_nodedup.pc\endcsname{0}
\expandafter\def\csname v@x.olmo.64.pool_evpc.pc\endcsname{2}
\expandafter\def\csname v@x.olmo.64.full_iter.pc\endcsname{100}
\expandafter\def\csname v@x.olmo.debate.range.pc\endcsname{0--12}
\expandafter\def\csname v@x.olmo.pool_ev_nodedup.range.pc\endcsname{0--58}
\expandafter\def\csname v@x.olmo.pool_evpc.range.pc\endcsname{2--80}
\expandafter\def\csname v@x.olmo.full_iter.range.pc\endcsname{99--100}
\expandafter\def\csname v@x.qwen.64.evpc.debate.pc\endcsname{$+$17 [8, 28]}
\expandafter\def\csname v@x.qwen.64.evpc.nodedup.pc\endcsname{$+$17 [8, 28]}
\expandafter\def\csname v@x.olmo.64.evpc.debate.pc\endcsname{$+$2 [0, 5]}
\expandafter\def\csname v@x.olmo.64.evpc.nodedup.pc\endcsname{$+$2 [0, 5]}
\expandafter\def\csname v@x.evpc.debate.range.pc\endcsname{$+$2 to $+$91}
\expandafter\def\csname v@x.full_iter.range.pc\endcsname{90--100}
\expandafter\def\csname v@x.view.complete.range.pc\endcsname{3--87}
\expandafter\def\csname v@x.view.incomplete.range.pc\endcsname{3--33}
\expandafter\def\csname v@x.view.share.range.pc\endcsname{0--100}
\expandafter\def\csname v@x.qwen.8.complete.pc\endcsname{100}
\expandafter\def\csname v@x.qwen.8.after.pc\endcsname{85}
\expandafter\def\csname v@x.qwen.16.complete.pc\endcsname{80}
\expandafter\def\csname v@x.qwen.16.after.pc\endcsname{75}
\expandafter\def\csname v@x.qwen.32.complete.pc\endcsname{47}
\expandafter\def\csname v@x.qwen.32.after.pc\endcsname{45}
\expandafter\def\csname v@x.qwen.64.complete.pc\endcsname{13}
\expandafter\def\csname v@x.qwen.64.after.pc\endcsname{20}
\expandafter\def\csname v@x.olmo.8.complete.pc\endcsname{76}
\expandafter\def\csname v@x.olmo.8.after.pc\endcsname{87}
\expandafter\def\csname v@x.olmo.16.complete.pc\endcsname{38}
\expandafter\def\csname v@x.olmo.16.after.pc\endcsname{76}
\expandafter\def\csname v@x.olmo.32.complete.pc\endcsname{14}
\expandafter\def\csname v@x.olmo.32.after.pc\endcsname{39}
\expandafter\def\csname v@x.olmo.64.complete.pc\endcsname{0}
\expandafter\def\csname v@x.olmo.64.after.pc\endcsname{3}
\expandafter\def\csname v@d2.prov.cite.hb.qwen.pc\endcsname{$-$1 [$-$4, 4]}
\expandafter\def\csname v@d2.prov.debate.hb.qwen.pc\endcsname{$+$82 [69, 92]}
\expandafter\def\csname v@d2.cite.debate.hb.qwen.pc\endcsname{$+$4 [$-$1, 11]}
\expandafter\def\csname v@d2.declared.hb.qwen.pc\endcsname{79}
\expandafter\def\csname v@d2.lexical.hb.qwen.pc\endcsname{94}
\expandafter\def\csname v@d2.final.prov.hb.qwen.pc\endcsname{95}
\expandafter\def\csname v@d2.final.cite.hb.qwen.pc\endcsname{91}
\expandafter\def\csname v@d2.debate.hb.qwen.16.pc\endcsname{8}
\expandafter\def\csname v@d2.debate_cite.hb.qwen.16.pc\endcsname{15}
\expandafter\def\csname v@d2.pool_evpc.hb.qwen.16.pc\endcsname{88}
\expandafter\def\csname v@d2.pool_evpc_cite.hb.qwen.16.pc\endcsname{93}
\expandafter\def\csname v@d2.pool_prov.hb.qwen.16.pc\endcsname{92}
\expandafter\def\csname v@d2.debate.hb.qwen.32.pc\endcsname{5}
\expandafter\def\csname v@d2.debate_cite.hb.qwen.32.pc\endcsname{7}
\expandafter\def\csname v@d2.pool_evpc.hb.qwen.32.pc\endcsname{91}
\expandafter\def\csname v@d2.pool_evpc_cite.hb.qwen.32.pc\endcsname{94}
\expandafter\def\csname v@d2.pool_prov.hb.qwen.32.pc\endcsname{94}
\expandafter\def\csname v@d2.prov.cite.hb.olmo.pc\endcsname{$-$1 [$-$4, 3]}
\expandafter\def\csname v@d2.prov.debate.hb.olmo.pc\endcsname{$+$76 [57, 90]}
\expandafter\def\csname v@d2.cite.debate.hb.olmo.pc\endcsname{$-$4 [$-$13, 4]}
\expandafter\def\csname v@d2.declared.hb.olmo.pc\endcsname{69}
\expandafter\def\csname v@d2.lexical.hb.olmo.pc\endcsname{72}
\expandafter\def\csname v@d2.final.prov.hb.olmo.pc\endcsname{91}
\expandafter\def\csname v@d2.final.cite.hb.olmo.pc\endcsname{81}
\expandafter\def\csname v@d2.debate.hb.olmo.16.pc\endcsname{21}
\expandafter\def\csname v@d2.debate_cite.hb.olmo.16.pc\endcsname{21}
\expandafter\def\csname v@d2.pool_evpc.hb.olmo.16.pc\endcsname{88}
\expandafter\def\csname v@d2.pool_evpc_cite.hb.olmo.16.pc\endcsname{96}
\expandafter\def\csname v@d2.pool_prov.hb.olmo.16.pc\endcsname{93}
\expandafter\def\csname v@d2.debate.hb.olmo.32.pc\endcsname{21}
\expandafter\def\csname v@d2.debate_cite.hb.olmo.32.pc\endcsname{12}
\expandafter\def\csname v@d2.pool_evpc.hb.olmo.32.pc\endcsname{92}
\expandafter\def\csname v@d2.pool_evpc_cite.hb.olmo.32.pc\endcsname{91}
\expandafter\def\csname v@d2.pool_prov.hb.olmo.32.pc\endcsname{93}
\expandafter\def\csname v@d2.prov.cite.pz.qwen.pc\endcsname{$-$16 [$-$20, $-$12]}
\expandafter\def\csname v@d2.prov.debate.pz.qwen.pc\endcsname{$+$61 [54, 68]}
\expandafter\def\csname v@d2.cite.debate.pz.qwen.pc\endcsname{$+$2 [$-$2, 5]}
\expandafter\def\csname v@d2.declared.pz.qwen.pc\endcsname{40}
\expandafter\def\csname v@d2.lexical.pz.qwen.pc\endcsname{57}
\expandafter\def\csname v@d2.final.prov.pz.qwen.pc\endcsname{70}
\expandafter\def\csname v@d2.final.cite.pz.qwen.pc\endcsname{85}
\expandafter\def\csname v@d2.debate.pz.qwen.16.pc\endcsname{10}
\expandafter\def\csname v@d2.debate_cite.pz.qwen.16.pc\endcsname{13}
\expandafter\def\csname v@d2.pool_evpc.pz.qwen.16.pc\endcsname{85}
\expandafter\def\csname v@d2.pool_evpc_cite.pz.qwen.16.pc\endcsname{89}
\expandafter\def\csname v@d2.pool_prov.pz.qwen.16.pc\endcsname{71}
\expandafter\def\csname v@d2.debate.pz.qwen.32.pc\endcsname{11}
\expandafter\def\csname v@d2.debate_cite.pz.qwen.32.pc\endcsname{12}
\expandafter\def\csname v@d2.pool_evpc.pz.qwen.32.pc\endcsname{88}
\expandafter\def\csname v@d2.pool_evpc_cite.pz.qwen.32.pc\endcsname{90}
\expandafter\def\csname v@d2.pool_prov.pz.qwen.32.pc\endcsname{75}
\expandafter\def\csname v@d2.prov.cite.pz.olmo.pc\endcsname{$-$15 [$-$20, $-$10]}
\expandafter\def\csname v@d2.prov.debate.pz.olmo.pc\endcsname{$+$50 [44, 57]}
\expandafter\def\csname v@d2.cite.debate.pz.olmo.pc\endcsname{$-$2 [$-$4, 1]}
\expandafter\def\csname v@d2.declared.pz.olmo.pc\endcsname{43}
\expandafter\def\csname v@d2.lexical.pz.olmo.pc\endcsname{63}
\expandafter\def\csname v@d2.final.prov.pz.olmo.pc\endcsname{70}
\expandafter\def\csname v@d2.final.cite.pz.olmo.pc\endcsname{79}
\expandafter\def\csname v@d2.debate.pz.olmo.16.pc\endcsname{8}
\expandafter\def\csname v@d2.debate_cite.pz.olmo.16.pc\endcsname{6}
\expandafter\def\csname v@d2.pool_evpc.pz.olmo.16.pc\endcsname{75}
\expandafter\def\csname v@d2.pool_evpc_cite.pz.olmo.16.pc\endcsname{70}
\expandafter\def\csname v@d2.pool_prov.pz.olmo.16.pc\endcsname{59}
\expandafter\def\csname v@d2.debate.pz.olmo.32.pc\endcsname{9}
\expandafter\def\csname v@d2.debate_cite.pz.olmo.32.pc\endcsname{7}
\expandafter\def\csname v@d2.pool_evpc.pz.olmo.32.pc\endcsname{78}
\expandafter\def\csname v@d2.pool_evpc_cite.pz.olmo.32.pc\endcsname{74}
\expandafter\def\csname v@d2.pool_prov.pz.olmo.32.pc\endcsname{55}
\expandafter\def\csname v@d2.prov.cite.mq.qwen.pc\endcsname{$-$3 [$-$9, 2]}
\expandafter\def\csname v@d2.prov.debate.mq.qwen.pc\endcsname{$+$44 [37, 50]}
\expandafter\def\csname v@d2.cite.debate.mq.qwen.pc\endcsname{$-$1 [$-$5, 2]}
\expandafter\def\csname v@d2.declared.mq.qwen.pc\endcsname{41}
\expandafter\def\csname v@d2.lexical.mq.qwen.pc\endcsname{57}
\expandafter\def\csname v@d2.final.prov.mq.qwen.pc\endcsname{80}
\expandafter\def\csname v@d2.final.cite.mq.qwen.pc\endcsname{34}
\expandafter\def\csname v@d2.debate.mq.qwen.16.pc\endcsname{12}
\expandafter\def\csname v@d2.debate_cite.mq.qwen.16.pc\endcsname{10}
\expandafter\def\csname v@d2.pool_evpc.mq.qwen.16.pc\endcsname{57}
\expandafter\def\csname v@d2.pool_evpc_cite.mq.qwen.16.pc\endcsname{57}
\expandafter\def\csname v@d2.pool_prov.mq.qwen.16.pc\endcsname{54}
\expandafter\def\csname v@d2.prov.cite.mq.olmo.pc\endcsname{$-$4 [$-$10, 2]}
\expandafter\def\csname v@d2.prov.debate.mq.olmo.pc\endcsname{$+$39 [32, 46]}
\expandafter\def\csname v@d2.cite.debate.mq.olmo.pc\endcsname{$+$3 [$-$2, 7]}
\expandafter\def\csname v@d2.declared.mq.olmo.pc\endcsname{42}
\expandafter\def\csname v@d2.lexical.mq.olmo.pc\endcsname{46}
\expandafter\def\csname v@d2.final.prov.mq.olmo.pc\endcsname{73}
\expandafter\def\csname v@d2.final.cite.mq.olmo.pc\endcsname{33}
\expandafter\def\csname v@d2.debate.mq.olmo.16.pc\endcsname{9}
\expandafter\def\csname v@d2.debate_cite.mq.olmo.16.pc\endcsname{12}
\expandafter\def\csname v@d2.pool_evpc.mq.olmo.16.pc\endcsname{55}
\expandafter\def\csname v@d2.pool_evpc_cite.mq.olmo.16.pc\endcsname{55}
\expandafter\def\csname v@d2.pool_prov.mq.olmo.16.pc\endcsname{51}
\expandafter\def\csname v@d2.prov.cite.hb.range.pc\endcsname{$-$1}
\expandafter\def\csname v@d2.declared.hb.range.pc\endcsname{69--79}
\expandafter\def\csname v@d2.lexical.hb.range.pc\endcsname{72--94}
\expandafter\def\csname v@d2.prov.cite.pz.range.pc\endcsname{$-$16 to $-$15}
\expandafter\def\csname v@d2.declared.pz.range.pc\endcsname{40--43}
\expandafter\def\csname v@d2.lexical.pz.range.pc\endcsname{57--63}
\expandafter\def\csname v@d2.prov.cite.mq.range.pc\endcsname{$-$4 to $-$3}
\expandafter\def\csname v@d2.declared.mq.range.pc\endcsname{41--42}
\expandafter\def\csname v@d2.lexical.mq.range.pc\endcsname{46--57}
\expandafter\def\csname v@d2.cite.prov.pz.range.pc\endcsname{15--16}
\expandafter\def\csname v@d2.prov.debate.range.pc\endcsname{$+$39 to $+$82}
\expandafter\def\csname v@d2.cite.debate.range.pc\endcsname{$-$4 to $+$4}
\expandafter\def\csname v@d2.gate.pc\endcsname{78}
\expandafter\def\csname v@d3.evpc.hist.hb.qwen.pc\endcsname{$+$84 [72, 93]}
\expandafter\def\csname v@d3.evpc.digest.hb.qwen.pc\endcsname{$+$85 [74, 94]}
\expandafter\def\csname v@d3.hist.debate.hb.qwen.pc\endcsname{$-$1 [$-$6, 3]}
\expandafter\def\csname v@d3.digest.cap.hb.qwen.pc\endcsname{$-$6 [$-$9, $-$3]}
\expandafter\def\csname v@d3.debate.hb.qwen.16.pc\endcsname{8}
\expandafter\def\csname v@d3.debate_hist.hb.qwen.16.pc\endcsname{5}
\expandafter\def\csname v@d3.pool_ev_cap.hb.qwen.16.pc\endcsname{16}
\expandafter\def\csname v@d3.pool_digest.hb.qwen.16.pc\endcsname{3}
\expandafter\def\csname v@d3.pool_evpc.hb.qwen.16.pc\endcsname{88}
\expandafter\def\csname v@d3.debate.hb.qwen.32.pc\endcsname{5}
\expandafter\def\csname v@d3.debate_hist.hb.qwen.32.pc\endcsname{6}
\expandafter\def\csname v@d3.pool_ev_cap.hb.qwen.32.pc\endcsname{4}
\expandafter\def\csname v@d3.pool_digest.hb.qwen.32.pc\endcsname{5}
\expandafter\def\csname v@d3.pool_evpc.hb.qwen.32.pc\endcsname{91}
\expandafter\def\csname v@d3.evpc.hist.hb.olmo.pc\endcsname{$+$62 [46, 77]}
\expandafter\def\csname v@d3.evpc.digest.hb.olmo.pc\endcsname{$+$75 [58, 89]}
\expandafter\def\csname v@d3.hist.debate.hb.olmo.pc\endcsname{$+$6 [$-$3, 17]}
\expandafter\def\csname v@d3.digest.cap.hb.olmo.pc\endcsname{$+$5 [0, 10]}
\expandafter\def\csname v@d3.debate.hb.olmo.16.pc\endcsname{21}
\expandafter\def\csname v@d3.debate_hist.hb.olmo.16.pc\endcsname{29}
\expandafter\def\csname v@d3.pool_ev_cap.hb.olmo.16.pc\endcsname{12}
\expandafter\def\csname v@d3.pool_digest.hb.olmo.16.pc\endcsname{15}
\expandafter\def\csname v@d3.pool_evpc.hb.olmo.16.pc\endcsname{88}
\expandafter\def\csname v@d3.debate.hb.olmo.32.pc\endcsname{21}
\expandafter\def\csname v@d3.debate_hist.hb.olmo.32.pc\endcsname{26}
\expandafter\def\csname v@d3.pool_ev_cap.hb.olmo.32.pc\endcsname{8}
\expandafter\def\csname v@d3.pool_digest.hb.olmo.32.pc\endcsname{15}
\expandafter\def\csname v@d3.pool_evpc.hb.olmo.32.pc\endcsname{92}
\expandafter\def\csname v@d3.evpc.hist.pz.qwen.pc\endcsname{$+$76 [69, 82]}
\expandafter\def\csname v@d3.evpc.digest.pz.qwen.pc\endcsname{$+$81 [74, 87]}
\expandafter\def\csname v@d3.hist.debate.pz.qwen.pc\endcsname{0 [$-$2, 2]}
\expandafter\def\csname v@d3.digest.cap.pz.qwen.pc\endcsname{$-$7 [$-$12, $-$3]}
\expandafter\def\csname v@d3.debate.pz.qwen.16.pc\endcsname{10}
\expandafter\def\csname v@d3.debate_hist.pz.qwen.16.pc\endcsname{11}
\expandafter\def\csname v@d3.pool_ev_cap.pz.qwen.16.pc\endcsname{14}
\expandafter\def\csname v@d3.pool_digest.pz.qwen.16.pc\endcsname{6}
\expandafter\def\csname v@d3.pool_evpc.pz.qwen.16.pc\endcsname{85}
\expandafter\def\csname v@d3.debate.pz.qwen.32.pc\endcsname{11}
\expandafter\def\csname v@d3.debate_hist.pz.qwen.32.pc\endcsname{11}
\expandafter\def\csname v@d3.pool_ev_cap.pz.qwen.32.pc\endcsname{13}
\expandafter\def\csname v@d3.pool_digest.pz.qwen.32.pc\endcsname{6}
\expandafter\def\csname v@d3.pool_evpc.pz.qwen.32.pc\endcsname{88}
\expandafter\def\csname v@d3.evpc.hist.pz.olmo.pc\endcsname{$+$66 [59, 73]}
\expandafter\def\csname v@d3.evpc.digest.pz.olmo.pc\endcsname{$+$72 [65, 78]}
\expandafter\def\csname v@d3.hist.debate.pz.olmo.pc\endcsname{$+$2 [0, 4]}
\expandafter\def\csname v@d3.digest.cap.pz.olmo.pc\endcsname{$+$1 [$-$1, 3]}
\expandafter\def\csname v@d3.debate.pz.olmo.16.pc\endcsname{8}
\expandafter\def\csname v@d3.debate_hist.pz.olmo.16.pc\endcsname{12}
\expandafter\def\csname v@d3.pool_ev_cap.pz.olmo.16.pc\endcsname{4}
\expandafter\def\csname v@d3.pool_digest.pz.olmo.16.pc\endcsname{4}
\expandafter\def\csname v@d3.pool_evpc.pz.olmo.16.pc\endcsname{75}
\expandafter\def\csname v@d3.debate.pz.olmo.32.pc\endcsname{9}
\expandafter\def\csname v@d3.debate_hist.pz.olmo.32.pc\endcsname{9}
\expandafter\def\csname v@d3.pool_ev_cap.pz.olmo.32.pc\endcsname{5}
\expandafter\def\csname v@d3.pool_digest.pz.olmo.32.pc\endcsname{7}
\expandafter\def\csname v@d3.pool_evpc.pz.olmo.32.pc\endcsname{78}
\expandafter\def\csname v@d3.evpc.hist.range.pc\endcsname{$+$62 to $+$84}
\expandafter\def\csname v@d3.evpc.digest.range.pc\endcsname{$+$72 to $+$85}
\expandafter\def\csname v@d3.hist.debate.range.pc\endcsname{$-$1 to $+$6}
\expandafter\def\csname v@d3.digest.cap.range.pc\endcsname{$-$7 to $+$5}
\expandafter\def\csname v@d3.hist.debate.max.pc\endcsname{$+$6}
\expandafter\def\csname v@d3.hist.range.pc\endcsname{5--29}
\expandafter\def\csname v@d3.digest.range.pc\endcsname{3--15}
\expandafter\def\csname v@d4.debate.qwen.4.pc\endcsname{20}
\expandafter\def\csname v@d4.pool_evpc.qwen.4.pc\endcsname{56}
\expandafter\def\csname v@d4.full_iter.qwen.4.pc\endcsname{77}
\expandafter\def\csname v@d4.evpc.debate.qwen.4.pc\endcsname{$+$36 [29, 42]}
\expandafter\def\csname v@d4.debate.qwen.8.pc\endcsname{13}
\expandafter\def\csname v@d4.pool_evpc.qwen.8.pc\endcsname{47}
\expandafter\def\csname v@d4.full_iter.qwen.8.pc\endcsname{78}
\expandafter\def\csname v@d4.evpc.debate.qwen.8.pc\endcsname{$+$34 [27, 40]}
\expandafter\def\csname v@d4.debate.qwen.16.pc\endcsname{11}
\expandafter\def\csname v@d4.pool_evpc.qwen.16.pc\endcsname{45}
\expandafter\def\csname v@d4.full_iter.qwen.16.pc\endcsname{70}
\expandafter\def\csname v@d4.evpc.debate.qwen.16.pc\endcsname{$+$34 [27, 40]}
\expandafter\def\csname v@d4.private.qwen.16.pc\endcsname{50}
\expandafter\def\csname v@d4.evpc.sag.qwen.pc\endcsname{11}
\expandafter\def\csname v@d4.full.sag.qwen.pc\endcsname{7}
\expandafter\def\csname v@d4.cross.debate.qwen.pc\endcsname{0 [$-$3, 2]}
\expandafter\def\csname v@d4.cross.evpc.qwen.pc\endcsname{$-$11 [$-$17, $-$6]}
\expandafter\def\csname v@d4.cross.full.qwen.pc\endcsname{$-$11 [$-$16, $-$6]}
\expandafter\def\csname v@d4.debate.olmo.4.pc\endcsname{19}
\expandafter\def\csname v@d4.pool_evpc.olmo.4.pc\endcsname{47}
\expandafter\def\csname v@d4.full_iter.olmo.4.pc\endcsname{65}
\expandafter\def\csname v@d4.evpc.debate.olmo.4.pc\endcsname{$+$29 [22, 36]}
\expandafter\def\csname v@d4.debate.olmo.8.pc\endcsname{14}
\expandafter\def\csname v@d4.pool_evpc.olmo.8.pc\endcsname{46}
\expandafter\def\csname v@d4.full_iter.olmo.8.pc\endcsname{68}
\expandafter\def\csname v@d4.evpc.debate.olmo.8.pc\endcsname{$+$32 [25, 39]}
\expandafter\def\csname v@d4.debate.olmo.16.pc\endcsname{12}
\expandafter\def\csname v@d4.pool_evpc.olmo.16.pc\endcsname{37}
\expandafter\def\csname v@d4.full_iter.olmo.16.pc\endcsname{64}
\expandafter\def\csname v@d4.evpc.debate.olmo.16.pc\endcsname{$+$25 [18, 32]}
\expandafter\def\csname v@d4.private.olmo.16.pc\endcsname{56}
\expandafter\def\csname v@d4.evpc.sag.olmo.pc\endcsname{10}
\expandafter\def\csname v@d4.full.sag.olmo.pc\endcsname{1}
\expandafter\def\csname v@d4.cross.debate.olmo.pc\endcsname{$+$3 [0, 7]}
\expandafter\def\csname v@d4.cross.evpc.olmo.pc\endcsname{$-$17 [$-$24, $-$11]}
\expandafter\def\csname v@d4.cross.full.olmo.pc\endcsname{$-$11 [$-$18, $-$5]}
\expandafter\def\csname v@d4.evpc.debate.range.pc\endcsname{$+$25 to $+$36}
\expandafter\def\csname v@r.test.facts.c0.hb.qwen.pc\endcsname{78}
\expandafter\def\csname v@r.test.facts.c8.hb.qwen.pc\endcsname{70}
\expandafter\def\csname v@r.test.facts.c0.pz.qwen.pc\endcsname{75}
\expandafter\def\csname v@r.test.facts.c8.pz.qwen.pc\endcsname{25}
\expandafter\def\csname v@r.test.facts.c0.hb.olmo.pc\endcsname{67}
\expandafter\def\csname v@r.test.facts.c8.hb.olmo.pc\endcsname{44}
\expandafter\def\csname v@r.test.facts.c0.pz.olmo.pc\endcsname{54}
\expandafter\def\csname v@r.test.facts.c8.pz.olmo.pc\endcsname{9}
\expandafter\def\csname v@r.test.facts.c0.range.pc\endcsname{54--78}
\expandafter\def\csname v@r.test.facts.c8.range.pc\endcsname{9--70}
\expandafter\def\csname v@r.test.argued.18.range.pc\endcsname{11--35}
\expandafter\def\csname v@r.test.argued.81.range.pc\endcsname{95--100}
\expandafter\def\csname v@r.test.facts.drop.range.pc\endcsname{8--49}
\expandafter\def\csname v@search.peer.acc.qwen.pc\endcsname{$+$19 [8, 29]}
\expandafter\def\csname v@search.peer.accp.qwen.pc\endcsname{0.1}
\expandafter\def\csname v@search.peer.tok.qwen.pc\endcsname{67 [65, 69]}
\expandafter\def\csname v@search.peer.tokp.qwen.pc\endcsname{0.0}
\expandafter\def\csname v@search.lead.acc.qwen.pc\endcsname{$+$13 [3, 22]}
\expandafter\def\csname v@search.lead.accp.qwen.pc\endcsname{0.5}
\expandafter\def\csname v@search.lead.tok.qwen.pc\endcsname{91 [91, 92]}
\expandafter\def\csname v@search.lead.tokp.qwen.pc\endcsname{0.0}
\expandafter\def\csname v@search.peer.acc.olmo.pc\endcsname{$+$32 [21, 43]}
\expandafter\def\csname v@search.peer.accp.olmo.pc\endcsname{0.0}
\expandafter\def\csname v@search.peer.tok.olmo.pc\endcsname{78 [76, 80]}
\expandafter\def\csname v@search.peer.tokp.olmo.pc\endcsname{0.0}
\expandafter\def\csname v@search.lead.acc.olmo.pc\endcsname{$+$23 [11, 35]}
\expandafter\def\csname v@search.lead.accp.olmo.pc\endcsname{0.0}
\expandafter\def\csname v@search.lead.tok.olmo.pc\endcsname{93 [93, 94]}
\expandafter\def\csname v@search.lead.tokp.olmo.pc\endcsname{0.0}
\expandafter\def\csname v@search.peer.gain.range.pc\endcsname{19--32}
\expandafter\def\csname v@search.peer.router.range.pc\endcsname{65--71}
\expandafter\def\csname v@search.peer.base.range.pc\endcsname{39--46}
\expandafter\def\csname v@search.lead.gain.range.pc\endcsname{13--23}
\expandafter\def\csname v@search.lead.router.range.pc\endcsname{50--51}
\expandafter\def\csname v@search.lead.base.range.pc\endcsname{27--38}
\expandafter\def\csname v@search.dev.gain.qwen.pc\endcsname{$+$14 [4, 23]}
\expandafter\def\csname v@search.dev.copy.qwen.pc\endcsname{$+$1 [$-$3, 5]}
\expandafter\def\csname v@search.dev.lead.qwen.pc\endcsname{$+$12 [2, 22]}
\expandafter\def\csname v@search.single.qwen.pc\endcsname{$+$45 [27, 64]}
\expandafter\def\csname v@search.shared.qwen.pc\endcsname{$+$11 [$-$1, 23]}
\expandafter\def\csname v@search.dev.gain.olmo.pc\endcsname{$+$15 [7, 24]}
\expandafter\def\csname v@search.dev.copy.olmo.pc\endcsname{$+$3 [$-$4, 9]}
\expandafter\def\csname v@search.dev.lead.olmo.pc\endcsname{0 [$-$9, 9]}
\expandafter\def\csname v@search.single.olmo.pc\endcsname{$+$44 [28, 59]}
\expandafter\def\csname v@search.shared.olmo.pc\endcsname{$+$30 [16, 42]}
\expandafter\def\csname v@search.dev.gain.range.pc\endcsname{$+$14 to $+$15}
\expandafter\def\csname v@search.dev.copy.range.pc\endcsname{$+$1 to $+$3}
\expandafter\def\csname v@search.dev.lead.range.pc\endcsname{0 to $+$12}
\expandafter\def\csname v@search.single.range.pc\endcsname{$+$44 to $+$45}
\expandafter\def\csname v@search.shared.range.pc\endcsname{$+$11 to $+$30}
\expandafter\def\csname v@search.dev.gain.4.range.pc\endcsname{$+$5 to $+$10}
\expandafter\def\csname v@search.dev.gain.8.range.pc\endcsname{$+$9 to $+$17}
\expandafter\def\csname v@search.dev.gain.16.range.pc\endcsname{$+$16 to $+$17}
\expandafter\def\csname v@search.dev.gain.32.range.pc\endcsname{$+$19 to $+$25}
\expandafter\def\csname v@arch.pool.hb.qwen.pc\endcsname{$+$80 [66, 91]}
\expandafter\def\csname v@arch.ow.hb.qwen.pc\endcsname{$+$67 [57, 77]}
\expandafter\def\csname v@arch.hier.hb.qwen.pc\endcsname{$+$60 [50, 69]}
\expandafter\def\csname v@arch.pool.pz.qwen.pc\endcsname{$+$78 [74, 82]}
\expandafter\def\csname v@arch.ow.pz.qwen.pc\endcsname{$+$49 [46, 53]}
\expandafter\def\csname v@arch.hier.pz.qwen.pc\endcsname{$+$41 [38, 45]}
\expandafter\def\csname v@arch.pool.hb.olmo.pc\endcsname{$+$56 [45, 66]}
\expandafter\def\csname v@arch.ow.hb.olmo.pc\endcsname{$+$54 [42, 66]}
\expandafter\def\csname v@arch.hier.hb.olmo.pc\endcsname{$+$53 [41, 64]}
\expandafter\def\csname v@arch.pool.pz.olmo.pc\endcsname{$+$70 [65, 74]}
\expandafter\def\csname v@arch.ow.pz.olmo.pc\endcsname{$+$45 [41, 49]}
\expandafter\def\csname v@arch.hier.pz.olmo.pc\endcsname{$+$32 [29, 36]}
\expandafter\def\csname v@arch.pool.gain.range.pc\endcsname{$+$56 to $+$80}
\expandafter\def\csname v@arch.ow.gain.range.pc\endcsname{$+$45 to $+$67}
\expandafter\def\csname v@arch.hier.gain.range.pc\endcsname{$+$32 to $+$60}
\expandafter\def\csname v@arch.star.gain.range.pc\endcsname{$+$60 to $+$69}
\expandafter\def\csname v@arch.decider.gain.range.pc\endcsname{$+$32 to $+$67}
\expandafter\def\csname v@arch.raw.max.pc\endcsname{5}
\expandafter\def\csname v@arch.evpc.range.pc\endcsname{36--68}
\expandafter\def\csname v@arch.ow.raw.max.pc\endcsname{5}
\expandafter\def\csname v@arch.ow.evpc.range.pc\endcsname{50--68}
\expandafter\def\csname v@arch.hier.raw.max.pc\endcsname{5}
\expandafter\def\csname v@arch.hier.evpc.range.pc\endcsname{36--62}
\expandafter\def\csname v@arch.star.evpc.range.pc\endcsname{65--72}
\expandafter\def\csname v@arch.star.raw.max.pc\endcsname{6}
\expandafter\def\csname v@arch.peer.evpc.range.pc\endcsname{79--90}
\expandafter\def\csname v@arch.peer.raw.range.pc\endcsname{5--23}
\expandafter\def\csname v@arch.below.peer.range.pc\endcsname{16--45}
\expandafter\def\csname v@arch.below.star.range.pc\endcsname{12--24}
\expandafter\def\csname v@arch.dev.pool.gain.range.pc\endcsname{$+$65 to $+$85}
\expandafter\def\csname v@arch.dev.ow.gain.range.pc\endcsname{$+$42 to $+$80}
\expandafter\def\csname v@arch.dev.hier.gain.range.pc\endcsname{$+$29 to $+$64}
\expandafter\def\csname v@arch.dev.star.gain.range.pc\endcsname{$+$57 to $+$82}
\expandafter\def\csname v@arch.dev.below.peer.range.pc\endcsname{6--47}
\expandafter\def\csname v@arch.reach.ow.range.pc\endcsname{81--91}
\expandafter\def\csname v@arch.reach.ow.raw.range.pc\endcsname{51--70}
\expandafter\def\csname v@arch.reach.hier.range.pc\endcsname{58--81}
\expandafter\def\csname v@arch.reach.hier.raw.range.pc\endcsname{21--49}
\expandafter\def\csname v@arch.tok.run.pool.range.pc\endcsname{17--23}
\expandafter\def\csname v@arch.tok.prompt.pool.range.pc\endcsname{12--19}
\expandafter\def\csname v@arch.tok.run.ow.range.pc\endcsname{87--92}
\expandafter\def\csname v@arch.tok.prompt.ow.range.pc\endcsname{16--27}
\expandafter\def\csname v@arch.tok.run.hier.range.pc\endcsname{88--91}
\expandafter\def\csname v@arch.tok.prompt.hier.range.pc\endcsname{50--58}

\makeatletter
\newcommand{\V}[1]{\ifcsname v@#1\endcsname\csname v@#1\endcsname\else\textbf{\textcolor{red}{??\detokenize{#1}}}%
  \PackageWarning{numbers}{undefined value #1}\fi}
\makeatother

\renewcommand{\algorithmiccomment}[1]{\hfill{\color{black!55}$\triangleright$~#1}}

\title{\textbf{Neuro-Inspired Inhibitory Message Passing for\\Accurate and Token-Efficient Multi-Agent Reasoning}}

\author[1]{Nima H. Siboni}
\author[1]{Magdalena Springer}
\author[1,*]{Vahid Rostami}
\affil[1]{Computational Systems Neuroscience, Institute of Zoology, University of Cologne, Germany}
\affil[*]{\fontsize{8}{10}\selectfont Corresponding author: \texttt{vrostami [at] uni-koeln [dot] de}}
\date{}

\begin{document}

\maketitle

\begin{abstract}
\looseness=-1
Multi-agent systems promise to solve problems by bringing together information beyond the reach of any individual
agent. Retrieval and tool use can distribute this information across agents, making communication essential to
collaboration. A central challenge is keeping crucial evidence visible in each agent's context as messages accumulate.
Repeated information can crowd out rare, decisive evidence.
Models of cortical circuits suggest a principle for addressing this imbalance: inhibition can selectively suppress
shared inputs while preserving sensitivity to weaker signals.
We adapt this principle into inhibitory message passing, a training-free communication router that cancels redundant
evidence separately for each evidence sentence and receiver. Each round, it forwards evidence sentences restated in
agents' messages, delivering each at most once to receivers whose own evidence lacks it and excluding the senders'
reasoning and answers.
Against standard message passing, accuracy in peer groups rose by \V{C1.pp.min}--\V{C1.pp.max} percentage
points across three benchmarks, and tokens per run fell by \V{tok.savings.range}.
Gains held across seven open models spanning four to thirty-two billion parameters and two commercial models, in
groups up to sixty-four agents, and across multi-agent architectures including peer-to-peer, orchestrator--worker,
and multilevel hierarchical organizations.
These findings support a scalable design pattern for more accurate and token-efficient multi-agent reasoning.
\end{abstract}

\section{Introduction}
\label{sec:intro}

A group of agents can collectively possess all the evidence needed to solve a task and still reach the wrong conclusion.
Retrieval and tool use give agents access to different evidence, but communication does not guarantee that these pieces
are combined \citep{anthropic2026multiagent}. Evidence shared by many agents can recur throughout a discussion,
while private evidence held by individual agents receives little attention or never reaches the decision maker. Hidden
profiles make this problem explicit: the shared evidence favours a wrong answer, and identifying the correct answer
requires combining private evidence \citep{stasser1985pooling}. Language-model groups exhibit this failure, sometimes
remaining less accurate than a single agent given all the evidence \citep{li2025hiddenbench,anthropic2026multiagent}.

Scaling multi-agent systems raises the stakes of communication design: coordination can yield diminishing returns, and errors
can propagate across agents \citep{kim2025scaling}. Standard message passing can amplify the imbalance between shared
and private evidence. Each sender's message reaches its receivers as written, carrying both the evidence it restates and
the sender's reasoning and answer, whether in a debate between peers or a worker's report to its lead \citep{du2024debate}.
Hearing every sender increases the input each receiver must process, and repetition can sway a model's judgment
\citep{xie2024adaptive,schuster2026whose}. Restricting the number of senders keeps inboxes smaller but risks excluding
the few agents holding decisive evidence. How can we design multi-agent communication that preserves accurate
collective reasoning as groups grow, despite increasing information load and reinforcement of shared mistakes?

\begin{figure*}[!t]
  \centering
  \includegraphics[width=\textwidth]{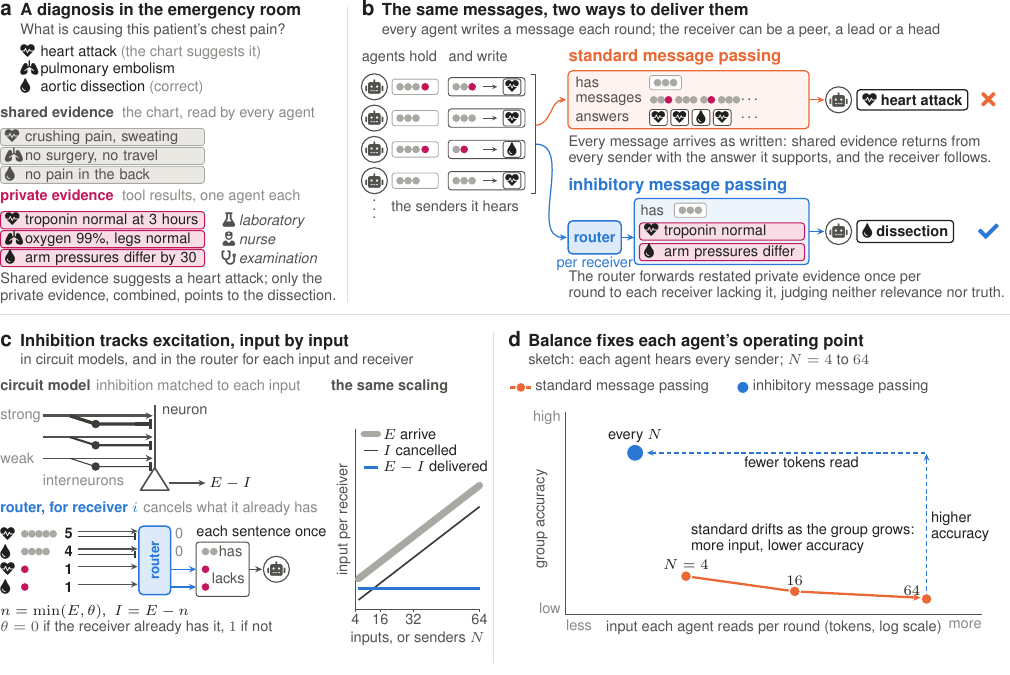}
  \caption{\textbf{Inhibitory message passing gives each agent the distributed evidence it lacks, once per round, and
  cancels what it already has.}
  (\textbf{a})~An illustrative task with distributed evidence. Every agent reads the chart (shared evidence), and three
  agents' own tools each return one result (private evidence, one holder each). The shared evidence suggests a heart
  attack; only the private evidence, combined, points to an aortic dissection.
  (\textbf{b})~The same messages, delivered two ways to a receiver, which can be a peer, a lead or a head. Under standard
  message passing the receiver gets every message as written, with its answer. Under inhibitory message passing a
  router forwards each restated private evidence sentence once per round to every receiver that lacks it
  (\cref{eq:inbox}); the nurse's result has not been restated yet.
  (\textbf{c})~Inhibition tracks excitation input by input, in a circuit model (top) and in the router (bottom;
  \cref{eq:channel}): what the receiver already has is cancelled ($\theta=0$), and each evidence sentence it
  lacks passes once ($\theta=1$). As inputs are added, excitation $E$ and inhibition $I$ grow together while $E-I$ stays flat.
  (\textbf{d})~The expected consequence, sketched for agents that hear every sender and lack a fixed set of evidence:
  as the group grows, standard message passing drifts toward more input and lower accuracy, while inhibitory message
  passing holds each agent at one operating point: the same, smaller input and the same, higher accuracy at every
  $N$. \Cref{fig:balance}a tests the input, \cref{fig:heldout} the accuracy.
  Orange: standard message passing; blue: inhibitory message passing; gray: shared evidence; red: private evidence.}
  \label{fig:principle}
\end{figure*}

Models of cortical circuits suggest a principle for regulating this information flow: inhibition can balance
excitation input by input, counteracting strong or shared drive while preserving responsiveness to weak and changing
inputs \citep{vanvreeswijk1996chaos,vogels2009gating,renart2010asynchronous,rostami2024spiking}. We propose
\textit{inhibitory message passing} as a communication design pattern based on this principle
(\cref{fig:principle}). It couples broad access to distributed evidence with cancellation tailored to each evidence
sentence and receiver. A training-free communication router forwards evidence sentences restated in agents' messages, delivering
each at most once per round to receivers whose own evidence does not contain it. Agents continue to reason and produce
answers, while the router excludes the senders' reasoning and answers. The router requires no additional model calls.

Across the evaluated benchmarks, inhibitory message passing improved group accuracy while reducing token use.
The gains persisted across the tested model families, group sizes and communication architectures
(\cref{fig:heldout}). Controls showed that limiting inbox size alone did not reproduce these gains. Preserving access
to evidence holders, cancelling redundant input for each receiver, and forwarding the evidence itself without the
senders' reasoning and answers all mattered (\cref{fig:balance,fig:controls}). Together, these results support a communication
design pattern for maintaining accurate and efficient collective reasoning as groups grow.

\section{Inhibitory message passing}
\label{sec:method}

\subsection{Distributed evidence}
\label{sec:setting}

In a multi-agent system, $N$ agents work on a task together, and each agent's context holds the task, the evidence the
agent has and whatever arrives from the others. The evidence is of two kinds. Shared evidence is what every agent has,
such as a chart they all read. Private evidence is what one agent alone has, such as the result of its own tool call or
search; we call that agent the holder of that evidence. In any architecture, agents pass on what they hold: each
message carries the evidence its sender relies on and the sender's reasoning and answer. The communication rule, which
decides what from other agents enters an agent's context, therefore determines what each agent can combine.
\Cref{fig:principle}a gives an example. Agents diagnosing a patient with chest pain all read the chart, which suggests
a heart attack, and three of them query their own tools, each obtaining one result. Together the results rule out a
heart attack and point to an aortic dissection, but no agent holds more than one of them: the shared evidence favours a
wrong answer, and only the private evidence combined reveals the correct one. Tasks of this kind are hidden profiles
\citep{stasser1985pooling}, and our benchmarks are built this way: every agent receives the same shared evidence, each
private evidence sentence is given to one holder, and the shared evidence alone leads to a wrong answer or to none
(\cref{app:data}).

The agents communicate in rounds: in round~0 each agent answers alone, and in each later round it reads its own
evidence, its previous message and an inbox, and writes a new message with its reasoning and answer. The two conditions
differ only in the inbox (\cref{fig:principle}b). Under standard message passing, the inbox holds the senders' messages
as written, so the shared evidence returns in almost every message with the answer it supports, and the receiver in
the example concludes heart attack, as most of its senders did. Under inhibitory message passing, a router builds each
inbox from the same messages. When a sender's message restates one of its own evidence sentences, the router forwards that
evidence sentence, once per round, to every receiver that lacks it, and nothing else of the message. What reaches the receiver is the private evidence of others, and with the normal troponin and the unequal arm
pressures in view, the same receiver concludes that the cause is a dissection. The router judges neither the relevance
nor the truth of what it forwards: it only matches what messages restate against the evidence each receiver already
has.

\subsection{The routing rule}
\label{sec:principle}
\label{sec:evidence}

Under both conditions every agent writes a message each round, and the group's answer is read from these messages.
The router changes only what a receiver gets from the senders it hears, by three rules (\cref{fig:controls}a).
(1)~Evidence, not messages: when a sender's message restates one of the sentences the sender itself holds, that
original sentence, not the message, is what can travel. (2)~Once per round: however many senders restate the same
sentence, the receiver gets it once. (3)~Only if missing: a sentence already among the receiver's own evidence is not
delivered. Nothing else of a message travels, neither the sender's
reasoning nor its answer. A message restates an evidence sentence when it repeats enough of the sentence's rare words,
the words that few other sentences of the task share (\cref{app:matching}), so the router makes no model calls and needs
no training.

Written out for round $t\ge1$, with $F_j$ the evidence sentences agent $j$ holds, $\operatorname{src}(m)\subseteq F_j$ those its message
$m$ restates, $C_j^{(t)}$ the sentences its messages have restated up to round $t$
($C_j^{(0)}=\operatorname{src}(m_j^{(0)})$), and
$\mathcal{S}_i^{(t)}$ the senders receiver $i$ hears in round $t$:
\begin{subequations}\label{eq:inbox}
\begin{align}
\text{standard:}\quad \mathrm{inbox}_i^{(t)} &= \big\{\, m_j^{(t-1)} : j\in\mathcal{S}_i^{(t)} \big\}, \label{eq:inbox-std}\\
\text{inhibitory:}\quad \mathrm{inbox}_i^{(t)} &= \Big(\bigcup_{j\in\mathcal{S}_i^{(t)}} C_j^{(t-1)}\Big)\setminus F_i,
\qquad C_j^{(t)} = C_j^{(t-1)}\cup\operatorname{src}\big(m_j^{(t)}\big). \label{eq:inbox-inh}
\end{align}
\end{subequations}
$C_j$ carries rule~(1), the union over senders rule~(2) and the subtraction of $F_i$ rule~(3), while standard message passing (\cref{eq:inbox-std}) delivers each
sender's latest message as written. Delivered sentences are not added to $F_i$, so a sentence the receiver lacks is delivered again every round as long as
some sender has restated it; this matters because an agent sees only its own evidence, its last message and the current inbox.

Counting arrivals shows what the last two rules buy. For receiver $i$ and sentence $u$
in round $t$, let $E_{iu}$ count the senders whose $C_j^{(t-1)}$ holds $u$, the copies that arrive; the router delivers $n_{iu}$
of them and cancels $I_{iu}$:
\begin{equation}
E_{iu}=\sum_{j\in\mathcal{S}_i^{(t)}}\mathbf{1}\big[u\in C_j^{(t-1)}\big],\qquad
n_{iu}=\min\big(E_{iu},\theta_{iu}\big),\qquad
I_{iu}=E_{iu}-n_{iu},
\label{eq:channel}
\end{equation}
where $\mathbf{1}[\cdot]$ is the indicator function and the threshold $\theta_{iu}=\mathbf{1}[u\notin F_i]$ is one
for a sentence the receiver lacks and zero for one it holds; the sentences with $n_{iu}=1$ are exactly the inbox of
\cref{eq:inbox-inh}. Above the threshold, cancellation grows one for one with the arrivals, so an inbox never holds
more than the sentences the receiver lacks, however many senders repeat them (\cref{app:bound}), and an agent that lacks
a fixed set of sentences, as in our benchmarks, reads the same inbox at every group size (\cref{fig:principle}d). This
bound is what lets the router hear every sender the architecture allows, so that every holder of a private evidence
sentence stays within reach (can be heard), whereas a fixed number of senders loses holders as the group grows (\cref{sec:balance}).
Standard message passing cancels nothing and delivers whole messages. \Cref{eq:channel} has the form of
excitation--inhibition balance taken input by input: inhibition that grows with the input it cancels, for each input
separately (\cref{sec:intro}, \cref{fig:principle}c).

Who hears whom is fixed by the architecture, and \cref{alg:router} writes out one round for any sender set. In a peer
group every other peer can be heard; standard message passing hears six random peers per round and the router all of
them (\cref{sec:design} says why). In orchestrator--worker and hierarchical teams, where one agent decides, a lead or head
hears the workers or team leads that report to it, a worker hears only the task its lead sends it, and a team lead's
$C_j$ also holds the sentences it received and restated, so evidence travels up to the deciding head (\cref{app:arch}).

\section{Experiments}
\label{sec:design}

A multi-agent system varies along several axes: its architecture (who hears whom and who decides), its group size, its
model and its task. Our experiments ask two questions: whether the gain of inhibitory message passing holds as these
axes change (\cref{sec:heldout,sec:orch}), and which of its rules carry it: rules~(2) and (3) with the reach they
afford (\cref{sec:balance}), then rule~(1), what travels (\cref{sec:specific}). Both build on a core comparison (\cref{tab:design}, top row) that changes only the communication rule: peer groups of 4 to
32 agents of Qwen3-8B \citep{yang2025qwen3} or Olmo-3-7B \citep{olmo2025olmo3} solve distributed-evidence tasks from
three benchmarks under standard versus inhibitory message passing.

\begin{figure*}[!t]
  \centering
  \includegraphics[width=\textwidth]{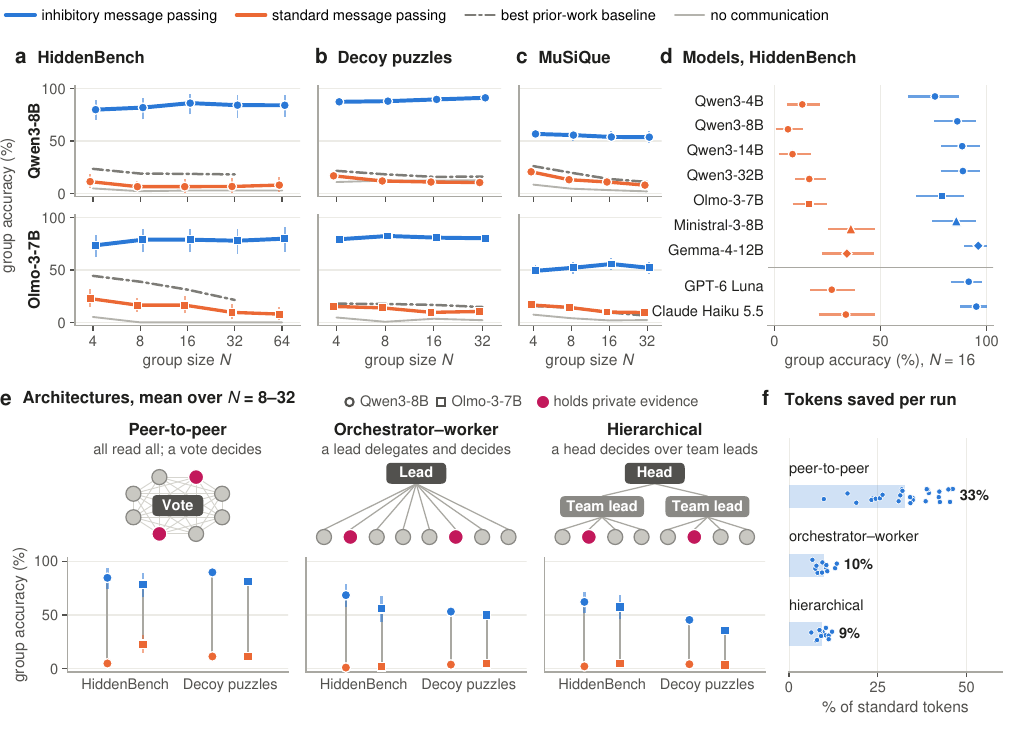}
  \caption{\textbf{Inhibitory message passing raised group accuracy across group size, model and architecture, at fewer
  tokens.} (\textbf{a}--\textbf{c})~Group accuracy against $N$, two main models; the prior-work baseline
  ran up to $N=32$. (\textbf{d})~HiddenBench at
  $N=16$: seven open models; below the rule, two commercial models. (\textbf{e})~Three architectures, group accuracy
  averaged over $N=8$--32. Each line runs from standard (orange) to inhibitory message passing (blue). Circles:
  Qwen3-8B; squares: Olmo-3-7B. Dark boxes mark where the group's answer is decided: a vote, the lead or the head. Here
  peer-to-peer groups hear every peer, unlike a--c. (\textbf{f})~Tokens saved per run by inhibitory message passing, in
  \% of standard message passing's tokens, counting every agent, the decider included. Dots: settings, for peer-to-peer
  those of a--c; bars: means.
  Bars in a--e: 95\% intervals.}
  \label{fig:heldout}
\end{figure*}

In round~0 each agent answers alone; in each of six
further rounds it reads its inbox and writes a message of at most 100 words. Under standard message passing the inbox
holds the latest messages of six random peers, and under the router the evidence sentences every peer's messages have
restated and the receiver lacks. The baseline hears six peers because an inbox of every peer's message grows with the
group: with messages cut to \V{bound.msg.tokens} tokens, it alone would hold \V{bound.inbox.64} tokens at $N=64$,
against a \V{bound.context}-token context (\cref{app:runs}), so a system that scales bounds the number of senders a
receiver hears. Six senders do not handicap the baseline: hearing every peer did not raise its accuracy
(\cref{tab:ablation}), and in the peer groups of \cref{fig:heldout}e both conditions hear every peer.

Because agents receive the same evidence and prompts in both conditions (\cref{tab:prompt}), any difference stems from the inbox: what it
carries and from whom (\cref{sec:balance} separates the two). One group solving one question is a run. We report
group accuracy, the share of runs whose final answer is correct (a plurality vote in peer groups, ties split evenly;
the answer of the lead or head in a team, the decider), and total tokens per run.

Three extensions move one axis at a time, keeping the others as in the core comparison unless \cref{tab:design} says
otherwise: group size, to 64 agents; model, with further open models from 4B to 32B parameters and two commercial
models, GPT-6 Luna and Claude Haiku 5.5; and architecture, with orchestrator--worker teams \citep{anthropic2025research},
in which a lead delegates subtasks and decides, and hierarchical teams, in which a head decides over team leads of
four workers each (\cref{app:arch}). A check then removes our hand from the assignment of holders: 32 agents find
their own evidence by search (\cref{app:search}). To see which rule of \cref{eq:inbox-inh} carries the gain, we then
remove them one at a time in the peer groups and models of the core comparison (\cref{sec:balance,sec:specific}).
\Cref{app:limits} lets the needed evidence grow and adds distractors and a planted false sentence.

\begin{table}[t]
  \caption{\textbf{One core comparison, extended one axis at a time.} Each row keeps the core comparison's settings
  except those listed; the model and architecture extensions and the search check also change the benchmarks and the
  group sizes.
  Every model ran with thinking off (\cref{app:runs,app:g5}). $N$: agents per group (the workers, in teams).}
  \label{tab:design}
  \centering
  \small
  \setlength{\tabcolsep}{4pt}
  \begin{tabular}{@{}>{\raggedright\arraybackslash}p{0.95in}>{\raggedright\arraybackslash}p{2.7in}%
    >{\raggedright\arraybackslash}p{1.6in}>{\raggedright\arraybackslash}p{1.0in}@{}}
    \toprule
    Extension & What changes & Question it answers & Shown in\\
    \midrule
    Core comparison & Standard vs.\ inhibitory message passing in peer groups; Qwen3-8B, Olmo-3-7B; HiddenBench,
    decoy puzzles, MuSiQue; $N=4$--32 & Does the router help a peer group? & Fig.~\ref{fig:heldout}a--c\\
    Group size & $N$ up to 64 on HiddenBench, same evidence & Does accuracy hold as the group grows? &
    Figs.~\ref{fig:heldout}a, \ref{fig:balance}a\\
    Model & Ministral-3-8B \citep{mistral2025ministral3} and Gemma-4-12B \citep{google2026gemma4} (HiddenBench,
    puzzles); Qwen3 at 4B, 14B and 32B, GPT-6 Luna, Claude Haiku 5.5 (HiddenBench) & Is the gain one model's
    weakness? & Fig.~\ref{fig:heldout}d; App.~\ref{app:models}\\
    Architecture & Orchestrator--worker and hierarchical teams (a lead hears its staff), beside peer groups that hear
    every peer; HiddenBench, puzzles; $N=8$--32 & Does the router matter when one agent decides? &
    Fig.~\ref{fig:heldout}e\\
    Check: own searches & Holders set by each agent's own search; MuSiQue; $N=32$; peer groups and a lead that
    delegates nothing & Did assigning the holders do the work? & App.~\ref{app:search}\\
    Taking the router apart & One rule of Eq.~\eqref{eq:inbox} removed, or the threshold of Eq.~\eqref{eq:channel}
    moved or replaced by a random sample of the same size; a full inbox history; an LLM summary of messages; $N=4$--32 & Which of the three rules carries the gain? &
    Figs.~\ref{fig:balance}b, \ref{fig:controls}, Table~\ref{tab:filters}; App.~\ref{app:controls}\\
    \bottomrule
  \end{tabular}
\end{table}

\begin{figure*}[t]
  \centering
  \includegraphics[width=\textwidth]{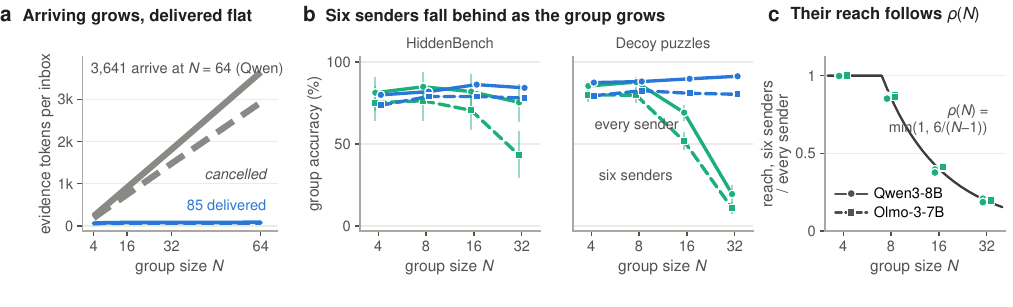}
  \caption{\textbf{Every holder within reach without the repeated input: cancellation absorbs the copies that hearing
  every sender brings, and six senders lose reach as the group grows.}
  (\textbf{a})~HiddenBench, inhibitory message passing: tokens of the evidence sentences per inbox that arrive
  from every sender (grey) and that are delivered (blue; \cref{eq:channel}, summed over sentences); the difference is
  cancelled because the receiver already holds those sentences, the scaling sketched in \cref{fig:principle}c.
  (\textbf{b})~Group accuracy of inhibitory message passing hearing every sender (blue, as in \cref{fig:heldout}a--b)
  or six random senders (green, 95\% intervals); below eight agents, six senders are every sender. (\textbf{c})~Reach
  of six senders relative to every sender on the three benchmarks, against $\rho(N)$ of \cref{eq:reach}: six senders lose
  exactly the reach that a random draw of six predicts. Solid lines and circles: Qwen3-8B; dashed lines and squares: Olmo-3-7B.}
  \label{fig:balance}
\end{figure*}

\begin{figure*}[!t]
  \centering
  \includegraphics[width=\textwidth]{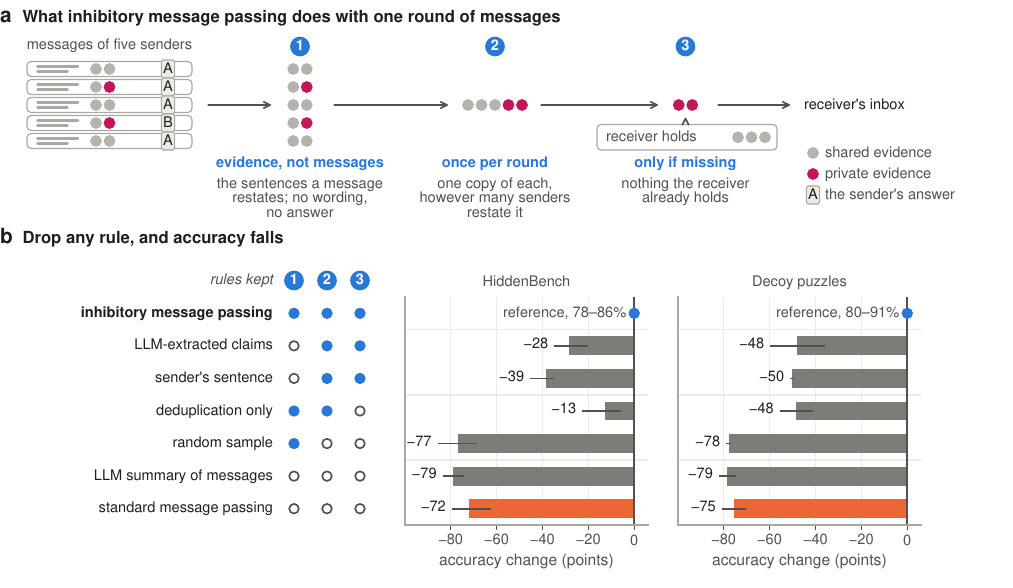}
  \caption{\textbf{Each of the router's three rules is needed.} (\textbf{a})~One round. Agents write and read
  messages as under standard message passing; the router delivers the original evidence sentences their messages
  restate (1), one copy of each (2), only to receivers that lack them (3), and nothing of the messages themselves.
  (\textbf{b})~Variants that drop one or more rules (open circles) lose group accuracy against inhibitory message
  passing: two main models at $N=16$ and 32, paired on items; bars, mean; lines, range. Deduplication only,
  the random sample and the LLM summary are cut to the router's inbox size.}
  \label{fig:controls}
\end{figure*}

The three benchmarks test complementary forms of evidence integration (\cref{app:data}): HiddenBench
\citep{li2025hiddenbench} poses multiple-choice hidden profiles whose shared evidence favours a decoy; the decoy puzzles,
generated with Enigmata \citep{chen2025enigmata}, require multi-step navigation, with shared roads that lead to a wrong
landmark or to none; and MuSiQue-Ans \citep{trivedi2022musique} requires multi-hop synthesis across paragraphs.
Puzzles and MuSiQue questions are screened per model so that an agent holding all the evidence solves them while the
shared evidence alone does not. Each benchmark was split once by item hash into a development set, on which we fixed
the protocol, chose the prior-work baselines and ran most controls of \cref{app:controls}, and a test set, on which the
comparisons of \cref{fig:heldout,fig:balance,fig:controls} ran. The figures also show agents answering alone (no communication) and the strongest of four prior-work baselines
(exchange-then-decide, share-all, one message per answer and anonymous message passing; \cref{app:arms}). The
confirmatory comparisons were pre-registered as one-sided Holm-corrected tests with paired item-bootstrap 95\% intervals,
after excluding items and runs with many unparsed answers
\citep{efron1993bootstrap,holm1979simple,nosek2018preregistration} (\cref{app:prereg}): of the \V{prereg.n} tests of
\cref{tab:confirm}, \V{prereg.pass} passed, and the exception concerns late-arriving evidence rather than the gain.

\section{Results}
\label{sec:results}

\subsection{Inhibitory message passing raised peer-group accuracy at every group size, at lower cost}
\label{sec:heldout}

In the core comparison, inhibitory message passing improved peer-group accuracy over standard message passing
at every group size while reducing token cost (\cref{fig:heldout}a--c). Averaged across group sizes, accuracy rose by
\V{C1.min.pc} to \V{C1.max.pc} percentage points across the three benchmarks and both main models. Prior-work
baselines closed little of this gap: the router still exceeded the strongest of them by \V{C2.min.pc} to
\V{C2.max.pc} points (\cref{tab:confirm}).

This advantage held or widened as groups grew and generated more copies of the shared evidence. On HiddenBench,
the router held Qwen3-8B accuracy steady at \V{Dr.hb.qwen.pool_evpc.pc}\% from 4 agents to the 64 of the
group-size extension
(\cref{fig:heldout}a), whereas standard message passing remained low for Qwen3-8B and degraded for Olmo-3-7B from
\V{D.hb.olmo.debate.4.pc}\% in 4-agent groups to \V{D.hb.olmo.debate.64.pc}\% at 64 agents. The same stability across
group scale appeared on the decoy puzzles and MuSiQue (\cref{fig:heldout}b--c).

Although the router hears every peer rather than six random senders, it still required fewer tokens per run,
because each missing evidence sentence is delivered at most once in place of the senders' reasoning and answers:
across the two main models, three benchmarks and all group sizes, total tokens per run fell by
\V{tok.savings.range} relative to standard message passing (\cref{fig:heldout}f).

\subsection{The gain held across models and architectures}
\label{sec:orch}

Scaling or swapping the language model did not remove the bottleneck of standard message passing. On HiddenBench at
$N=16$, the router outperformed standard message passing in all seven open models, Qwen3 from 4B to 32B
parameters among them (\V{size.evpc.range.pc}\% against \V{size.debate.range.pc}\%), and in both commercial models,
GPT-6 Luna and Claude Haiku 5.5 (\cref{fig:heldout}d; \cref{app:models}); in the pre-registered commercial evaluations
across group sizes, group accuracy rose by \V{g5.diff.est.pc} and \V{g5h.diff.est.pc} percentage points.

Centralizing the decision in an orchestrator lead or a hierarchical head did not rescue standard message passing
either (\cref{fig:heldout}e), as the same communication rule carries the staff's reports up to each lead. Averaged over group
sizes on HiddenBench and the decoy puzzles, every decider scored \V{arch.raw.max.pc}\% or less under standard message
passing and \V{arch.evpc.range.pc}\% under the router. At $N=32$, cancellation shrank the decider's
prompt by roughly four fifths in orchestrator--worker teams and by almost half in hierarchical teams (\cref{tab:arch-inbox}), which
cut total run tokens by about a tenth once the workers' own task prompts are counted (\cref{fig:heldout}f).

Within inhibitory message passing, peer plurality voting remained \V{arch.below.peer.range.pc} percentage points
more accurate than single-decider teams, even though the orchestrator lead's final inbox contained
\V{arch.reach.ow.range.pc}\% of the private evidence sentences (\cref{fig:arch-reach}). How the group combines
evidence and selects its answer therefore remains consequential after communication improves.

Assigning the holders was not what made the difference. When 32 agents instead searched MuSiQue for their own
paragraphs, so that an evidence sentence could have one holder, several or none, the router still exceeded standard message
passing by \V{search.peer.gain.range.pc} percentage points at fewer tokens (\cref{app:search,fig:search}).

\subsection{Cancellation absorbs the copies, so every holder can be heard}
\label{sec:balance}

The two rules of \cref{eq:inbox-inh}, once per round and only if missing, are what kept the inbox, and with it
accuracy, flat as the group grew. On HiddenBench, the number of times each shared evidence sentence arrived at a Qwen3-8B inbox rose from \V{ei.qwen.pool.S.E.4} at
$N=4$ to \V{ei.qwen.pool.S.E.64} at $N=64$, yet the router cancelled every one of them, because every receiver already
held these sentences. Delivered tokens therefore stayed flat while arriving ones grew, the scaling sketched in
\cref{fig:principle}c: at $N=64$, \V{ei.qwen.tok.E.64} tokens of evidence sentences arrived per inbox and
\V{ei.qwen.tok.n.64} were delivered (\cref{fig:balance}a). At the same time, each missing private evidence sentence reached \V{ei.qwen.pool.H.n.range.pc}\% of
the receivers lacking it at every group size and failed to arrive only when its holder had not yet restated it.

Removing both rules (once per round and only if missing) shows what the cancelled copies would have done. With all
restated copies passed through, accuracy at $N=16$ and 32 fell to
\V{fil.nodedup.hb.range.pc}\% on HiddenBench and \V{fil.nodedup.pz.range.pc}\% on the decoy puzzles, at
\V{fil.nodedup.tok.ratio.range} the router's tokens (\cref{tab:filters}). On the puzzles, this accuracy
fell steadily as $N$ grew from 4 to 32 for both models, whereas the router remained stable (\cref{app:copies}). Even in single-receiver tests without
multi-agent interaction, repeating each shared sentence eight times reduced accuracy for both models on HiddenBench and
the puzzles (\cref{app:reader}): repeated copies alone degrade reasoning.

Keeping once per round but dropping only if missing separates the two: under deduplication only, each restated sentence arrives once
but nothing is subtracted, so one copy of every shared evidence sentence still arrives. Cut to the router's inbox size
(the number of sentences the router delivers that round), it reached \V{fil.capdedup.hb.range.pc}\% on HiddenBench and only
\V{fil.capdedup.pz.range.pc}\% on the puzzles, because the shared evidence consumed the slots; without a size limit it
came within \V{fil.ev.gap.range.pc} points of the router at \V{fil.ev.tok.more.range}\% more tokens (\cref{app:filters}).
With one holder per private evidence sentence, the once-per-round rule already removes all but one copy of each shared
evidence sentence, and the only-if-missing rule removes that last copy (\cref{app:dedup}); the first accounts for most
of the gain over all restated copies and the second for the rest.

Because cancellation keeps inboxes compact, the router can afford to hear every peer rather than a random subset
$\mathcal{S}_i^{(t)}$ of $M=6$ peers. With six random senders, a missing evidence sentence held by one agent reaches a
receiver in a given round with probability
\begin{equation}
r=d\,\rho(N),\qquad \rho(N)=\min\big(1,M/(N-1)\big),
\label{eq:reach}
\end{equation}
where $d$ is the probability that the holder has restated the evidence sentence so far and $\rho(N)$ the probability
that the holder is among the $M$ senders drawn from the $N-1$ peers. Hearing every sender keeps $\rho=1$ at every scale, while six senders let $\rho(N)$ decay with $N$: the measured
reach with six senders (the share of receivers lacking an evidence sentence that received it), relative to every sender, tracked
$\rho(N)$ to within \V{pred.Texp.rho.maxdiff} in every setting, and accuracy with six senders matched hearing every
sender in small groups but fell steadily behind as $N$ grew (\cref{fig:balance}b--c). Cancellation thus resolves the dilemma of the introduction: every
holder stays within reach without the repeated input that reach would otherwise bring.

\subsection{What must travel: the original evidence sentence, not the sender's restatement of it}
\label{sec:specific}

Rules (2) and (3) decide how many copies get through; rule (1) decides what each copy is. Before testing it, we ask
whether a filter that knows less could do the router's job at the same inbox size. None did
(\cref{fig:controls}b). A random sample of the same size from all arriving copies, with no threshold per evidence
sentence, rarely selected the rare missing evidence sentences (\cref{app:filters}) and fell
\V{fil.cap.gap.abs.range.pc} percentage points below the router, and an LLM summary of each round's messages, written
within the router's inbox length in words, reached \V{fil.digest.range.pc}\%, no better than the random sample; the
opposite remedy, a full inbox history added to standard message passing, reached \V{fil.hist.range.pc}\% at more
than twice its tokens (\cref{app:filters}).

\label{sec:source}The text that carries each evidence sentence then mattered (\cref{fig:controls}b). Forwarding the
sender's sentence that restates each evidence sentence instead of the original evidence sentence, with the selection
rule unchanged, lowered accuracy (\V{z5.own.range.pc} points). Claims an LLM extracted from the messages, without
seeing the original sentences, fell short in the same way (\V{z5.claims.range.pc} points), whereas a neutral LLM paraphrase of each evidence sentence nearly matched
the original (\V{z5.para.range.pc} points; \cref{app:source}): what matters is where the text comes from, not its exact words. The logs
show why (\cref{app:source}): \V{z5.view.wrong.range.pc}\% of the
senders' sentences came from agents then arguing for a wrong answer, and \V{z5.view.held.range.pc}\% restated, beside
the missing evidence sentence, one the receiver already held, which carried the decoy back into the inbox.

\label{sec:answers}Reintroducing the senders' reasoning and answers beside the routed evidence sentences had the same effect as
forwarding the senders' sentences: at
$N=16$ on HiddenBench it lowered group accuracy from \V{h1.hb16.qwen.pool_evpc.pc}\% to
\V{h1.hb16.qwen.pool_evpc_raw.pc}\% for Qwen3-8B and from \V{h1.hb16.olmo.pool_evpc.pc}\% to
\V{h1.hb16.olmo.pool_evpc_raw.pc}\% for Olmo-3-7B (\cref{app:answers}). By contrast,
forwarding only the sentences the latest message restates, $\operatorname{src}(m_j^{(t-1)})$ in place of the sentences
$j$ has restated so far, $C_j^{(t-1)}$, changed accuracy by \V{u.last.range.pc} points (\cref{app:source}). The evidence itself and the absence of answers matter far
more than the sentences restated in earlier rounds.

\section{Related work}
\label{sec:related}

Like human groups, language-model collectives overweight shared information and underuse evidence held by a small minority
\citep{gigone1993common,lu2012hidden,li2025hiddenbench}; agents withhold information \citep{cemri2025mast} or gather
enough of it yet fail to combine it \citep{zhang2026silobench}. Even when every agent sees the whole problem, debate often does no
better than independent sampling or self-correction at an equal or lower budget
\citep{wang2023selfconsistency,huang2024selfcorrect,smit2024mad,choi2025debate,bertalanic2026consensus}. Recent
interventions preserve dissent, optimize or prune the communication graph, prompt agents to surface private evidence,
choose who speaks, or route messages by stated needs \citep{nguyen2026dar,zhuge2024gptswarm,zhang2025agentprune,
zhang2025gdesigner,li2025hiddenbench,babu2026consilience,zeng2025s2mad,tambwekar2026proxifield};
others compress messages or merge similar ones for each receiver \citep{huang2026pact,guan2026moc}. None removes the
evidence a receiver already holds or delivers each evidence sentence once however many senders repeat it, although the
value of a message depends on what its receiver can already see \citep{xie2026bayes}.

Individual components of our router have precedents: quotes verified against the source text \citep{khan2024debating},
shared blackboards \citep{erman1980hearsay,han2025blackboard}, evidence pooled for a judge without the agents' answers
\citep{ji2026anonymous}, and debate without speaker labels \citep{choi2026identity}. Counting a repeated source once is
the norm of probabilistic inference and decentralized data fusion \citep{pearl1988probabilistic,grime1994data}; its
violation when agents reuse each other's reports is called ``data incest'' \citep{krishnamurthy2014incest,jain2025interacting}. People neglect such
correlations \citep{enke2019correlation}, and LLM aggregators count a restated copy as a fraction of a new observation
unless prompted otherwise \citep{gao2026copies}. Brain-inspired designs have treated agents as cortical areas in a
sparse debate graph \citep{sun2025cortexdebate}. Our contribution combines source grounding, deduplication, and
receiver-specific subtraction into a per-sentence, per-receiver communication rule (\cref{app:related}) and demonstrates
its scaling and architectural generality.

\section{Discussion}
\label{sec:discussion}

As a group of agents grows, its communication has to keep the evidence within reach while limiting the repetition. In
our experiments, inhibitory message passing improved accuracy and reduced token use by forwarding the evidence
sentences that receivers lacked and cancelling the repeated copies, and no filter that merely shortened the inbox
reproduced the gain (\cref{fig:controls}). A communication rule, then, has to account for two things at once: the
evidence a message refers to and the evidence its receiver already holds.

What travels matters as much as how much. Forwarding the original evidence sentence lets a receiver weigh it without
the sender's preferred answer: the sender's sentence and the claims an LLM extracted from the messages carried that
answer along and cost accuracy, whereas a neutral paraphrase did not. Letting the senders' reasoning and answers back in beside the routed evidence
undid much of the gain, as knowing the other members' preferences impedes human groups on hidden profiles
\citep{mojzisch2010knowing}. Excluding the senders' conclusions thus closes one route by which a shared mistake
spreads. Current models know that consensus is not evidence but do not act on that knowledge unprompted
\citep{anthropic2026multiagent}; the router does not ask them to.

For system design, the rule separates the transport of evidence from the exchange of answers (\cref{tab:pattern}).
Applying it needs a reliable link between a message and the evidence it refers to. Word overlap with the holder's own
sentences provides that link without asking anything of the agents; where documents already carry identifiers, a
source tag can provide it directly, as with the declared tags of \cref{app:tags}. With that link in place,
repeated copies of one source count once while independent sources stay distinct, and a receiver's prompt stops
growing with the group. Shorter inputs help in their own right, since models use long contexts less reliably even
when the extra text only duplicates the relevant passages \citep{liu2024lost,levy2024same}, but only if the filter
keeps the missing evidence: delegation, full inbox histories and LLM summaries did not substitute for the
cancellation. The gains across models and architectures suggest that the rule can serve as one reusable component in
several forms of agent organization.

\section{Limitations}
\label{sec:limitations}

Our experiments concern tasks in which agents combine distributed evidence to reach a scored answer, and most scaling
comparisons keep the required evidence fixed as the group grows. The router bounds the repetition; the distinct
evidence a receiver must process can still grow. When the required evidence grew with the group, accuracy declined
as delivery and integration both became limiting (\cref{app:grow}). The results therefore establish scalability with
respect to repeated evidence; growth of the underlying information is further work.

The rule also depends on identifying which evidence a message refers to and what the receiver already holds. Our
implementation matches messages against known evidence sentences (\cref{app:matching}). Less structured settings
need reliable provenance, including the distinction between repeated reports of one observation and independent
observations of the same claim; errors in it would suppress useful evidence or let redundant input accumulate.

The router does not assess the truth or credibility of what it forwards: a planted false sentence cost accuracy
(\cref{app:falsehood}), and the rule has no way to withdraw evidence that is later corrected or outdated. Learning
source reliability and revising or withdrawing evidence are open extensions.

Finally, the router forwards original evidence and excludes the senders' reasoning and answers. Coding, planning and
distributed computation may require workers to exchange intermediate results that no original evidence contains,
and our experiments do not establish when excluding them would impede collaboration. Extending the rule to such
tasks means preserving useful intermediate results while still controlling repetition and the spread of mistaken
conclusions.

\section{Conclusion}
\label{sec:conclusion}

The circuit models that gave the router its rule also suggest where to look next. The router adapts a
simple principle from them, in which inhibition grows with the excitation it meets and a neuron responds to the small difference between the two: strong or shared drive is largely counteracted,
while a weak input still gets through
\citep{vanvreeswijk1996chaos,vogels2009gating,renart2010asynchronous,rostami2024spiking}. \Cref{eq:channel} is this
balance taken input by input, with an evidence sentence as the input and a receiver as the neuron, and
our controls test only this arithmetic: cancellation that grows with repetition carried most of the
gain, its dependence on what the receiver holds carried the rest, and a cap on inbox size alone, the analogue of
balancing excitation and inhibition only in sum, reproduced neither effect. The cortex is far richer than this cartoon,
in its interneuron types, its timescales, its divisive as well as subtractive inhibition and its plasticity
\citep{isaacson2011inhibition,vogels2011inhibitory}; none of that is modelled here, and the connection is a shared
computational principle, not a shared mechanism (\cref{app:mapping}). Those features are also the next steps for
message passing between agents: inhibition that is learned could cancel, per receiver, the sources that have misled it
before; inhibition that divides could weight a sentence by its independent sources; inhibition over several timescales
could let delivered evidence fade or be withdrawn. One piece of this arithmetic already changes what a group of agents
can do.

\section*{Use of large language models}
We conceived the study, designed the experiments, and developed the analyses and scientific interpretation. We used
Claude Opus 5.5 and Claude Fable 5.1 for code generation, debugging, and assistance with manuscript drafting. The scientific
arguments, figure designs, and narrative structure were developed by the authors through substantive writing and
iterative revision. We reviewed and validated the implementation and reported findings and take full responsibility
for the scientific content and conclusions.

\bibliography{references}

@inproceedings{du2024debate,
  author        = {Du, Yilun and Li, Shuang and Torralba, Antonio and Tenenbaum, Joshua B. and Mordatch, Igor},
  title         = {Improving Factuality and Reasoning in Language Models through Multiagent Debate},
  booktitle     = {Proceedings of the 41st International Conference on Machine Learning},
  series        = {Proceedings of Machine Learning Research},
  volume        = {235},
  pages         = {11733--11763},
  publisher     = {PMLR},
  year          = {2024},
  url           = {https://proceedings.mlr.press/v235/du24e.html},
  eprint        = {2305.14325},
  archivePrefix = {arXiv}
}

@inproceedings{choi2025debate,
  author        = {Choi, Hyeong Kyu and Zhu, Xiaojin and Li, Sharon},
  title         = {Debate or Vote: Which Yields Better Decisions in Multi-Agent Large Language Models?},
  booktitle     = {Advances in Neural Information Processing Systems},
  volume        = {38},
  year          = {2025},
  doi           = {10.52202/085713-3405},
  eprint        = {2508.17536},
  archivePrefix = {arXiv}
}

@inproceedings{smit2024mad,
  author        = {Smit, Andries Petrus and Grinsztajn, Nathan and Duckworth, Paul and Barrett, Thomas D. and Pretorius, Arnu},
  title         = {Should we be going {MAD}? {A} Look at Multi-Agent Debate Strategies for {LLMs}},
  booktitle     = {Proceedings of the 41st International Conference on Machine Learning},
  series        = {Proceedings of Machine Learning Research},
  volume        = {235},
  pages         = {45883--45905},
  publisher     = {PMLR},
  year          = {2024},
  url           = {https://proceedings.mlr.press/v235/smit24a.html},
  eprint        = {2311.17371},
  archivePrefix = {arXiv}
}

@inproceedings{wang2023selfconsistency,
  author        = {Wang, Xuezhi and Wei, Jason and Schuurmans, Dale and Le, Quoc V. and Chi, Ed H. and Narang, Sharan and Chowdhery, Aakanksha and Zhou, Denny},
  title         = {Self-Consistency Improves Chain of Thought Reasoning in Language Models},
  booktitle     = {International Conference on Learning Representations},
  year          = {2023},
  url           = {https://openreview.net/forum?id=1PL1NIMMrw},
  eprint        = {2203.11171},
  archivePrefix = {arXiv}
}

@article{wynn2025talk,
  author        = {Wynn, Andrea and Satija, Harsh and Hadfield, Gillian},
  title         = {Talk Isn't Always Cheap: Understanding Failure Modes in Multi-Agent Debate},
  journal       = {arXiv preprint arXiv:2509.05396},
  year          = {2025},
  eprint        = {2509.05396},
  archivePrefix = {arXiv}
}

@article{nguyen2026dar,
  author        = {Nguyen, Manh and Nguyen, Anh and Nguyen, Dung and Venkatesh, Svetha and Le, Hung},
  title         = {Hear Both Sides: Efficient Multi-Agent Debate via Diversity-Aware Message Retention},
  journal       = {arXiv preprint arXiv:2603.20640},
  year          = {2026},
  eprint        = {2603.20640},
  archivePrefix = {arXiv}
}

@inproceedings{weng2025conformity,
  author        = {Weng, Zhiyuan and Chen, Guikun and Wang, Wenguan},
  title         = {Do as We Do, Not as You Think: the Conformity of Large Language Models},
  booktitle     = {International Conference on Learning Representations},
  year          = {2025},
  url           = {https://openreview.net/forum?id=st77ShxP1K},
  eprint        = {2501.13381},
  archivePrefix = {arXiv}
}

@article{kim2025scaling,
  author        = {Kim, Yubin and Gu, Ken and Park, Chanwoo and Park, Chunjong and Schmidgall, Samuel and Heydari, A. Ali and Yan, Yao and Zhang, Zhihan and Zhuang, Yuchen and Liu, Yun and Malhotra, Mark and Liang, Paul Pu and Park, Hae Won and Yang, Yuzhe and Xu, Xuhai and Du, Yilun and Patel, Shwetak and Althoff, Tim and McDuff, Daniel and Liu, Xin},
  title         = {Towards a Science of Scaling Agent Systems},
  journal       = {arXiv preprint arXiv:2512.08296},
  year          = {2025},
  eprint        = {2512.08296},
  archivePrefix = {arXiv}
}

@inproceedings{khan2024debating,
  author        = {Khan, Akbir and Hughes, John and Valentine, Dan and Ruis, Laura and Sachan, Kshitij and Radhakrishnan, Ansh and Grefenstette, Edward and Bowman, Samuel R. and Rockt{\"a}schel, Tim and Perez, Ethan},
  title         = {Debating with More Persuasive {LLMs} Leads to More Truthful Answers},
  booktitle     = {Proceedings of the 41st International Conference on Machine Learning},
  series        = {Proceedings of Machine Learning Research},
  volume        = {235},
  pages         = {23662--23733},
  publisher     = {PMLR},
  year          = {2024},
  url           = {https://proceedings.mlr.press/v235/khan24a.html},
  eprint        = {2402.06782},
  archivePrefix = {arXiv}
}

@inproceedings{zeng2025s2mad,
  author        = {Zeng, Yuting and Huang, Weizhe and Jiang, Lei and Liu, Tongxuan and Jin, Xitai and Tiana, Chen Tianying and Li, Jing and Xu, Xiaohua},
  title         = {{S$^2$-MAD}: Breaking the Token Barrier to Enhance Multi-Agent Debate Efficiency},
  booktitle     = {Proceedings of the 2025 Conference of the Nations of the Americas Chapter of the Association for Computational Linguistics: Human Language Technologies (Volume 1: Long Papers)},
  pages         = {9393--9408},
  publisher     = {Association for Computational Linguistics},
  year          = {2025},
  doi           = {10.18653/v1/2025.naacl-long.475},
  eprint        = {2502.04790},
  archivePrefix = {arXiv}
}

@inproceedings{zhang2025agentprune,
  author        = {Zhang, Guibin and Yue, Yanwei and Li, Zhixun and Yun, Sukwon and Wan, Guancheng and Wang, Kun and Cheng, Dawei and Yu, Jeffrey Xu and Chen, Tianlong},
  title         = {Cut the Crap: An Economical Communication Pipeline for {LLM}-based Multi-Agent Systems},
  booktitle     = {International Conference on Learning Representations},
  year          = {2025},
  url           = {https://openreview.net/forum?id=LkzuPorQ5L},
  eprint        = {2410.02506},
  archivePrefix = {arXiv}
}

@inproceedings{sun2025cortexdebate,
  author    = {Sun, Yiliu and Zhao, Zicheng and Wan, Sheng and Gong, Chen},
  title     = {{CortexDebate}: Debating Sparsely and Equally for Multi-Agent Debate},
  booktitle = {Findings of the Association for Computational Linguistics: ACL 2025},
  pages     = {9503--9523},
  publisher = {Association for Computational Linguistics},
  year      = {2025},
  doi       = {10.18653/v1/2025.findings-acl.495}
}

@misc{anthropic2025research,
  author       = {{Anthropic}},
  title        = {How we built our multi-agent research system},
  howpublished = {Engineering at Anthropic (blog post)},
  month        = jun,
  year         = {2025},
  url          = {https://www.anthropic.com/engineering/multi-agent-research-system},
  note         = {Published 13 June 2025}
}

@inproceedings{li2025hiddenbench,
  author        = {Li, Yuxuan and Naito, Aoi and Shirado, Hirokazu},
  title         = {Systematic Failures in Collective Reasoning under Distributed Information in Multi-Agent {LLMs}},
  booktitle     = {Proceedings of the 43rd International Conference on Machine Learning},
  series        = {Proceedings of Machine Learning Research},
  volume        = {306},
  pages         = {69250--69268},
  publisher     = {PMLR},
  year          = {2026},
  url           = {https://proceedings.mlr.press/v306/li26ej.html},
  eprint        = {2505.11556},
  archivePrefix = {arXiv},
  note          = {Code: \url{https://github.com/Yassellee/HiddenBench_ICML}}
}

@article{babu2026consilience,
  author        = {Babu, Abhijith and Kaur, Ramneet and Pramanik, Vishal and Kotevska, Olivera and Bastian, Nathaniel D. and Jha, Susmit and Raj, Sunny and Wu, Yanzhao and Jha, Sumit Kumar and Roy, Anirban},
  title         = {{Consilience}: Conformally Calibrated Communication Control for Hidden-Profile Multi-Agent Reasoning},
  journal       = {arXiv preprint arXiv:2608.20564},
  year          = {2026},
  eprint        = {2608.20564},
  archivePrefix = {arXiv}
}

@article{li2026infodelphi,
  author        = {Li, Yuante and Tao, Yicheng and Zhang, Kate and Wang, Taozhi and Gu, Gefei and Zhou, Yaxin},
  title         = {Diverse Evidence, Better Forecasts: Multi-Agent Deliberation Under Information Asymmetry},
  journal       = {arXiv preprint arXiv:2607.01661},
  year          = {2026},
  eprint        = {2607.01661},
  archivePrefix = {arXiv}
}

@article{wan2026illusion,
  author        = {Wan, Herun and Wu, Jiaying and Luo, Minnan and Li, Fanxiao and Wang, Ningnan and Chen, Nancy F. and Kan, Min-Yen},
  title         = {The Deliberative Illusion: Diagnosing Factual Attrition and Stance Homogenization in Multi-Agent {LLM} Deliberation},
  journal       = {arXiv preprint arXiv:2606.03032},
  year          = {2026},
  eprint        = {2606.03032},
  archivePrefix = {arXiv}
}

@article{han2025blackboard,
  author        = {Han, Bochen and Zhang, Songmao},
  title         = {Exploring Advanced {LLM} Multi-Agent Systems Based on Blackboard Architecture},
  journal       = {arXiv preprint arXiv:2507.01701},
  year          = {2025},
  eprint        = {2507.01701},
  archivePrefix = {arXiv}
}

@misc{anthropic2026multiagent,
  author        = {{Anthropic}},
  title         = {Patterns and problems in emerging multiagent systems},
  howpublished  = {Research report, Anthropic Frontier Red Team},
  month         = aug,
  year          = {2026},
  url           = {https://www.anthropic.com/research/multiagent-systems},
  note          = {Published 13 August 2026; corresponding author: C. Zou}
}

@article{stasser1985pooling,
  author        = {Stasser, Garold and Titus, William},
  title         = {Pooling of unshared information in group decision making: Biased information sampling during discussion},
  journal       = {Journal of Personality and Social Psychology},
  volume        = {48},
  number        = {6},
  pages         = {1467--1478},
  year          = {1985},
  doi           = {10.1037/0022-3514.48.6.1467}
}

@article{gigone1993common,
  author        = {Gigone, Daniel and Hastie, Reid},
  title         = {The common knowledge effect: Information sharing and group judgment},
  journal       = {Journal of Personality and Social Psychology},
  volume        = {65},
  number        = {5},
  pages         = {959--974},
  year          = {1993},
  doi           = {10.1037/0022-3514.65.5.959}
}

@article{lu2012hidden,
  author        = {Lu, Li and Yuan, Y. Connie and McLeod, Poppy Lauretta},
  title         = {Twenty-Five Years of Hidden Profiles in Group Decision Making: A Meta-Analysis},
  journal       = {Personality and Social Psychology Review},
  volume        = {16},
  number        = {1},
  pages         = {54--75},
  year          = {2012},
  doi           = {10.1177/1088868311417243}
}

@article{trivedi2022musique,
  author        = {Trivedi, Harsh and Balasubramanian, Niranjan and Khot, Tushar and Sabharwal, Ashish},
  title         = {{MuSiQue}: Multihop Questions via Single-hop Question Composition},
  journal       = {Transactions of the Association for Computational Linguistics},
  volume        = {10},
  pages         = {539--554},
  year          = {2022},
  doi           = {10.1162/tacl_a_00475},
  eprint        = {2108.00573},
  archivePrefix = {arXiv}
}

@article{erman1980hearsay,
  author        = {Erman, Lee D. and Hayes-Roth, Frederick and Lesser, Victor R. and Reddy, D. Raj},
  title         = {The {Hearsay-II} Speech-Understanding System: Integrating Knowledge to Resolve Uncertainty},
  journal       = {ACM Computing Surveys},
  volume        = {12},
  number        = {2},
  pages         = {213--253},
  year          = {1980},
  doi           = {10.1145/356810.356816}
}

@article{ji2026anonymous,
  author        = {Ji, Jia-Hao and Li, Sijie and Cheng, Jiabei and She, Zixi and Yu, Jin-Tai and Yuan, Zhiyuan},
  title         = {Candidate supply and answer selection shape the value of {LLM} judging in multi-agent systems},
  journal       = {arXiv preprint arXiv:2608.25937},
  year          = {2026},
  eprint        = {2608.25937},
  archivePrefix = {arXiv},
  primaryClass  = {cs.AI}
}

@article{jain2025interacting,
  author        = {Jain, Adit and Krishnamurthy, Vikram},
  title         = {Interacting Large Language Model Agents. {B}ayesian Social Learning Based Interpretable Models},
  journal       = {IEEE Access},
  volume        = {13},
  pages         = {25465--25504},
  year          = {2025},
  doi           = {10.1109/ACCESS.2025.3538599},
  eprint        = {2411.01271},
  archivePrefix = {arXiv}
}

@article{mojzisch2010knowing,
  author  = {Mojzisch, Andreas and Schulz-Hardt, Stefan},
  title   = {Knowing others' preferences degrades the quality of group decisions},
  journal = {Journal of Personality and Social Psychology},
  volume  = {98},
  number  = {5},
  pages   = {794--808},
  year    = {2010},
  doi     = {10.1037/a0017627}
}

@book{pearl1988probabilistic,
  author    = {Pearl, Judea},
  title     = {Probabilistic Reasoning in Intelligent Systems: Networks of Plausible Inference},
  publisher = {Morgan Kaufmann},
  year      = {1988},
  doi       = {10.1016/C2009-0-27609-4}
}

@article{grime1994data,
  author  = {Grime, S. and Durrant-Whyte, H. F.},
  title   = {Data fusion in decentralized sensor networks},
  journal = {Control Engineering Practice},
  volume  = {2},
  number  = {5},
  pages   = {849--863},
  year    = {1994},
  doi     = {10.1016/0967-0661(94)90349-2}
}

@article{vanvreeswijk1996chaos,
  author        = {van Vreeswijk, C. and Sompolinsky, H.},
  title         = {Chaos in Neuronal Networks with Balanced Excitatory and Inhibitory Activity},
  journal       = {Science},
  volume        = {274},
  number        = {5293},
  pages         = {1724--1726},
  year          = {1996},
  doi           = {10.1126/science.274.5293.1724}
}

@article{rostami2024spiking,
  author        = {Rostami, Vahid and Rost, Thomas and Schmitt, Felix Johannes and van Albada, Sacha Jennifer and Riehle, Alexa and Nawrot, Martin Paul},
  title         = {Spiking attractor model of motor cortex explains modulation of neural and behavioral variability by prior target information},
  journal       = {Nature Communications},
  volume        = {15},
  number        = {1},
  pages         = {6304},
  year          = {2024},
  doi           = {10.1038/s41467-024-49889-4}
}

@article{hassabis2017neuroscience,
  author        = {Hassabis, Demis and Kumaran, Dharshan and Summerfield, Christopher and Botvinick, Matthew},
  title         = {Neuroscience-Inspired Artificial Intelligence},
  journal       = {Neuron},
  volume        = {95},
  number        = {2},
  pages         = {245--258},
  year          = {2017},
  doi           = {10.1016/j.neuron.2017.06.011}
}

@article{renart2010asynchronous,
  author  = {Renart, Alfonso and de la Rocha, Jaime and Bartho, Peter and Hollender, Liad and Parga, N{\'e}stor and Reyes, Alex and Harris, Kenneth D.},
  title   = {The asynchronous state in cortical circuits},
  journal = {Science},
  volume  = {327},
  number  = {5965},
  pages   = {587--590},
  year    = {2010},
  doi     = {10.1126/science.1179850}
}

@article{isaacson2011inhibition,
  author  = {Isaacson, Jeffry S. and Scanziani, Massimo},
  title   = {How inhibition shapes cortical activity},
  journal = {Neuron},
  volume  = {72},
  number  = {2},
  pages   = {231--243},
  year    = {2011},
  doi     = {10.1016/j.neuron.2011.09.027}
}

@article{vogels2009gating,
  author  = {Vogels, Tim P. and Abbott, L. F.},
  title   = {Gating multiple signals through detailed balance of excitation and inhibition in spiking networks},
  journal = {Nature Neuroscience},
  volume  = {12},
  number  = {4},
  pages   = {483--491},
  year    = {2009},
  doi     = {10.1038/nn.2276}
}

@article{vogels2011inhibitory,
  author  = {Vogels, T. P. and Sprekeler, H. and Zenke, F. and Clopath, C. and Gerstner, W.},
  title   = {Inhibitory plasticity balances excitation and inhibition in sensory pathways and memory networks},
  journal = {Science},
  volume  = {334},
  number  = {6062},
  pages   = {1569--1573},
  year    = {2011},
  doi     = {10.1126/science.1211095}
}

@article{yang2025qwen3,
  author        = {Yang, An and Li, Anfeng and Yang, Baosong and Zhang, Beichen and Hui, Binyuan and Zheng, Bo and Yu, Bowen and Gao, Chang and Huang, Chengen and Lv, Chenxu and Zheng, Chujie and Liu, Dayiheng and Zhou, Fan and Huang, Fei and Hu, Feng and others},
  title         = {{Qwen3} Technical Report},
  journal       = {arXiv preprint arXiv:2505.09388},
  year          = {2025},
  url           = {https://arxiv.org/abs/2505.09388},
  eprint        = {2505.09388},
  archivePrefix = {arXiv}
}

@article{olmo2025olmo3,
  author        = {{Team Olmo} and Ettinger, Allyson and Bertsch, Amanda and Kuehl, Bailey and Graham, David and Heineman, David and Groeneveld, Dirk and Brahman, Faeze and Timbers, Finbarr and Ivison, Hamish and Morrison, Jacob and Poznanski, Jake and Lo, Kyle and Soldaini, Luca and Jordan, Matt and others},
  title         = {{Olmo} 3},
  journal       = {arXiv preprint arXiv:2512.13961},
  year          = {2025},
  url           = {https://arxiv.org/abs/2512.13961},
  eprint        = {2512.13961},
  archivePrefix = {arXiv}
}

@article{mistral2025ministral3,
  author        = {Liu, Alexander H. and Khandelwal, Kartik and Subramanian, Sandeep and Jouault, Victor and Rastogi, Abhinav and Sad{\'e}, Adrien and Jeffares, Alan and Jiang, Albert and Cahill, Alexandre and Gavaudan, Alexandre and Sablayrolles, Alexandre and H{\'e}liou, Am{\'e}lie and You, Amos and Ehrenberg, Andy and Lo, Andy and others},
  title         = {{Ministral} 3},
  journal       = {arXiv preprint arXiv:2601.08584},
  year          = {2026},
  url           = {https://arxiv.org/abs/2601.08584},
  eprint        = {2601.08584},
  archivePrefix = {arXiv}
}

@article{google2026gemma4,
  author        = {{Gemma Team} and El Abd, Sherif and Aggarwal, Vaibhav and Algayres, Robin and Andreev, Alek and Bachem, Olivier and Ballantyne, Ian and Brick, Cormac and C{\u{a}}rbune, Victor and Casbon, Michelle and Chaturvedi, Mayank and Chawla, Aditya and Cotruta, Victor and Coucke, Alice and Culliton, Phil and others},
  title         = {{Gemma} 4 Technical Report},
  journal       = {arXiv preprint arXiv:2607.02770},
  year          = {2026},
  url           = {https://arxiv.org/abs/2607.02770},
  eprint        = {2607.02770},
  archivePrefix = {arXiv}
}

@inproceedings{chen2025enigmata,
  author        = {Chen, Jiangjie and He, Qianyu and Yuan, Siyu and Chen, Aili and Cai, Zhicheng and Dai, Weinan and Yu, Hongli and Chen, Jiaze and Li, Xuefeng and Yu, Qiying and Zhou, Hao and Wang, Mingxuan},
  title         = {{Enigmata}: Scaling Logical Reasoning in Large Language Models with Synthetic Verifiable Puzzles},
  booktitle     = {Advances in Neural Information Processing Systems},
  volume        = {38},
  pages         = {4037--4085},
  year          = {2025},
  doi           = {10.52202/085713-0128},
  eprint        = {2505.19914},
  archivePrefix = {arXiv}
}

@book{efron1993bootstrap,
  author        = {Efron, Bradley and Tibshirani, Robert J.},
  title         = {An Introduction to the Bootstrap},
  series        = {Monographs on Statistics and Applied Probability},
  volume        = {57},
  publisher     = {Chapman \& Hall},
  address       = {New York},
  year          = {1993}
}

@article{holm1979simple,
  author        = {Holm, Sture},
  title         = {A Simple Sequentially Rejective Multiple Test Procedure},
  journal       = {Scandinavian Journal of Statistics},
  volume        = {6},
  number        = {2},
  pages         = {65--70},
  year          = {1979},
  url           = {https://www.jstor.org/stable/4615733}
}

@article{nosek2018preregistration,
  author        = {Nosek, Brian A. and Ebersole, Charles R. and DeHaven, Alexander C. and Mellor, David T.},
  title         = {The preregistration revolution},
  journal       = {Proceedings of the National Academy of Sciences},
  volume        = {115},
  number        = {11},
  pages         = {2600--2606},
  year          = {2018},
  doi           = {10.1073/pnas.1708274114}
}

@article{tambwekar2026proxifield,
  author        = {Tambwekar, Pradyumna and Feng, Yenchia and Patel, Deep and Maamari, Karime},
  title         = {{Proxifield}: Decentralized Multi-Agent Communication through Semantic Proximity},
  journal       = {arXiv preprint arXiv:2609.20889},
  year          = {2026},
  eprint        = {2609.20889},
  archivePrefix = {arXiv}
}

@inproceedings{bertalanic2026consensus,
  author        = {Bertalani{\v{c}}, Bla{\v{z}} and Fortuna, Carolina},
  title         = {The Cost of Consensus: Isolated Self-Correction Prevails Over Unguided Homogeneous Multi-Agent Debate},
  booktitle     = {Proceedings of the ACM Conference on AI and Agentic Systems},
  pages         = {311--329},
  publisher     = {Association for Computing Machinery},
  year          = {2026},
  doi           = {10.1145/3786335.3813137},
  eprint        = {2605.00914},
  archivePrefix = {arXiv}
}

@article{robertson2009bm25,
  author        = {Robertson, Stephen and Zaragoza, Hugo},
  title         = {The Probabilistic Relevance Framework: {BM25} and Beyond},
  journal       = {Foundations and Trends in Information Retrieval},
  volume        = {3},
  number        = {4},
  pages         = {333--389},
  year          = {2009},
  doi           = {10.1561/1500000019}
}

@misc{saboo2025adk,
  author        = {Shubham Saboo},
  title         = {Developer's guide to multi-agent patterns in {ADK}},
  howpublished  = {Google Developers Blog},
  month         = dec,
  year          = {2025},
  url           = {https://developers.googleblog.com/en/developers-guide-to-multi-agent-patterns-in-adk/},
  note          = {Published 16 December 2025; the Parallel Fan-Out/Gather pattern}
}

@inproceedings{zhang2026silobench,
  author        = {Zhang, Yuzhe and Liu, Feiran and Shan, Yi and Huang, Xinyi and Yang, Xin and Zhu, Yueqi and Cheng, Xuxin and Liu, Cao and Zeng, Ke and Zhang, Terry Jingchen and Jiang, Wenyuan},
  title         = {{Silo-Bench}: A Scalable Environment for Evaluating Distributed Coordination in Multi-Agent {LLM} Systems},
  booktitle     = {Proceedings of the 64th Annual Meeting of the Association for Computational Linguistics (Volume 1: Long Papers)},
  pages         = {29379--29398},
  publisher     = {Association for Computational Linguistics},
  year          = {2026},
  doi           = {10.18653/v1/2026.acl-long.1354},
  url           = {https://aclanthology.org/2026.acl-long.1354/},
  eprint        = {2603.01045},
  archivePrefix = {arXiv}
}

@inproceedings{choi2026identity,
  author        = {Choi, Hyeong Kyu and Zhu, Xiaojin and Li, Sharon},
  title         = {When Identity Skews Debate: Anonymization for Bias-Reduced Multi-Agent Reasoning},
  booktitle     = {Proceedings of the 64th Annual Meeting of the Association for Computational Linguistics (Volume 1: Long Papers)},
  pages         = {14284--14311},
  publisher     = {Association for Computational Linguistics},
  year          = {2026},
  doi           = {10.18653/v1/2026.acl-long.650},
  url           = {https://aclanthology.org/2026.acl-long.650/},
  eprint        = {2510.07517},
  archivePrefix = {arXiv}
}

@inproceedings{guan2026moc,
  author        = {Guan, Yao and Wang, Lin and Lu, Zhihui and Wang, Ziyi and Yan, Wenzhu and Duan, Qiang},
  title         = {{MOC}: Multi-Order Communication in {LLM}-based Multi-Agent Systems},
  booktitle     = {Proceedings of the 43rd International Conference on Machine Learning},
  series        = {Proceedings of Machine Learning Research},
  volume        = {306},
  pages         = {37603--37619},
  publisher     = {PMLR},
  year          = {2026},
  url           = {https://proceedings.mlr.press/v306/guan26i.html},
  eprint        = {2606.02359},
  archivePrefix = {arXiv}
}

@inproceedings{xie2026bayes,
  author        = {Xie, Yi and Zhou, Zhanke and Fan, Yi and Ge, Yong and Han, Bo and Liu, Bo},
  title         = {Bayes-Sufficient Compression Is Not Enough: How Does Communication Help Multi-Agent Systems?},
  booktitle     = {Advances in Neural Information Processing Systems},
  volume        = {39},
  year          = {2026},
  url           = {https://neurips.cc/virtual/2026/poster/155601},
  eprint        = {2610.03769},
  archivePrefix = {arXiv}
}

@article{huang2026pact,
  author        = {Huang, Chen and Wu, Yuhao and Zhang, Wenxuan},
  title         = {What Should Agents Say? Action-state Communication for Efficient Multi-Agent Systems},
  journal       = {arXiv preprint arXiv:2606.05304},
  year          = {2026},
  eprint        = {2606.05304},
  archivePrefix = {arXiv}
}

@article{gao2026copies,
  author        = {Gao, Jianxin and Li, Runze and Yu, Tianyi and Ren, Liangwei and Chen, Bohan and Wang, Zining},
  title         = {Copies or Sources? Measuring How {LLM} Aggregators Count Restated Evidence in Multi-Agent Systems},
  journal       = {arXiv preprint arXiv:2610.06192},
  year          = {2026},
  eprint        = {2610.06192},
  archivePrefix = {arXiv}
}

@article{enke2019correlation,
  author        = {Enke, Benjamin and Zimmermann, Florian},
  title         = {Correlation Neglect in Belief Formation},
  journal       = {The Review of Economic Studies},
  volume        = {86},
  number        = {1},
  pages         = {313--332},
  year          = {2019},
  doi           = {10.1093/restud/rdx081}
}

@inproceedings{huang2024selfcorrect,
  author        = {Huang, Jie and Chen, Xinyun and Mishra, Swaroop and Zheng, Huaixiu Steven and Yu, Adams Wei and Song, Xinying and Zhou, Denny},
  title         = {Large Language Models Cannot Self-Correct Reasoning Yet},
  booktitle     = {International Conference on Learning Representations},
  year          = {2024},
  url           = {https://openreview.net/forum?id=IkmD3fKBPQ},
  eprint        = {2310.01798},
  archivePrefix = {arXiv}
}

@article{krishnamurthy2014incest,
  author        = {Krishnamurthy, Vikram and Hoiles, William},
  title         = {Online Reputation and Polling Systems: Data Incest, Social Learning, and Revealed Preferences},
  journal       = {IEEE Transactions on Computational Social Systems},
  volume        = {1},
  number        = {3},
  pages         = {164--179},
  year          = {2014},
  doi           = {10.1109/TCSS.2014.2377891}
}

@inproceedings{xie2024adaptive,
  author        = {Xie, Jian and Zhang, Kai and Chen, Jiangjie and Lou, Renze and Su, Yu},
  title         = {Adaptive Chameleon or Stubborn Sloth: Revealing the Behavior of Large Language Models in Knowledge Conflicts},
  booktitle     = {International Conference on Learning Representations},
  year          = {2024},
  url           = {https://openreview.net/forum?id=auKAUJZMO6},
  eprint        = {2305.13300},
  archivePrefix = {arXiv}
}

@inproceedings{schuster2026whose,
  author        = {Schuster, Jakob and Gautam, Vagrant and Markert, Katja},
  title         = {Whose Facts Win? {LLM} Source Preferences under Knowledge Conflicts},
  booktitle     = {Proceedings of the 64th Annual Meeting of the Association for Computational Linguistics (Volume 1: Long Papers)},
  pages         = {29430--29459},
  publisher     = {Association for Computational Linguistics},
  year          = {2026},
  doi           = {10.18653/v1/2026.acl-long.1357},
  eprint        = {2601.03746},
  archivePrefix = {arXiv}
}

@article{liu2024lost,
  author        = {Liu, Nelson F. and Lin, Kevin and Hewitt, John and Paranjape, Ashwin and Bevilacqua, Michele and Petroni, Fabio and Liang, Percy},
  title         = {Lost in the Middle: How Language Models Use Long Contexts},
  journal       = {Transactions of the Association for Computational Linguistics},
  volume        = {12},
  pages         = {157--173},
  year          = {2024},
  doi           = {10.1162/tacl_a_00638}
}

@inproceedings{levy2024same,
  author        = {Levy, Mosh and Jacoby, Alon and Goldberg, Yoav},
  title         = {Same Task, More Tokens: the Impact of Input Length on the Reasoning Performance of Large Language Models},
  booktitle     = {Proceedings of the 62nd Annual Meeting of the Association for Computational Linguistics (Volume 1: Long Papers)},
  pages         = {15339--15353},
  publisher     = {Association for Computational Linguistics},
  year          = {2024},
  doi           = {10.18653/v1/2024.acl-long.818}
}

@inproceedings{cemri2025mast,
  author        = {Cemri, Mert and Pan, Melissa Z. and Yang, Shuyi and Agrawal, Lakshya A. and Chopra, Bhavya and Tiwari, Rishabh and Keutzer, Kurt and Parameswaran, Aditya and Klein, Dan and Ramchandran, Kannan and Zaharia, Matei and Gonzalez, Joseph E. and Stoica, Ion},
  title         = {Why Do Multi-Agent {LLM} Systems Fail?},
  booktitle     = {Advances in Neural Information Processing Systems (Datasets and Benchmarks Track)},
  volume        = {38},
  pages         = {135676--135729},
  year          = {2025},
  doi           = {10.52202/085713-4082},
  url           = {https://neurips.cc/virtual/2025/poster/121528},
  eprint        = {2503.13657},
  archivePrefix = {arXiv}
}

@inproceedings{zhang2025gdesigner,
  author        = {Zhang, Guibin and Yue, Yanwei and Sun, Xiangguo and Wan, Guancheng and Yu, Miao and Fang, Junfeng and Wang, Kun and Chen, Tianlong and Cheng, Dawei},
  title         = {{G-Designer}: Architecting Multi-agent Communication Topologies via Graph Neural Networks},
  booktitle     = {Proceedings of the 42nd International Conference on Machine Learning},
  series        = {Proceedings of Machine Learning Research},
  volume        = {267},
  pages         = {76678--76692},
  publisher     = {PMLR},
  year          = {2025},
  url           = {https://openreview.net/forum?id=LpE54NUnmO},
  eprint        = {2410.11782},
  archivePrefix = {arXiv}
}

@inproceedings{zhuge2024gptswarm,
  author        = {Zhuge, Mingchen and Wang, Wenyi and Kirsch, Louis and Faccio, Francesco and Khizbullin, Dmitrii and Schmidhuber, J{\"u}rgen},
  title         = {{GPTSwarm}: Language Agents as Optimizable Graphs},
  booktitle     = {Proceedings of the 41st International Conference on Machine Learning},
  series        = {Proceedings of Machine Learning Research},
  volume        = {235},
  pages         = {62743--62767},
  publisher     = {PMLR},
  year          = {2024},
  url           = {https://proceedings.mlr.press/v235/zhuge24a.html},
  eprint        = {2402.16823},
  archivePrefix = {arXiv}
}

@inproceedings{heinzelman1999spin,
  author        = {Heinzelman, Wendi Rabiner and Kulik, Joanna and Balakrishnan, Hari},
  title         = {Adaptive Protocols for Information Dissemination in Wireless Sensor Networks},
  booktitle     = {Proceedings of the 5th Annual ACM/IEEE International Conference on Mobile Computing and Networking (MobiCom)},
  pages         = {174--185},
  publisher     = {ACM},
  year          = {1999},
  doi           = {10.1145/313451.313529}
}
\bibliographystyle{plainnat}

\newpage
\appendix
\crefalias{section}{appendix}
\crefalias{subsection}{appendix}
\counterwithin{figure}{section}
\counterwithin{table}{section}
\counterwithin{algorithm}{section}

\FloatBarrier
\section{Benchmarks, models and conditions}

\subsection{Benchmarks}
\label{app:data}

Every item gives all agents the same shared evidence and gives each private evidence sentence to one agent, its holder.
The three benchmarks differ in what the shared evidence does (\cref{tab:bench}). In HiddenBench \citep{li2025hiddenbench}
and in the decoy puzzles, navigation stories generated with Enigmata \citep{chen2025enigmata}, it favours a wrong
answer, so a group that follows what everyone knows goes wrong together; in MuSiQue-Ans \citep{trivedi2022musique} it
merely falls short, and the private paragraphs have to be combined across hops. Each benchmark is split once by item
hash into a development set and a test set. Puzzles and MuSiQue questions were screened for each model with
single-agent calls so that every kept item needs the private evidence: with every evidence sentence the model had to
answer at least half of its samples correctly, and with the shared evidence alone at most a quarter.

\begin{table}[!htb]
\caption{\textbf{Three benchmarks, one construction.} Every agent holds the shared evidence; each private evidence
sentence has one holder.}
\label{tab:bench}
\centering
\small
\begin{tabular}{@{}>{\raggedright\arraybackslash}p{0.12\textwidth}>{\raggedright\arraybackslash}p{0.28\textwidth}%
  >{\raggedright\arraybackslash}p{0.18\textwidth}>{\raggedright\arraybackslash}p{0.14\textwidth}>{\raggedright\arraybackslash}p{0.16\textwidth}@{}}
\toprule
Benchmark & Shared evidence & Private evidence & Items (development / test) & Answer \\
\midrule
HiddenBench & three to eight sentences (median four) that favour one wrong option, the decoy & three or four sentences & 24 / 41 tasks & one of three or four options \\
Decoy puzzles & roads that lead to a wrong landmark (strong decoy) or to none (weak decoy) & four roads & 48 / 100 items per decoy strength & the landmark reached \\
MuSiQue-Ans & four paragraphs that do not answer the question & two to four supporting paragraphs & screened per model & free text, scored by normalized exact match \\
\bottomrule
\end{tabular}
\end{table}

\subsection{Models and runs}
\label{app:runs}

Every open model runs in vLLM at temperature 1 with thinking off and a \V{bound.context}-token context; each reply is
capped at 320 tokens and each delivered message at \V{bound.msg.tokens} tokens, so an inbox that held every peer's
message would outgrow the context in a large group. The peer groups that hear every peer under both conditions
(\cref{fig:heldout}e) therefore ran with contexts of \V{bound.every.context.range} tokens. The two commercial models ran
through their providers' interfaces with the same temperature and reply cap and with reasoning off. The main models
are Qwen3-8B and Olmo-3-7B-Instruct; \cref{app:models} adds five open and two commercial models. Unless a table says
otherwise, we ran 4 seeds on HiddenBench and 3 elsewhere, and the conditions of a seed share its round~0, so that every
comparison is paired. Accuracies are in percent and differences in percentage points throughout.

\subsection{Conditions}
\label{app:arms}

Four conditions in \cref{tab:arms} adapt published remedies for groups that follow their shared evidence: delaying the
vote until the discussion ends, telling every agent to share all it has, hearing one voice per distinct answer, and
hiding the senders' answers. The strongest of them on development items, the best prior-work baseline of
\cref{fig:heldout} and \cref{tab:confirm}, was one message per answer for both main models on every benchmark, except
Olmo-3-7B on HiddenBench, where it was exchange-then-decide. The appendix tables also show two references: no
communication, and full information, in which every agent holds all the evidence and nothing needs to travel.

\begin{table}[!htb]
\caption{\textbf{Conditions.} The last four rows adapt published remedies: exchange-then-decide and share-all from
HiddenBench \citep{li2025hiddenbench} (the share-all wording is ours), one message per answer from the answer-level
filter that diversity-aware message retention (DAR) tested beside its own \citep{nguyen2026dar}, and anonymous message
passing from an anonymous evidence board \citep{ji2026anonymous}. ``6 senders'' in a table marks the router's inbox
built from six random peers.}
\label{tab:arms}
\centering
\small
\begin{tabular}{@{}>{\raggedright\arraybackslash}p{0.25\textwidth}>{\raggedright\arraybackslash}p{0.27\textwidth}>{\raggedright\arraybackslash}p{0.40\textwidth}@{}}
\toprule
Condition & Who a receiver hears & What a receiver sees \\
\midrule
no communication & nobody & its own evidence \\
full information & nobody & all evidence and its last message; ``one answer'': all evidence, answering afresh each round \\
standard message passing & 6 random peers & their latest messages, as written \\
\textbf{inhibitory message passing} & every peer with a parsed answer & evidence sentences that peers' messages restate and it lacks, each once per round; no answers \\
exchange-then-decide & 6 random peers & messages; votes only after the last round \\
one message per answer & one peer per distinct answer & messages as written \\
anonymous message passing & 6 random peers & messages without the senders' answers \\
share-all & 6 random peers & messages, each agent told ``Share all the information you have.'' \\
\bottomrule
\end{tabular}
\end{table}

\FloatBarrier
\section{Prompts}
\label{app:prompts}

Every agent reads the same prompt template in every condition (\cref{tab:prompt}); the conditions differ only in what
fills the inbox. The router adds no instruction: the sentences it forwards arrive under the heading where peers'
messages would.

\begin{table}[!htb]
\caption{\textbf{One prompt template for every condition.} Angle brackets mark what each call fills in. Standard
message passing fills the inbox with the sampled peers' last messages and their answers, inhibitory message passing with
the evidence sentences that peers' messages restate and the receiver lacks, without answers; an empty inbox reads
``(no new messages)''. On MuSiQue the agent gives an answer instead of an option. Two controls add one sentence each:
share-all (\cref{tab:arms}) and declared tags (\cref{app:tags}).}
\label{tab:prompt}
\centering
\scriptsize
\begin{tabular}{@{}p{0.97\textwidth}@{}}
\toprule
\textit{Round 0: each agent answers alone}\par\smallskip
\ttfamily $\langle$task description$\rangle$\par\smallskip
Information available to you:\par
- $\langle$evidence sentence$\rangle$\par
- $\langle$evidence sentence$\rangle$ \ldots\par\smallskip
Options: $\langle$option$\rangle$, $\langle$option$\rangle$, \ldots\par\smallskip
Which option do you choose? Reason in at most 100 words, then end with a last line of the form "Final answer:
$\langle$option$\rangle$". \\
\midrule
\textit{Rounds 1 to 6: the same task description, evidence and options, then}\par\smallskip
\ttfamily Your current choice: $\langle$answer of the previous round$\rangle$\par
Your last message: $\langle$message of the previous round$\rangle$\par\smallskip
Messages from other participants:\par
[1] $\langle$a peer's message$\rangle$ (choice: $\langle$its answer$\rangle$)\hfill\textrm{\textit{standard message
passing}}\par
[1] $\langle$a restated evidence sentence the receiver lacks$\rangle$\hfill\textrm{\textit{inhibitory message passing}}\par
\ldots\par\smallskip
Using their messages as additional information, reconsider your choice. Reason in at most 100 words, then end with a
last line of the form "Final answer: $\langle$option$\rangle$". \\
\bottomrule
\end{tabular}
\end{table}

\FloatBarrier
\section{The routing rule in detail}
\label{app:theory}

\Cref{alg:router} writes out one round of \cref{eq:inbox-inh} for any sender set. The rest of this section gives the
matching rule that detects a restatement, the bound it puts on an inbox, what each cancellation rule removes, and a
key that needs no matching at all.

\begin{algorithm}[!htb]
  \caption{\textbf{Inhibitory message passing sends each restated evidence sentence once per round to every receiver
  that lacks it.} One round $t\ge1$; $C_j$, the sentences $j$ has restated so far, starts from its round-0 message.
  The architecture gives the senders $\mathcal{S}_i^{(t)}$: in a peer group every other agent with a parsed answer, in
  a team the staff that report to a lead; a team lead's $C_j$ also holds the received sentences it restates
  (\cref{app:arch}).}
  \label{alg:router}
  \small
  \begin{algorithmic}[1]
    \FOR{each receiver $i$, with senders $\mathcal{S}_i^{(t)}$}
      \STATE $\mathrm{inbox}_i^{(t)} \gets \big(\bigcup_{j\in\mathcal{S}_i^{(t)}}C_j^{(t-1)}\big)\setminus F_i$
        \COMMENT{evidence sentences the receiver lacks, each once, no answers}
      \STATE agent $i$ reads the question, $F_i$, $m_i^{(t-1)}$ and $\mathrm{inbox}_i^{(t)}$, then writes $m_i^{(t)}$
    \ENDFOR
    \FOR{each agent $j$}
      \STATE $C_j^{(t)} \gets C_j^{(t-1)} \cup \operatorname{src}\big(m_j^{(t)}\big)$
        \COMMENT{sentences $j$ has restated so far; $\operatorname{src}(m_j^{(t)}) = \{u\in F_j : m_j^{(t)} \text{ restates } u\}$}
    \ENDFOR
  \end{algorithmic}
\end{algorithm}

\subsection{Matching}
\label{app:matching}

A message restates one of its sender's evidence sentences when it repeats enough of that sentence's distinctive words.
Words are weighted by inverse document frequency over the item's $U$ evidence sentences,
$\mathrm{idf}=\ln\frac{U+1}{\mathrm{df}+0.5}$, and a message sentence counts as a restatement when it covers at least
half of the evidence sentence's idf weight, or holds at least two of its rare words (words in at most two of the item's
evidence sentences), at least one of them absent from the question and options. The router forwards whole sentences,
so evidence stated together is not split, and it runs centrally, although the rule could run at each receiver. It
identifies a piece of evidence by its sentence, which stands in for the evidence's source only when the sentence has a
single source.

\subsection{A bounded inbox}
\label{app:bound}

Since $n_{iu}\le\theta_{iu}$ in \cref{eq:channel}, the evidence sentences in an inbox take at most
\[
\sum_u n_{iu}\,|u| \;\le\; \sum_{u\notin F_i}|u|
\]
tokens, with $|u|$ the tokens of sentence $u$: a bound set by what the receiver lacks, not by $N$. Without cancellation
they take $\sum_u E_{iu}\,|u|$ tokens, where $E_{iu}$ counts the senders that have restated $u$ and grows with $N$ for
a shared evidence sentence; under standard message passing the inbox holds $|\mathcal{S}_i^{(t)}|$ whole messages.

\subsection{What rules (2) and (3) each remove}
\label{app:dedup}

When each private evidence sentence has one holder, as in every comparison with assigned holders, it arrives
at most once ($E_{iu}\le1$), so cancellation acts only on the shared evidence sentences, which every receiver holds:
once per round removes all but one copy of each, and only if missing removes that last copy. Once per round does more
only when a missing evidence sentence has several holders, as when the agents search for their own evidence
(\cref{app:search}).

\subsection{Declared tags as the key}
\label{app:tags}

\begin{figure}[!htb]
  \centering
  \includegraphics[width=\textwidth]{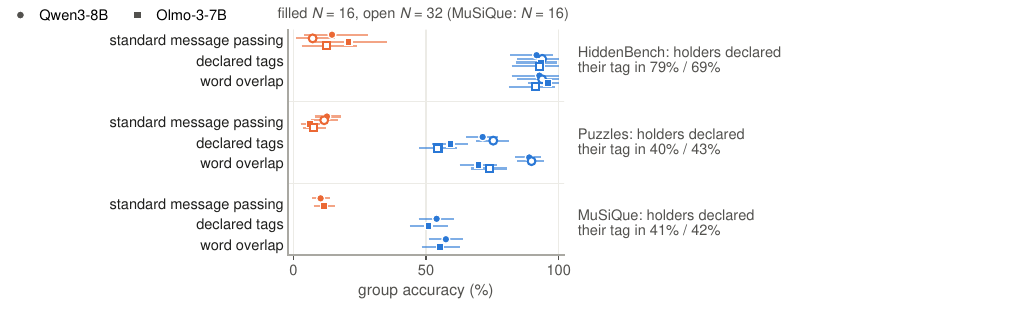}
  \caption{\textbf{Declared tags could serve as the router's key on HiddenBench and MuSiQue but not on the puzzles.}
  Development items. Left: group accuracy with the router keyed on declared tags or on word overlap, and with standard
  message passing shown the tags. Filled: $N=16$; open: $N=32$; MuSiQue only at $N=16$. Right: share of the holders'
  first messages that declared their evidence sentence's tag. 95\% intervals.}
  \label{fig:tags}
\end{figure}

Word overlap is not the only possible key. To test one that needs no matching, every evidence sentence carried a source
tag such as [S3], and every agent was told: ``When you state a fact, cite its tag.'' The router then forwarded each
evidence sentence whose tag a sender had declared. The tag key matched word overlap on HiddenBench
(\V{d2.prov.cite.hb.range.pc} points) and MuSiQue (\V{d2.prov.cite.mq.range.pc}, $N=16$) but fell short on the puzzles
(\V{d2.prov.cite.pz.range.pc}; \cref{fig:tags}).

\subsection{The rule as a design pattern}
\label{app:pattern}

For system designers, \cref{tab:pattern} states the rule in the form of a design pattern: the problem it addresses,
the solution, its consequences, and what it does not do.

\begin{table}[!htb]
  \caption{\textbf{Inhibitory message passing is a design pattern for routing evidence between
  agents.} Token savings: the two main models, against standard message passing.}
  \label{tab:pattern}
  \centering
  \small
  \begin{tabular}{@{}>{\bfseries}l p{0.8\textwidth}@{}}
    \toprule
    Problem & In any system whose agents pass on what they hold, messages repeat the evidence every agent has, while an
    evidence sentence one agent holds reaches few receivers. The repetition grows with the group.\\
    Solution & Each round, forward the evidence sentences each sender's messages restate once to each receiver that lacks them, keyed on
    the source (here word overlap with the holder's sentence; a source tag where available). Omit the senders' reasoning and answers.\\
    Consequences & Higher accuracy; \V{tok.savings.range} fewer tokens per run in peer groups; a shorter prompt for
    the decider; no model calls.\\
    Applies to & Evidence that can be traced to original sentences, passed between agents in any architecture.\\
    Tested on & Peer-to-peer, orchestrator--worker and hierarchical systems; 4 to 64 agents; nine models; evidence
    assigned to agents or found by their own searches (peer groups and a lead that delegates nothing); tasks with a
    scored answer.\\
    Does not do & Check whether an evidence sentence is true. Combine the evidence for the receiver. Carry results agents derive, such as
    code or plans.\\
    \bottomrule
  \end{tabular}
\end{table}

\FloatBarrier
\section{Pre-registered tests}
\label{app:prereg}

The confirmatory claims rest on \V{prereg.n} tests fixed before the test-set runs (\cref{tab:confirm}); \V{prereg.pass}
passed, and the exception is a test of timing, not of the gain. For each main model and benchmark, three tests ask
whether inhibitory message passing beats standard message passing by at least 5 points averaged over $N$, beats the
best prior-work baseline averaged over $N$, and beats its own inbox built from six random senders at the largest $N$.
A fourth test, on HiddenBench at $N=4$, gives each holder its private evidence sentence only in round~3 and asks whether
the router loses a smaller share of its gain over no communication than standard message passing does.

This late-evidence test failed for Qwen3-8B ($p=\V{C4p.hb.qwen}$) because the share it compares is a ratio, and
standard message passing had almost no gain to lose: it gained \V{late.qwen.debate.K0.pc} points over no communication
with early evidence and \V{late.qwen.debate.K3.pc} with late evidence, while the router kept
\V{late.qwen.pool.K3.pc} of its \V{late.qwen.pool.K0.pc} points (Olmo-3-7B: \V{late.olmo.pool.K3.pc} of
\V{late.olmo.pool.K0.pc}). The router's gain survives late evidence; the index measures it poorly when the baseline
has nothing to lose.

Each design was fixed before its test-set runs: the core comparison, the architectures, fan-out/gather (a lead that
reads every worker and delegates nothing; \cref{app:arch}), the agents' own searches and the two commercial models
(\cref{app:models}). The Qwen3 sizes, MuSiQue at $N=32$ and the filters of \cref{tab:filters} ran on test items without
a confirmatory test. Tests are one-sided at $\alpha=0.05$ with Holm's correction over the tests of each
pre-registration (per model for the architectures and own searches); items are the unit, seeds are averaged, and
intervals are paired item bootstraps with 10{,}000 resamples. An item with more than 20\% unparsed answers and a run
with more than 5\% were excluded; keeping every run changed no conclusion, and no design changed after it was fixed.

\begin{table}[!htb]
  \caption{\textbf{Every pre-registered test passed on test items except the late-evidence test for Qwen3-8B on
  HiddenBench.} Estimates with 95\% intervals, after the exclusions. Rows under ``Peers'' give the router's gain in group
  accuracy in the core comparison, in percentage points, over the named condition averaged over $N$ ($N=4$--32; MuSiQue
  4--16), or at the largest $N$ over six senders. The late-evidence row, the primacy index, gives the share of benefit lost when private evidence arrives late, standard
  message passing minus the router. Architecture and fan-out/gather rows average over $N=8$, 16 and 32; in the
  architecture rows for peer-to-peer groups, under both communication rules each agent hears every peer, and the two
  conditions ran on different GPUs, so they do not share round~0. ``All restated copies'' is the router without
  cancellation (\cref{app:copies}). Own-search rows
  are at $N=32$; a token ratio below 1 means fewer tokens. \checkmark/$\times$: passed/failed.}
  \label{tab:confirm}
  \centering
  \scriptsize
  \setlength{\tabcolsep}{3pt}
  \begin{tabular}{@{}l@{\hspace{4pt}}l@{\hspace{6pt}}l@{}}
\toprule
 & Qwen3-8B & Olmo-3-7B \\
\midrule
\multicolumn{3}{@{}l}{\emph{Peers: HiddenBench}} \\
over standard message passing & $+$75 [62, 86]\,\checkmark & $+$60 [50, 71]\,\checkmark  \\
over the best prior-work baseline & $+$63 [52, 72]\,\checkmark & $+$44 [33, 54]\,\checkmark  \\
every sender over six senders & $+$11 [5, 18]\,\checkmark & $+$34 [22, 47]\,\checkmark  \\
late evidence (primacy index) & 0.98 [$-$1.05, 2.21]\,$\times$ & 0.61 [0.13, 0.96]\,\checkmark \\
\midrule
\multicolumn{3}{@{}l}{\emph{Peers: Decoy puzzles}} \\
over standard message passing & $+$77 [72, 81]\,\checkmark & $+$70 [66, 74]\,\checkmark  \\
over the best prior-work baseline & $+$74 [69, 78]\,\checkmark & $+$65 [61, 70]\,\checkmark  \\
every sender over six senders & $+$71 [65, 77]\,\checkmark & $+$71 [66, 76]\,\checkmark  \\
\midrule
\multicolumn{3}{@{}l}{\emph{Peers: MuSiQue}} \\
over standard message passing & $+$41 [36, 45]\,\checkmark & $+$39 [34, 44]\,\checkmark  \\
over the best prior-work baseline & $+$36 [31, 40]\,\checkmark & $+$39 [35, 44]\,\checkmark  \\
every sender over six senders & $+$8 [5, 11]\,\checkmark & $+$11 [7, 15]\,\checkmark  \\
\midrule
\multicolumn{3}{@{}l}{\emph{Architectures, over standard message passing}} \\
Peer-to-peer, HiddenBench & $+$80 [66, 91]\,\checkmark & $+$56 [45, 66]\,\checkmark  \\
Peer-to-peer, Decoy puzzles & $+$78 [74, 82]\,\checkmark & $+$70 [65, 74]\,\checkmark  \\
Orchestrator--worker, HiddenBench & $+$67 [57, 77]\,\checkmark & $+$54 [42, 66]\,\checkmark  \\
Orchestrator--worker, Decoy puzzles & $+$49 [46, 53]\,\checkmark & $+$45 [41, 49]\,\checkmark  \\
Hierarchical, HiddenBench & $+$60 [50, 69]\,\checkmark & $+$53 [41, 64]\,\checkmark  \\
Hierarchical, Decoy puzzles & $+$41 [38, 45]\,\checkmark & $+$32 [29, 36]\,\checkmark  \\
\midrule
\multicolumn{3}{@{}l}{\emph{Fan-out/gather: one lead reads every worker's report}} \\
HiddenBench, over standard message passing & $+$69 [56, 81]\,\checkmark & $+$63 [52, 74]\,\checkmark  \\
HiddenBench, over all restated copies ($N=32$) & $+$52 [41, 64]\,\checkmark & $+$59 [47, 71]\,\checkmark  \\
Decoy puzzles, over standard message passing & $+$60 [56, 64]\,\checkmark & $+$61 [57, 65]\,\checkmark  \\
Decoy puzzles, over all restated copies ($N=32$) & $+$46 [40, 52]\,\checkmark & $+$59 [54, 64]\,\checkmark  \\
\midrule
\multicolumn{3}{@{}l}{\emph{Holders from the agents' own searches: MuSiQue, $N=32$}} \\
Peers, over standard message passing & $+$19 [8, 29]\,\checkmark & $+$32 [21, 43]\,\checkmark  \\
Peers, tokens per run (ratio) & 0.67 [0.65, 0.69]\,\checkmark & 0.78 [0.76, 0.80]\,\checkmark \\
Fan-out/gather lead, over standard message passing & $+$13 [3, 22]\,\checkmark & $+$23 [11, 35]\,\checkmark  \\
Fan-out/gather lead, tokens per run (ratio) & 0.91 [0.91, 0.92]\,\checkmark & 0.93 [0.93, 0.94]\,\checkmark \\
\bottomrule
\end{tabular}

\end{table}

\FloatBarrier
\section{Other models}
\label{app:models}
\label{app:g5}

Neither another model family nor a larger model removed the bottleneck of standard message passing. Ministral-3-8B
\citep{mistral2025ministral3}, Gemma-4-12B \citep{google2026gemma4} and Qwen3 at 4B, 14B and 32B ran on HiddenBench,
the first two also on the decoy puzzles, without a pre-registered test, and for every model and every $N$
the router led standard message passing (\cref{tab:heldout-rep}). For Qwen3-4B, -14B and -32B it led by
\V{size.gain.range.pc} points at every $N$ from 4 to 32 (95\% lower bounds at least \V{size.gain.lo.min.pc} points) at
\V{size.tok.ratio.range.pc}\% of the tokens of standard message passing.

The two commercial models, GPT-6 Luna and Claude Haiku 5.5, had a pre-registered test of their own on HiddenBench,
with 4 seeds, $N=4$ to 32 and the exclusion rules of \cref{app:prereg}: a gain of at least 5 points averaged
over $N$. GPT-6 Luna gained \V{g5.diff.pooled.pc} points ($p<\V{g5.diff.p}$) and Claude Haiku 5.5
\V{g5h.diff.pooled.pc} ($p<\V{g5h.diff.p}$); without exclusions they gained \V{g5.allruns.pc} and \V{g5h.allruns.pc}
(\cref{tab:api}). The exclusions matter here because GPT-6 Luna often abstained in words the answer matcher leaves
unparsed and Claude Haiku 5.5's safety classifier declined some calls; both count as unparsed answers.

\begin{table}[!htb]
\caption{Ministral-3-8B, Gemma-4-12B and Qwen3 at 4B, 14B and 32B: group accuracy under the two
conditions and with full information.}
\label{tab:heldout-rep}
\centering
\scriptsize
\setlength{\tabcolsep}{3pt}
\begin{tabular}{@{}lllcccc@{}}
\toprule
Benchmark & Model & Condition & $N=4$ & $N=8$ & $N=16$ & $N=32$ \\
\midrule
HiddenBench & Ministral-3-8B & standard message passing & 43 & 40 & 36 & 37  \\
 &  & \textbf{inhibitory message passing} & 84 & 82 & 86 & 87  \\
 &  & full information & 66 & 67 & 70 & 80  \\
HiddenBench & Gemma-4-12B & standard message passing & 55 & 39 & 34 & 28  \\
 &  & \textbf{inhibitory message passing} & 95 & 93 & 96 & 96  \\
 &  & full information & 91 & 94 & 97 & 95  \\
HiddenBench & Qwen3-4B & standard message passing & 24 & 17 & 13 & 11  \\
 &  & \textbf{inhibitory message passing} & 74 & 74 & 76 & 69  \\
 &  & full information & 62 & 67 & 71 & 70  \\
HiddenBench & Qwen3-14B & standard message passing & 19 & 10 & 9 & 7  \\
 &  & \textbf{inhibitory message passing} & 88 & 87 & 88 & 89  \\
 &  & full information & 78 & 84 & 81 & 83  \\
HiddenBench & Qwen3-32B & standard message passing & 31 & 20 & 16 & 13  \\
 &  & \textbf{inhibitory message passing} & 91 & 90 & 89 & 89  \\
 &  & full information & 89 & 91 & 92 & 92  \\
\midrule
Decoy puzzles & Ministral-3-8B & standard message passing & 11 & 13 & 9 & 7  \\
 &  & \textbf{inhibitory message passing} & 95 & 95 & 96 & 98  \\
 &  & full information & 94 & 96 & 98 & 99  \\
Decoy puzzles & Gemma-4-12B & standard message passing & 37 & 31 & 25 & 21  \\
 &  & \textbf{inhibitory message passing} & 99 & 99 & 99 & 99  \\
 &  & full information & 100 & 100 & 100 & 100  \\
\bottomrule
\end{tabular}

\end{table}

\begin{table}[!htb]
\caption{\textbf{Two commercial models on HiddenBench.} Group accuracy per condition and $N$, and the
router's gain over standard message passing in percentage points with its 95\% interval; \V{g5.items} tasks for GPT-6
Luna and \V{g5h.items} for Claude Haiku 5.5 after the exclusions.}
\label{tab:api}
\centering
\scriptsize
\setlength{\tabcolsep}{4pt}
\begin{tabular}{@{}llcccc@{}}
\toprule
Model & Condition & $N=4$ & $N=8$ & $N=16$ & $N=32$ \\
\midrule
GPT-6 Luna & standard message passing & 39 & 31 & 27 & 18  \\
 & inhibitory message passing & 94 & 93 & 92 & 91  \\
 & full information & 94 & 96 & 96 & 95  \\
 & gain of the router [95\% CI] & $+$55 [41, 68] & $+$61 [47, 74] & $+$65 [52, 76] & $+$74 [63, 83]  \\
\midrule
Claude Haiku 5.5 & standard message passing & 48 & 36 & 34 & 30  \\
 & inhibitory message passing & 96 & 97 & 95 & 95  \\
 & full information & 91 & 92 & 92 & 92  \\
 & gain of the router [95\% CI] & $+$48 [35, 60] & $+$60 [47, 73] & $+$61 [48, 75] & $+$65 [51, 78]  \\
\bottomrule
\end{tabular}

\end{table}

\FloatBarrier
\section{Team architectures}
\label{app:arch}

In a team, \cref{eq:inbox} holds at every level: a lead or head hears the staff that report to it, and a worker hears
only the task its lead sends. Every round runs in phases, the workers first, then any team leads, then the decider,
each phase reading the reports of the one before. In an orchestrator--worker team the lead writes up to three tasks,
one per line, and its workers receive them in turn; in a hierarchical team the head gives tasks to team leads in the
same way, and each team lead to four workers. Every lead and head holds the question, the options and the shared
evidence. Under the router a report is replaced by the evidence sentences it restates that the receiver lacks, and a
team lead's report also carries the received sentences it restates, so evidence sentences travel up to the decider
verbatim. Fan-out/gather, which \citet{saboo2025adk} describes as a single pass, is here a lead that delegates
nothing: in each round it reads every worker's report, and the workers read its last message.

The router raised group accuracy in every architecture and at every $N$ (\cref{tab:arch}, every run;
\cref{tab:confirm} and the levels of \cref{sec:orch} apply the pre-registered exclusions, which move the gains by a
few points). What changed for the decider shows in its inbox (\cref{tab:arch-inbox}). Under standard message passing
it held \V{arch.copies.raw.range} copies of the shared evidence, more as $N$ grew, and an orchestrator--worker lead saw
\V{arch.reach.ow.raw.range.pc}\% of the private evidence sentences; under the router it held no copy, and the lead saw
\V{arch.reach.ow.range.pc}\%. At $N=32$ the decider's prompt shrank to \V{arch.tok.prompt.ow.range} of its length in
orchestrator--worker teams and \V{arch.tok.prompt.hier.range} in hierarchical teams, and whole runs to
\V{arch.tok.run.ow.range} and \V{arch.tok.run.hier.range}, since the decider's prompts are only
\V{arch.tok.decider.share.range} of a run's tokens under standard message passing. The fan-out/gather lead reached
\V{j3.evpc.range.pc}\% with the router and \V{j3.raw.range.pc}\% without it, and a lead given all restated copies,
still without reasoning and answers, scored \V{J3b.min.pc}--\V{J3b.max.pc} points below the router at $N=32$
(pre-registered, \cref{tab:confirm}): removing the copies helped beyond removing the answers.

\begin{figure}[!htb]
  \centering
  \includegraphics[width=\textwidth]{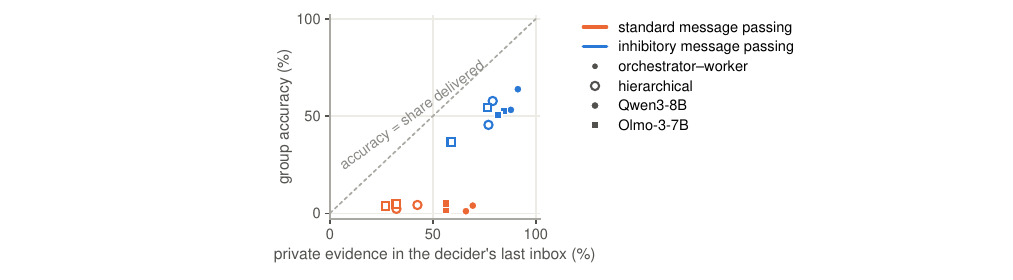}
  \caption{\textbf{Most private evidence sentences reached the decider under the router, yet its accuracy stayed below
  the share delivered.} Share of private evidence in the decider's last inbox against group accuracy, mean over
  $N=8$, 16 and 32. Filled: orchestrator--worker; open: hierarchical. Circles: Qwen3-8B; squares: Olmo-3-7B. Orange: standard
  message passing; blue: inhibitory message passing. Dashed diagonal: accuracy equal to the share delivered.}
  \label{fig:arch-reach}
\end{figure}

Delivery is not the whole story. Under the router each decider still stayed \V{arch.below.peer.range.pc} points below a
peer group on the same items (\cref{tab:arch}), although most private evidence sentences reached it
(\cref{fig:arch-reach}): a single decider is more exposed to its own reasoning errors than a vote.

\begin{table}[!htb]
\caption{\textbf{Inhibitory message passing raised group accuracy in every architecture and at every group
size.} Group accuracy over every run; \cref{tab:confirm} applies the exclusions. Peer-to-peer groups hear every
peer and vote; otherwise the decider answers.}
\label{tab:arch}
\centering
\scriptsize
\setlength{\tabcolsep}{3pt}
\begin{tabular}{@{}ll@{\hspace{8pt}}rrr@{\hspace{10pt}}rrr@{}}
\toprule
 & & \multicolumn{3}{c}{HiddenBench, $N=$} & \multicolumn{3}{c}{Decoy puzzles, $N=$} \\
\cmidrule(lr){3-5}\cmidrule(l){6-8}
Architecture & Condition & 8 & 16 & 32 & 8 & 16 & 32 \\
\midrule
\multicolumn{8}{@{}l}{\emph{Qwen3-8B}} \\
Peer-to-peer & standard & 7 & 4 & 2 & 11 & 12 & 12  \\
 & \textbf{inhibitory} & 76 & 80 & 80 & 88 & 90 & 91  \\
Fan-out/gather & standard & 4 & 2 & 2 & 6 & 6 & 5  \\
 & all restated copies & 43 & 30 & 20 & 22 & 20 & 19  \\
 & \textbf{inhibitory} & 67 & 68 & 68 & 66 & 66 & 65  \\
Orchestrator--worker & standard & 1 & 1 & 1 & 5 & 3 & 4  \\
 & \textbf{inhibitory} & 62 & 66 & 63 & 53 & 50 & 56  \\
Hierarchical & standard & 3 & 2 & 1 & 3 & 6 & 4  \\
 & \textbf{inhibitory} & 58 & 55 & 60 & 47 & 45 & 45  \\
\midrule
\multicolumn{8}{@{}l}{\emph{Olmo-3-7B}} \\
Peer-to-peer & standard & 18 & 23 & 24 & 13 & 12 & 10  \\
 & \textbf{inhibitory} & 73 & 73 & 74 & 82 & 80 & 81  \\
Fan-out/gather & standard & 3 & 1 & 1 & 6 & 7 & 6  \\
 & all restated copies & 17 & 9 & 7 & 7 & 7 & 6  \\
 & \textbf{inhibitory} & 57 & 60 & 62 & 69 & 66 & 65  \\
Orchestrator--worker & standard & 4 & 1 & 1 & 5 & 5 & 5  \\
 & \textbf{inhibitory} & 54 & 54 & 50 & 53 & 50 & 49  \\
Hierarchical & standard & 5 & 5 & 4 & 4 & 2 & 4  \\
 & \textbf{inhibitory} & 51 & 53 & 59 & 39 & 34 & 37  \\
\bottomrule
\end{tabular}

\end{table}

\begin{table}[!htb]
\caption{\textbf{Under the router, the decider's inbox held no copy of the shared evidence and more private evidence, and
its prompt was shorter.} The percentage of private evidence sentences in the decider's last inbox, and the copies of
shared evidence sentences (summed over the shared evidence sentences) and the decider's prompt tokens, averaged over
rounds. Every run.}
\label{tab:arch-inbox}
\centering
\scriptsize
\setlength{\tabcolsep}{3pt}
\begin{tabular}{@{}lll@{\hspace{8pt}}rrr@{\hspace{10pt}}rrr@{}}
\toprule
 & & & \multicolumn{3}{c}{HiddenBench, $N=$} & \multicolumn{3}{c}{Decoy puzzles, $N=$} \\
\cmidrule(lr){4-6}\cmidrule(l){7-9}
Decider & Measure & Condition & 8 & 16 & 32 & 8 & 16 & 32 \\
\midrule
\multicolumn{9}{@{}l}{\emph{Qwen3-8B}} \\
Lead & private evidence reached & standard & 62 & 68 & 68 & 67 & 70 & 70  \\
 &  & \textbf{inhibitory} & 91 & 91 & 91 & 88 & 88 & 88  \\
 & shared-evidence copies & standard & 13.5 & 27.0 & 58.3 & 18.4 & 38.4 & 76.0 \\
 &  & \textbf{inhibitory} & 0.0 & 0.0 & 0.0 & 0.0 & 0.0 & 0.0 \\
 & prompt tokens & standard & 1,196 & 1,834 & 3,134 & 1,108 & 1,809 & 3,156 \\
 &  & \textbf{inhibitory} & 662 & 659 & 665 & 511 & 510 & 516 \\
Head & private evidence reached & standard & 25 & 32 & 40 & 35 & 44 & 49  \\
 &  & \textbf{inhibitory} & 76 & 80 & 81 & 77 & 76 & 77  \\
 & shared-evidence copies & standard & 2.6 & 5.4 & 10.9 & 3.7 & 7.3 & 14.8 \\
 &  & \textbf{inhibitory} & 0.0 & 0.0 & 0.0 & 0.0 & 0.0 & 0.0 \\
 & prompt tokens & standard & 719 & 861 & 1,144 & 589 & 730 & 1,023 \\
 &  & \textbf{inhibitory} & 650 & 655 & 664 & 512 & 513 & 522 \\
\midrule
\multicolumn{9}{@{}l}{\emph{Olmo-3-7B}} \\
Lead & private evidence reached & standard & 51 & 56 & 63 & 53 & 58 & 58  \\
 &  & \textbf{inhibitory} & 85 & 84 & 84 & 82 & 82 & 81  \\
 & shared-evidence copies & standard & 10.7 & 21.0 & 44.0 & 15.6 & 31.8 & 63.9 \\
 &  & \textbf{inhibitory} & 0.0 & 0.0 & 0.0 & 0.0 & 0.0 & 0.0 \\
 & prompt tokens & standard & 1,101 & 1,580 & 2,536 & 963 & 1,472 & 2,506 \\
 &  & \textbf{inhibitory} & 681 & 679 & 679 & 496 & 495 & 497 \\
Head & private evidence reached & standard & 22 & 32 & 42 & 21 & 27 & 33  \\
 &  & \textbf{inhibitory} & 74 & 77 & 78 & 60 & 58 & 59  \\
 & shared-evidence copies & standard & 2.7 & 5.6 & 11.4 & 3.5 & 6.9 & 14.0 \\
 &  & \textbf{inhibitory} & 0.0 & 0.0 & 0.0 & 0.0 & 0.0 & 0.0 \\
 & prompt tokens & standard & 746 & 888 & 1,178 & 581 & 714 & 985 \\
 &  & \textbf{inhibitory} & 670 & 676 & 678 & 487 & 488 & 491 \\
\bottomrule
\end{tabular}

\end{table}

\FloatBarrier
\section{Holders from the agents' own searches}
\label{app:search}

\begin{figure}[!htb]
  \centering
  \includegraphics[width=\textwidth]{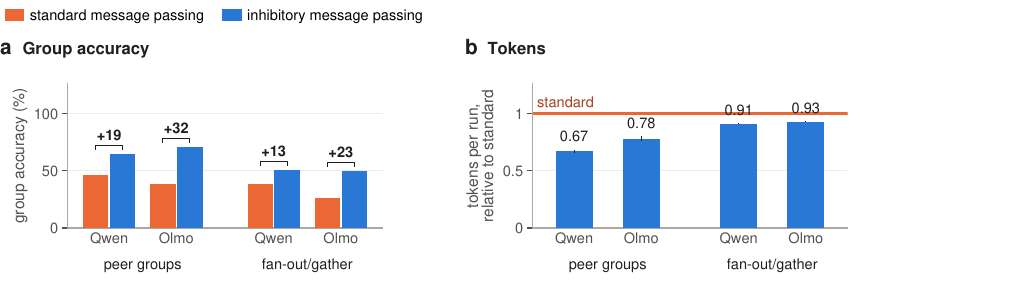}
  \caption{\textbf{When the agents' own searches set the holders, inhibitory message passing raised group accuracy at
  fewer tokens.} MuSiQue, $N=32$. (\textbf{a})~Group accuracy of peer groups and of a fan-out/gather lead;
  \cref{tab:confirm} gives the intervals. (\textbf{b})~Tokens per run relative to standard message passing, with 95\%
  intervals.}
  \label{fig:search}
\end{figure}

Here no one assigns the holders. For each MuSiQue question, each of 32 agents wrote one search query of at most 12
words, BM25 \citep{robertson2009bm25} ranked the question's 20 supporting and distractor paragraphs, and the agent held
its top three in every round and condition. A group of $N$ is the first $N$ of these agents, so a supporting paragraph
can have one holder, several or none, and once per round now has work to do: in a one-seed pilot at
$N=32$, forwarding all restated copies matched the router's accuracy at \V{search.dev.copytok.range} times its tokens.
Questions were kept when the model scored at most 25\% closed book and with one agent's paragraphs, and at least 50\%
with the union of 16 agents' paragraphs: \V{search.test.qwen} questions for Qwen3-8B and \V{search.test.olmo} for
Olmo-3-7B. On these, every pre-registered test at $N=32$ passed (\cref{tab:confirm}): the router raised group accuracy
by \V{search.peer.gain.range.pc} points in peer groups and \V{search.lead.gain.range.pc} for a fan-out/gather lead, at
fewer tokens (\cref{fig:search}).

\FloatBarrier
\section{Taking the router apart}
\label{app:controls}

\Cref{sec:balance,sec:specific} remove the router's rules one at a time. This section adds what the main text leaves
out, in four questions: what repeated copies do to a group and to a single receiver, whether a smaller inbox alone
explains the gain, whether the text that carries the evidence matters, and what the senders' reasoning and answers do
once they are let back in. \Cref{tab:variants} defines every variant by the rule it drops and what the receiver gets,
and \cref{tab:ablation} gives the levels of the variants it lists, on development items. Unless a caption says otherwise, all
used both main models on HiddenBench and the decoy puzzles at $N=16$ and 32.

\begin{table}[!htb]
\caption{\textbf{The variants of the router.} Each keeps the router's matching and its sender set and changes one
thing. The rules are those of \cref{eq:inbox-inh}: (1) evidence, not messages; (2) once per round; (3) only if
missing.}
\label{tab:variants}
\centering
\small
\begin{tabular}{@{}>{\raggedright\arraybackslash}p{0.26\textwidth}>{\raggedright\arraybackslash}p{0.14\textwidth}>{\raggedright\arraybackslash}p{0.52\textwidth}@{}}
\toprule
Variant & Drops & What the receiver gets \\
\midrule
LLM-extracted claims & (1) & the factual claims an LLM lists from each sender's message, without its opinions or conclusions \\
sender's sentence & (1) & the sentence of the sender's latest message that restated each routed evidence sentence \\
sender's sentence, all copies & (1), (2), (3) & those sentences from every sender, without cancellation \\
LLM paraphrase of the evidence & nothing; the text reworded & each routed evidence sentence reworded once by the same model \\
latest message only & nothing; the record shortened & only the sentences restated in the sender's latest message, not all it has restated \\
+ answers only & nothing; answers added & the routed sentences and each sender's answer \\
+ reasoning and answers & nothing; messages added & the routed sentences and each sender's whole message \\
deduplication only & (3) & each distinct arriving sentence once, held or not, cut at random to the router's inbox size \\
deduplication only, no size limit & (3) & the same, uncut \\
random sample & (2), (3) & as many sentences as the router would deliver that round, drawn at random from all arriving copies \\
all restated copies & (2), (3) & every arriving copy, once per sender that restated it \\
LLM summary of messages & (1), (2), (3) & one LLM summary of every agent's last message and answer, written within the router's inbox length \\
+ full inbox history & (1), (2), (3) & standard message passing, with all of the receiver's earlier inboxes \\
\bottomrule
\end{tabular}
\end{table}

\begin{table}[!htb]
\caption{\textbf{Taking the router apart on development items} (group accuracy, Qwen3-8B / Olmo-3-7B; HB:
HiddenBench; puzzles by decoy strength and MQ (MuSiQue) at $N=16$). The third row is standard message passing hearing
every peer (\cref{sec:design}); the indented rows are variants of the router (\cref{tab:variants}). $^\dagger$:
exploratory, run after the design was fixed.}
\label{tab:ablation}
\centering
\scriptsize
\setlength{\tabcolsep}{3pt}
\begin{tabular}{@{}lccccc@{}}
\toprule
Condition & HB $N{=}16$ & HB $N{=}32$ & puzzles strong & puzzles weak & MQ $N{=}16$ \\
\midrule
no communication & 2 / 11 & 1 / 9 & 0 / 0 & 16 / 7 & 5 / 1  \\
standard message passing & 8 / 21 & 5 / 21 & 0 / 1 & 19 / 15 & 12 / 9  \\
standard message passing with every sender$^\dagger$ & 2 / 23 & 3 / 20 & 1 / 5 & 17 / 14 & --  \\
inhibitory message passing, 6 senders & 88 / 81 & 79 / 57 & 45 / 36 & 74 / 52 & 52 / 42  \\
\textbf{inhibitory message passing} & 88 / 88 & 91 / 92 & 87 / 76 & 84 / 74 & 57 / 55  \\
\quad + answers only$^\dagger$ & 84 / 90 & 84 / 94 & 81 / 62 & 67 / 57 & 42 / 30  \\
\quad + reasoning and answers$^\dagger$ & 44 / 54 & 31 / 48 & 3 / 9 & 40 / 16 & --  \\
\quad deduplication only, no size limit$^\dagger$ & 85 / 86 & 86 / 88 & 77 / 62 & 75 / 59 & 52 / 44  \\
\quad all restated copies$^\dagger$ & 70 / 54 & 60 / 55 & 1 / 0 & 34 / 13 & --  \\
full information & 91 / 91 & -- & 96 / 81 & 91 / 87 & 81 / 75  \\
\bottomrule
\end{tabular}

\end{table}

\subsection{Repeated copies}
\label{app:copies}

\begin{figure}[!htb]
  \centering
  \includegraphics[width=\textwidth]{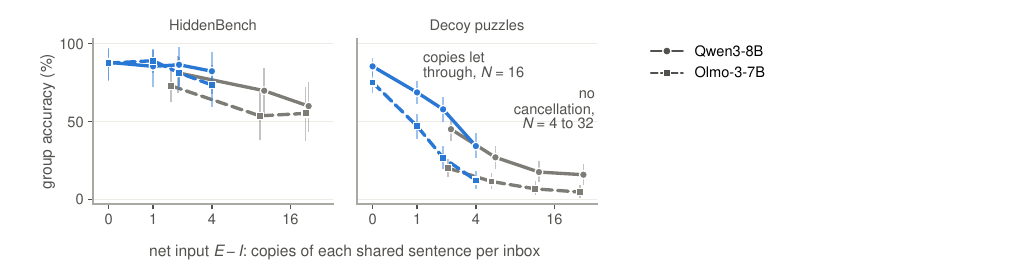}
  \caption{\textbf{Accuracy fell as more copies of each shared evidence sentence reached a receiver, most clearly on the
  puzzles.} Development items, exploratory. Group accuracy against the copies of each shared evidence sentence per inbox
  that pass the router, without cancellation as the group grows (gray; $N=4$ to 32) and with $k=1$, 2 or 4 copies let
  through at $N=16$ (blue; $k=0$ is the router). The x axis is logarithmic in 1 plus the copies. 95\% intervals.}
  \label{fig:netinput}
\end{figure}

Under all restated copies, each restated evidence sentence arrives once per sender that restated it. On the
puzzles, group accuracy then fell from \V{h3.qwen.nd.4.pc}\% at $N=4$ to \V{h3.qwen.nd.32.pc}\% at $N=32$ for
Qwen3-8B and from \V{h3.olmo.nd.4.pc}\% to \V{h3.olmo.nd.32.pc}\% for Olmo-3-7B, while the router held
\V{h3.qwen.pool.range.pc}\% and \V{h3.olmo.pool.range.pc}\%. The damage starts early: letting one, two or four copies
of each shared evidence sentence through the router at $N=16$ also lowered accuracy, from the first copy on the puzzles
and by four copies on HiddenBench (\cref{fig:netinput}).

\label{app:reader}The group is not needed for this effect. Single receivers answered once from a fixed inbox on
HiddenBench and the puzzles: each was an agent without private evidence, taken from a run of the router at
$N=16$ with its own evidence and round-0 answer (four HiddenBench tasks whose inboxes would overflow the
context were left out), and its inbox held every private evidence sentence once and each shared evidence sentence up
to eight times, with no answers. Eight copies of each shared evidence sentence lowered the share of correct receivers
in all four settings: on HiddenBench from \V{r.test.facts.c0.hb.qwen.pc}\% to \V{r.test.facts.c8.hb.qwen.pc}\% for
Qwen3-8B and from \V{r.test.facts.c0.hb.olmo.pc}\% to \V{r.test.facts.c8.hb.olmo.pc}\% for Olmo-3-7B, and on the
puzzles from \V{r.test.facts.c0.pz.qwen.pc}\% to \V{r.test.facts.c8.pz.qwen.pc}\% and from
\V{r.test.facts.c0.pz.olmo.pc}\% to \V{r.test.facts.c8.pz.olmo.pc}\%.

\subsection{Simpler filters}
\label{app:filters}

A smaller inbox alone does not explain the gain. The random sample, which gives each receiver as many sentences as the
router would deliver that round but draws them from all arriving copies, rarely delivered what a receiver lacked:
\V{kv.miss.cap.range.pc}\% of a receiver's missing evidence per round, against
\V{kv.miss.evpc.range.pc}\% under the router, because the copies of shared evidence sentences crowd out the rare ones;
an LLM summary of every message, written within the router's inbox length, did no better (\cref{tab:filters}).
Deduplication only, no size limit, which keeps once per round but delivers one copy of every shared evidence sentence
to receivers that hold it, cost \V{fil.ev.tok.more.range}\% more tokens than the router and \V{fil.ev.gap.range.pc}
points of accuracy, and all restated copies cost \V{fil.nodedup.tok.ratio.range} the router's tokens. The
opposite remedy fails as well: a full inbox history, which lets each receiver under standard message passing reread all
its earlier inboxes, cost \V{fil.hist.tok.ratio.range} the tokens of standard message passing and changed its accuracy
by \V{fil.hist.debate.range.pc} points.

\begin{table}[!htb]
\caption{\textbf{Simpler filters}: group accuracy (\%) and tokens per run of the random sample,
deduplication only and the LLM summary of messages, which keep the router's inbox size, and of ``+ full inbox
history'', all restated copies and deduplication only, no size limit, which do not; Qwen3-8B and Olmo-3-7B on
HiddenBench and the puzzles at $N=16$ and 32. Levels are per condition after its own exclusions; the differences quoted
in the text are paired on the items both conditions keep, so they can differ from the difference of two levels by a
few points.}
\label{tab:filters}
\centering
\scriptsize
\setlength{\tabcolsep}{4pt}
\begin{tabular}{@{}lcccccccc@{}}
\toprule
 & \multicolumn{4}{c}{Qwen3-8B} & \multicolumn{4}{c}{Olmo-3-7B} \\
\cmidrule(lr){2-5}\cmidrule(l){6-9}
 & \multicolumn{2}{c}{HiddenBench} & \multicolumn{2}{c}{puzzles} & \multicolumn{2}{c}{HiddenBench} & \multicolumn{2}{c}{puzzles} \\
Condition & $N=16$ & $N=32$ & $N=16$ & $N=32$ & $N=16$ & $N=32$ & $N=16$ & $N=32$ \\
\midrule
\multicolumn{9}{@{}l}{\emph{group accuracy}} \\
standard message passing & 6 & 7 & 11 & 10 & 16 & 10 & 10 & 10  \\
+ full inbox history & 7 & 5 & 12 & 12 & 19 & 22 & 10 & 10  \\
all restated copies & 53 & 42 & 24 & 20 & 48 & 44 & 8 & 7  \\
LLM summary of messages & 3 & 3 & 8 & 9 & 2 & 4 & 6 & 5  \\
random sample & 6 & 1 & 12 & 12 & 11 & 4 & 5 & 4  \\
deduplication only & 81 & 77 & 48 & 48 & 56 & 64 & 28 & 25  \\
deduplication only, no size limit & 75 & 72 & 84 & 84 & 70 & 70 & 68 & 68  \\
inhibitory message passing & 86 & 84 & 90 & 91 & 79 & 78 & 81 & 80  \\
\midrule
\multicolumn{9}{@{}l}{\emph{tokens per run (thousands)}} \\
standard message passing & 108 & 216 & 98 & 195 & 98 & 194 & 84 & 168 \\
+ full inbox history & 235 & 472 & 248 & 491 & 202 & 400 & 214 & 425 \\
all restated copies & 158 & 520 & 149 & 509 & 150 & 487 & 130 & 439 \\
LLM summary of messages & 75 & 149 & 56 & 111 & 70 & 138 & 49 & 98 \\
random sample & 65 & 131 & 50 & 99 & 63 & 126 & 47 & 94 \\
deduplication only & 66 & 132 & 53 & 106 & 64 & 127 & 49 & 98 \\
deduplication only, no size limit & 75 & 151 & 62 & 125 & 73 & 147 & 56 & 112 \\
inhibitory message passing & 66 & 132 & 53 & 108 & 64 & 128 & 48 & 97 \\
\bottomrule
\end{tabular}

\end{table}

\subsection{The text that carries the evidence}
\label{app:source}

The logs show why the sender's sentence cost accuracy. Each delivered sentence restated an evidence sentence the receiver
lacked, but \V{z5.view.wrong.range.pc}\% of them came from a sender then backing a wrong answer, and
\V{z5.view.held.range.pc}\% also restated an evidence sentence the receiver already held. Some restate the original
faithfully; others add a verdict, reverse a hazard or attach evidence about another option (\cref{tab:uex}). On test
items, against the original evidence sentence, the sender's sentence lowered accuracy (\V{z5.own.range.pc} points), as
did the LLM-extracted claims (\V{z5.claims.range.pc}), whose extractor never saw the original sentences and kept the
senders' inferences; the LLM paraphrase of the evidence stayed within \V{z5.para.range.pc} points of the original (the
paraphrases were checked for length and kept numbers, not for meaning). Two further variants locate the effect: latest message only changed accuracy by \V{u.last.range.pc} points, so the sentences restated in earlier
rounds matter little, and cancellation still helped the sender's sentence, which scored \V{u.ownnd.range.pc} points
above the sender's sentence, all copies.

\begin{table}[!htb]
  \caption{\textbf{What the sender's sentence delivers in place of the original evidence sentence}: four (evidence
  sentence, restating sentence) pairs drawn at random (Qwen3-8B, HiddenBench development, $N=16$). They illustrate a
  failure mode, not its frequency.}
  \label{tab:uex}
  \centering\scriptsize
\begin{tabular}{p{0.46\textwidth}p{0.46\textwidth}}
\toprule
Evidence sentences restated (what inhibitory message passing forwards) & The sender's sentence that restated them (what the sender's sentence forwards) \\
\midrule
Station Delta is the oldest building and sometimes has plumbing issues, but is currently empty. Maintenance crew at Station Delta inspected all major systems this morning and confirmed water is safe, backup lights work, and no signs of contamination were found. & Delta offers reliable power, safe water, and is currently empty---making it the safest and most functional option. \\
The Community Center basement has had a chemical leak that was just discovered today; the evacuation was for this, and it's only safe for short visits, not day-long events. & The Community Center is now confirmed safe for a full day, with no chemical leak impact on main areas. \\
The City Library has a large conference room and claims to be ready; it also has a known backup generator for power. Library staff admit privately that their backup generator only has enough fuel for 2 hours and cannot be refueled due to a citywide gasoline shortage. & The Library's generator only lasts 2 hours, insufficient for the full day. \\
Due to minor flooding on Oakcrest road, buses can only reach Riverside Park from the north side. & Although buses can only reach Riverside from the north due to Oakcrest road flooding, that route is clear. \\
\bottomrule
\end{tabular}

\end{table}

\subsection{Reasoning and answers beside the evidence}
\label{app:answers}

Letting the senders' reasoning and answers back in beside the routed evidence sentences undoes much of the gain at
$N=32$ and on the puzzles as at $N=16$ on HiddenBench (\cref{tab:ablation}, row ``+ reasoning and answers''). Single
receivers (\cref{app:reader}) show the pull directly: with every private evidence sentence in the inbox,
they were correct in \V{r.test.argued.81.range.pc}\% of cases when eight messages with reasoning and answers backed the
correct option and one the decoy, and in \V{r.test.argued.18.range.pc}\% when one backed the correct option and eight
the decoy.

\FloatBarrier
\section{Evidence growth and evidence quality}
\label{app:limits}

The following development-set tests examine how accuracy changes when the required evidence grows, distractors
accumulate, or a false sentence is planted. Inhibitory message passing retained its advantage over standard message
passing with growing required evidence (\cref{app:grow}), distractor paragraphs (\cref{app:private}) and a planted
false sentence (\cref{app:falsehood}).

\subsection{Needed evidence that grows with the group}
\label{app:grow}

In HiddenBench, the decoy puzzles and MuSiQue, at most four agents hold private evidence, so a larger group adds only
agents with shared evidence. To let the needed evidence grow with the group, we generated elimination problems: the
group picks the one option (a supplier, a venue, a suspect, \ldots) that no evidence sentence rules out. Each wrong
option is ruled out by one private evidence sentence (``Supplier Onyx failed the March safety inspection, so it cannot
get the contract''), nothing positive is said of the answer, and the shared evidence favours a decoy. Each of $H=N/4$
holders holds one elimination, so an agent knows the answer only when it knows all $H$. Items were screened per model:
with every evidence sentence the model had to answer at least three quarters of its samples correctly, and with the
shared evidence or one holder's evidence at most a quarter, naming the decoy from the shared evidence; 48 items
per group size were kept.

\begin{figure}[!htb]
  \centering
  \includegraphics[width=0.56\textwidth]{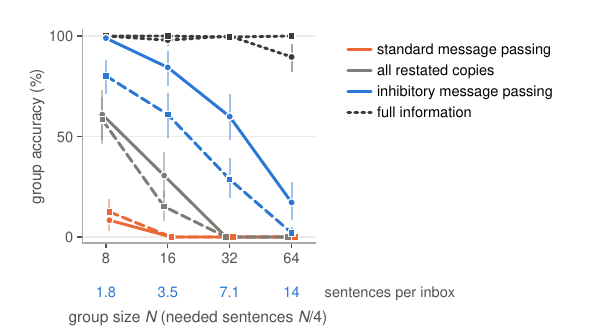}
  \caption{\textbf{When the needed evidence grows with the group, delivery and integration fall behind.} Elimination
  problems, development items, $H=N/4$: group accuracy under standard message passing, all restated copies, inhibitory
  message passing and full information, with 95\% intervals. Blue numbers: missing evidence sentences the router
  delivered per inbox, Qwen3-8B. Solid lines and circles: Qwen3-8B; dashed lines and squares: Olmo-3-7B.}
  \label{fig:grow}
\end{figure}

The router stayed above standard message passing and all restated copies, but its accuracy fell as $H$ grew
(\cref{fig:grow}), for Qwen3-8B from \V{x.qwen.8.pool_evpc.pc}\% at $N=8$ to \V{x.qwen.64.pool_evpc.pc}\% at $N=64$.
Unlike in \cref{fig:balance}a, where each receiver lacked a fixed set of evidence sentences, the evidence delivered per
inbox grew with the group, and the logs show two limits. At $N=8$ the inbox held every missing evidence sentence in
\V{x.qwen.8.complete.pc}\% of Qwen3-8B's receiver-rounds, and \V{x.qwen.8.after.pc}\% of those ended correct; at $N=64$
an inbox held \V{x.qwen.64.hidden} of the up to sixteen missing evidence sentences on average, only
\V{x.qwen.64.complete.pc}\% held all of them, and only \V{x.qwen.64.after.pc}\% of those ended correct. Delivery was
incomplete, and receivers that had every evidence sentence still rarely combined them; the logs do not separate the
two. We did not test regimes where the private evidence exceeds one context window.

\subsection{Distractor paragraphs}
\label{app:private}

On MuSiQue, each agent beyond the holders also held a distractor paragraph from the question's source
that neither answers it nor is shared, so a group of $N$ holds $N-H$ distracting paragraphs. The router led standard
message passing by \V{d4.evpc.debate.range.pc} points at every $N$ from 4 to 16, but its accuracy fell with $N$, from
\V{d4.pool_evpc.qwen.4.pc}\% to \V{d4.pool_evpc.qwen.16.pc}\% for Qwen3-8B and from \V{d4.pool_evpc.olmo.4.pc}\% to
\V{d4.pool_evpc.olmo.16.pc}\% for Olmo-3-7B. The distractors hurt whether or not they are routed: against the same
items without them at $N=16$, they changed the router's accuracy by \V{d4.cross.evpc.qwen.pc} and
\V{d4.cross.evpc.olmo.pc} points and that of full information by \V{d4.cross.full.qwen.pc} and
\V{d4.cross.full.olmo.pc}.

\subsection{A planted falsehood}
\label{app:falsehood}

The planted-falsehood test adds one false ``official update'' favouring the decoy to one agent's evidence in the
HiddenBench profile at $N=16$. The router forwarded it only when restated. Although accuracy fell,
inhibitory message passing remained substantially more accurate than standard message passing in both models
(\cref{tab:falsehood}), and its loss was smaller than that of full information. The router preserves an advantage in
this test while leaving evidence verification to the receiving agents.

\begin{table}[!htb]
\caption{\textbf{A planted falsehood} (HiddenBench development, $N=16$): group accuracy without $\to$ with the false
sentence, and the change in percentage points (95\% interval).}
\label{tab:falsehood}
\centering
\small
\resizebox{\textwidth}{!}{\begin{tabular}{@{}lcccc@{}}
\toprule
 & \multicolumn{2}{c}{Qwen3-8B} & \multicolumn{2}{c}{Olmo-3-7B} \\
Condition & reference $\to$ planted & change & reference $\to$ planted & change \\
\midrule
no communication & 2 $\to$ 3 & $+$2 [0, 4] & 11 $\to$ 11 & 0 [0, 0]  \\
standard message passing & 8 $\to$ 5 & $-$3 [$-$6, 0] & 21 $\to$ 23 & $+$2 [$-$7, 9]  \\
inhibitory message passing, 6 senders & 88 $\to$ 78 & $-$10 [$-$18, $-$3] & 81 $\to$ 66 & $-$15 [$-$24, $-$8]  \\
\textbf{inhibitory message passing} & 88 $\to$ 82 & $-$6 [$-$16, 4] & 88 $\to$ 79 & $-$8 [$-$16, $-$2]  \\
full information, one answer & 81 $\to$ 47 & $-$34 [$-$47, $-$21] & 74 $\to$ 34 & $-$41 [$-$56, $-$26]  \\
full information & 91 $\to$ 57 & $-$34 [$-$53, $-$16] & 91 $\to$ 42 & $-$48 [$-$64, $-$33]  \\
\bottomrule
\end{tabular}
}
\end{table}

\FloatBarrier
\section{The circuit analogy and its limits}
\label{app:mapping}

The analogy of \cref{sec:principle} and \cref{fig:principle}c is between rules for controlling input, not between
mechanisms. In circuit models, inhibition can track excitation input by input (detailed balance) or only in total
(global balance) \citep{vogels2009gating}; \cref{tab:mapping} maps these and the related circuit-model concepts onto
the router.

\begin{table}[!htb]
\caption{Computational analogy between cortical circuit models and evidence routing.}
\label{tab:mapping}
\centering
\small
\begin{tabular}{@{}>{\raggedright\arraybackslash}p{0.42\textwidth}>{\raggedright\arraybackslash}p{0.50\textwidth}@{}}
\toprule
Circuit-model concept & Router counterpart \\
\midrule
neuron or local circuit & receiving agent \\
shared feedforward input & evidence sentence initially held by multiple agents \\
input to a subset of neurons & evidence sentence initially held by only some agents \\
recurrent excitation & reinforcing influence of peers' reasoning and answers \\
excitation--inhibition balance input by input (detailed balance) & cancellation computed per evidence sentence and receiver \\
balance of total excitation and inhibition (global balance) & limit on total inbox size, independent of content \\
\bottomrule
\end{tabular}
\end{table}

The analogy stops at this arithmetic. Agents reason in language and the router operates on evidence sentences, not on
currents: restatements are matched by word overlap, cancellation uses the identity of a sentence and the receiver's
initial evidence, and a sentence pools copies from many senders, so it does not denote one projection. Omitting
reasoning and answers removes a source of recurrent influence within the analogy; cancellation per evidence sentence
is its only subtractive operation.

\FloatBarrier
\section{Extended related work}
\label{app:related}

Each operation of the router has a precedent, listed in \cref{tab:operations} with what it leaves open; what the router
adds is their combination, per evidence sentence and per receiver, in the communication rule between reasoning agents.
Beyond these precedents, debating agents conform to their peers \citep{weng2025conformity,wynn2025talk}, and hidden
profiles are studied on HiddenBench \citep{li2025hiddenbench,babu2026consilience,tambwekar2026proxifield}, in
forecasting with evidence split across agents \citep{li2026infodelphi}, in deliberation that loses distributed
evidence \citep{wan2026illusion} and with data partitioned across up to 100 agents \citep{zhang2026silobench}. Circuit
models of excitation--inhibition balance motivate a computational design, not a neural claim
\citep{hassabis2017neuroscience}.

\begin{table}[!htb]
\caption{Precedents of each operation of inhibitory message passing.}
\label{tab:operations}
\centering
\scriptsize
\begin{tabular}{@{}>{\raggedright\arraybackslash}p{0.17\textwidth}>{\raggedright\arraybackslash}p{0.44\textwidth}>{\raggedright\arraybackslash}p{0.33\textwidth}@{}}
\toprule
Operation & Precedent & What it leaves open \\
\midrule
forward verbatim evidence & debaters with a quote tool; quotes help the debater of the correct answer \citep{khan2024debating} &
  two debaters and a judge; no group of deciders, no scaling \\
drop the answers & an anonymous evidence board read by one judge \citep{ji2026anonymous}; removing self and peer labels
  curbs identity-driven sycophancy \citep{choi2026identity} & one judge, or labels removed while answers still circulate;
  not tested with distributed evidence \\
subtract what the receiver holds & a receiver discards peers' answers similar to its own or earlier ones \citep{zeng2025s2mad};
  learned pruning of edges \citep{zhang2025agentprune}; per-receiver merging of similar messages \citep{guan2026moc};
  sensor nodes advertise data so that each receives only what it lacks \citep{heinzelman1999spin} & LLM methods filter on
  answers, similarity or learned value, for cost; SPIN (sensor protocols for information via negotiation) moves sensor data, not evidence among reasoning agents \\
pool & blackboards \citep{erman1980hearsay,han2025blackboard} & every receiver reads the whole board; a redundant entry is
  removed for all receivers, not per receiver \\
count each piece of evidence once & the probabilistic norm and the channel filter of decentralized fusion
  \citep{pearl1988probabilistic,grime1994data}; incest removal \citep{krishnamurthy2014incest}; ``data incest'' among
  LLM agents \citep{jain2025interacting}; people neglect correlated sources \citep{enke2019correlation}; LLM aggregators
  count restated copies as partial new evidence \citep{gao2026copies} & the fusion filters are exact on trees only; among
  LLM agents, measured at one aggregator and corrected by prompting in four-agent teams; no communication rule removes copies
  before every receiver \\
brain-inspired communication & agents as cortical areas in a sparse, trust-weighted debate graph
  \citep{sun2025cortexdebate} & topology, not what a message carries \\
\bottomrule
\end{tabular}
\end{table}

\FloatBarrier
\section{Compute}
\label{app:compute}

The open-model runs used about \V{gpuh.h100.total} H100 GPU-hours and \V{gpuh.a30.total} A30 GPU-hours on one cluster,
and \V{gpuh.task.total} task-hours on a second cluster's A6000, A40 and H200 GPUs (a task-hour is one task's wall time
on one GPU). The commercial models ran on their providers' servers.

\end{document}